%% file: main.tex
\pdfoutput=1
 \documentclass[10pt,twocolumn,letterpaper]{article}

\usepackage{cvpr}
\usepackage{times}
\usepackage{epsfig}
\usepackage{graphicx}
\usepackage{amsmath}
\usepackage{amssymb}

\iftrue 
        \newcommand{\cutsectionup}{\vspace*{-0.08in}}
        \newcommand{\cutsectiondown}{\vspace*{-0.06in}}

        \newcommand{\cutsubsectionup}{\vspace*{-0.05in}}
        \newcommand{\cutsubsectiondown}{\vspace*{-0.04in}}

        \newcommand{\cuthalfcaptionup}{\vspace*{-0.15in}}
        \newcommand{\cuthalfcaptiondown}{\vspace*{-0.25in}}

        \newcommand{\cutcaptionup}{\vspace*{-0.1in}}
        \newcommand{\cutcaptiondown}{\vspace*{-0.15in}}

        \newcommand{\cuthalftablecaptionup}{\vspace*{-0in}}
        \newcommand{\cuthalftablecaptiondown}{\vspace*{-0.1in}}

        \newcommand{\cuttablecaptionup}{\vspace*{-0.05in}}
        \newcommand{\cuttablecaptiondown}{\vspace*{-0.1in}}

        \newcommand{\cutequationup}{\vspace*{-0.05in}}
        \newcommand{\cutequationdown}{\vspace*{-0in}}

        \newcommand{\cuttableup}{\vspace*{-0in}}
        \newcommand{\cuttabledown}{\vspace*{-0.21in}}
        
        \newcommand{\cutabstractup}{\vspace*{-0.12in}}
        \newcommand{\cutabstractdown}{\vspace*{-0.3in}}

\else 
        \newcommand{\cutsectionup}{}
        \newcommand{\cutsectiondown}{}

        \newcommand{\cutsubsectionup}{}
        \newcommand{\cutsubsectiondown}{}

        \newcommand{\cuthalfcaptionup}{}
        \newcommand{\cuthalfcaptiondown}{}

        \newcommand{\cutcaptionup}{}
        \newcommand{\cutcaptiondown}{}

        \newcommand{\cutequationup}{}
        \newcommand{\cutequationdown}{}

        \newcommand{\cuttableup}{}
        \newcommand{\cuttabledown}{}

        \newcommand{\cutabstractup}{}
        \newcommand{\cutabstractdown}{}

\fi

\newcommand{\addparagraphup}{\vspace*{0.01in}}

\usepackage{multirow}
\usepackage{makecell}
\usepackage{bm}
\usepackage{caption}
\usepackage[table]{xcolor}
\definecolor{Color0}{rgb}{.875,.875,.875}
\definecolor{Color1}{rgb}{1,.75,.75}
\definecolor{Color2}{rgb}{1,.875,.75}
\definecolor{Color3}{rgb}{.875,1,.875}
\definecolor{Color4}{rgb}{.875,.875,1}
\definecolor{ColorA}{rgb}{1,1,.875}
\definecolor{ColorB}{rgb}{.625,.625,.625}

\usepackage{balance}

\def\onewidth{\linewidth}
\def\twowidth{\linewidth}
\def\threewidth{.75\linewidth}

\def\fivewidth{.32\linewidth}
\def\sixwidth{.32\linewidth}

\def\clswidth{.24\linewidth}
\def\clsheight{.33\paperheight}

\def\catnumone{264}
\def\catnumtwo{530}
\def\catnumthree{1088}
\def\catnumfour{1117}
\def\catnumfive{1021}
\def\catnumsix{1369}
\def\catwordone{Cardigan Welsh corgi}
\def\catwordtwo{digital clock}
\def\catwordthree{foxhound}
\def\catwordfour{wildcat}
\def\catwordfive{shark}
\def\catwordsix{frozen dessert}

\def\printappendix{}
\ifdefined\printappendix
    \def\appendixword{Appendix}
\else
    \def\appendixword{Supplementary material}
\fi

\ifdefined\showupdated
    
\else
    
\fi

\usepackage[pagebackref=true,breaklinks=true,letterpaper=true,colorlinks,bookmarks=false]{hyperref}

\cvprfinalcopy 

\def\cvprPaperID{2307} 

\ifcvprfinal\fi
\begin{document}

\title{Hierarchical Novelty Detection for Visual Object Recognition}

\author{Kibok Lee$^{*}$\quad Kimin Lee$^{\dagger}$\quad Kyle Min$^{*}$\quad Yuting Zhang$^{*}$\quad Jinwoo Shin$^{\dagger}$\quad Honglak Lee$^{\ddagger*}$\\
$^{*}$University of Michigan, Ann Arbor, MI, USA\\
$^{\dagger}$Korea Advanced Institute of Science and Technology, Daejeon, Korea\\
$^{\ddagger}$Google Brain, Mountain View, CA, USA\\
}

\maketitle


\input{0_abs}

\input{1_intro}

\input{2_related}

\input{3_approach}

\input{4_hnd}

\input{5_zsl}
\balance
\input{6_conclusion}

\cutsectionup
\section*{Acknowledgements}
\cutsectiondown

This work was supported in part by Software R\&D Center, Samsung Electronics Co., Ltd., Kwanjeong Educational Foundation Scholarship, Sloan Research Fellowship, and DARPA Explainable AI (XAI) program \#313498.
We also thank Zeynep Akata, Yongqin Xian, Junhyuk Oh, Lajanugen Logeswaran, Sungryull Sohn, Jongwook Choi, and Yijie Guo for helpful discussions.

\clearpage

{\small
\bibliographystyle{ieee}
\bibliography{ref}
}
\clearpage

\ifdefined\printappendix
    \numberwithin{table}{section}
    \numberwithin{figure}{section}
    \numberwithin{equation}{section}
    
    \onecolumn
    \begin{appendix}
    \part*{Appendix}
    \input{a_hnd}
    \input{b_qual_smp}

    \clearpage
    \input{c_qual_cls}
    \clearpage
    \input{d_zsl}
    \end{appendix}
\fi
\end{document}

%% file: 0_abs.tex
\begin{abstract}
\cutabstractup
Deep neural networks have achieved impressive success in large-scale visual object recognition tasks with a predefined set of classes.
However, recognizing objects of novel classes unseen during training still remains challenging.
The problem of detecting such novel classes has been addressed in the literature, but most prior works have focused on providing simple binary or regressive decisions, e.g., the output would be ``known,'' ``novel,'' or corresponding confidence intervals.
In this paper, we study more informative novelty detection schemes based on a hierarchical classification framework.
For an object of a novel class, we aim for finding its closest super class in the hierarchical taxonomy of known classes.
To this end, we propose two different approaches termed top-down and flatten methods, and their combination as well.
The essential ingredients of our methods are confidence-calibrated classifiers, data relabeling, and the leave-one-out strategy for modeling novel classes under the hierarchical taxonomy.
Furthermore, our method can generate a hierarchical embedding that leads to improved generalized zero-shot learning performance in combination with other commonly-used semantic embeddings.
\cutabstractdown
\end{abstract}

%% file: 1_intro.tex
\cutsectionup
\section{Introduction} \label{sec:intro}
\cutsectiondown

\begin{figure}
\centering
\includegraphics[width=\onewidth]{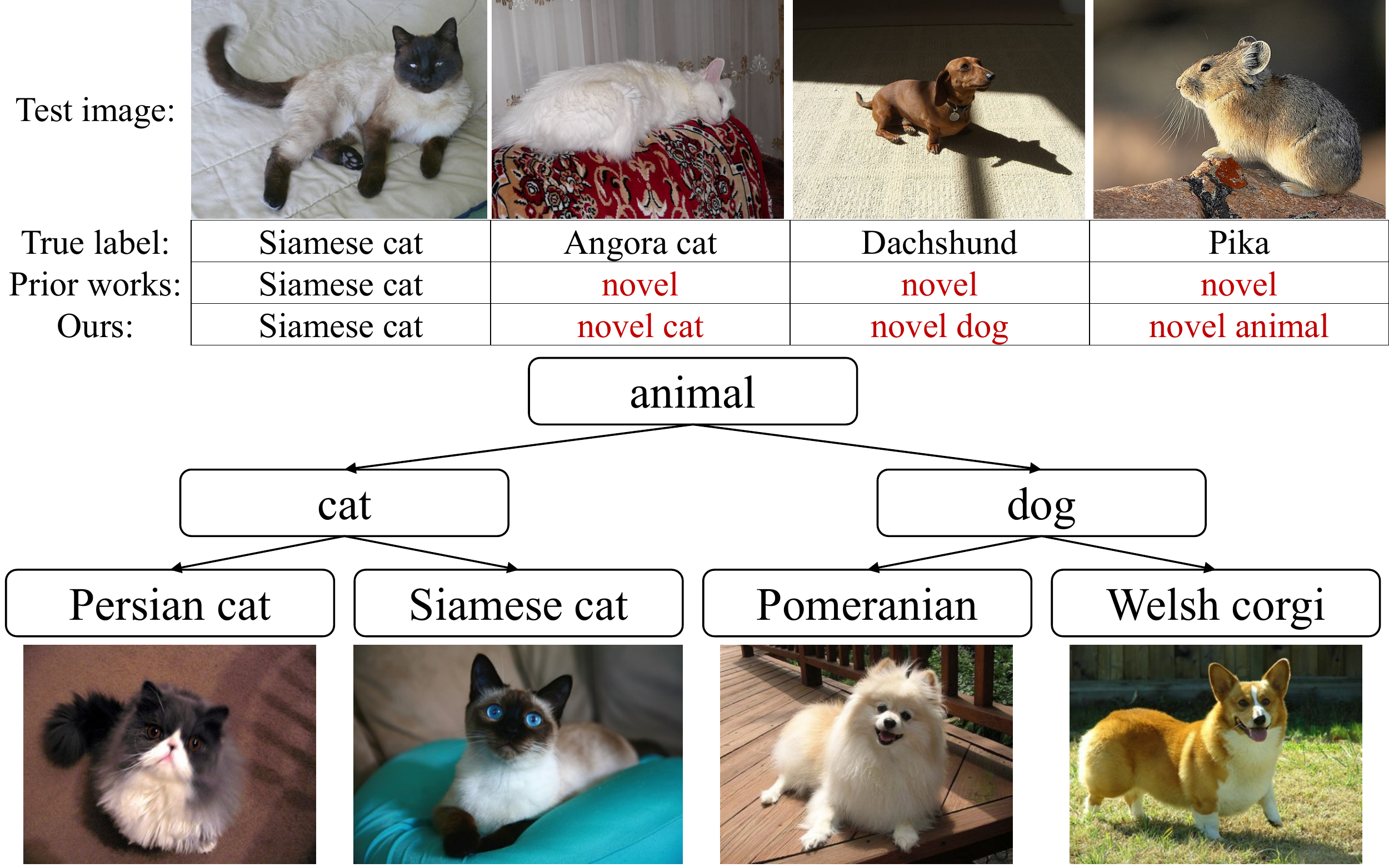}
\cuthalfcaptionup
\vspace*{-0.05in}
\caption{
An illustration of our proposed hierarchical novelty detection task.
In contrast to prior novelty detection works, we aim to find the most specific class label of a novel data on the taxonomy built with known classes.
}
\cuthalfcaptiondown
\label{fig:tasks}
\end{figure}

Object recognition in large-scale image datasets has achieved impressive performance with deep convolutional neural networks (CNNs)~\cite{he2015delving, he2016deep, simonyan2015very, szegedy2015going}.
The standard CNN architectures
are learned to recognize a predefined set of classes seen during training.
However, in practice,
a new type of objects could emerge (e.g., a new kind of consumer product).
Hence, it is desirable to extend the CNN architectures for detecting the novelty of an object (i.e., deciding if the object does not match any previously trained object classes).
There have been recent efforts toward developing efficient novelty detection methods
~\cite{bendale2015towards, hendrycks2016baseline, lakshminarayanan2016simple, li2017dropout, pimentel2014review}, 
but most of the existing methods measure only the model uncertainty, i.e., confidence score, which is often too ambiguous for practical use.
For example, suppose one trains a classifier on an animal image dataset as in Figure~\ref{fig:tasks}.
A standard novelty detection method
can be applied to a cat-like image to evaluate its novelty, but
such a method would not tell whether the novel object is a new species of cat unseen in the training set or a new animal species.

To address this issue, we design a new classification framework for more informative novelty detection by utilizing a hierarchical taxonomy,
where the taxonomy can be 
extracted from the natural language information, e.g., WordNet hierarchy~\cite{miller1995wordnet}.
Our approach is also motivated by a strong empirical correlation between hierarchical semantic relationships and the visual appearance of objects~\cite{deng2010does}.
Under our scheme, a taxonomy is built with the hypernym-hyponym relationships
between known classes such that objects from novel classes are expected to be classified into the most relevant label, i.e., the closest class in the taxonomy.
%
For example, as illustrated in Figure~\ref{fig:tasks}, our goal is to distinguish ``new cat,'' ``new dog,'' and ``new animal,'' which cannot be achieved in the standard novelty detection tasks. 
We call this problem \emph{hierarchical novelty detection} task.

In contrast to standard object recognition tasks with a \emph{closed set} of classes, our proposed framework can be useful for extending the domain of classes to an \emph{open set} with taxonomy information (i.e., dealing with any objects unseen in training). 
In practical application scenarios, our framework can be potentially useful for automatically or interactively organizing a customized taxonomy (e.g., company's product catalog, wildlife monitoring, personal photo library) by suggesting closest categories for an image from novel categories (e.g., new consumer products, unregistered animal species, untagged scenes or places).

We propose two different approaches for hierarchical novelty detection:
top-down and flatten methods.
In the top-down method, each super class has a confidence-calibrated classifier which detects a novel class if the posterior categorical distribution is close to a uniform distribution.
Such a classifier was recently studied
for a standard novelty detection task~\cite{lee2018training}, and we extend it for detecting novel classes under our hierarchical novelty detection framework.
On the other hand, the flatten method  computes a softmax probability distribution of all disjoint classes.
Then, it 
predicts the most likely fine-grained label, either a known class or a novel class.
Although the flatten method simplifies the full hierarchical structure,
it outperforms the top-down method for datasets of a large hierarchical depth.

Furthermore, we combine two methods for utilizing their complementary benefits:
top-down methods naturally leverage the hierarchical structure information, 
but the classification performance might be degraded due to the error aggregation.
On the contrary, flatten methods have a single classification rule that avoids the error aggregation, but the classifier's flat structure does not utilize the full information of hierarchical taxonomy.
We empirically show that combining the top-down and flatten models further improves hierarchical novelty detection performance.

Our method can also be useful for generalized zero-shot learning (GZSL)~\cite{chao2016empirical, xian2017zero} tasks. 
GZSL is a classification task with classes both seen and unseen during training, given that semantic side information for all test classes is provided.
We show that our method can generate a hierarchical embedding that leads to improved GZSL performance in combination with other commonly used semantic embeddings.



%% file: 2_related.tex
\cutsectionup
\section{Related work} \label{sec:related}
\cutsectiondown

\addparagraphup
\noindent{\bf Novelty detection.}
For robust prediction, it is desirable to detect a test sample if it looks unusual or significantly differs from the representative training data.
Novelty detection is a task recognizing such abnormality of data~(see \cite{hodge2004survey,pimentel2014review} for a survey).
Recent novelty detection approaches leverage the output of deep neural network classification models. 
A confidence score about novelty can be measured by taking the maximum predicted probability~\cite{hendrycks2016baseline}, ensembling such outputs from multiple models~\cite{lakshminarayanan2016simple},
or synthesizing a score based on the predicted categorical distribution~\cite{bendale2015towards}.
There have also been recent efforts toward confidence-calibrated novelty detection, i.e., calibrating how much the model is certain with its novelty detection,
by postprocessing~\cite{liang2018enhancing}
or learning with joint objective~\cite{lee2018training}.

\addparagraphup
\noindent{\bf Object recognition with taxonomy.}
Incorporating the hierarchical taxonomy for object classification has been investigated in the literature,
either to improve classification performance~\cite{deng2014large, yan2015hd}, or to extend the classification tasks to obtain more informative results~\cite{deng2012hedging, zhao2017open}.
Specifically for the latter purpose, Deng et al.~\cite{deng2012hedging} gave some reward to super class labels in a taxonomy and  maximized the expected reward.
Zhao et al.~\cite{zhao2017open} proposed an open set scene parsing framework, where the hierarchy of labels is used to estimate the similarity between the predicted label and the ground truth.
In contemporary work, Simeone et al.~\cite{simeone2017hierarchical} proposed a hierarchical classification and novelty detection task for
the music genre classification, but their settings are different from ours:
in their task,
novel classes do not belong to any node in the taxonomy.
Thus, their method cannot distinguish the difference between novel classes similar to some known classes.
To the best of our knowledge, our work is the first to propose a unified framework for hierarchical novelty detection and visual object recognition. 

\addparagraphup
\noindent{\bf Generalized zero-shot learning (GZSL).}
We remark that GZSL~\cite{chao2016empirical, xian2017zero} can be thought as addressing a similar task as ours.
While the standard ZSL tasks test classes unseen during training only,
GZSL tasks test both seen and unseen classes
such that the novelty is automatically detected if the predicted label is not a seen class.
However,
ZSL and GZSL tasks are valid only under the assumption that specific
semantic information of all test classes is given, e.g., attributes~\cite{akata2015evaluation, lampert2013attribute, rohrbach2013transfer} or text description~\cite{changpinyo2016synthesized, frome2013devise, fu2016semi, norouzi2013zero, reed2016learning, socher2013zero} of the objects.
Therefore, GZSL cannot recognize a novel class if prior knowledge about the specific novel class is not provided, i.e., it is limited to classifying objects with prior knowledge, regardless of their novelty.
Compared to GZSL, the advantages of the proposed hierarchical novelty detection are that
1)~it does not require any prior knowledge on novel classes but only utilizes the taxonomy of known classes,
2)~a reliable super class label can be more useful and human-friendly than an error-prone prediction over excessively subdivided classes, and
3)~high-quality taxonomies are available off-the-shelf and they are better interpretable than latent semantic embeddings.
In Section~\ref{sec:zsl}, we also show that our models for hierarchical novelty detection can also generate a hierarchical embedding such that combination with other semantic embeddings improves the GZSL performance.


%% file: 3_approach.tex
\cutsectionup
\section{Approach} \label{sec:approach}
\cutsectiondown

In this section, we define terminologies to describe hierarchical taxonomy and then propose models for
our proposed hierarchical novelty detection task.

\cutsubsectionup
\subsection{Taxonomy} \label{sec:taxonomy}
\cutsubsectiondown

\begin{figure}[t]
\centering
\includegraphics[width=\twowidth]{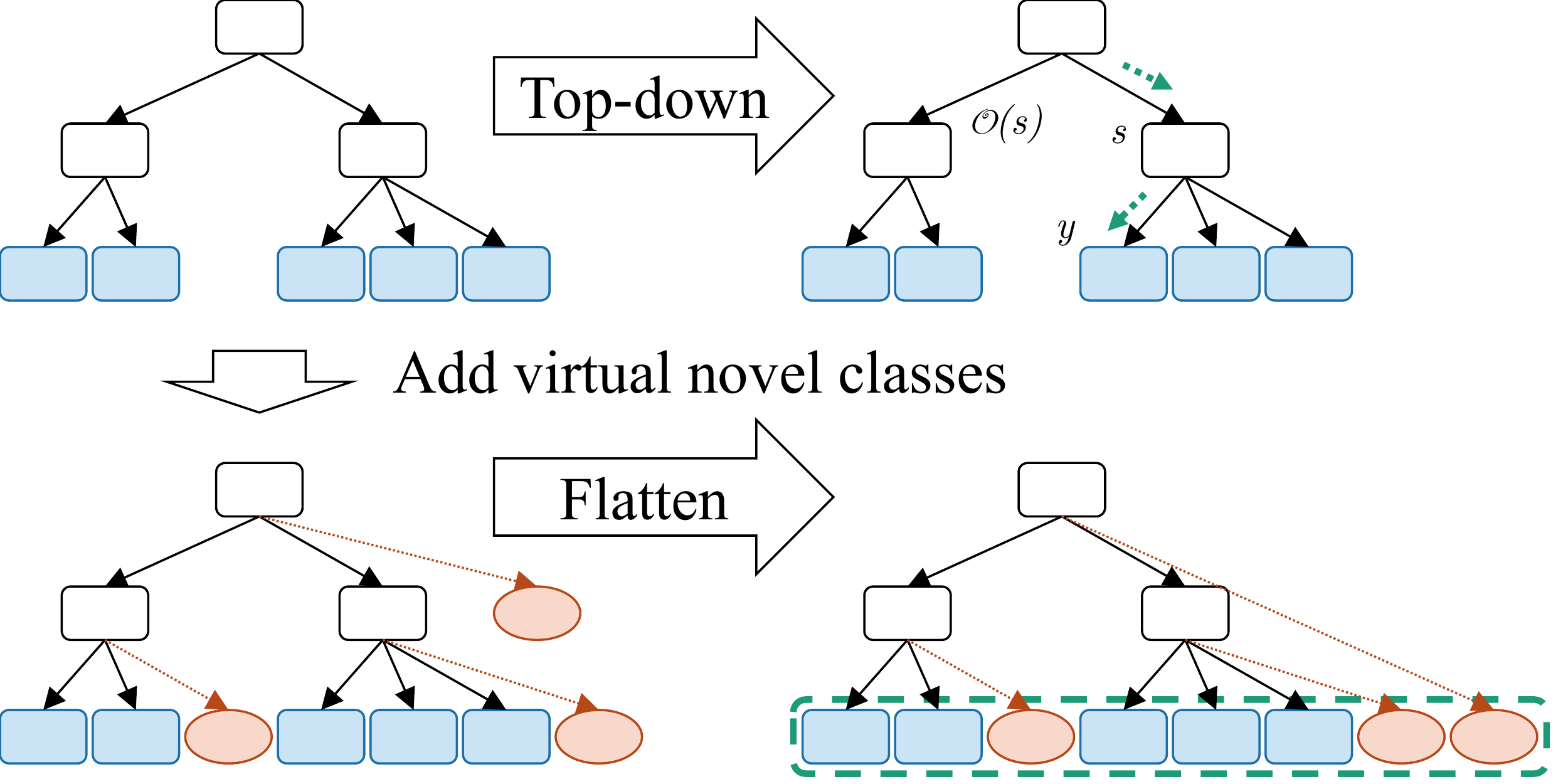}
\vspace*{-0.15in}
\cuthalfcaptionup
\caption{Illustration of two proposed approaches.
In the top-down method, classification starts from the root class, and propagates to one of its children until the prediction arrives at a known leaf class (blue) or stops if the prediction is not confident, which means that the prediction is a novel class whose closest super class is the predicted class.
In the flatten method, we add a virtual novel class (red) under each super class as a representative of all novel classes, and then flatten the structure for classification.
}
\cuthalfcaptiondown
\label{fig:approach}
\end{figure}

A taxonomy represents a hierarchical relationship between classes, where each node in the taxonomy corresponds to a class or a set of indistinguishable classes.\footnote{
For example, if a class has only one known child class, these two classes are indistinguishable as they are trained with exactly the same data.
}
We define three types of classes as follows:
1)~\emph{known leaf classes} are nodes with no child, which are known and seen during training, 
2)~\emph{super classes} are ancestors of the leaf classes, which are also known, and
3)~\emph{novel classes} are unseen during training, so they do not explicitly appear in the taxonomy.\footnote{
We note that ``novel'' in our task is similar but different from ``unseen'' commonly referred in ZSL works;
while class-specific semantic information for unseen classes must be provided in ZSL, such information for novel classes is not required in our task.
}
We note that all known leaf and novel classes have no child and are disjoint, i.e., they are neither ancestor nor descendant of each other.
In the example in Figure~\ref{fig:tasks}, four species of cats and dogs are leaf classes, ``cat,'' ``dog,'' and ``animal'' are super classes, and any other classes unseen during training, e.g., ``Angora cat,'' ``Dachshund,'' and ``Pika'' are novel classes.

In the proposed hierarchical novelty detection framework, we first build a taxonomy with known leaf classes and their super classes.
At test time,
we aim to predict the most fine-grained label in the taxonomy.
For instance,
if an image is predicted as novel, we try to assign one of the super classes, implying that the input is in a novel class whose closest known class in the taxonomy is that super class.
%
%
%

To represent the hierarchical relationship, let
$\mathcal{T}$ be the taxonomy of known classes, and for a class $y$,
$\mathcal{P}(y)$ be the set of parents, 
$\mathcal{C}(y)$ be the set of children, 
$\mathcal{A}(y)$ be the set of ancestors including itself, and 
$\mathcal{N}(y)$ be the set of novel classes whose closest known class is $y$.
Let $\mathcal{L}(\mathcal{T})$ be the set of all descendant known leaves under $\mathcal{T}$,
such that $\mathcal{T} \backslash \mathcal{L}(\mathcal{T})$ is the set of all super classes in $\mathcal{T}$.

As no prior knowledge of $\mathcal{N}(y)$ is provided during training and testing, all classes in $\mathcal{N}(y)$ are indistinguishable in our hierarchical novelty detection framework.
Thus, we treat $\mathcal{N}(y)$ as a single class in our analysis.

\cutsubsectionup
\subsection{Top-down method} \label{sec:td}
\cutsubsectiondown


A natural way to perform classification using a hierarchical taxonomy is following \emph{top-down} classification decisions starting from the root class, as shown in the top of Figure~\ref{fig:approach}.
Let $(x,y) \sim Pr(x,y|s)$ be a pair of an image and its label sampled from data distribution at a super class $s$, where $y \in \mathcal{C}(s) \cup \mathcal{N}(s)$.
Then, the classification rule is defined as
\cutequationup
\begin{align*}
\hat{y} = 
\begin{cases}
\underset{y'}{\arg\max} ~
Pr(y'|x,s;\theta_{s}) &\text{if confident}, \\
\qquad\qquad \mathcal{N}(s) &\text{otherwise,}
\end{cases}
\end{align*}
\cutequationdown
where $\theta_{s}$ is the model parameters of $\mathcal{C}(s)$
and $Pr(\,\cdot\, |x,s;\theta_{s})$ is the posterior categorical distribution given an image $x$ at a super class $s$.
The top-down classification stops at $s$ if the prediction is a known leaf class or the classifier is not confident with the prediction (i.e., the predicted class is in 
$\mathcal{N}(s)$).
We measure the prediction confidence using the KL divergence with respect to the uniform distribution:
intuitively, a confidence-calibrated classifier generates near-uniform posterior probability vector if the classifier is not confident about its prediction.
Hence, we interpret that the prediction is confident at a super class $s$ if
\cutequationup
\begin{align*}
D_{KL} (U(\cdot |s) \parallel Pr(\cdot|x,s;\theta_{s})) \geq \lambda_{s},
\end{align*}
\cutequationdown
where $\lambda_{s}$ is a threshold, $D_{KL}$ denotes the KL divergence, and $U(\cdot|s)$ is the uniform distribution when the classification is made under a super class $s$.
To train such confidence-calibrated classifiers, we leverage classes disjoint from the class $s$. 
Let $\mathcal{O}(s)$ be such a set of all known classes except for $s$ and its descendents.
%
Then, the objective function of our top-down classification model at a super class $s$ is
\cutequationup
\begin{align}
&\min_{\theta} ~
\mathbb{E}_{Pr(x,y|s)} \left[ -\log Pr (y|x,s; \theta_{s}) \right] \nonumber\\
&\quad+ \mathbb{E}_{Pr(x,y|\mathcal{O}(s))} \left[ D_{KL} \left( U (\cdot|s) \parallel Pr(\cdot|x,s; \theta_{s}) \right) \right],
\end{align}
\cutequationdown
where $Pr(x,y|\mathcal{O}(s))$ denotes the data distribution
of $\mathcal{O}(s)$.

However, under the above top-down scheme, the classification error might aggregate as the hierarchy goes deeper.
For example, if one of the classifiers has poor performance, then the overall classification performance of all descendent classes should be low.
In addition, the taxonomy is not necessarily a tree but a directed-acyclic graph (DAG), i.e., a class could belong to multiple parents, which could lead to incorrect classification.\footnote{
For example, if there are multiple paths to a class in a taxonomy, then the class may belong to (i.e., be a descendant of) multiple children at some super class $s$, which may lead to low KL divergence from the uniform distribution and the image could be incorrectly classified as $\mathcal{N}(s)$.
}
%
In the next section, we propose flatten approaches, which overcome the error aggregation issue.
Nevertheless, the top-down method can be used for extracting good visual features for boosting the performance of the flatten method, as we show in Section~\ref{sec:hnd}.

\cutsubsectionup
\subsection{Flatten method} \label{sec:fl}
\cutsubsectiondown
We now propose to represent all probabilities of known leaf and novel classes in a single probability vector, i.e., we \emph{flatten} the hierarchy, as described on the bottom of Figure~\ref{fig:approach}.
The key idea is that a probability of a super class $s$ can be represented as
$Pr(s|x) = \sum_{y \in \mathcal{C}(s)} Pr(y|x) + Pr(\mathcal{N}(s)|x)$,
such that from the root node, we have
$\sum_{l \in \mathcal{L}(\mathcal{T}) } Pr(l|x) + \sum_{s \in \mathcal{T} \backslash \mathcal{L}(\mathcal{T})} Pr(\mathcal{N}(s)|x) = 1$,
where $l$ and $s$ are summed over all known leaf classes and super classes, respectively. 
Note that $\mathcal{N}(s)$ is considered as a single novel class under the super class $s$, as discussed in Section~\ref{sec:taxonomy}.
Thus, as described in Figure~\ref{fig:approach}, one can virtually add an extra child for each super class to denote all novel classes under it.
Let $(x,y) \sim Pr(x,y)$ be a pair of an image and its most fine-grained label sampled from data distribution.
Then, the classification rule is
\begin{align*}
\hat{y} = \underset{y'}{\arg\max}~ Pr(y'|x;\theta),
\end{align*}
where $y'$ is either a known leaf or novel class.
Here, a problem is that we have no training data 
from novel classes.
To address this, we propose two approaches to model the score (i.e., posterior probability) of novel classes.

\begin{figure}[t]
\centering
\includegraphics[width=\threewidth]{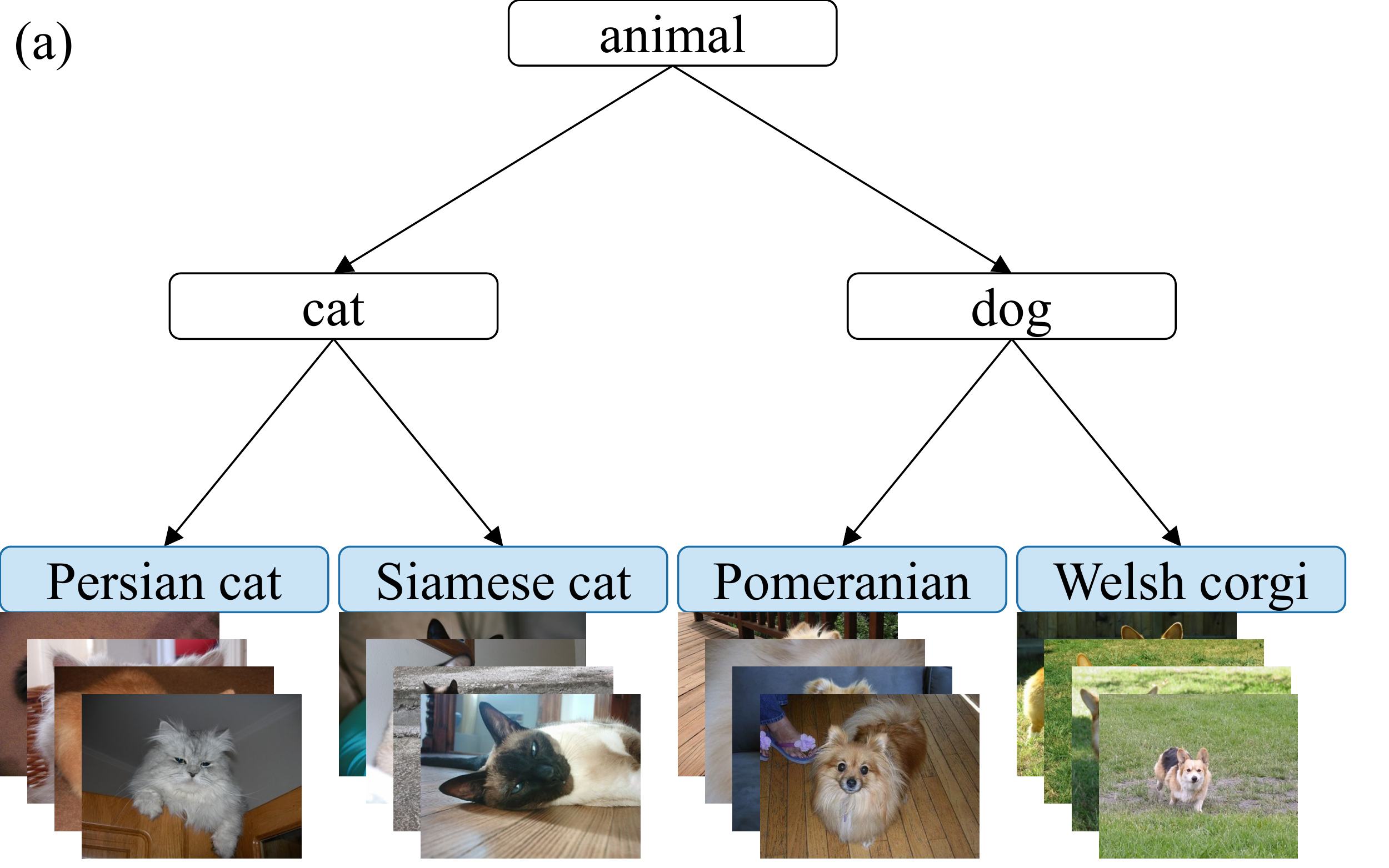}
\includegraphics[width=\threewidth]{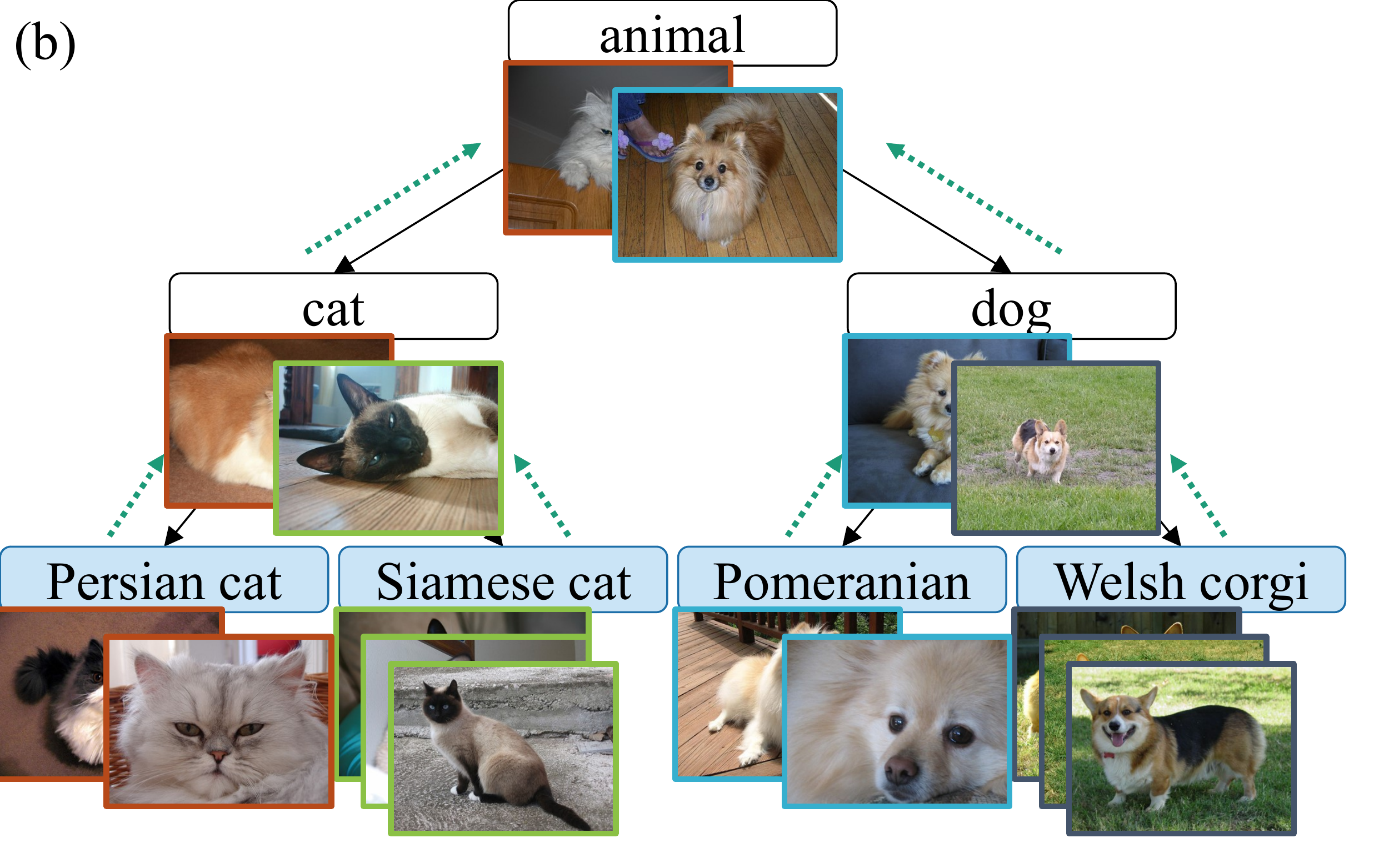}
\includegraphics[width=\threewidth]{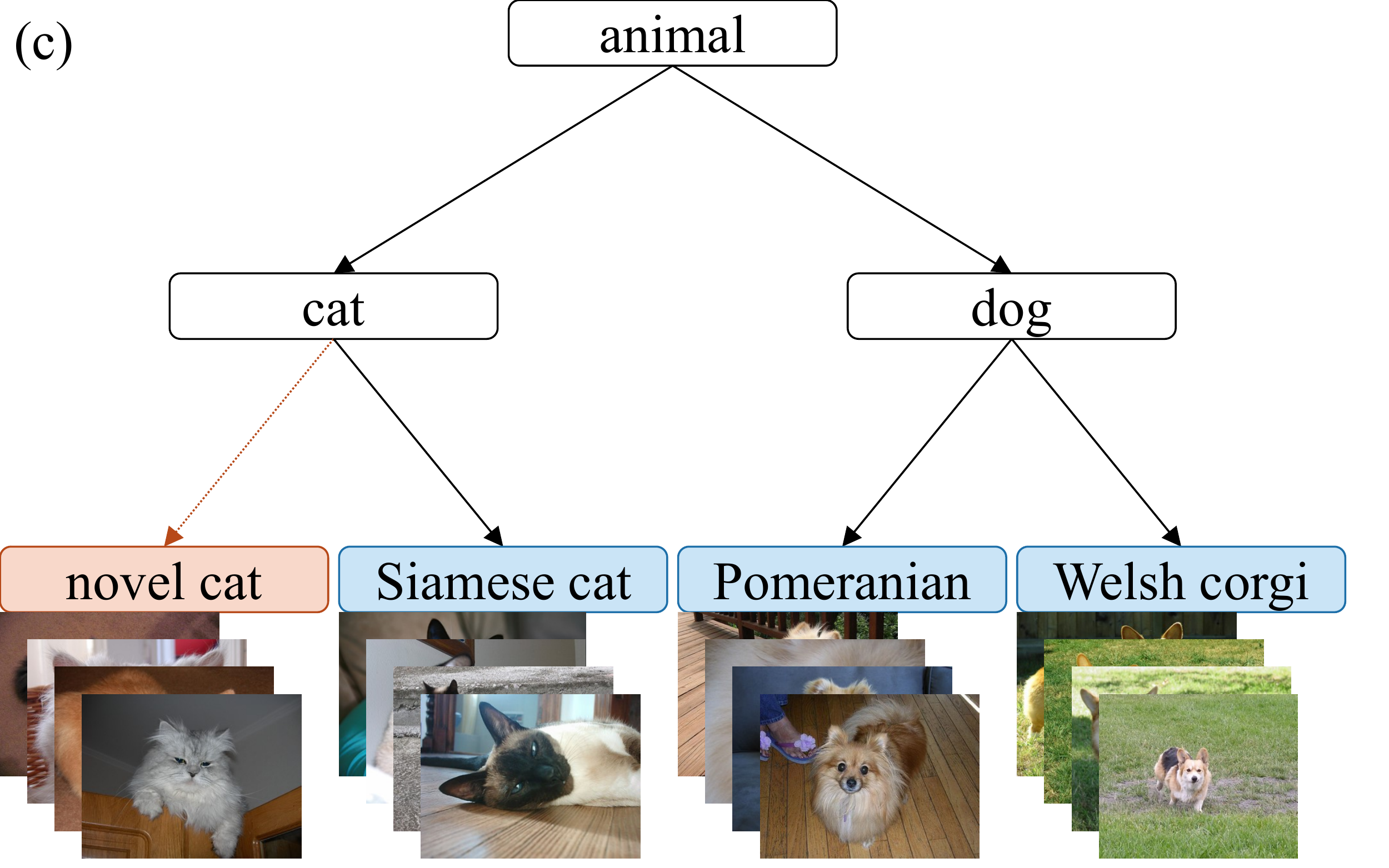}
\includegraphics[width=\threewidth]{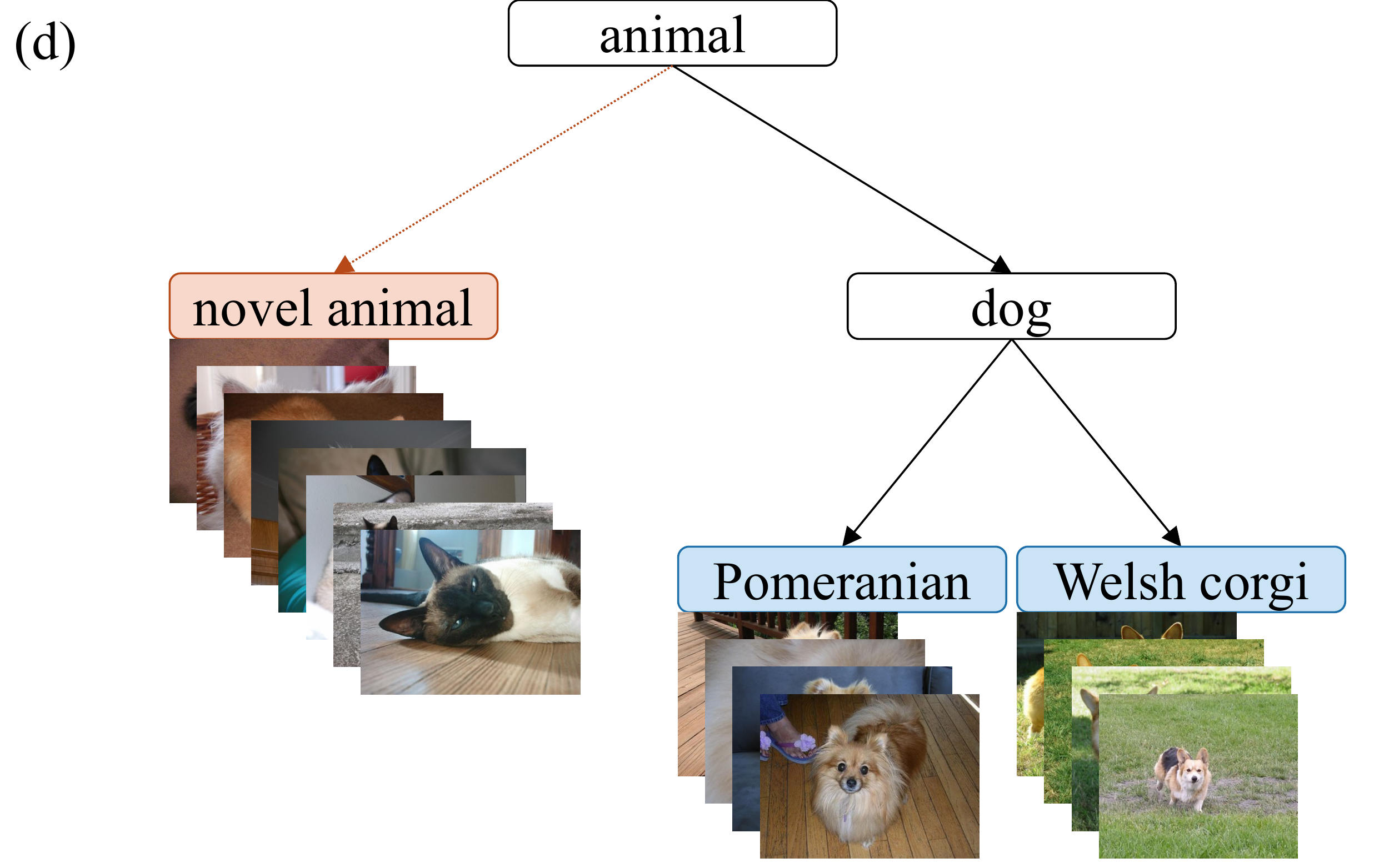}
\cuthalfcaptionup
\vspace*{0.03in}
\caption{
Illustration of strategies to train novel class scores in flatten methods.
(a) shows the training images in the taxonomy.
(b) shows relabeling strategy.
Some training images are relabeled to super classes in a bottom-up manner.
(c--d) shows leave-one-out (LOO) strategy.
To learn a novel class score under a super class, one of its children is temporarily removed such that its descendant known leaf classes are treated as novel during training.
}
\cuthalfcaptiondown
\label{fig:flatten}
\end{figure}

\addparagraphup
\noindent{\bf Data relabeling.}
A naive strategy is to relabel some training samples to its ancestors in hierarchy.
Then, the images relabeled to a super class are considered as novel class images under the super class.
This can be viewed as a supervised learning with both fine-grained and coarse-grained classes where they are considered to be disjoint, and one can optimize an objective function of a simple cross entropy function over all known leaf classes and novel classes:
\begin{align}
\min_{\theta} ~
\mathbb{E}_{Pr(x,y)} \left[ -\log Pr (y|x; \theta_{\mathcal{T}}) \right].
\end{align}
In our experiments, each training image is randomly relabeled recursively in a bottom-up manner with a probability of $r$, where $0 < r < 1$ is termed a relabeling rate.
An example of relabeling is illustrated in Figure~\ref{fig:flatten}~(b).

\addparagraphup
\noindent{\bf Leave-one-out strategy.}
A more sophisticated way to model novel classes is to temporarily remove a portion of taxonomy during training:
specifically, for a training label $y$, we recursively remove one of its ancestor $a \in \mathcal{A}(y)$ from the taxonomy $\mathcal{T}$ in a hierarchical manner.
To represent a deficient taxonomy, we define $\mathcal{T} \backslash a$ as a taxonomy where $a$ and its descendants are removed from the original taxonomy $\mathcal{T}$.
At each stage of removal, the training label $y$ becomes a novel class of the parent of $a$ in $\mathcal{T} \backslash a$, i.e., $\mathcal{N}(\mathcal{P}(a))$.
Figure~\ref{fig:flatten}~(a, c--d) illustrates this idea with an example:
in Figure~\ref{fig:flatten}~(a),
when $y$ is ``Persian cat,''
the set of its ancestor is $\mathcal{A}(y)=$\{~``Persian cat,'' ``cat,'' ``animal''~\}.
In Figure~\ref{fig:flatten}~(c),
images under $a=$``Persian cat'' belong to $\mathcal{N}(\mathcal{P}(a))=$``novel cat'' in $\mathcal{T} \backslash a$.
Similarly, in Figure~\ref{fig:flatten}~(d),
images under $a=$``cat'' belong to $\mathcal{N}(\mathcal{P}(a))=$``novel animal'' in $\mathcal{T} \backslash a$.
As we leave a class out to learn a novel class, we call this \emph{leave-one-out} (LOO) method.
With some notation abuse for simplicity, the objective function of the LOO model is then
\cutequationup
\begin{align}
\min_{\theta} ~
&\mathbb{E}_{Pr(x,y)} \bigg[
-\log Pr (y|x; \theta_{\mathcal{L}(\mathcal{T})}) \nonumber \\
&\qquad + \sum_{a \in \mathcal{A}(y)} -\log Pr (\mathcal{N}(\mathcal{P}(a))|x; {\theta_{\mathcal{T} \backslash a}}) \bigg],
\end{align}
\cutequationdown
where the first term is the standard cross entropy loss with the known leaf classes, and the second term is the summation of losses with $\mathcal{N}(\mathcal{P}(a))$ and the leaves under $\mathcal{T} \backslash a$.
We provide further implementation details in \appendixword.

As we mentioned earlier, the flatten methods can be combined with the top-down method in sequence:
the top-down method first extracts multiple softmax probability vectors from visual features, and then the concatenation of all probabilities can be used as an input of the LOO model.
We name the combined method TD+LOO for conciseness.

%% file: 4_hnd.tex
\cutsectionup
\section{Evaluation: Hierarchical novelty detection} \label{sec:hnd}
\cutsectiondown

We present the hierarchical novelty detection performance of our proposed methods 
combined with CNNs on ImageNet~\cite{deng2009imagenet},
Animals with Attributes 2 (AwA2)~\cite{lampert2013attribute, xian2017zero}, and Caltech-UCSD Birds (CUB)~\cite{welinder2010caltech}, where they represent visual object datasets with deep, coarse-grained, and fine-grained taxonomy, respectively.
Experimental results on CIFAR-100~\cite{krizhevsky2009learning} can be found in \appendixword, where the overall trends of results are similar to others. 

\input{qual_smp/qual_smp_0}

\cutsubsectionup
\subsection{Evaluation setups} \label{sec:setup}
\cutsubsectiondown

\addparagraphup
\noindent{\bf Compared algorithms.}
As a baseline, we modify the dual accuracy reward trade-off search (DARTS) algorithm~\cite{deng2012hedging} 
for our purpose.
Note that DARTS gives some rewards to labels in hierarchy, where fine-grained prediction gets higher reward.
Under this algorithm, for a novel class, its closest super class in the taxonomy would give the maximum reward.
At test time, 
the modified DARTS generates expected rewards for all known leaf and novel classes, so prediction can be done in the same way as the flatten methods.

As our proposed methods, Relabel, LOO, and TD+LOO are compared.
For a fair comparison in terms of the model capacity, deep Relabel and LOO models are also experimented, where a deep model is a stack of fully connected layers followed by rectified linear units (ReLU).
We do not report the performance of the pure top-down method
since 1)~one can combine it with the LOO method for better performance as mentioned in Section~\ref{sec:td}, and 
2)~fair comparisons between the pure top-down method and others are not easy.
Intuitively, the confidence threshold $\lambda_s$ in Section~\ref{sec:td} can be tuned:
for example, the novel class score bias in the flatten method would improve the novel class detection accuracy, but large $\lambda_s$ does not guarantee the best novel class performance in the top-down method because hierarchical classification results would tend to stop at the root class.

\addparagraphup
\noindent{\bf Datasets.}
ImageNet~\cite{deng2009imagenet} consists of 22k object classes where the taxonomy of the classes is built with the hypernym-hyponym relationships in WordNet~\cite{miller1995wordnet}.
We take 1k mutually exclusive classes in ILSVRC 2012 as known leaf classes, which are a subset of the ImageNet.\footnote{
Except ``teddy bear,'' all classes in ILSVRC 2012 are in ImageNet.
}
Based on the hypernym-hyponym relationships in WordNet, we initially obtained 860 super classes of 1k known leaf classes, and then merged indistinguishable super classes.
Specifically, if a super class has only one child or shares exactly the same descendant leaf classes, it is merged with classes connected to the class.
After merging, the resultant taxonomy is a DAG and has 396 super classes where all super classes have at least two children and have different set of descendant leaf classes.
On the other hand, the rest of 21k classes can be used as novel classes for testing.
Among them, we discarded super classes, classes under 1k known leaf classes, and classes with less than 50 images for reliable performance measure.
After filtering classes, we obtain about 16k novel classes.
ILSVRC 2012 has about 1.3M training images and another 50k images in 1k known leaf classes.
We put the 50k images aside from training and used for test, and we sampled another 50k images from 1.3M training images for validation.
For novel classes, we sampled 50 images from each class.
In summary, we have about 1.2M training images, 50k validation images, and 50k test images from known leaf classes, and 800k test images from novel classes.

AwA2~\cite{lampert2013attribute, xian2017zero} consists of 40 known leaf classes and 10 novel classes with 37k images, and CUB~\cite{welinder2010caltech} consists of 150 known leaf classes and 50 novel classes with 12k images.
Similar to ImageNet, the taxonomy of each dataset is built with the hypernym-hyponym relationships in WordNet.
The resultant taxonomy is a tree and has 21 and 43 super classes for AwA2 and CUB, respectively.

\addparagraphup
\noindent{\bf Training.}
We take ResNet-101~\cite{he2016deep} as a visual feature extractor (i.e., the penultimate layer of the CNNs before the classification layer) for all compared methods.
The CNNs are pretrained with ILSVRC 2012 1k classes, where they do not contain any novel classes of datasets experimented.
Then, the final classification layer of the CNNs is replaced with our proposed models.
Note that CNNs and our proposed models can be trained in an end-to-end manner, but we take and freeze the pretrained parameters in all layers except for the final layer for the sake of faster training.

For ImageNet, we use mini-batch SGD with 5k center-cropped data per batch.
As a regularization, L2 norm weight decay with parameter $10^{-2}$ is applied.
The initial learning rate is $10^{-2}$ and it decays at most two times when loss improvement is less than 2\% compared to the last epoch.

For AwA2 and CUB, the experiments are done in the same environment with the above except that the models are trained with the full-batch GD and Adam optimizer~\cite{kingma2014adam}.


\addparagraphup
\noindent{\bf Metrics.}
We first consider the top-1 accuracy by counting the number of predicted labels exactly matching the ground truth.
Note that we have two types of classes in test datasets, i.e., known and novel classes.
Performances on two types of classes are in trade-off relation, i.e., if one tunes model parameters for favoring novel classes, the accuracy of known classes would be decreased.
Specifically, by adding some positive bias to the novel class scores (e.g., logits in the softmax), one can increase novel class accuracy while decreasing known class accuracy, or vice versa.
Hence, for a fair comparison,
we measure the novel class accuracy with respect to some fixed known class accuracy, e.g., 50\%.
As a more informative evaluation metric,
we also measure the area under known-novel class accuracy curve (AUC).
Varying the novel class score bias, a curve of known class accuracy versus novel class accuracy can be drawn, which depicts the relationship between the known class accuracy and the novel class accuracy.
The AUC is the area under this curve, which is independent of the novel class score bias.

\cutsubsectionup
\subsection{Experimental results} \label{sec:exp}
\cutsubsectiondown

We first compare the hierarchical novelty detection results of the baseline method and our proposed methods qualitatively with test images on ImageNet in Figure~\ref{fig:qual_smp_0}.
We remark that our proposed methods can provide informative prediction results by utilizing the taxonomy of the dataset.
In Figure~\ref{fig:qual_smp_0}~(a), LOO and TD+LOO find the ground truth label (the most fine-grained label in the taxonomy), while DARTS classifies it as ``beagle,'' which is in fact visually similar to ``American foxhound.''
In Figure~\ref{fig:qual_smp_0}~(b), none of the methods find the ground truth, but the prediction of TD+LOO is the most informative, as it is the closest label in the hierarchy.
In Figure~\ref{fig:qual_smp_0}~(c--d), only the prediction of TD+LOO is correct, but the rest of the methods also give a reasonable amount of information.
More qualitative results can be found in \appendixword.

\begin{figure*}[t]
\centering\setlength{\tabcolsep}{0cm}
\begin{tabular}{ccc}
(a) ImageNet & (b) AwA2 & (c) CUB \cr
\includegraphics[width=\fivewidth]{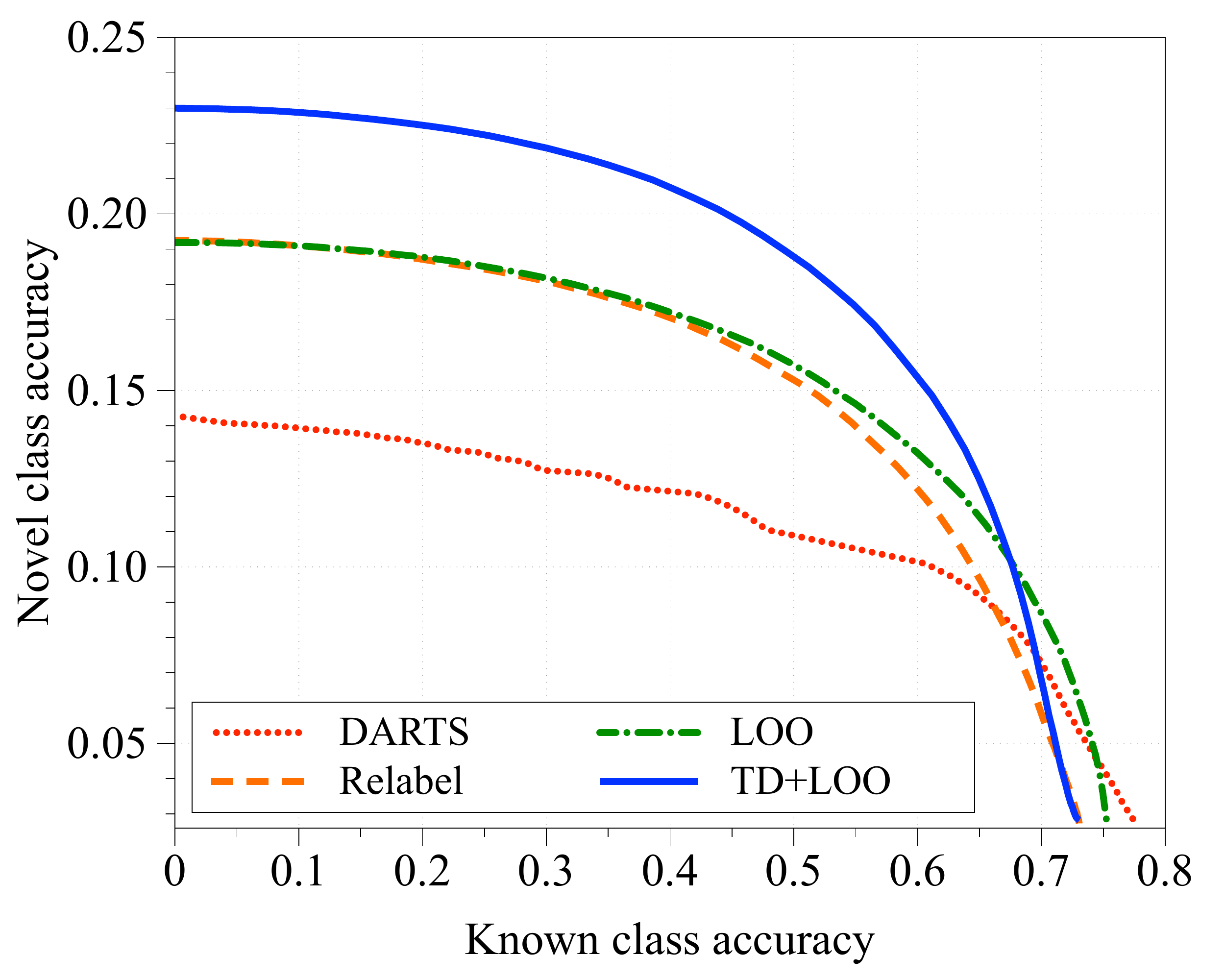} &
\includegraphics[width=\fivewidth]{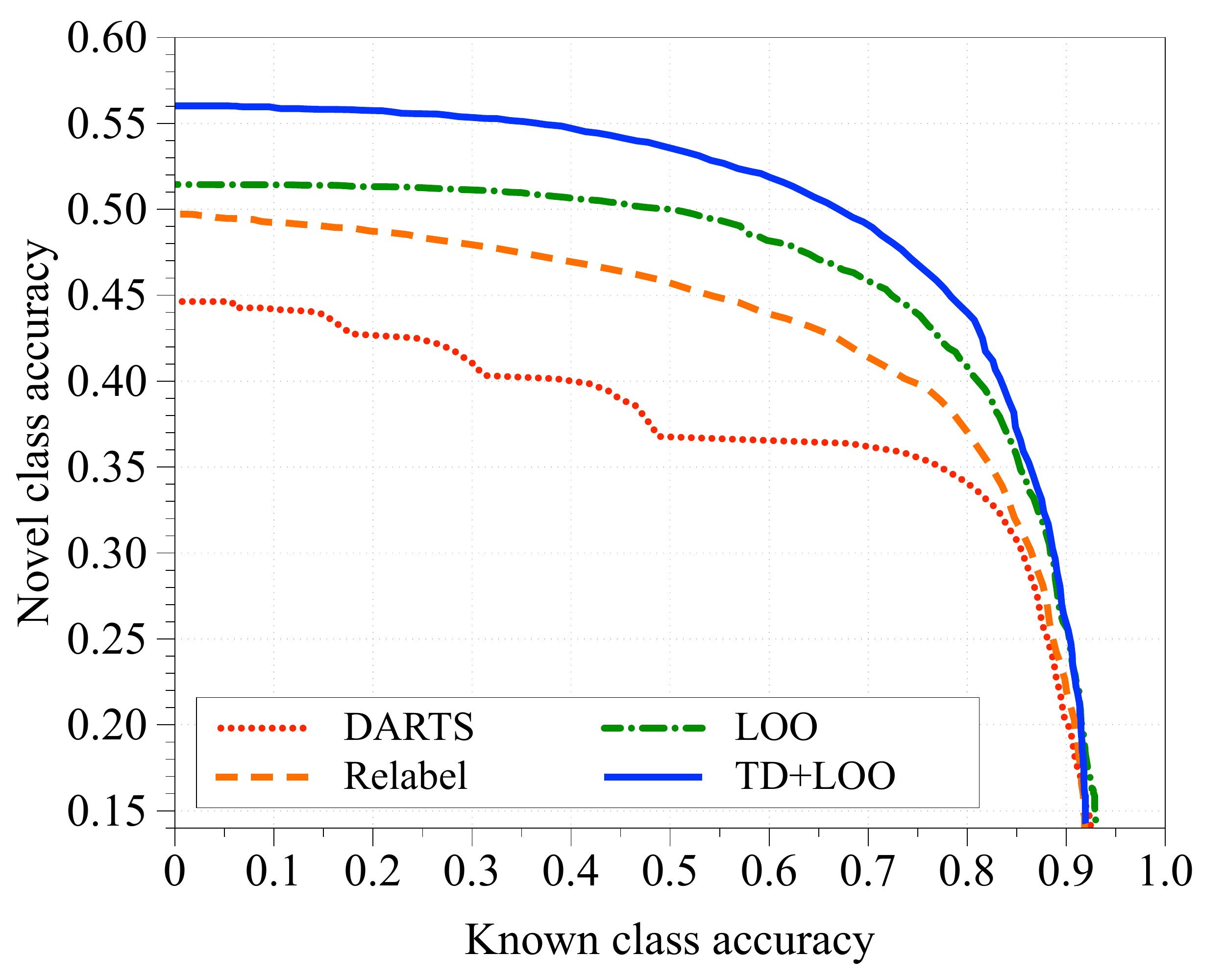} &
\includegraphics[width=\fivewidth]{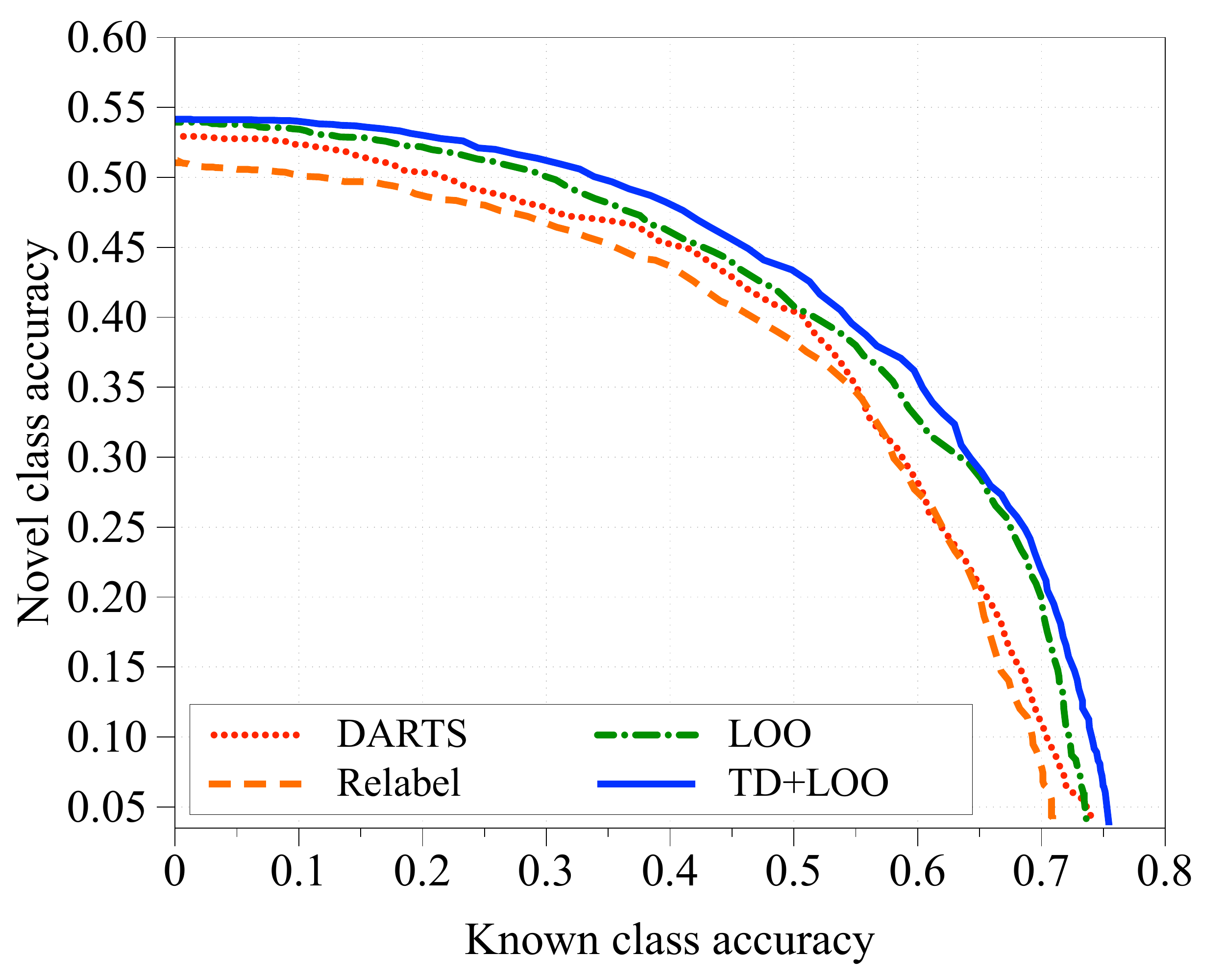} \cr
\end{tabular}
\cutcaptionup
\caption{
Known-novel class accuracy curves obtained by varying the novel class score bias on ImageNet, AwA2, and CUB.
In most regions, our proposed methods outperform the baseline method.
}
\cutcaptiondown
\label{fig:comparison}
\end{figure*}

\begin{table}[t]
\small\centering\setlength{\tabcolsep}{0cm}
\cuthalftablecaptionup
\caption{
Hierarchical novelty detection results on ImageNet, AwA2, and CUB.
For a fair comparison, 50\% of known class accuracy is guaranteed by adding a bias to all novel class scores (logits).
The AUC is obtained by varying the bias.
Known-novel class accuracy curve is shown in Figure~\ref{fig:comparison}.
Values in bold indicate the best performance.
}
\cuthalftablecaptiondown
\cuttableup
\begin{tabular}{
|>{\centering}m{1.8cm}|
|>{\centering}m{1cm}|>{\centering}m{1cm}|
|>{\centering}m{1cm}|>{\centering}m{1cm}|
|>{\centering}m{1cm}|>{\centering}m{1cm}|
}
\hline
\multirow{2}{*}{Method} & \multicolumn{2}{c||}{ImageNet} & \multicolumn{2}{c||}{AwA2} & \multicolumn{2}{c|}{CUB} \cr
\cline{2-7}
 & Novel & AUC & Novel & AUC & Novel & AUC \cr
\hline
\hline
DARTS~\cite{deng2012hedging} & 
10.89 & 8.83 & 
36.75 & 35.14 & 
40.42 & 30.07 \cr
\hline
Relabel & 
15.29 & 11.51 & 
45.71 & 40.28 & 
38.23 & 28.75 \cr
\hline
LOO & 
15.72 & 12.00 & 
50.00 & 43.63 & 
40.78 & 31.92 \cr
\hline
TD+LOO & 
{\bf 18.78} & {\bf 13.98} & 
{\bf 53.57} & {\bf 46.77} & 
{\bf 43.29} & {\bf 33.16} \cr
\hline
\end{tabular}
\cuttabledown
\label{tbl:hnd}
\end{table}

Table~\ref{tbl:hnd} shows the hierarchical novelty detection performance on ImageNet, AwA2, and CUB.
One can note that the proposed methods significantly outperform the baseline method in most cases, except the case of Relabel on CUB, because validation could not find the best relabeling rate for test.
Also, we remark that LOO outperforms Relabel.
The main difference between two methods is that Relabel gives a penalty to the original label if it is relabeled during training, which turns out to be harmful for the performance.
Finally, TD+LOO exhibits the best performance, which implies that the multiple softmax probability vectors extracted from the top-down method are more useful than the vanilla visual features extracted from the state-of-the-art CNNs in the hierarchical novelty detection tasks.
Figure~\ref{fig:comparison} shows the known-novel class accuracy curves by varying the bias added to the novel class scores.
Our proposed methods have higher novel class accuracy than the baseline in most regions.

%% file: qual_smp/qual_smp_0.tex
\begin{figure*}[t]
\footnotesize\centering\setlength{\tabcolsep}{0cm}
\begin{tabular}{
>{\centering}m{1.12cm}>{\centering}m{0.4cm}>{\centering}m{0.4cm}m{2.32cm}
>{\centering}m{1.12cm}>{\centering}m{0.4cm}>{\centering}m{0.4cm}m{2.32cm}
>{\centering}m{1.12cm}>{\centering}m{0.4cm}>{\centering}m{0.4cm}m{2.32cm}
>{\centering}m{1.12cm}>{\centering}m{0.4cm}>{\centering}m{0.4cm}m{2.32cm}
}
\multicolumn{4}{c}{(a)} & 
\multicolumn{4}{c}{(b)} & 
\multicolumn{4}{c}{(c)} & 
\multicolumn{4}{c}{(d)} \cr
\multicolumn{4}{c}{\includegraphics[width=4.24cm, height=3.18cm, keepaspectratio]{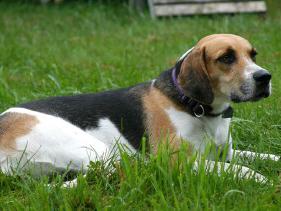}} & 
\multicolumn{4}{c}{\includegraphics[width=4.24cm, height=3.18cm, keepaspectratio]{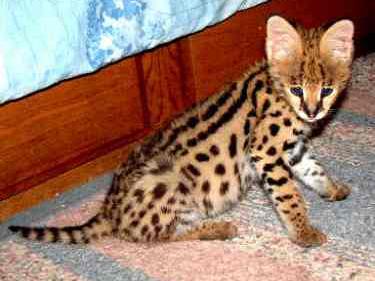}} & 
\multicolumn{4}{c}{\includegraphics[width=4.24cm, height=3.18cm, keepaspectratio]{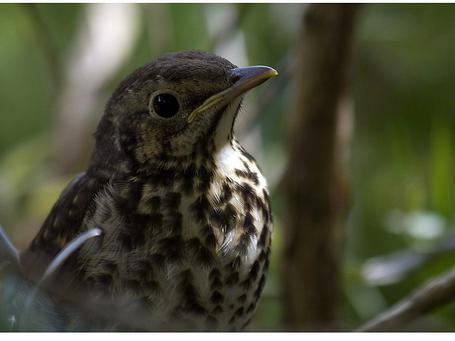}} & 
\multicolumn{4}{c}{\includegraphics[width=4.24cm, height=3.18cm, keepaspectratio]{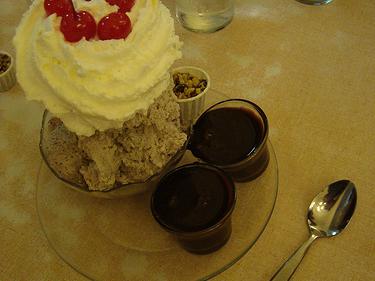}} \cr
\multicolumn{4}{l}{\cellcolor{ColorA} Novel class: American foxhound} & 
\multicolumn{4}{l}{\cellcolor{ColorA} Novel class: serval} & 
\multicolumn{4}{l}{\cellcolor{ColorA} Novel class: song thrush} & 
\multicolumn{4}{l}{\cellcolor{ColorA} Novel class: ice-cream sundae} \cr
\cellcolor{ColorB} Method & \cellcolor{ColorB} $\epsilon$ & \cellcolor{ColorB} A & \multicolumn{1}{c}{\cellcolor{ColorB} Word} & 
\cellcolor{ColorB} Method & \cellcolor{ColorB} $\epsilon$ & \cellcolor{ColorB} A & \multicolumn{1}{c}{\cellcolor{ColorB} Word} & 
\cellcolor{ColorB} Method & \cellcolor{ColorB} $\epsilon$ & \cellcolor{ColorB} A & \multicolumn{1}{c}{\cellcolor{ColorB} Word} & 
\cellcolor{ColorB} Method & \cellcolor{ColorB} $\epsilon$ & \cellcolor{ColorB} A & \multicolumn{1}{c}{\cellcolor{ColorB} Word} \cr
\cellcolor{Color0} GT & \cellcolor{Color0} & \cellcolor{Color0} & \cellcolor{Color0} foxhound & 
\cellcolor{Color0} GT & \cellcolor{Color0} & \cellcolor{Color0} & \cellcolor{Color0} wildcat & 
\cellcolor{Color0} GT & \cellcolor{Color0} & \cellcolor{Color0} & \cellcolor{Color0} thrush & 
\cellcolor{Color0} GT & \cellcolor{Color0} & \cellcolor{Color0} & \cellcolor{Color0} frozen dessert \cr
\cellcolor{Color1} DARTS & \cellcolor{Color1} 2 & \cellcolor{Color1} N & \cellcolor{Color1} beagle & 
\cellcolor{Color1} DARTS & \cellcolor{Color1} 3 & \cellcolor{Color1} N & \cellcolor{Color1} Egyptian cat & 
\cellcolor{Color1} DARTS & \cellcolor{Color1} 3 & \cellcolor{Color1} N & \cellcolor{Color1} hummingbird & 
\cellcolor{Color1} DARTS & \cellcolor{Color1} 4 & \cellcolor{Color1} Y & \cellcolor{Color1} food, nutrient \cr
\cellcolor{Color2} Relabel & \cellcolor{Color2} 1 & \cellcolor{Color2} Y & \cellcolor{Color2} hound dog & 
\cellcolor{Color2} Relabel & \cellcolor{Color2} 2 & \cellcolor{Color2} N & \cellcolor{Color2} domestic cat & 
\cellcolor{Color2} Relabel & \cellcolor{Color2} 2 & \cellcolor{Color2} Y & \cellcolor{Color2} bird & 
\cellcolor{Color2} Relabel & \cellcolor{Color2} 1 & \cellcolor{Color2} N & \cellcolor{Color2} ice cream \cr
\cellcolor{Color3} LOO & \cellcolor{Color3} 0 & \cellcolor{Color3} Y & \cellcolor{Color3} foxhound & 
\cellcolor{Color3} LOO & \cellcolor{Color3} 2 & \cellcolor{Color3} Y & \cellcolor{Color3} feline & 
\cellcolor{Color3} LOO & \cellcolor{Color3} 1 & \cellcolor{Color3} Y & \cellcolor{Color3} oscine bird & 
\cellcolor{Color3} LOO & \cellcolor{Color3} 1 & \cellcolor{Color3} Y & \cellcolor{Color3} dessert \cr
\cellcolor{Color4} TD+LOO & \cellcolor{Color4} 0 & \cellcolor{Color4} Y & \cellcolor{Color4} foxhound & 
\cellcolor{Color4} TD+LOO & \cellcolor{Color4} 1 & \cellcolor{Color4} Y & \cellcolor{Color4} cat & 
\cellcolor{Color4} TD+LOO & \cellcolor{Color4} 0 & \cellcolor{Color4} Y & \cellcolor{Color4} thrush & 
\cellcolor{Color4} TD+LOO & \cellcolor{Color4} 0 & \cellcolor{Color4} Y & \cellcolor{Color4} frozen dessert \cr
\multicolumn{4}{c}{\includegraphics[width=4.24cm, height=3.6cm, keepaspectratio]{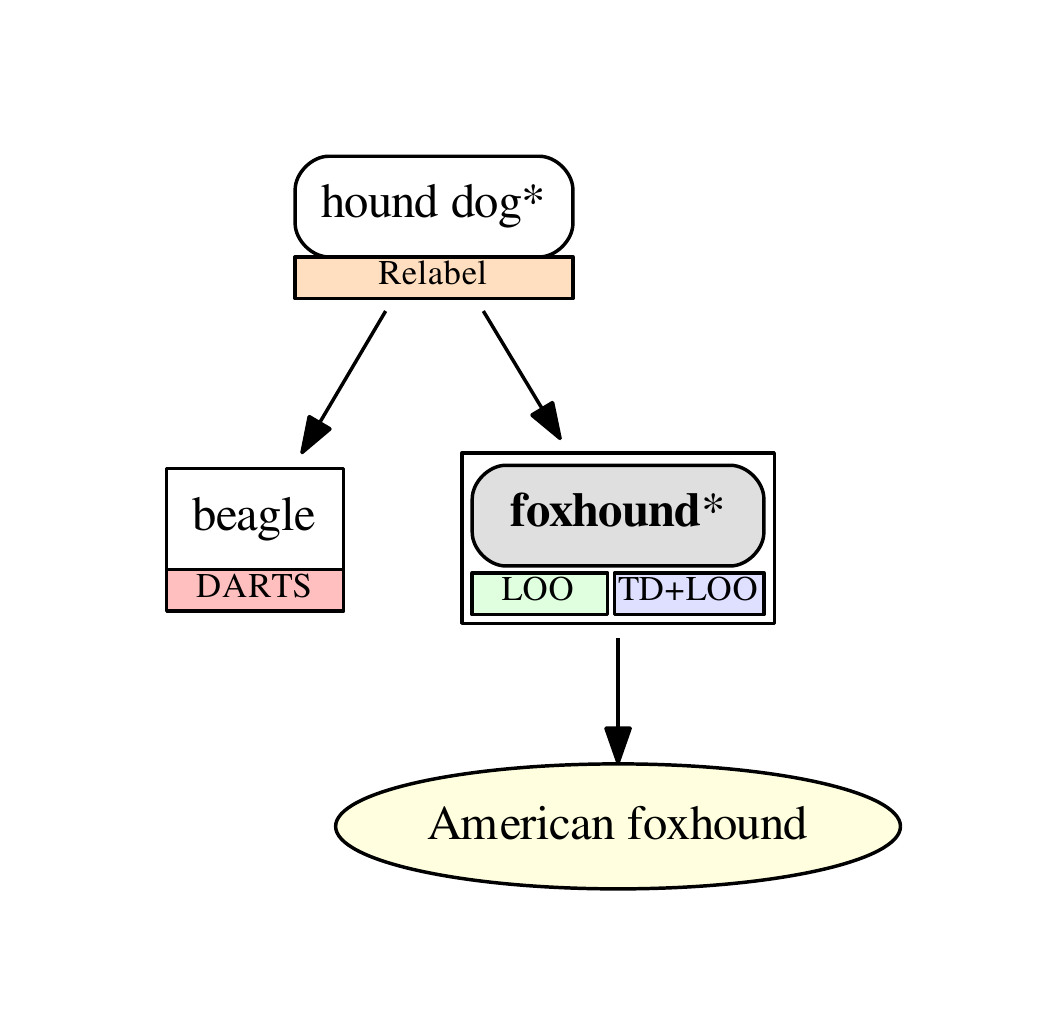}} & 
\multicolumn{4}{c}{\includegraphics[width=4.24cm, height=3.6cm, keepaspectratio]{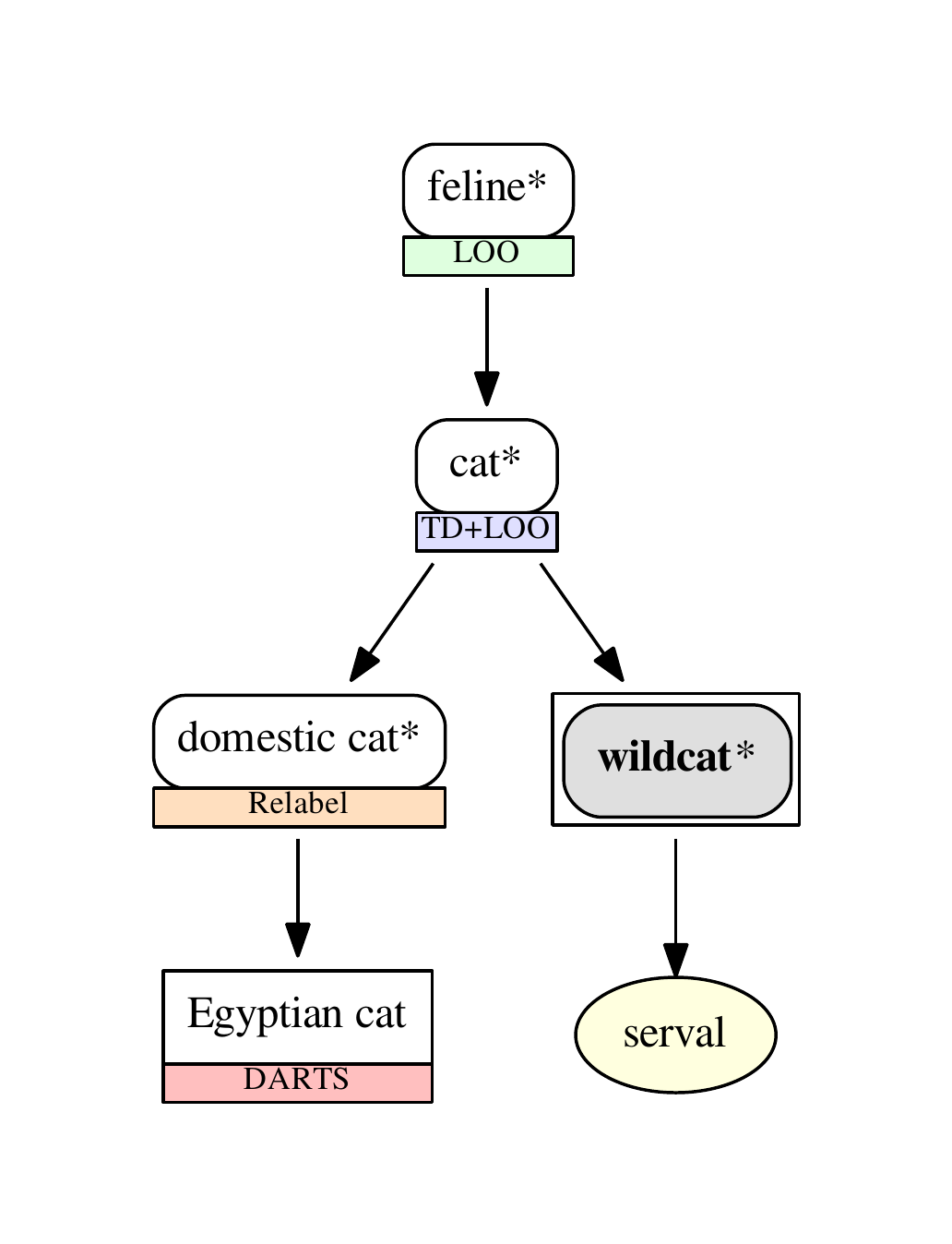}} & 
\multicolumn{4}{c}{\includegraphics[width=4.24cm, height=3.6cm, keepaspectratio]{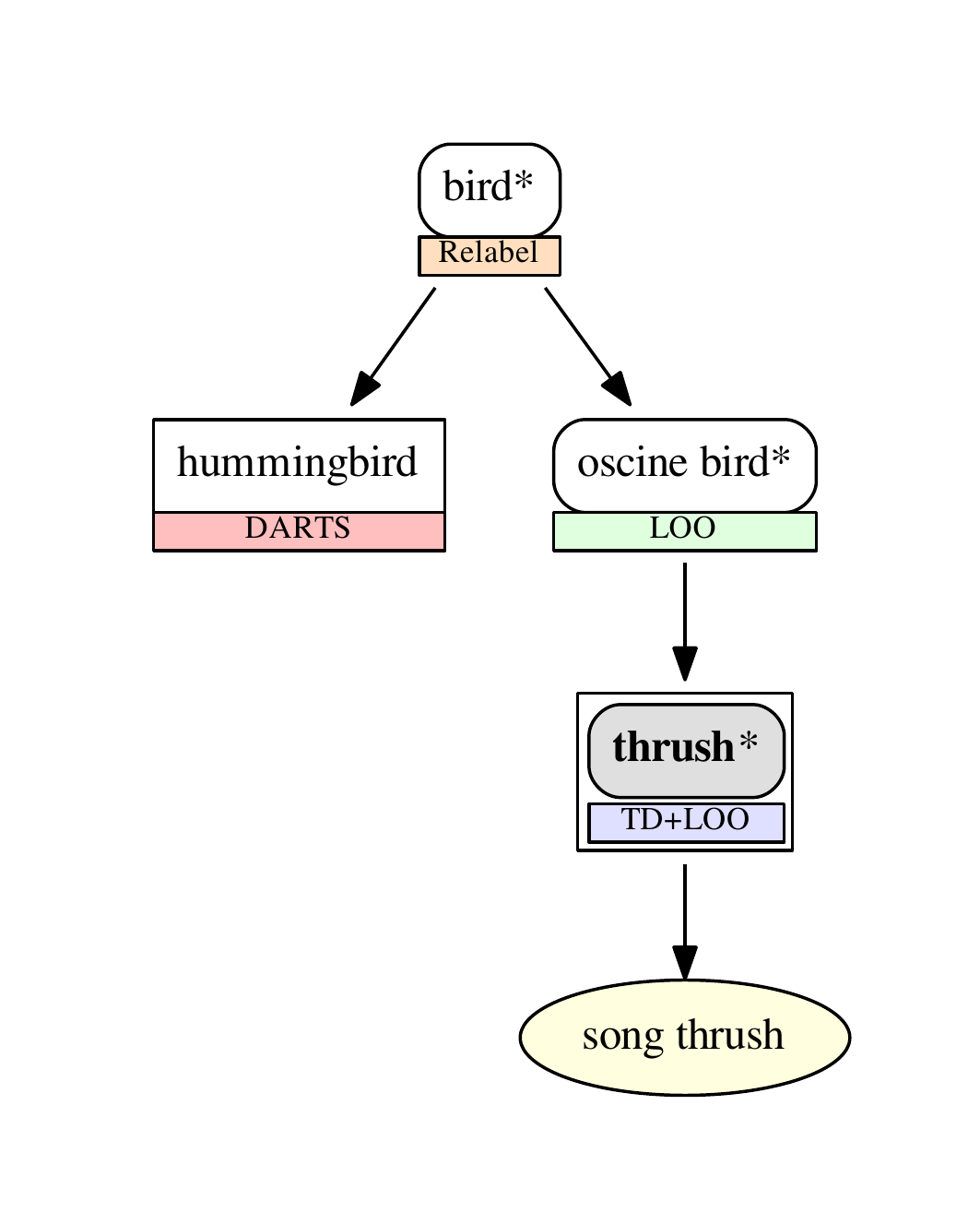}} & 
\multicolumn{4}{c}{\includegraphics[width=4.24cm, height=3.6cm, keepaspectratio]{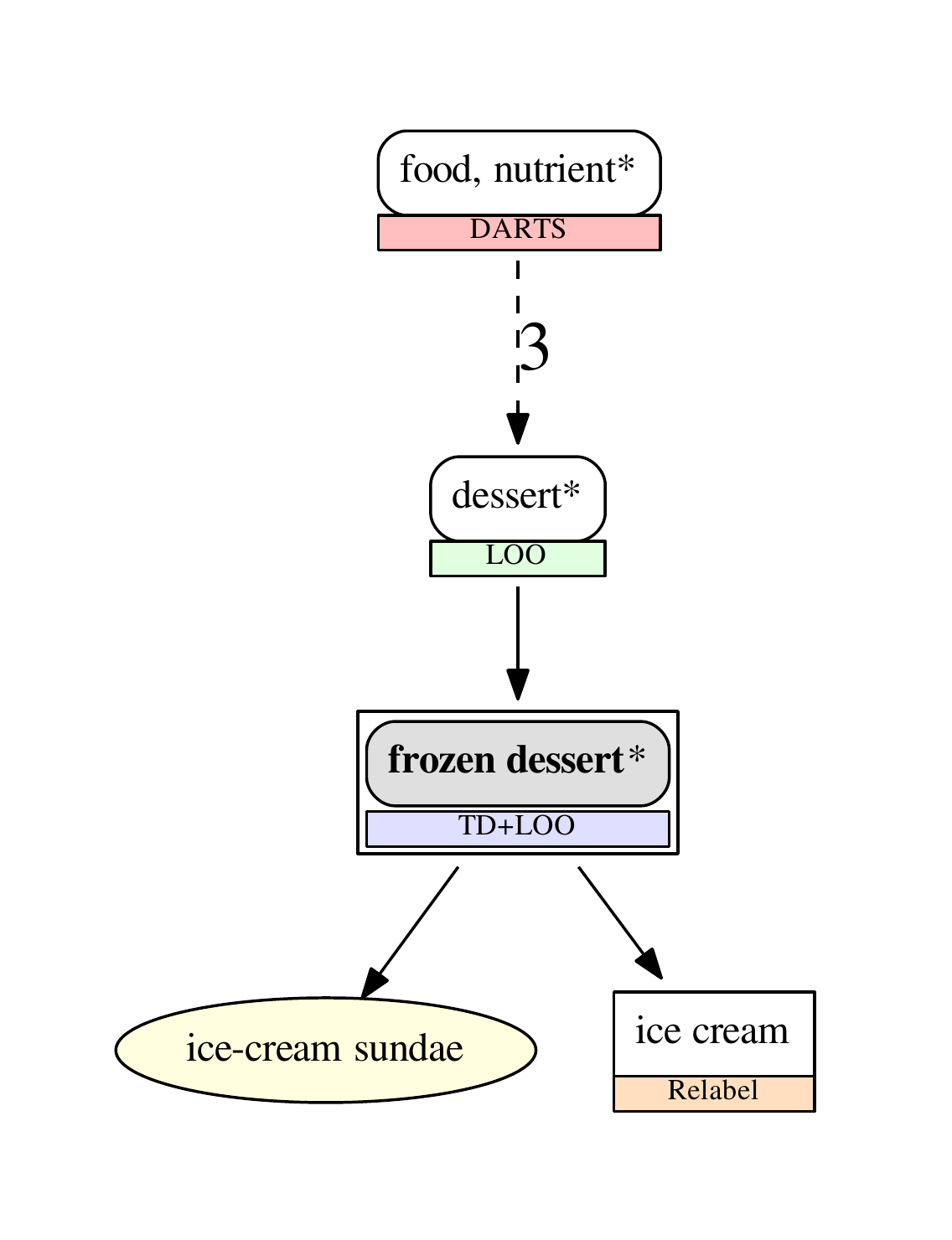}} \cr
\end{tabular}
\vspace{-0.2in}
\caption{Qualitative results of hierarchical novelty detection on ImageNet.
``GT'' is the closest known ancestor of the novel class, which is the expected prediction,
``DARTS'' is the baseline method proposed in \cite{deng2012hedging} where we modify the method for our purpose, and the others are our proposed methods.
``$\epsilon$'' is the distance between the prediction and GT,
``A'' indicates whether the prediction is an ancestor of GT, and
``Word'' is the English word of the predicted label.
Dashed edges represent multi-hop connection, where the number indicates the number of edges between classes.
If the prediction is on a super class (marked with * and rounded), then the test image is classified as a novel class whose closest class in the taxonomy is the super class.
}
\vspace{-0.2in}
\label{fig:qual_smp_0}
\end{figure*}

%% file: 5_zsl.tex
\cutsectionup
\section{Evaluation: Generalized zero-shot learning} \label{sec:zsl}

We present the GZSL performance of the combination of the hierarchical embedding obtained by 
the top-down method 
and other semantic embeddings on Animals with Attributes (AwA1 and AwA2)\footnote{AwA1 is similar to AwA2, but images in AwA1 are no longer available due to the public copyright license issue. We used precomputed CNN features for AwA1, which is available at
\url{http://datasets.d2.mpi-inf.mpg.de/xian/xlsa17.zip}.
} 
~\cite{lampert2013attribute, xian2017zero}
and Caltech-UCSD Birds (CUB)~\cite{welinder2010caltech}.

\cutsubsectionup
\subsection{Evaluation setups}
\cutsubsectiondown

\addparagraphup
\noindent{\bf Hierarchical embeddings for GZSL.}
GZSL requires an output semantic embedding built with side information, e.g., attributes labeled by human, or word embedding trained with a large text corpus.
In addition to those two commonly used semantic embeddings, Akata et al.~\cite{akata2015evaluation} proposed to use hierarchical relationships between all classes, including classes unseen during training.
Specifically, they measured the shortest path distance between classes in the taxonomy built with both known and novel classes, and took the vector of distance values as an output embedding.
We refer to this embedding as Path.

On the other hand,
motivated by the effectiveness of the features extracted from the top-down method shown in Section~\ref{sec:exp},
we set the enumeration of the ideal multiple softmax probability vectors as the semantic embedding:
let $\mathcal{C}(s)[i]$ be the $i$-th child of a super class $s$.
Then, for a label $y$ and a super class $s$, the $i$-th element of an ideal output probability vector $t^{(y,s)} \in [0,1]^{|\mathcal{C}(s)|}$ is
\cutequationup
\begin{align*}
t^{(y,s)}[i] =
\begin{cases}
1 &\text{ if $y$ belongs to $\mathcal{C}(s)[i]$} \\
0 &\text{ if $y$ belongs to $\mathcal{C}(s)[j]$ where $i \neq j$} \\
\frac{1}{|\mathcal{C}(s)|} &\text{ if $y$ is novel or does not belong to $s$}
\end{cases}
\end{align*}
\cutequationdown
where $|\mathcal{C}(s)|$ is the number of known child classes under $s$.
The resultant output embedding is the concatenation of them with respect to the super classes, i.e., the ground truth semantic vector of a class $y$ is $t^y = [\dots, t^{(y,s)}, \dots]$, and we call this embedding TD.
See \appendixword~for an example of the ideal output probability vector $t^{y}$.

Since classes sharing the same closest super class have exactly the same desired output probability vector, we made random guess for fine-grained classification in the experiments only with TD.

\addparagraphup
\noindent{\bf Datasets.}
AwA1 and AwA2~\cite{lampert2013attribute, xian2017zero} consists of 40 seen classes and 10 unseen classes with 37k images, and CUB~\cite{welinder2010caltech} consists of 150 seen classes and 50 unseen classes with 12k images,\footnote{
In GZSL, we have semantic information of unseen classes.
In this sense, although unseen classes are not used for training, they are \emph{known} as such a class-specific semantic information is required.
}
where the taxonomy can be built in the same way with Section~\ref{sec:hnd}.

\begin{figure*}[t]
\centering\setlength{\tabcolsep}{0cm}
\begin{tabular}{ccc}
(a) AwA1 & (b) AwA2 & (c) CUB \cr
\includegraphics[width=\sixwidth]{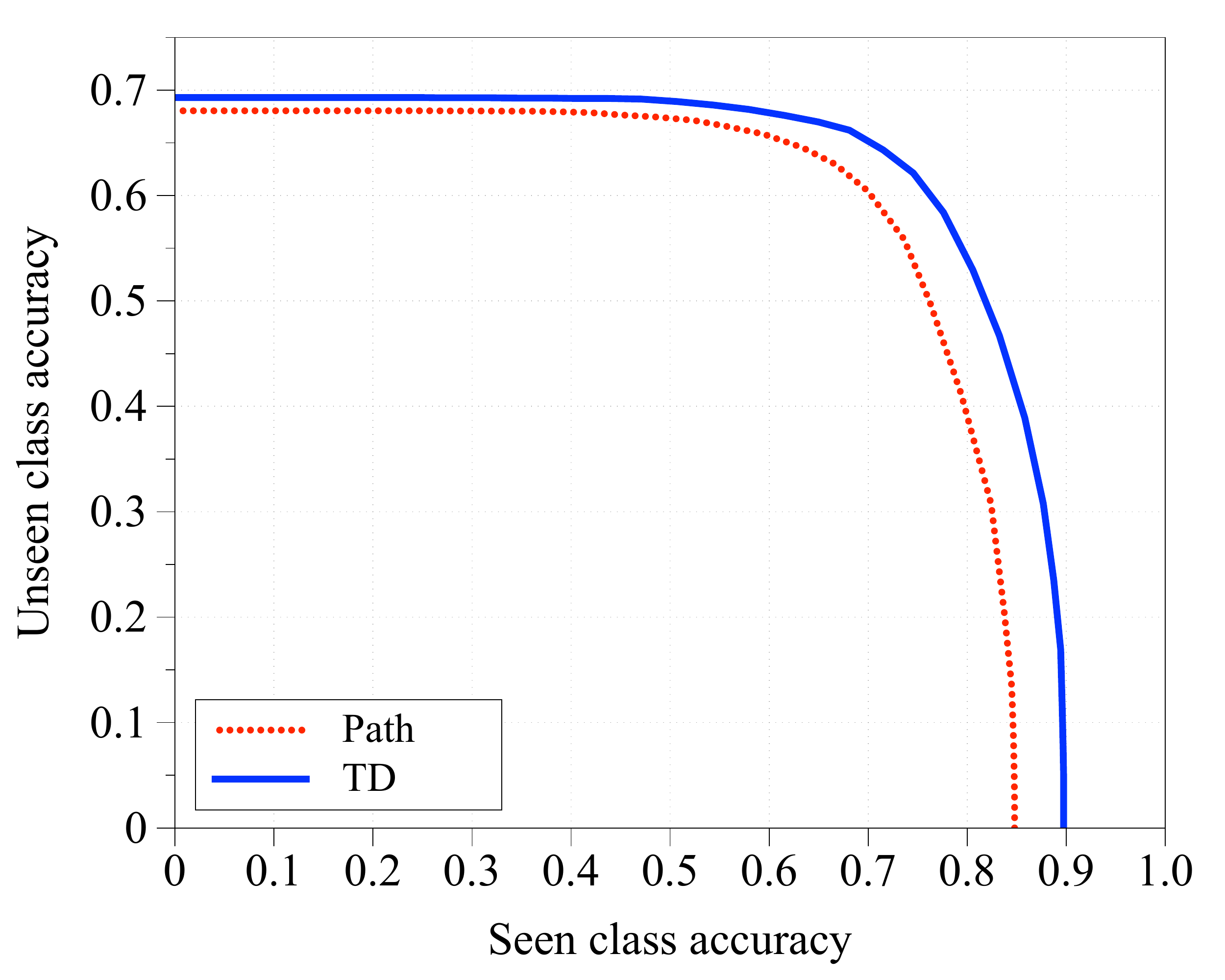} &
\includegraphics[width=\sixwidth]{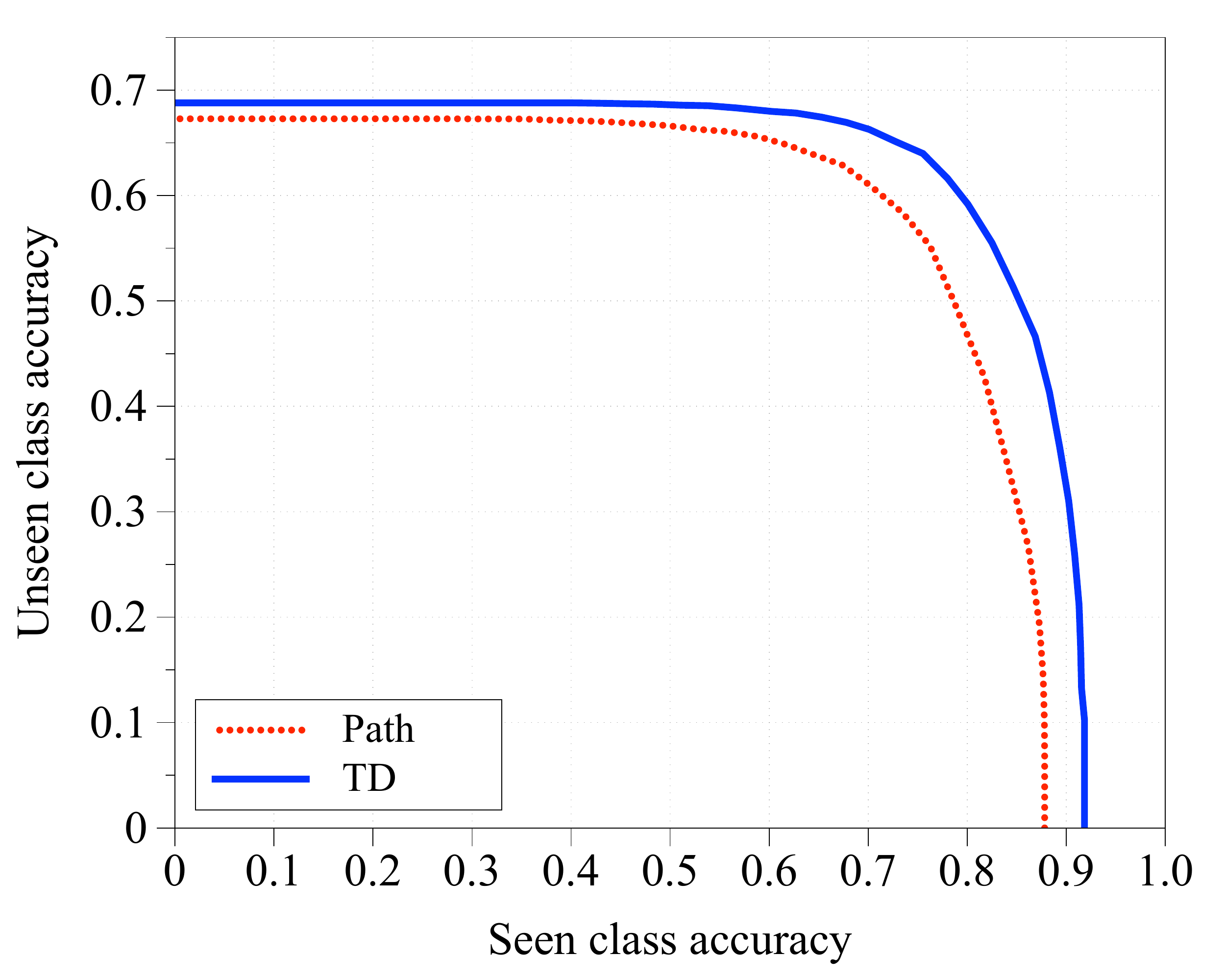} &
\includegraphics[width=\sixwidth]{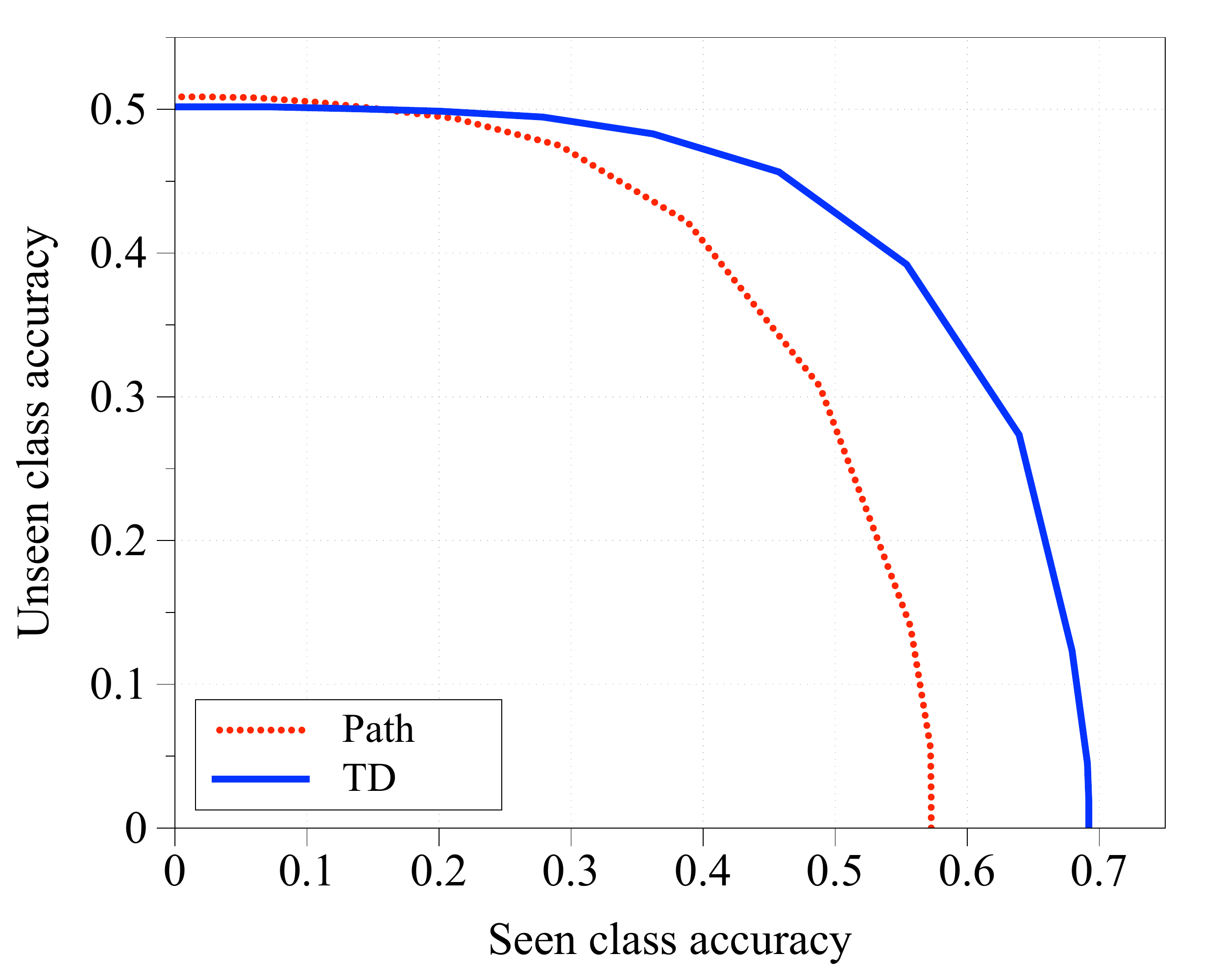} \cr
\end{tabular}
\cutcaptionup
\caption{
Seen-unseen class accuracy curves of the best combined models obtained by varying the unseen class score bias on AwA1, AwA2, and CUB.
``Path'' is the hierarchical embedding proposed in \cite{akata2015evaluation}, and ``TD'' is the embedding of the multiple softmax probability vector obtained from the proposed top-down method.
In most regions, TD outperforms Path.
}
\cutcaptiondown
\label{fig:comparison_zsl}
\end{figure*}

\begin{table}[t]
\small\centering\setlength{\tabcolsep}{0cm}
\cuthalftablecaptionup
\caption{
ZSL and GZSL performance of semantic embedding models and their combinations on AwA1, AwA2, and CUB.
``Att'' stands for continuous attributes labeled by human,
``Word'' stands for word embedding trained with the GloVe objective~\cite{pennington2014glove},
and ``Hier'' stands for the hierarchical embedding, where
``Path'' is proposed in \cite{akata2015evaluation}, and
``TD'' is output of the proposed top-down method.
``Unseen'' is the accuracy when only unseen classes are tested, and 
``AUC'' is the area under the seen-unseen curve 
where the unseen class score bias is varied for computation.
The curve used to obtain AUC is shown in Figure~\ref{fig:comparison_zsl}.
Values in bold indicate the best performance among the combined models.
}
\cuthalftablecaptiondown
\cuttableup
\begin{tabular}{
|>{\centering}m{0.75cm}|>{\centering}m{0.75cm}|>{\centering}m{0.75cm}|
|>{\centering}m{0.95cm}|>{\centering}m{0.95cm}|
|>{\centering}m{0.95cm}|>{\centering}m{0.95cm}|
|>{\centering}m{0.95cm}|>{\centering}m{0.95cm}|
}
\hline
\multicolumn{3}{|c||}{Embedding} & \multicolumn{2}{c||}{AwA1} & \multicolumn{2}{c||}{AwA2} & \multicolumn{2}{c|}{CUB} \cr
\hline
Att & Word & Hier & Unseen & AUC & Unseen & AUC & Unseen & AUC \cr
\hline
\hline
$\checkmark$ & & & 
65.29 & 50.02 & 63.87 & 51.27 & 50.05 & 23.60 \cr
& $\checkmark$ & & 
51.87 & 39.67 & 54.77 & 42.21 & 27.28 & 11.47 \cr
$\checkmark$ & $\checkmark$ & & 
67.80 & 52.84 & 65.76 & 53.18 & 49.83 & 24.13 \cr
\hline
& & Path & 
42.57 & 30.58 & 44.34 & 33.44 & 24.22 & 8.38 \cr
$\checkmark$ & & Path & 
67.09 & 51.45 & 66.58 & 53.50 & 50.25 & 23.70 \cr
& $\checkmark$ & Path & 
52.89 & 40.66 & 55.28 & 42.86 & 27.72 & 11.65 \cr
$\checkmark$ & $\checkmark$ & Path & 
68.04 & 53.21 & 67.28 & 54.31 & {\bf 50.87} & 24.20 \cr
\hline
& & TD & 
33.86 & 25.56 & 31.84 & 24.97 & 13.09 & 7.20 \cr
$\checkmark$ & & TD & 
66.13 & 54.66 & 66.86 & 57.49 & 50.17 & 30.31 \cr
& $\checkmark$ & TD & 
56.14 & 46.28 & 59.67 & 49.39 & 29.05 & 16.73 \cr
$\checkmark$ & $\checkmark$ & TD & 
{\bf 69.23} & {\bf 57.67} & {\bf 68.80} & {\bf 59.24} & 50.17 & {\bf 30.31} \cr
\hline
\end{tabular}
\cuttabledown
\label{tbl:zsl}
\end{table}

\addparagraphup
\noindent{\bf Training.}
We note that the performance of combined models is reported in \cite{akata2015evaluation}, but the numbers are outdated, due to the old CNNs and ZSL models.
Thus, instead of making direct comparison with theirs, we construct the environment following the state-of-the-art setting and compare the performance gain obtained by combining different hierarchical embedding models with other semantic embeddings.
We take ResNet-101 as a pretrained visual feature extractor, and we apply deep embedding model proposed in \cite{zhang2016learning} for training attribute and word embedding models, where it learns to map semantic embeddings to the visual feature embedding with two fully connected layers and ReLU between them.
As a combination strategy, we calculate prediction scores of each model and then use their weighted sum for final decision, where the weights are cross-validated.
See \cite{akata2015evaluation} for more details about the combination strategy.

\addparagraphup
\noindent{\bf Metrics.}
The ZSL performance is measured by testing unseen classes only, and the GZSL performance is measured by the area under seen-unseen curve (AUC) following the idea in \cite{chao2016empirical}.
We measure the class-wise accuracy rather than the sample-wise accuracy to avoid the effect of imbalanced test dataset, as suggested in \cite{xian2017zero}. 

\cutsubsectionup
\subsection{Experimental results}
\cutsubsectiondown

Table~\ref{tbl:zsl} shows the performance of the attribute and word embedding models, and two different hierarchical embedding models, i.e., Path and TD, and their combinations on AwA1, AwA2, and CUB.
In Table~\ref{tbl:zsl}, one can note that the standalone performance of TD is not better than Path, as it does not distinguish unseen classes sharing the same closest super class.
In the same reason, the improvement on ZSL performance with the combined models is fairly small in the combination with TD.
However, in the GZSL task, TD shows significantly better performance in the combined models, which means that our proposed top-down embedding is better when distinguishing both seen classes and unseen classes together.
Compared to the best single semantic embedding model (with attributes), the combination with TD leads to absolute improvement of AUC by 7.65\%, 7.97\%, and 6.71\% on AwA1, AwA2 and CUB, respectively, which is significantly better than that of Path.

%% file: 6_conclusion.tex
\cutsectionup
\section{Conclusion} \label{sec:conclusion}
\cutsectiondown

We propose a new 
hierarchical novelty detection framework, which performs object classification and hierarchical novelty detection by predicting the closest super class in a taxonomy.
We propose several methods for the hierarchical novelty detection task and evaluate their performance. 
In addition, the hierarchical embedding learned with our model can be combined with 
other semantic embeddings such as attributes and words
to improve generalized zero-shot learning performance.
As future work, augmenting textual information about labels for hierarchical novelty detection would be an interesting extension of this work.

%% file: a_hnd.tex
\cutsectionup
\section{More on hierarchical novelty detection} \label{sec:a_hnd}

\cutsubsectionup
\subsection{Details about objectives} \label{sec:a_flatten}
\cutsubsectiondown

We present the exact objective functions and softmax probabilities we propose in the paper.
Let $S(k) = S(y=k|x)$ be an unnormalized softmax score of the $k$-th class (which can be either known or novel), e.g., $S(k) = \exp \left( w_{k}^\top f(x) + b_{k} \right)$, where $f$ is a visual feature extractor.

\addparagraphup
\noindent{\bf Top-down.}
The objective function of the top-down method is
\cutequationup
\begin{align}
\min_{\theta} ~
\mathbb{E}_{Pr(x,y|s)} \left[ -\log Pr (y|x,s; \theta_{s}) \right]
+ \mathbb{E}_{Pr(x,y|\mathcal{O}(s))} \left[ D_{KL} \left( U (\cdot|s) \parallel Pr(\cdot|x,s; \theta_{s}) \right) \right].
\end{align}
\cutequationdown
The softmax probability used in this objective is
\cutequationup
\begin{align*}
Pr (y|x,s; \theta_{s}) = 
\frac{S(y)}
{\sum_{y' \in {\mathcal{C}(s)}} S(y')}.
\end{align*}
\cutequationdown

\addparagraphup
\noindent{\bf Relabel.}
Since super classes in taxonomy have training data by data relabeling, the objective is a standard cross entropy loss over all super and leaf classes:
\cutequationup
\begin{align}
\min_{\theta} ~
\mathbb{E}_{Pr(x,y)} \left[ -\log Pr (y|x; \theta_{\mathcal{T}}) \right].
\end{align}
\cutequationdown
The softmax probability used in this objective is
\cutequationup
\begin{align*}
Pr (y|x; \theta_{\mathcal{T}}) = 
\frac{S(y)}{\sum_{y' \in \mathcal{T}} S(y')} =
\frac{S(y)}
{\sum_{l \in \mathcal{L}(\mathcal{T})} S(l)
+ \sum_{s \in \mathcal{T} \backslash \mathcal{L}(\mathcal{T})} S(\mathcal{N}(s))}.
\end{align*}
\cutequationdown
Here, $\mathcal{T} \backslash \mathcal{L}(\mathcal{T})$ represents all super classes in $\mathcal{T}$.

\addparagraphup
\noindent{\bf LOO.}
We note that there is a notation abuse in the second term of the objective function of LOO for simplity; without notation abuse, the exact objective is
\cutequationup
\begin{align}
\min_{\theta} ~
\mathbb{E}_{Pr(x,y)} \bigg[
-\log Pr (y|x; \theta_{\mathcal{L}(\mathcal{T})})
 + \sum_{a \in \mathcal{A}(y)} -\log Pr (\mathcal{N}(\mathcal{P}(a))|x; {\theta_{\mathcal{N}(\mathcal{P}(a)) \cup \mathcal{L}(\mathcal{T} \backslash a)}}) \bigg].
\end{align}
\cutequationdown
The softmax probabilities are defined as:
\cutequationup
\begin{align*}
Pr (y|x; \theta_{\mathcal{L(T)}})
&= \frac{S(y)}{\sum_{l \in \mathcal{L}(\mathcal{T})} S(l)}, \\
Pr (\mathcal{N}(\mathcal{P}(a))|x; {\theta_{\mathcal{N}(\mathcal{P}(a)) \cup \mathcal{L}(\mathcal{T} \backslash a)}}) &= 
\frac{S(\mathcal{N}(\mathcal{P}(a)))}
{S(\mathcal{N}(\mathcal{P}(a)) +
\sum_{l \in {\mathcal{L}(\mathcal{T} \backslash a)}} S(l)}.
\end{align*}
\cutequationdown

\cutsubsectionup
\subsection{Hyperparameter search} \label{sec:hyper}
\cutsubsectiondown

A difficulty in hierarchical novelty detection is that there are no validation data from novel classes for hyperparameter search.
Similar to the training strategy, we leverage known class data for validation:
specifically, for the top-down method, the novelty detection performance of each classifier is measured with $\mathcal{O}(s)$,
i.e., for each classifier in a super class $s$, known leaf classes that do not belong to $s$ are considered as novel classes.
\cutequationup
\begin{align*}
\hat{y} = 
\begin{cases}
\underset{y'}{\arg\max} ~
Pr(y'|x,s;\theta_{s}) &\text{if } D_{KL} (U(\cdot|s) \parallel Pr(\cdot|x,s;\theta_{s})) \geq \lambda_{s}, \\
\qquad\qquad \mathcal{N}(s) &\text{otherwise,}
\end{cases}
\end{align*}
\cutequationdown
where $\lambda_{s}$ is chosen to maximize the harmonic mean of the known class accuracy and the novelty detection accuracy.
Note that $\lambda_{s}$ can be tuned for each classifier.

For validating flatten methods, we discard logits of ancestors of the label of training data in a hierarchical manner.
Mathematically, at the stage of removal of an ancestor $a \in \mathcal{A}(y)$, we do classification on $\theta_{\mathcal{T} \backslash a}$:
\cutequationup
\begin{align*}
\hat{y} = \underset{y'}{\arg\max} Pr(y'|x; \theta_{\mathcal{T} \backslash a}),
\end{align*}
\cutequationdown
where the ground truth is $\mathcal{N}(\mathcal{P}(a))$ at the stage.
The hyperparameters with the best validation AUC are chosen.

\addparagraphup
\noindent{\bf Model-specific description.}
DARTS has an accuracy guarantee as a hyperparameter.
We took the same candidates in the original paper, \{0\%, 10\%, \dots, 80\%, 85\%, 90\%, 95\%, 99\%\}, and found the best accuracy guarantee, which turned out to be 90\% for ImageNet and CUB, and 99\% for AwA2.
Similarly, for Relabel, we evaluated relabeling rate from 5\% to 95\%, and found that 30\%, 25\%, and 15\% are the best for ImageNet, AwA2, and CUB, respectively.
For the top-down method and LOO, the ratio of two loss terms can be tuned, but the performance was less sensitive to the ratio, so we kept 1:1 ratio.
For TD+LOO, we extracted the multiple softmax probability vectors from the top-down model 
and then trained the LOO.

There are some more strategies to improve the performance:
the proposed losses can be computed in a class-wise manner, i.e., weighted by the number of descendant classes, which is helpful when the taxonomy is highly imbalanced, e.g., ImageNet.
Also, the log of softmax and/or ReLU can be applied to the output of the top-down model.
We note that stacking layers to increase model capacity improves the performance of Relabel, while it does not for LOO.

\subsection{Experimental results on CIFAR-100} \label{sec:cifar}

We provide experimental results on CIFAR-100~\cite{krizhevsky2009learning}.
The compared algorithms are the same with the other experiments, 
and we tune the hyperparameters following the same procedure used for the other datasets described in Section~\ref{sec:hyper}.

\addparagraphup
\noindent{\bf Dataset.} 
The CIFAR-100 dataset~\cite{krizhevsky2009learning} consists of 50k training and 10k test images.
It has 20 super classes containing 5 leaf classes each, so one can naturally define the taxonomy of CIFAR-100 as the rooted tree of height two.
We randomly split the classes into two known leaf classes and three novel classes at each super class, such that we have 40 known leaf classes and 60 novel classes.
To build a validation set, we pick 50 images per known leaf class from the training set.

\addparagraphup
\noindent{\bf Preprocessing.} 
CIFAR-100 images have smaller size than natural images in other datasets, so we first train a shallower network, ResNet-18 with 40 known leaf classes.
Pretraining is done with only training images, without any information about novel classes.
And then, the last fully connected layer of the CNNs is replaced with our proposed methods.
We use 100 training data per batch.
As a regularization, L2 norm weight decay with parameter $10^{-2}$ is applied.
The initial learning rate is $10^{-2}$ and it decays at most two times when loss improvement is less than 2\% compared to the last epoch.

\addparagraphup
\noindent{\bf Experimental results.}
Table~\ref{tbl:cifar100} compares the baseline and our proposed methods.
One can note that the proposed methods outperform the baseline in both novel class accuracy and AUC.
However, unlike the results on the other datasets, TD+LOO does not outperform the vanilla LOO method, as one can expect that the vectors extracted from the top-down method might not be useful in the case of CIFAR-100 since its taxonomy is too simple and thus not informative.

\begin{table}[h]
\centering\setlength{\tabcolsep}{0cm}
\cuttablecaptionup
\caption{
Hierarchical novelty detection results on CIFAR-100.
For a fair comparison, 50\% of known class accuracy is guaranteed by adding a bias to all novel class scores (logits).
The AUC is obtained by varying the bias.
}
\cuttablecaptiondown
\cuttableup
\begin{tabular}{
|>{\centering}m{1.8cm}|>{\centering}m{1.65cm}|>{\centering}m{1.65cm}|
}
\hline
Method & Novel & AUC \cr
\hline
\hline
DARTS~\cite{deng2012hedging} & 22.38 & 17.84 \cr
\hline
Relabel & 22.58 & 18.31 \cr
\hline
LOO & {\bf 23.68} & {\bf 18.93} \cr
\hline
TD+LOO & 22.79 &  18.54 \cr
\hline
\end{tabular}
\cuttabledown
\label{tbl:cifar100}
\end{table}

%% file: b_qual_smp.tex
\cutsectionup
\section{Sample-wise qualitative results} \label{sec:qual_smp}
\cutsectiondown

In this section, we show sample-wise qualitative results on ImageNet.
We compared four different methods:
DARTS~\cite{deng2012hedging} is the baseline method where we modify the method for our purpose, and the others, Relabel, LOO, and TD+LOO, are our proposed methods.
In Figure~\ref{fig:qual_smp_1}--\ref{fig:qual_smp_8}, 
we put each test image at the top, 
a table of the classification results in the middle, 
and a sub-taxonomy representing the hierarchical relationship between classes appeared in the classification results at the bottom.
In tables, we provide the true label of the test image at the first row, which is either a novel class or a known leaf class.
In the ``Method'' column in tables, ``GT'' is the ground truth label for hierarchical novelty detection:
if the true label of the test image is a novel class, ``GT'' is the closest known ancestor of the novel class, which is the expected prediction;
otherwise, ``GT'' is the true label of the test image.
Each method has its own background color in both tables and sub-taxonomies.
In sub-taxonomies, the novel class is shown in ellipse shape if exists, GT is double-lined, and the name of the methods is displayed below its prediction.
Dashed edges represent multi-hop connection, where the number indicates the number of edges between classes:
for example, a dashed edge labeled with 3 implies that two classes exist in the middle of the connection.
Note that some novel classes have multiple ground truth labels if they have multiple paths to the taxonomy.

Figure~\ref{fig:qual_smp_1}--\ref{fig:qual_smp_2} show the hierarchical novelty detection results of known leaf classes, and
Figure~\ref{fig:qual_smp_3}--\ref{fig:qual_smp_8} show that of novel classes.
In general, while DARTS tends to produce a coarse-grained label, our proposed models try to find a fine-grained label.
In most cases, our prediction is not too far from the ground truth. 

\input{qual_smp/qual_smp_1}
\input{qual_smp/qual_smp_2}
\input{qual_smp/qual_smp_3}
\input{qual_smp/qual_smp_4}
\input{qual_smp/qual_smp_5}
\input{qual_smp/qual_smp_6}
\input{qual_smp/qual_smp_7}
\input{qual_smp/qual_smp_8}

%% file: qual_smp/qual_smp_1.tex
\begin{figure*}[t]
\footnotesize\centering\setlength{\tabcolsep}{0cm}
\begin{tabular}{
>{\centering}m{1.12cm}>{\centering}m{0.4cm}>{\centering}m{0.4cm}m{2.32cm}
>{\centering}m{1.12cm}>{\centering}m{0.4cm}>{\centering}m{0.4cm}m{2.32cm}
>{\centering}m{1.12cm}>{\centering}m{0.4cm}>{\centering}m{0.4cm}m{2.32cm}
>{\centering}m{1.12cm}>{\centering}m{0.4cm}>{\centering}m{0.4cm}m{2.32cm}
}
\multicolumn{4}{c}{(a)} & 
\multicolumn{4}{c}{(b)} & 
\multicolumn{4}{c}{(c)} & 
\multicolumn{4}{c}{(d)} \cr
\multicolumn{4}{c}{\includegraphics[width=4.24cm, height=3.18cm, keepaspectratio]{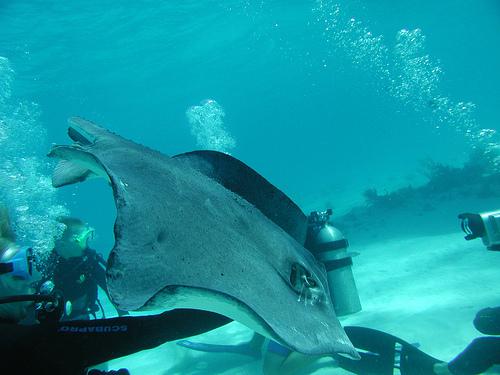}} & 
\multicolumn{4}{c}{\includegraphics[width=4.24cm, height=3.18cm, keepaspectratio]{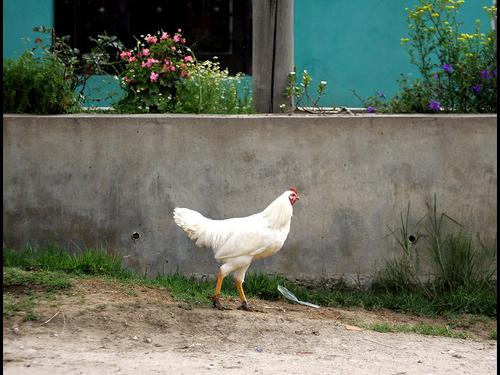}} & 
\multicolumn{4}{c}{\includegraphics[width=4.24cm, height=3.18cm, keepaspectratio]{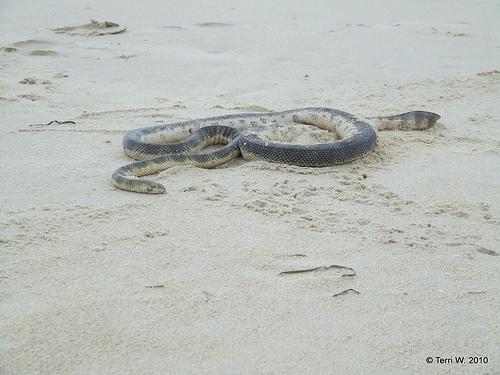}} & 
\multicolumn{4}{c}{\includegraphics[width=4.24cm, height=3.18cm, keepaspectratio]{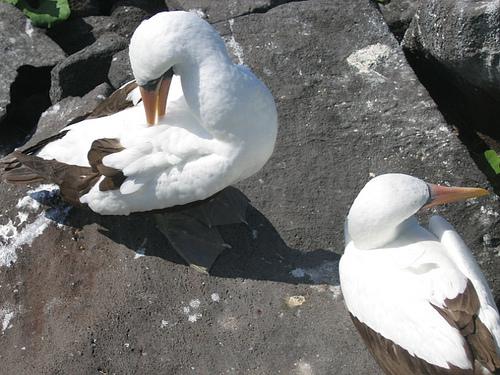}} \cr
\multicolumn{4}{l}{\cellcolor{ColorA} Known class: stingray} & 
\multicolumn{4}{l}{\cellcolor{ColorA} Known class: hen} & 
\multicolumn{4}{l}{\cellcolor{ColorA} Known class: sea snake} & 
\multicolumn{4}{l}{\cellcolor{ColorA} Known class: albatross} \cr
\cellcolor{ColorB} Method & \cellcolor{ColorB} $\epsilon$ & \cellcolor{ColorB} A & \multicolumn{1}{c}{\cellcolor{ColorB} Word} & 
\cellcolor{ColorB} Method & \cellcolor{ColorB} $\epsilon$ & \cellcolor{ColorB} A & \multicolumn{1}{c}{\cellcolor{ColorB} Word} & 
\cellcolor{ColorB} Method & \cellcolor{ColorB} $\epsilon$ & \cellcolor{ColorB} A & \multicolumn{1}{c}{\cellcolor{ColorB} Word} & 
\cellcolor{ColorB} Method & \cellcolor{ColorB} $\epsilon$ & \cellcolor{ColorB} A & \multicolumn{1}{c}{\cellcolor{ColorB} Word} \cr
\cellcolor{Color0} GT & \cellcolor{Color0} & \cellcolor{Color0} & \cellcolor{Color0} stingray & 
\cellcolor{Color0} GT & \cellcolor{Color0} & \cellcolor{Color0} & \cellcolor{Color0} hen & 
\cellcolor{Color0} GT & \cellcolor{Color0} & \cellcolor{Color0} & \cellcolor{Color0} sea snake & 
\cellcolor{Color0} GT & \cellcolor{Color0} & \cellcolor{Color0} & \cellcolor{Color0} albatross \cr
\cellcolor{Color1} DARTS & \cellcolor{Color1} 1 & \cellcolor{Color1} Y & \cellcolor{Color1} ray & 
\cellcolor{Color1} DARTS & \cellcolor{Color1} 2 & \cellcolor{Color1} N & \cellcolor{Color1} cock & 
\cellcolor{Color1} DARTS & \cellcolor{Color1} 1 & \cellcolor{Color1} Y & \cellcolor{Color1} snake & 
\cellcolor{Color1} DARTS & \cellcolor{Color1} 2 & \cellcolor{Color1} Y & \cellcolor{Color1} aquatic bird \cr
\cellcolor{Color2} Relabel & \cellcolor{Color2} 4 & \cellcolor{Color2} N & \cellcolor{Color2} tiger shark & 
\cellcolor{Color2} Relabel & \cellcolor{Color2} 1 & \cellcolor{Color2} Y & \cellcolor{Color2} bird & 
\cellcolor{Color2} Relabel & \cellcolor{Color2} 2 & \cellcolor{Color2} N & \cellcolor{Color2} colubrid snake & 
\cellcolor{Color2} Relabel & \cellcolor{Color2} 1 & \cellcolor{Color2} Y & \cellcolor{Color2} seabird \cr
\cellcolor{Color3} LOO & \cellcolor{Color3} 2 & \cellcolor{Color3} Y & \cellcolor{Color3} elasmobranch & 
\cellcolor{Color3} LOO & \cellcolor{Color3} 0 & \cellcolor{Color3} Y & \cellcolor{Color3} hen & 
\cellcolor{Color3} LOO & \cellcolor{Color3} 1 & \cellcolor{Color3} Y & \cellcolor{Color3} snake & 
\cellcolor{Color3} LOO & \cellcolor{Color3} 1 & \cellcolor{Color3} Y & \cellcolor{Color3} seabird \cr
\cellcolor{Color4} TD+LOO & \cellcolor{Color4} 1 & \cellcolor{Color4} Y & \cellcolor{Color4} ray & 
\cellcolor{Color4} TD+LOO & \cellcolor{Color4} 0 & \cellcolor{Color4} Y & \cellcolor{Color4} hen & 
\cellcolor{Color4} TD+LOO & \cellcolor{Color4} 1 & \cellcolor{Color4} Y & \cellcolor{Color4} snake & 
\cellcolor{Color4} TD+LOO & \cellcolor{Color4} 0 & \cellcolor{Color4} Y & \cellcolor{Color4} albatross \cr
\multicolumn{4}{c}{\includegraphics[width=4.24cm, height=3.6cm, keepaspectratio]{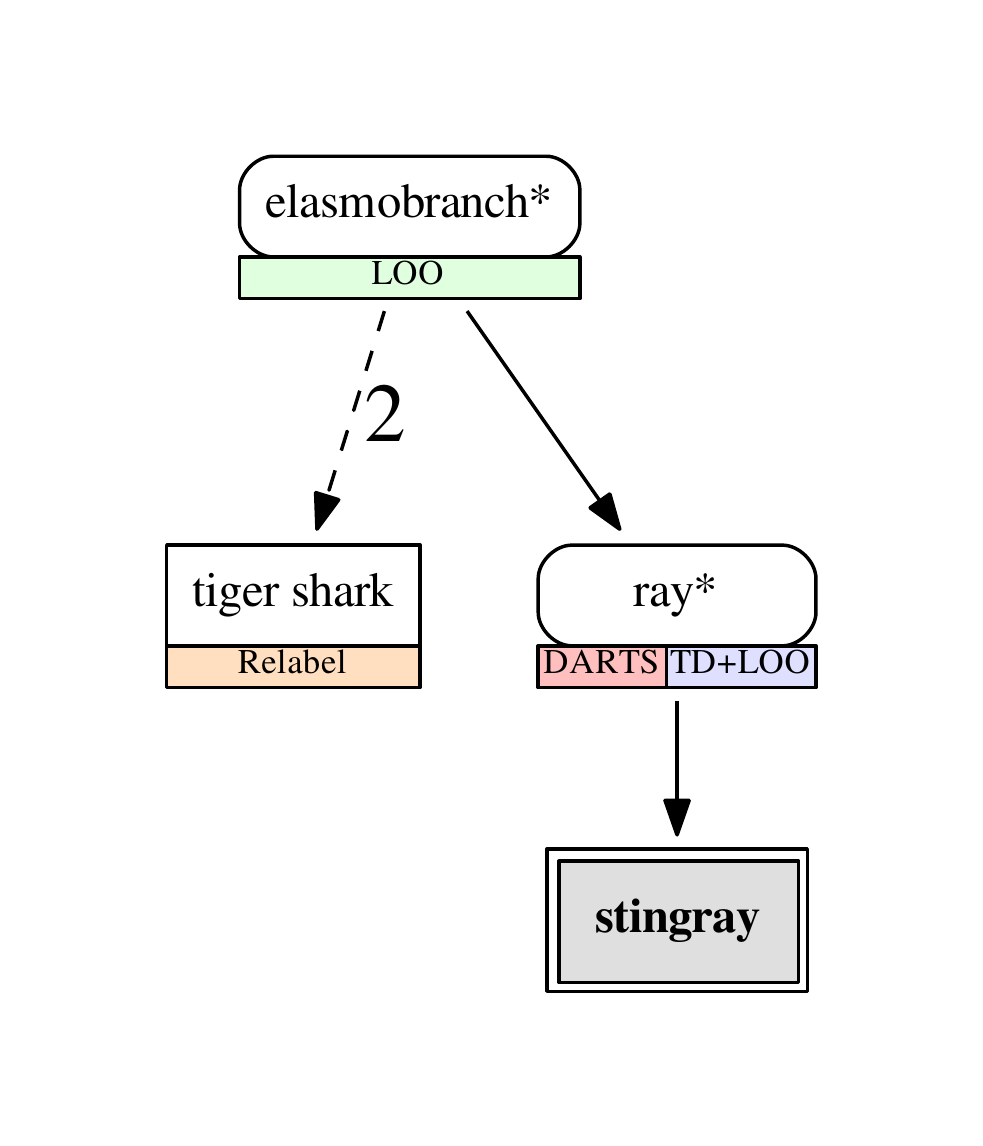}} & 
\multicolumn{4}{c}{\includegraphics[width=4.24cm, height=3.6cm, keepaspectratio]{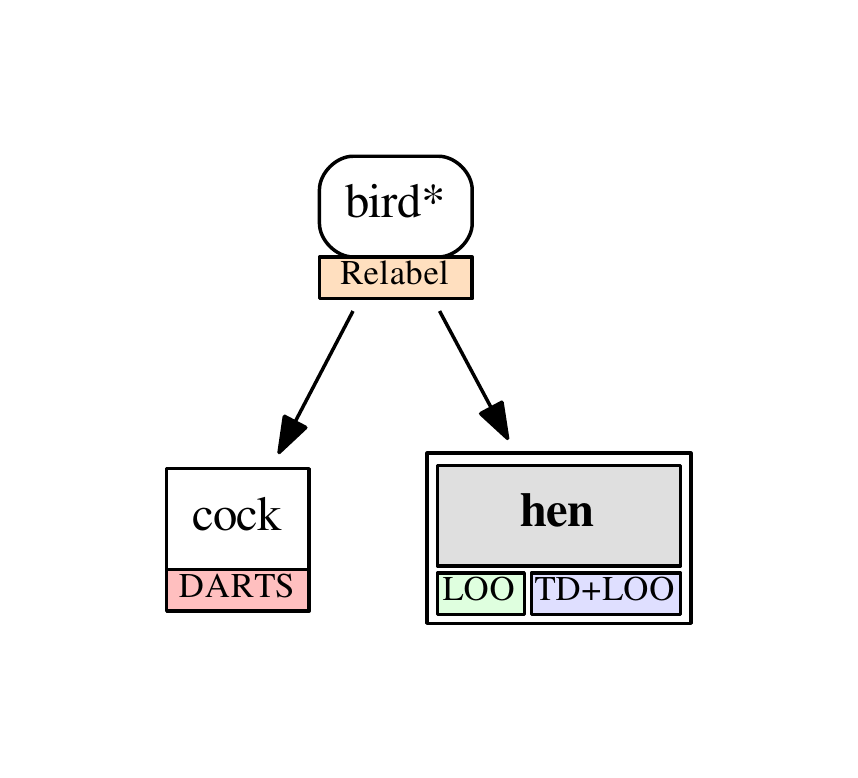}} & 
\multicolumn{4}{c}{\includegraphics[width=4.24cm, height=3.6cm, keepaspectratio]{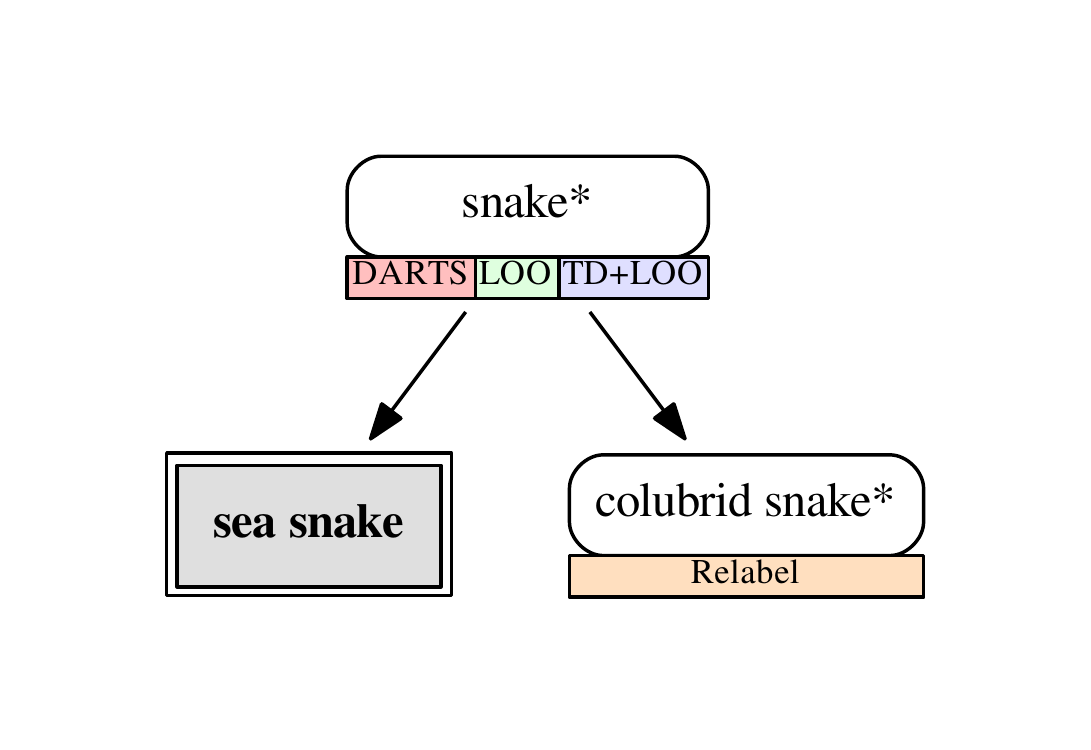}} & 
\multicolumn{4}{c}{\includegraphics[width=4.24cm, height=3.6cm, keepaspectratio]{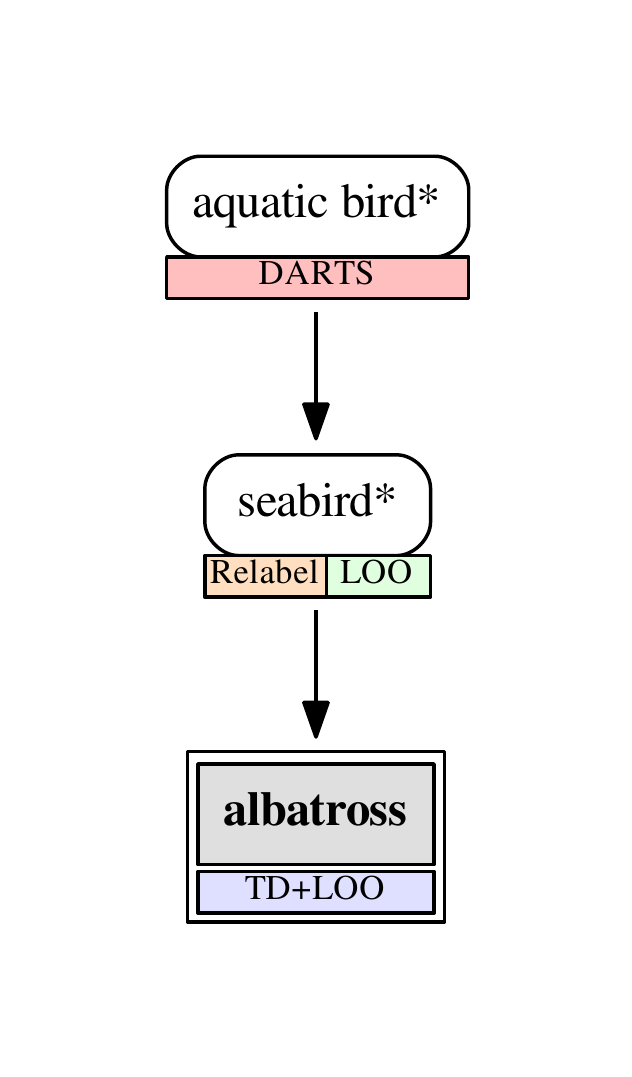}} \cr
\multicolumn{4}{c}{(e)} & 
\multicolumn{4}{c}{(f)} & 
\multicolumn{4}{c}{(g)} & 
\multicolumn{4}{c}{(h)} \cr
\multicolumn{4}{c}{\includegraphics[width=4.24cm, height=3.18cm, keepaspectratio]{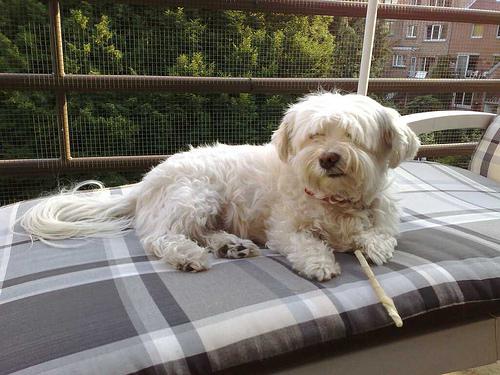}} & 
\multicolumn{4}{c}{\includegraphics[width=4.24cm, height=3.18cm, keepaspectratio]{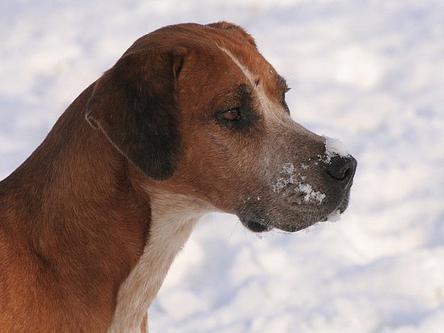}} & 
\multicolumn{4}{c}{\includegraphics[width=4.24cm, height=3.18cm, keepaspectratio]{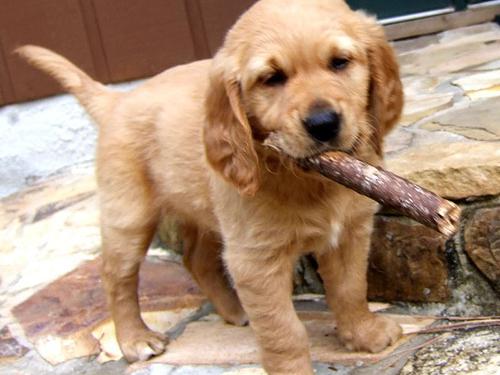}} & 
\multicolumn{4}{c}{\includegraphics[width=4.24cm, height=3.18cm, keepaspectratio]{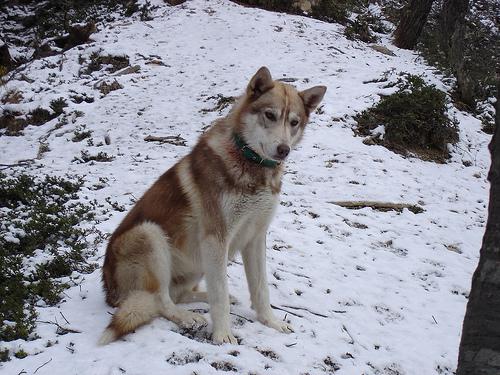}} \cr
\multicolumn{4}{l}{\cellcolor{ColorA} Known class: Maltese dog} & 
\multicolumn{4}{l}{\cellcolor{ColorA} Known class: English foxhound} & 
\multicolumn{4}{l}{\cellcolor{ColorA} Known class: golden retriever} & 
\multicolumn{4}{l}{\cellcolor{ColorA} Known class: Siberian husky} \cr
\cellcolor{ColorB} Method & \cellcolor{ColorB} $\epsilon$ & \cellcolor{ColorB} A & \multicolumn{1}{c}{\cellcolor{ColorB} Word} & 
\cellcolor{ColorB} Method & \cellcolor{ColorB} $\epsilon$ & \cellcolor{ColorB} A & \multicolumn{1}{c}{\cellcolor{ColorB} Word} & 
\cellcolor{ColorB} Method & \cellcolor{ColorB} $\epsilon$ & \cellcolor{ColorB} A & \multicolumn{1}{c}{\cellcolor{ColorB} Word} & 
\cellcolor{ColorB} Method & \cellcolor{ColorB} $\epsilon$ & \cellcolor{ColorB} A & \multicolumn{1}{c}{\cellcolor{ColorB} Word} \cr
\cellcolor{Color0} GT & \cellcolor{Color0} & \cellcolor{Color0} & \cellcolor{Color0} Maltese dog & 
\cellcolor{Color0} GT & \cellcolor{Color0} & \cellcolor{Color0} & \cellcolor{Color0} English foxhound & 
\cellcolor{Color0} GT & \cellcolor{Color0} & \cellcolor{Color0} & \cellcolor{Color0} golden retriever & 
\cellcolor{Color0} GT & \cellcolor{Color0} & \cellcolor{Color0} & \cellcolor{Color0} Siberian husky \cr
\cellcolor{Color1} DARTS & \cellcolor{Color1} 5 & \cellcolor{Color1} N & \cellcolor{Color1} Tibetan terrier & 
\cellcolor{Color1} DARTS & \cellcolor{Color1} 4 & \cellcolor{Color1} N & \cellcolor{Color1} {\scriptsize Rhodesian ridgeback} & 
\cellcolor{Color1} DARTS & \cellcolor{Color1} 2 & \cellcolor{Color1} Y & \cellcolor{Color1} sporting dog & 
\cellcolor{Color1} DARTS & \cellcolor{Color1} 2 & \cellcolor{Color1} Y & \cellcolor{Color1} working dog \cr
\cellcolor{Color2} Relabel & \cellcolor{Color2} 4 & \cellcolor{Color2} N & \cellcolor{Color2} terrier & 
\cellcolor{Color2} Relabel & \cellcolor{Color2} 3 & \cellcolor{Color2} Y & \cellcolor{Color2} hunting dog & 
\cellcolor{Color2} Relabel & \cellcolor{Color2} 1 & \cellcolor{Color2} Y & \cellcolor{Color2} retriever & 
\cellcolor{Color2} Relabel & \cellcolor{Color2} 3 & \cellcolor{Color2} N & \cellcolor{Color2} Eskimo dog \cr
\cellcolor{Color3} LOO & \cellcolor{Color3} 0 & \cellcolor{Color3} Y & \cellcolor{Color3} Maltese dog & 
\cellcolor{Color3} LOO & \cellcolor{Color3} 1 & \cellcolor{Color3} Y & \cellcolor{Color3} foxhound & 
\cellcolor{Color3} LOO & \cellcolor{Color3} 0 & \cellcolor{Color3} Y & \cellcolor{Color3} golden retriever & 
\cellcolor{Color3} LOO & \cellcolor{Color3} 1 & \cellcolor{Color3} Y & \cellcolor{Color3} sled dog \cr
\cellcolor{Color4} TD+LOO & \cellcolor{Color4} 0 & \cellcolor{Color4} Y & \cellcolor{Color4} Maltese dog & 
\cellcolor{Color4} TD+LOO & \cellcolor{Color4} 1 & \cellcolor{Color4} Y & \cellcolor{Color4} foxhound & 
\cellcolor{Color4} TD+LOO & \cellcolor{Color4} 0 & \cellcolor{Color4} Y & \cellcolor{Color4} golden retriever & 
\cellcolor{Color4} TD+LOO & \cellcolor{Color4} 1 & \cellcolor{Color4} Y & \cellcolor{Color4} sled dog \cr
\multicolumn{4}{c}{\includegraphics[width=4.24cm, height=3.6cm, keepaspectratio]{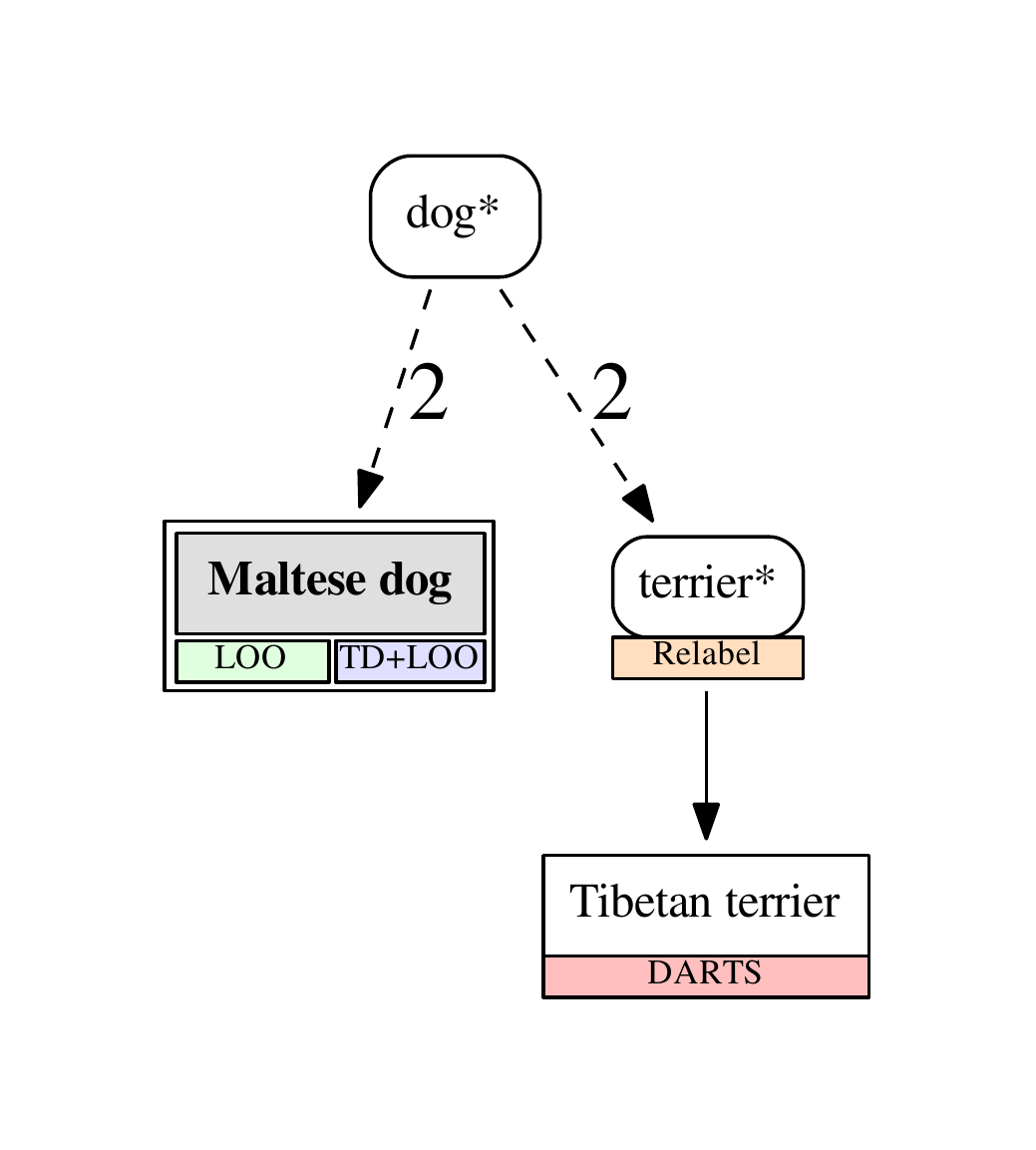}} & 
\multicolumn{4}{c}{\includegraphics[width=4.24cm, height=3.6cm, keepaspectratio]{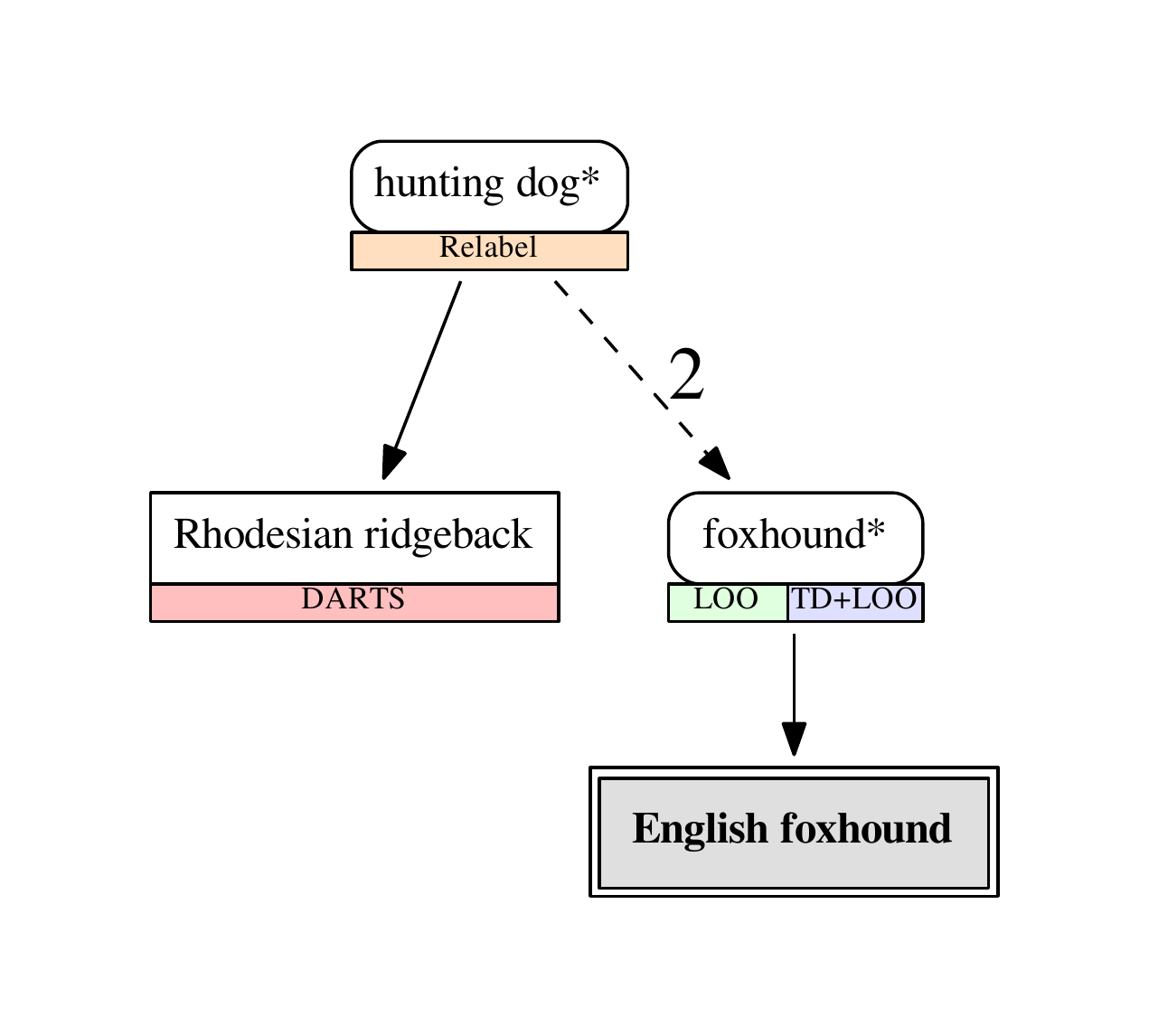}} & 
\multicolumn{4}{c}{\includegraphics[width=4.24cm, height=3.6cm, keepaspectratio]{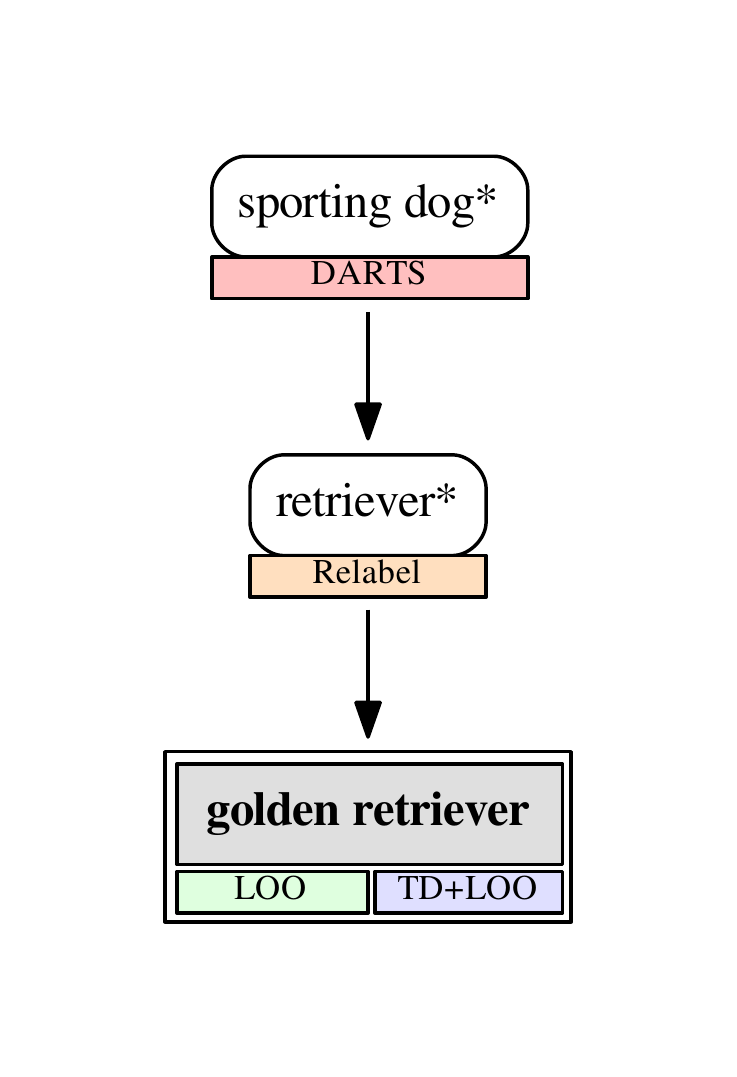}} & 
\multicolumn{4}{c}{\includegraphics[width=4.24cm, height=3.6cm, keepaspectratio]{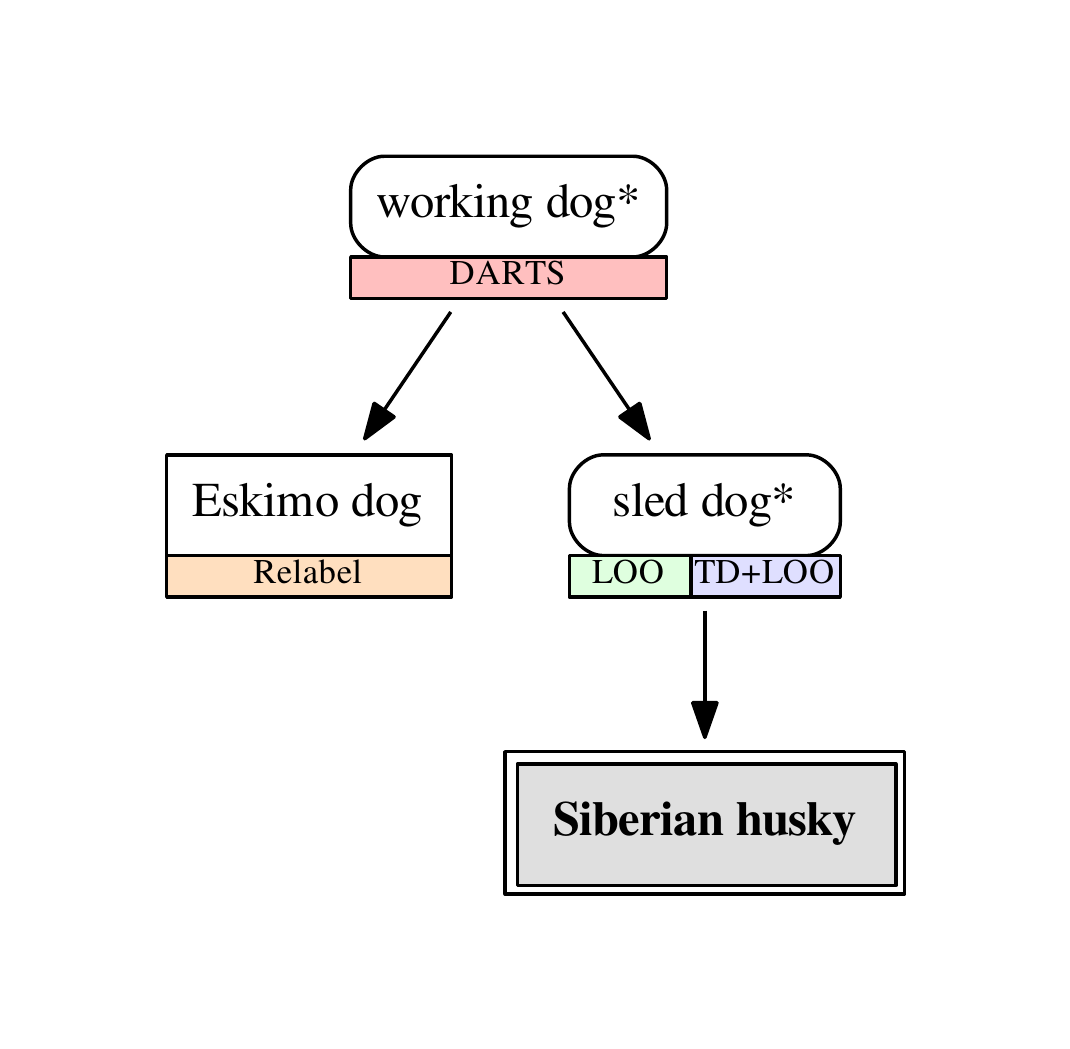}} \cr
\end{tabular}
\caption{Qualitative results of hierarchical novelty detection on ImageNet.
``GT'' is the true known leaf class, which is the expected prediction,
``DARTS'' is the baseline method proposed in \cite{deng2012hedging} where we modify the method for our purpose, and the others are our proposed methods.
``$\epsilon$'' is the distance between the prediction and GT,
``A'' indicates whether the prediction is an ancestor of GT, and
``Word'' is the English word of the predicted label.
Dashed edges represent multi-hop connection, where the number indicates the number of edges between classes.
If the prediction is on a super class (marked with * and rounded), then the test image is classified as a novel class whose closest class in the taxonomy is the super class.
}
\label{fig:qual_smp_1}
\end{figure*}

%% file: qual_smp/qual_smp_2.tex
\begin{figure*}[t]
\footnotesize\centering\setlength{\tabcolsep}{0cm}
\begin{tabular}{
>{\centering}m{1.12cm}>{\centering}m{0.4cm}>{\centering}m{0.4cm}m{2.32cm}
>{\centering}m{1.12cm}>{\centering}m{0.4cm}>{\centering}m{0.4cm}m{2.32cm}
>{\centering}m{1.12cm}>{\centering}m{0.4cm}>{\centering}m{0.4cm}m{2.32cm}
>{\centering}m{1.12cm}>{\centering}m{0.4cm}>{\centering}m{0.4cm}m{2.32cm}
}
\multicolumn{4}{c}{(a)} & 
\multicolumn{4}{c}{(b)} & 
\multicolumn{4}{c}{(c)} & 
\multicolumn{4}{c}{(d)} \cr
\multicolumn{4}{c}{\includegraphics[width=4.24cm, height=3.18cm, keepaspectratio]{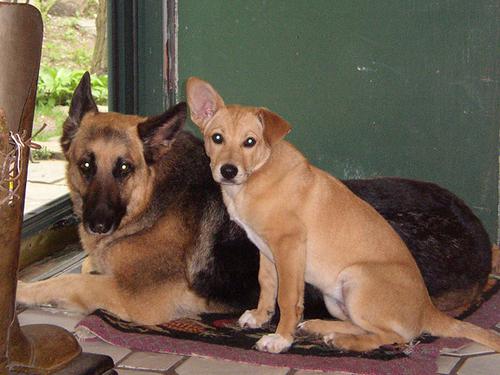}} & 
\multicolumn{4}{c}{\includegraphics[width=4.24cm, height=3.18cm, keepaspectratio]{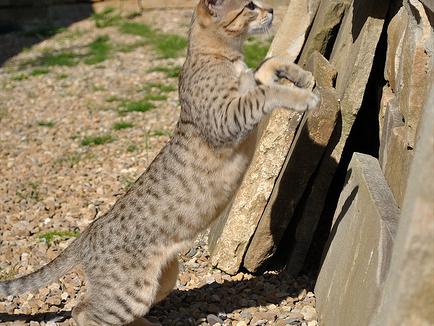}} & 
\multicolumn{4}{c}{\includegraphics[width=4.24cm, height=3.18cm, keepaspectratio]{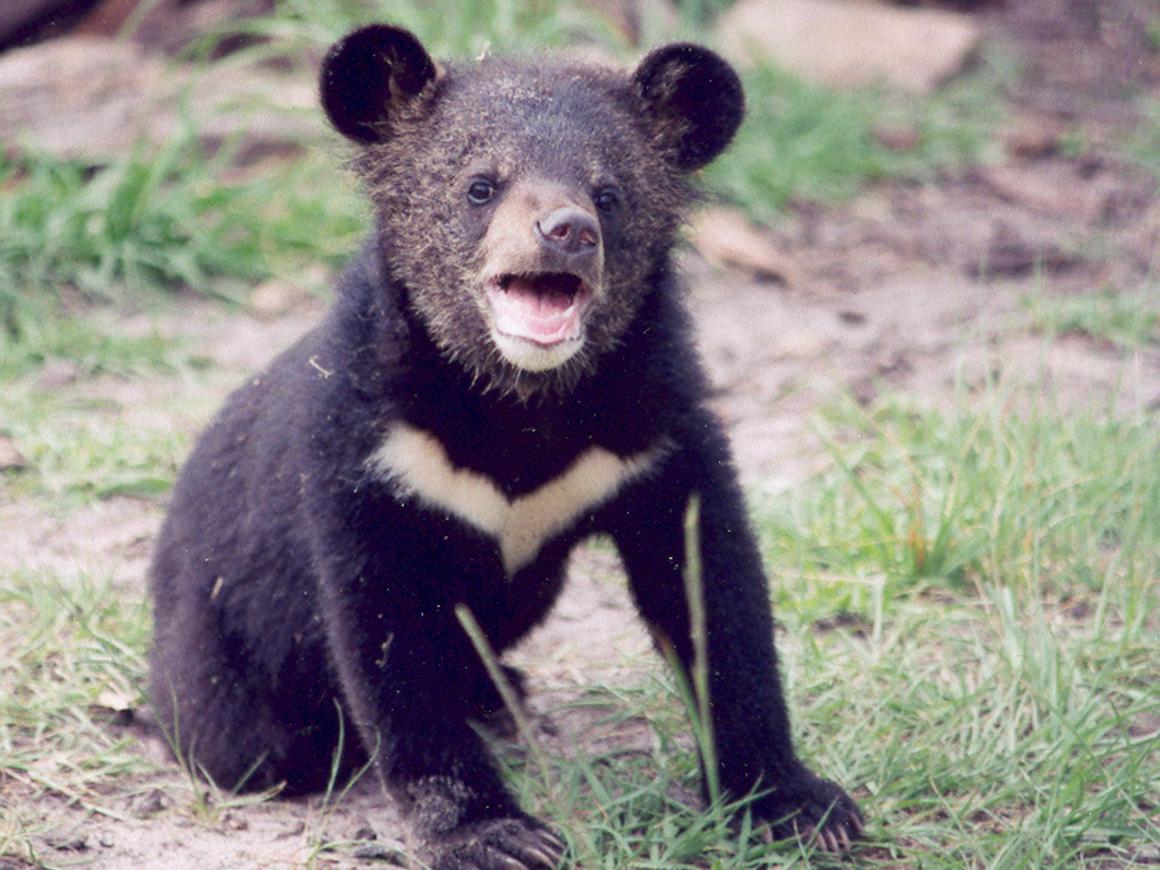}} & 
\multicolumn{4}{c}{\includegraphics[width=4.24cm, height=3.18cm, keepaspectratio]{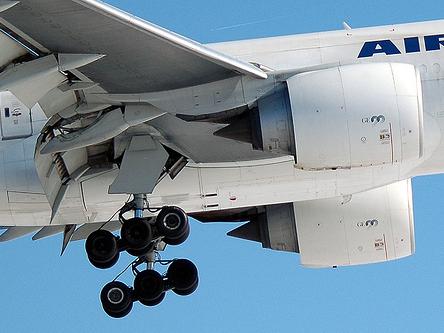}} \cr
\multicolumn{4}{l}{\cellcolor{ColorA} Known class: dingo} & 
\multicolumn{4}{l}{\cellcolor{ColorA} Known class: Egyptian cat} & 
\multicolumn{4}{l}{\cellcolor{ColorA} Known class: {\scriptsize American black bear}} & 
\multicolumn{4}{l}{\cellcolor{ColorA} Known class: airliner} \cr
\cellcolor{ColorB} Method & \cellcolor{ColorB} $\epsilon$ & \cellcolor{ColorB} A & \multicolumn{1}{c}{\cellcolor{ColorB} Word} & 
\cellcolor{ColorB} Method & \cellcolor{ColorB} $\epsilon$ & \cellcolor{ColorB} A & \multicolumn{1}{c}{\cellcolor{ColorB} Word} & 
\cellcolor{ColorB} Method & \cellcolor{ColorB} $\epsilon$ & \cellcolor{ColorB} A & \multicolumn{1}{c}{\cellcolor{ColorB} Word} & 
\cellcolor{ColorB} Method & \cellcolor{ColorB} $\epsilon$ & \cellcolor{ColorB} A & \multicolumn{1}{c}{\cellcolor{ColorB} Word} \cr
\cellcolor{Color0} GT & \cellcolor{Color0} & \cellcolor{Color0} & \cellcolor{Color0} dingo & 
\cellcolor{Color0} GT & \cellcolor{Color0} & \cellcolor{Color0} & \cellcolor{Color0} Egyptian cat & 
\cellcolor{Color0} GT & \cellcolor{Color0} & \cellcolor{Color0} & \cellcolor{Color0} {\scriptsize American black bear} & 
\cellcolor{Color0} GT & \cellcolor{Color0} & \cellcolor{Color0} & \cellcolor{Color0} airliner \cr
\cellcolor{Color1} DARTS & \cellcolor{Color1} 5 & \cellcolor{Color1} N & \cellcolor{Color1} shepherd dog & 
\cellcolor{Color1} DARTS & \cellcolor{Color1} 2 & \cellcolor{Color1} Y & \cellcolor{Color1} cat & 
\cellcolor{Color1} DARTS & \cellcolor{Color1} 0 & \cellcolor{Color1} Y & \cellcolor{Color1} {\scriptsize American black bear} & 
\cellcolor{Color1} DARTS & \cellcolor{Color1} 8 & \cellcolor{Color1} N & \cellcolor{Color1} wing \cr
\cellcolor{Color2} Relabel & \cellcolor{Color2} 3 & \cellcolor{Color2} N & \cellcolor{Color2} dog & 
\cellcolor{Color2} Relabel & \cellcolor{Color2} 4 & \cellcolor{Color2} N & \cellcolor{Color2} lynx & 
\cellcolor{Color2} Relabel & \cellcolor{Color2} 2 & \cellcolor{Color2} Y & \cellcolor{Color2} carnivore & 
\cellcolor{Color2} Relabel & \cellcolor{Color2} 2 & \cellcolor{Color2} N & \cellcolor{Color2} warplane \cr
\cellcolor{Color3} LOO & \cellcolor{Color3} 1 & \cellcolor{Color3} Y & \cellcolor{Color3} wild dog & 
\cellcolor{Color3} LOO & \cellcolor{Color3} 3 & \cellcolor{Color3} Y & \cellcolor{Color3} feline & 
\cellcolor{Color3} LOO & \cellcolor{Color3} 1 & \cellcolor{Color3} Y & \cellcolor{Color3} bear & 
\cellcolor{Color3} LOO & \cellcolor{Color3} 1 & \cellcolor{Color3} Y & \cellcolor{Color3} {\scriptsize heavier-than-air craft} \cr
\cellcolor{Color4} TD+LOO & \cellcolor{Color4} 0 & \cellcolor{Color4} Y & \cellcolor{Color4} dingo & 
\cellcolor{Color4} TD+LOO & \cellcolor{Color4} 3 & \cellcolor{Color4} N & \cellcolor{Color4} wildcat & 
\cellcolor{Color4} TD+LOO & \cellcolor{Color4} 1 & \cellcolor{Color4} Y & \cellcolor{Color4} bear & 
\cellcolor{Color4} TD+LOO & \cellcolor{Color4} 1 & \cellcolor{Color4} Y & \cellcolor{Color4} {\scriptsize heavier-than-air craft} \cr
\multicolumn{4}{c}{\includegraphics[width=4.24cm, height=3.6cm, keepaspectratio]{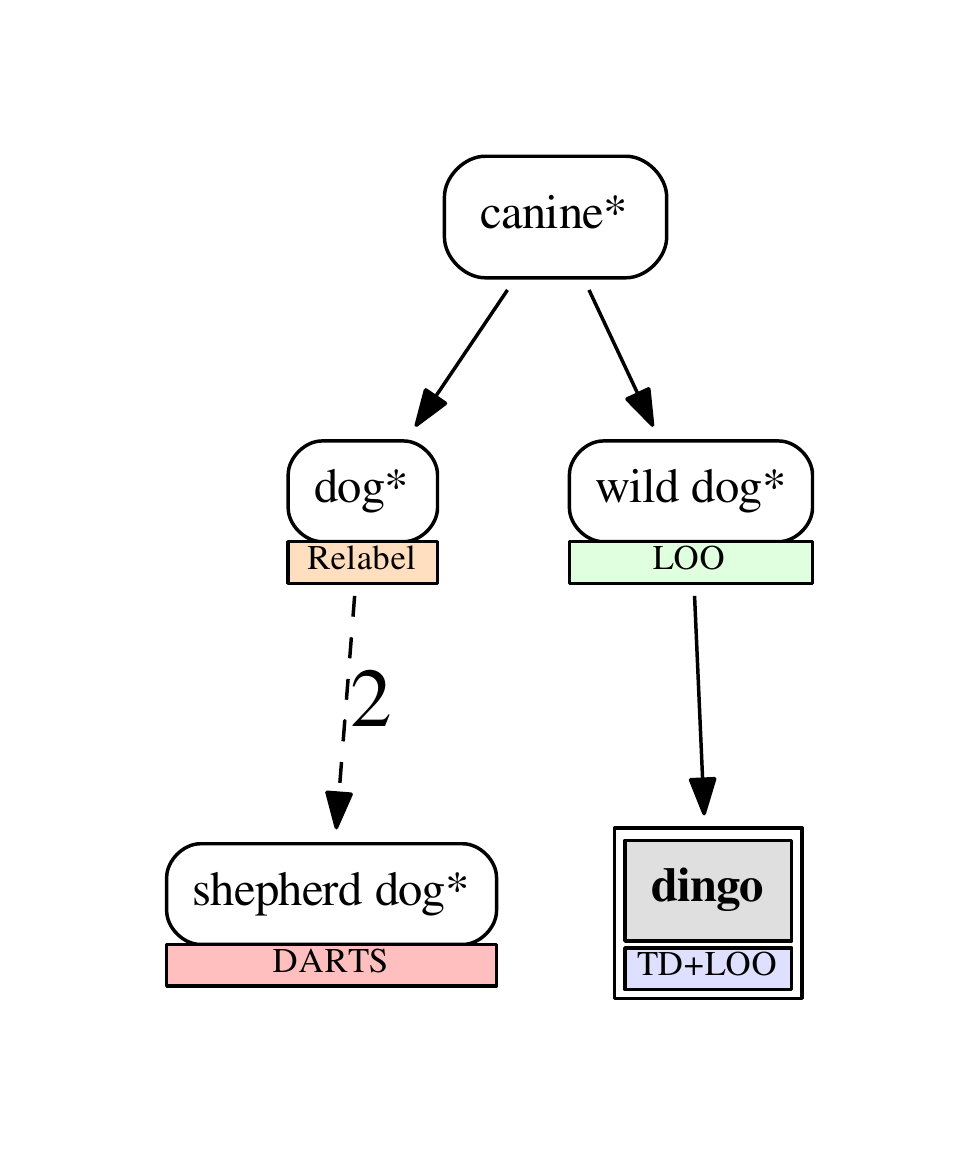}} & 
\multicolumn{4}{c}{\includegraphics[width=4.24cm, height=3.6cm, keepaspectratio]{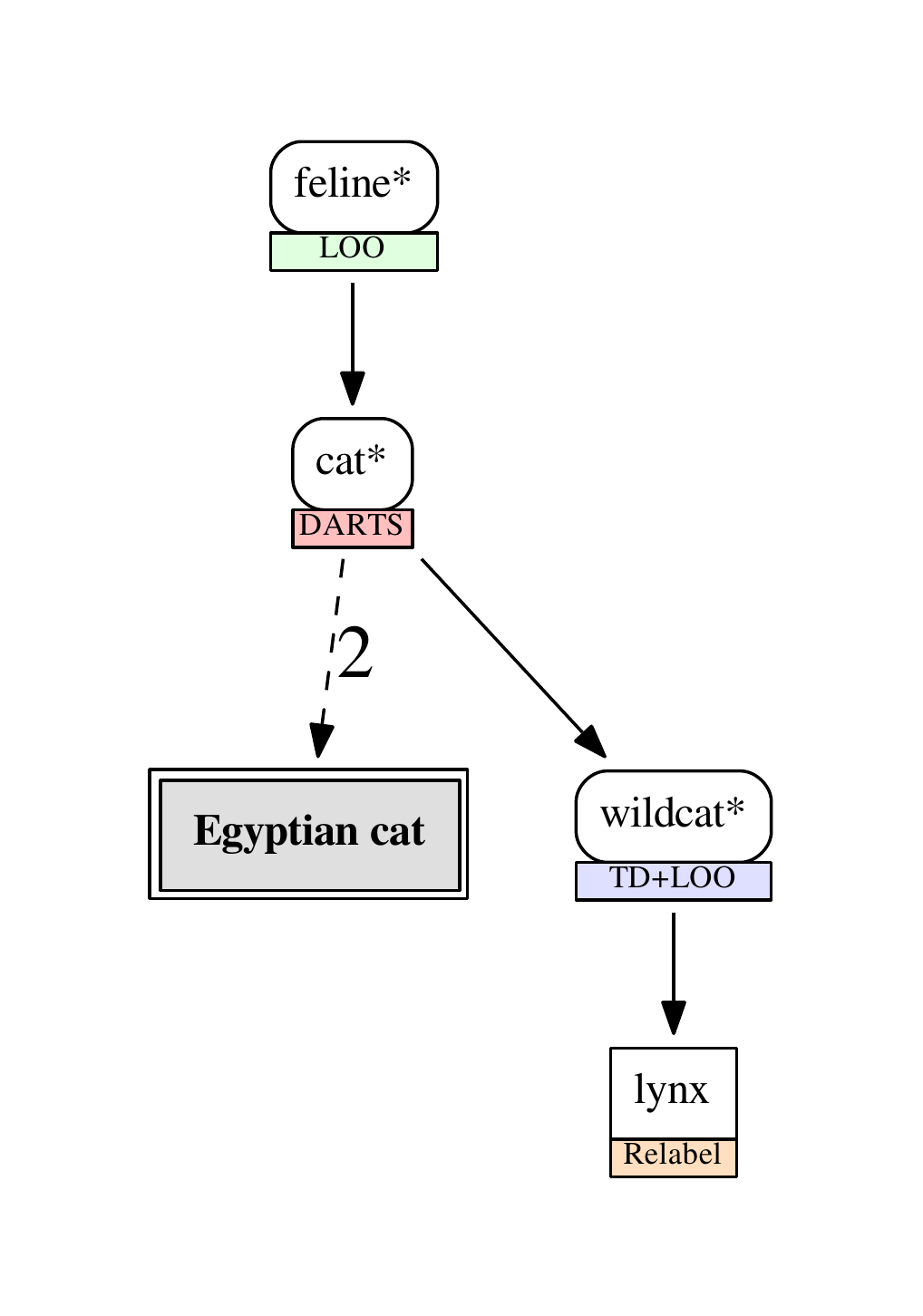}} & 
\multicolumn{4}{c}{\includegraphics[width=4.24cm, height=3.6cm, keepaspectratio]{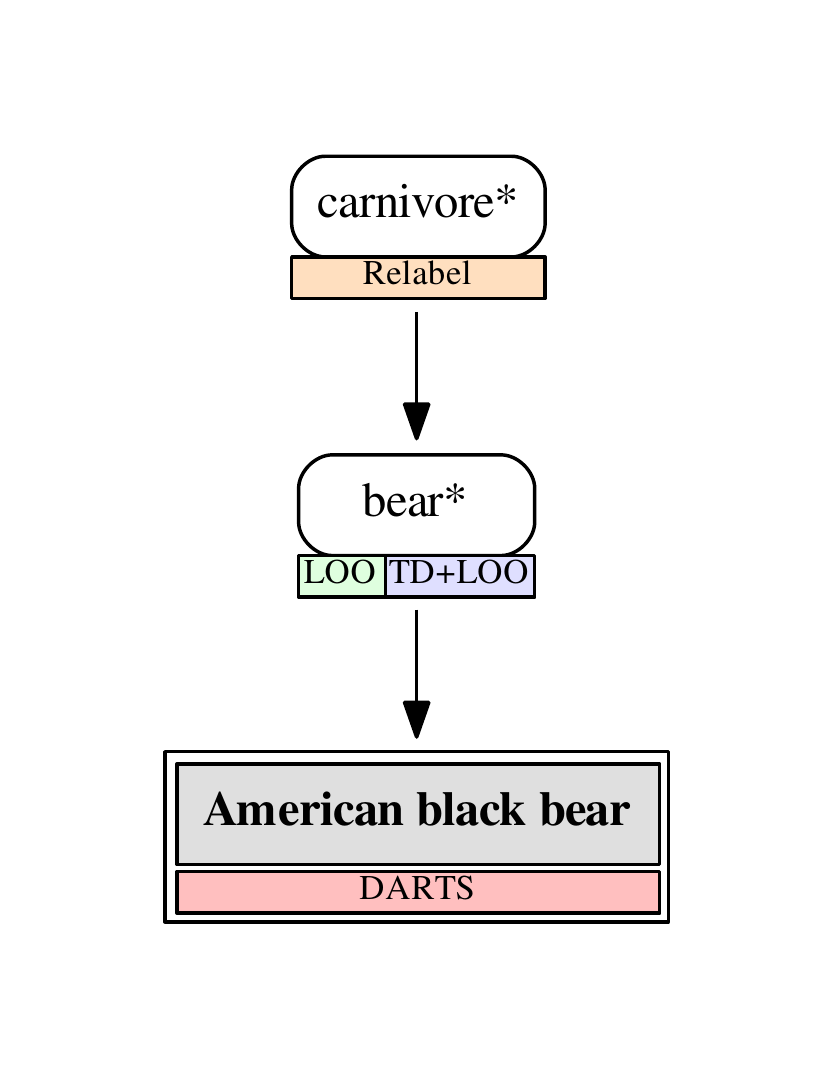}} & 
\multicolumn{4}{c}{\includegraphics[width=4.24cm, height=3.6cm, keepaspectratio]{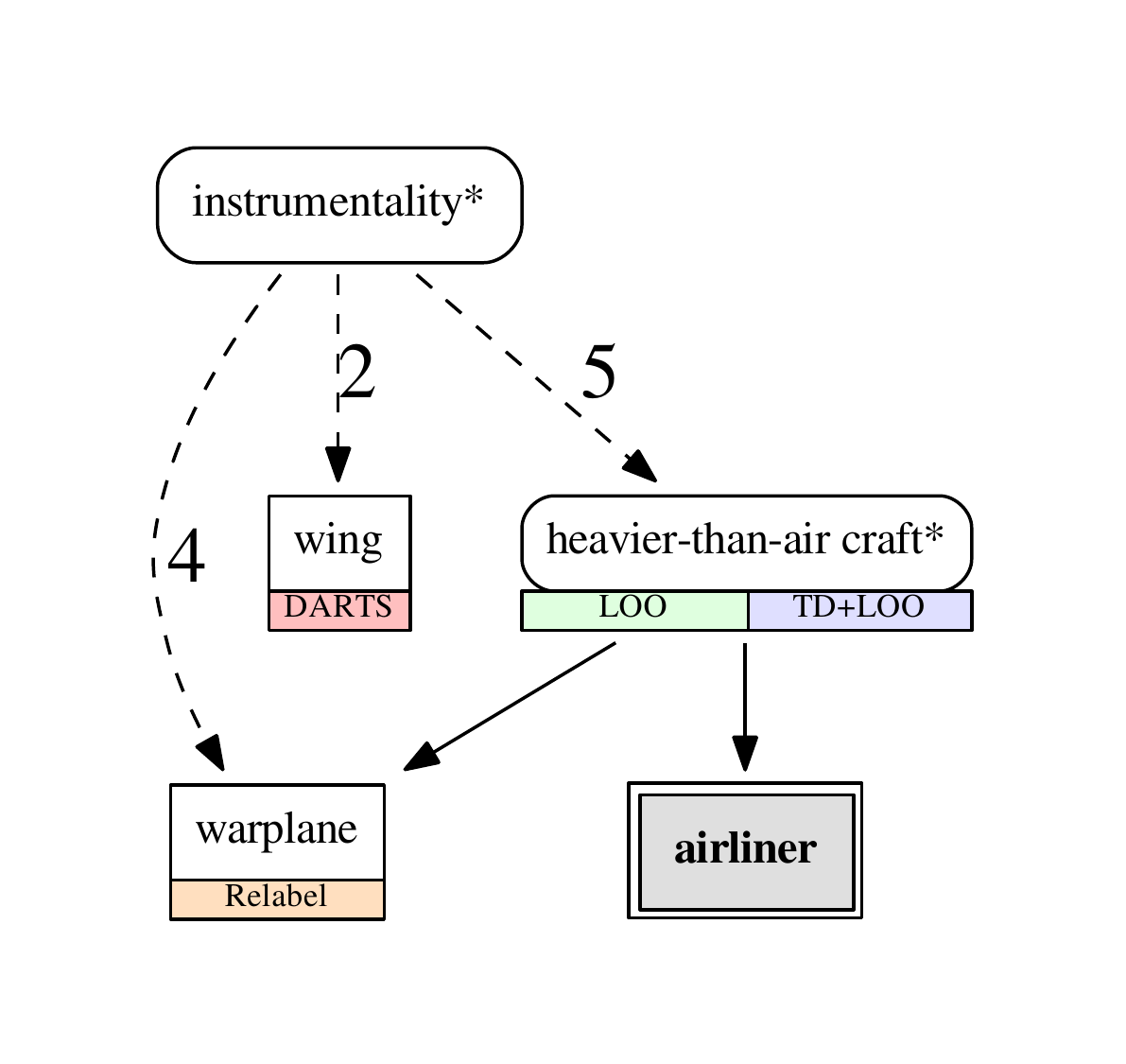}} \cr
\multicolumn{4}{c}{(e)} & 
\multicolumn{4}{c}{(f)} & 
\multicolumn{4}{c}{(g)} & 
\multicolumn{4}{c}{(h)} \cr
\multicolumn{4}{c}{\includegraphics[width=4.24cm, height=3.18cm, keepaspectratio]{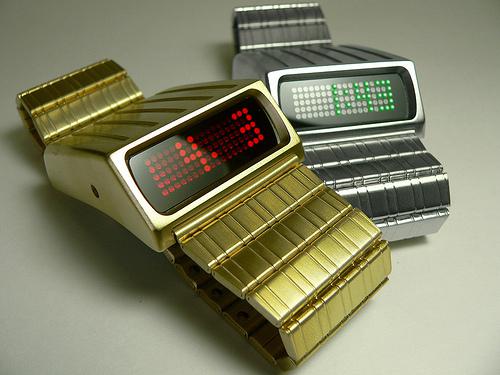}} & 
\multicolumn{4}{c}{\includegraphics[width=4.24cm, height=3.18cm, keepaspectratio]{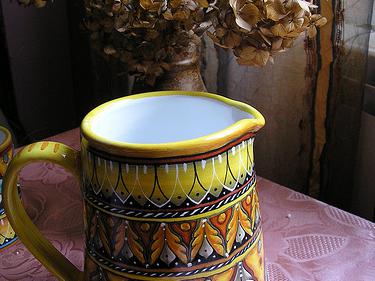}} & 
\multicolumn{4}{c}{\includegraphics[width=4.24cm, height=3.18cm, keepaspectratio]{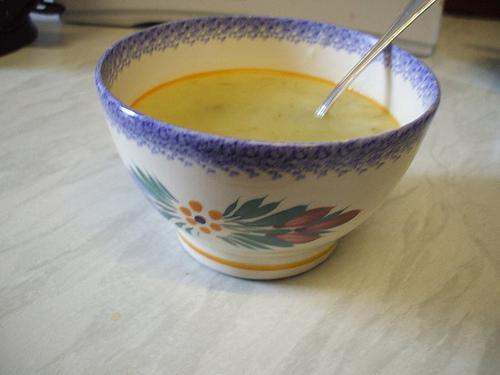}} & 
\multicolumn{4}{c}{\includegraphics[width=4.24cm, height=3.18cm, keepaspectratio]{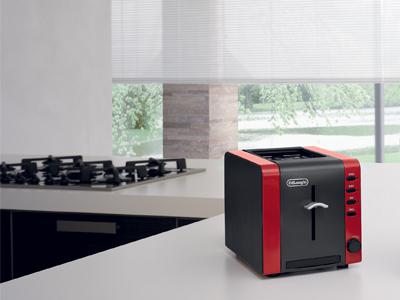}} \cr
\multicolumn{4}{l}{\cellcolor{ColorA} Known class: digital clock} & 
\multicolumn{4}{l}{\cellcolor{ColorA} Known class: pitcher} & 
\multicolumn{4}{l}{\cellcolor{ColorA} Known class: soup bowl} & 
\multicolumn{4}{l}{\cellcolor{ColorA} Known class: toaster} \cr
\cellcolor{ColorB} Method & \cellcolor{ColorB} $\epsilon$ & \cellcolor{ColorB} A & \multicolumn{1}{c}{\cellcolor{ColorB} Word} & 
\cellcolor{ColorB} Method & \cellcolor{ColorB} $\epsilon$ & \cellcolor{ColorB} A & \multicolumn{1}{c}{\cellcolor{ColorB} Word} & 
\cellcolor{ColorB} Method & \cellcolor{ColorB} $\epsilon$ & \cellcolor{ColorB} A & \multicolumn{1}{c}{\cellcolor{ColorB} Word} & 
\cellcolor{ColorB} Method & \cellcolor{ColorB} $\epsilon$ & \cellcolor{ColorB} A & \multicolumn{1}{c}{\cellcolor{ColorB} Word} \cr
\cellcolor{Color0} GT & \cellcolor{Color0} & \cellcolor{Color0} & \cellcolor{Color0} digital clock & 
\cellcolor{Color0} GT & \cellcolor{Color0} & \cellcolor{Color0} & \cellcolor{Color0} pitcher & 
\cellcolor{Color0} GT & \cellcolor{Color0} & \cellcolor{Color0} & \cellcolor{Color0} soup bowl & 
\cellcolor{Color0} GT & \cellcolor{Color0} & \cellcolor{Color0} & \cellcolor{Color0} toaster \cr
\cellcolor{Color1} DARTS & \cellcolor{Color1} 3 & \cellcolor{Color1} N & \cellcolor{Color1} digital watch & 
\cellcolor{Color1} DARTS & \cellcolor{Color1} 7 & \cellcolor{Color1} N & \cellcolor{Color1} drum & 
\cellcolor{Color1} DARTS & \cellcolor{Color1} 1 & \cellcolor{Color1} Y & \cellcolor{Color1} bowl & 
\cellcolor{Color1} DARTS & \cellcolor{Color1} 9 & \cellcolor{Color1} N & \cellcolor{Color1} furniture \cr
\cellcolor{Color2} Relabel & \cellcolor{Color2} 3 & \cellcolor{Color2} Y & \cellcolor{Color2} {\scriptsize measuring instrument} & 
\cellcolor{Color2} Relabel & \cellcolor{Color2} 1 & \cellcolor{Color2} Y & \cellcolor{Color2} vessel & 
\cellcolor{Color2} Relabel & \cellcolor{Color2} 1 & \cellcolor{Color2} Y & \cellcolor{Color2} bowl & 
\cellcolor{Color2} Relabel & \cellcolor{Color2} 7 & \cellcolor{Color2} N & \cellcolor{Color2} instrumentality \cr
\cellcolor{Color3} LOO & \cellcolor{Color3} 2 & \cellcolor{Color3} Y & \cellcolor{Color3} timepiece & 
\cellcolor{Color3} LOO & \cellcolor{Color3} 6 & \cellcolor{Color3} N & \cellcolor{Color3} {\scriptsize percussion instrument} & 
\cellcolor{Color3} LOO & \cellcolor{Color3} 11 & \cellcolor{Color3} N & \cellcolor{Color3} punch & 
\cellcolor{Color3} LOO & \cellcolor{Color3} 1 & \cellcolor{Color3} Y & \cellcolor{Color3} kitchen appliance \cr
\cellcolor{Color4} TD+LOO & \cellcolor{Color4} 0 & \cellcolor{Color4} Y & \cellcolor{Color4} digital clock & 
\cellcolor{Color4} TD+LOO & \cellcolor{Color4} 0 & \cellcolor{Color4} Y & \cellcolor{Color4} pitcher & 
\cellcolor{Color4} TD+LOO & \cellcolor{Color4} 11 & \cellcolor{Color4} N & \cellcolor{Color4} punch & 
\cellcolor{Color4} TD+LOO & \cellcolor{Color4} 0 & \cellcolor{Color4} Y & \cellcolor{Color4} toaster \cr
\multicolumn{4}{c}{\includegraphics[width=4.24cm, height=3.6cm, keepaspectratio]{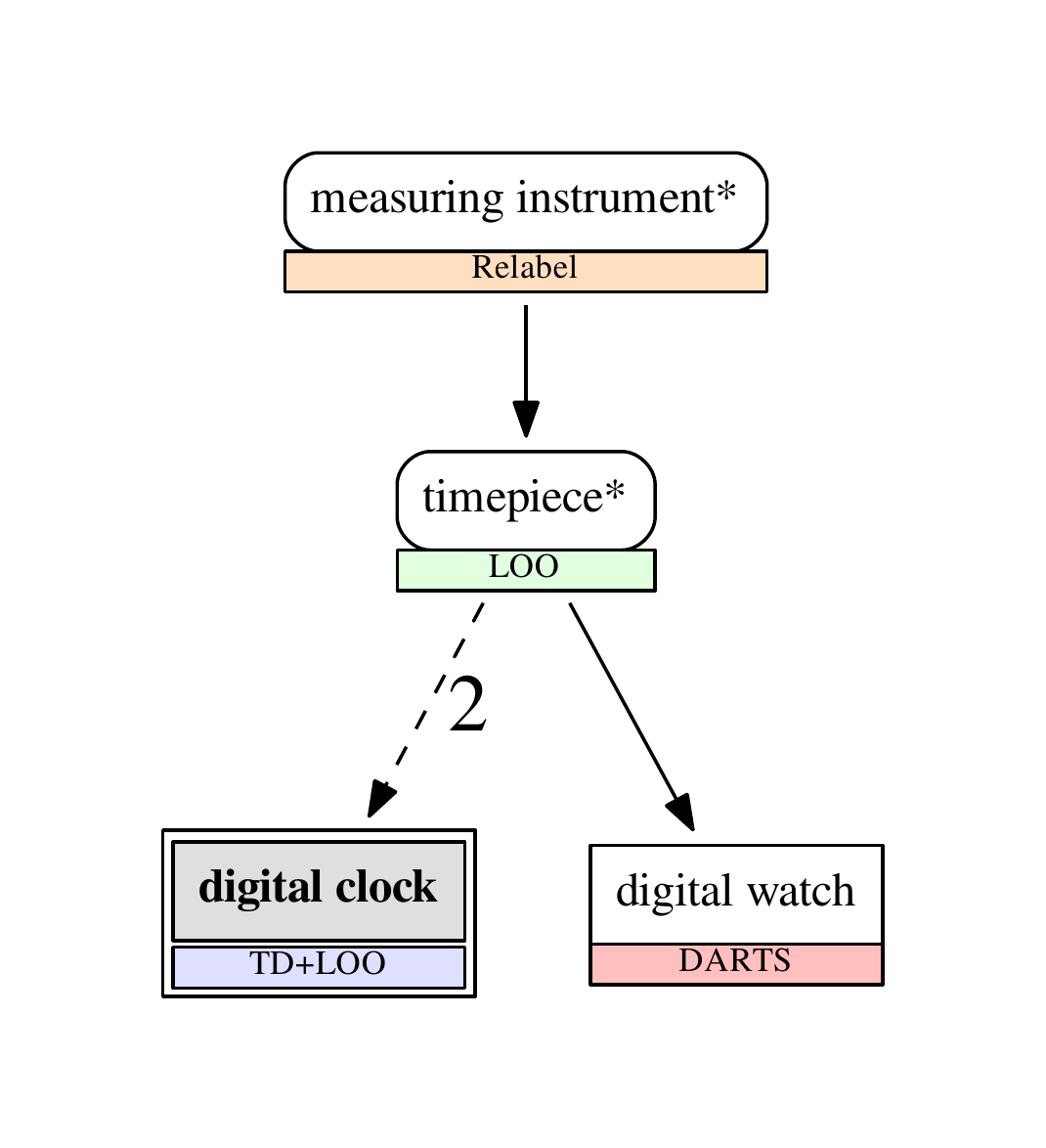}} & 
\multicolumn{4}{c}{\includegraphics[width=4.24cm, height=3.6cm, keepaspectratio]{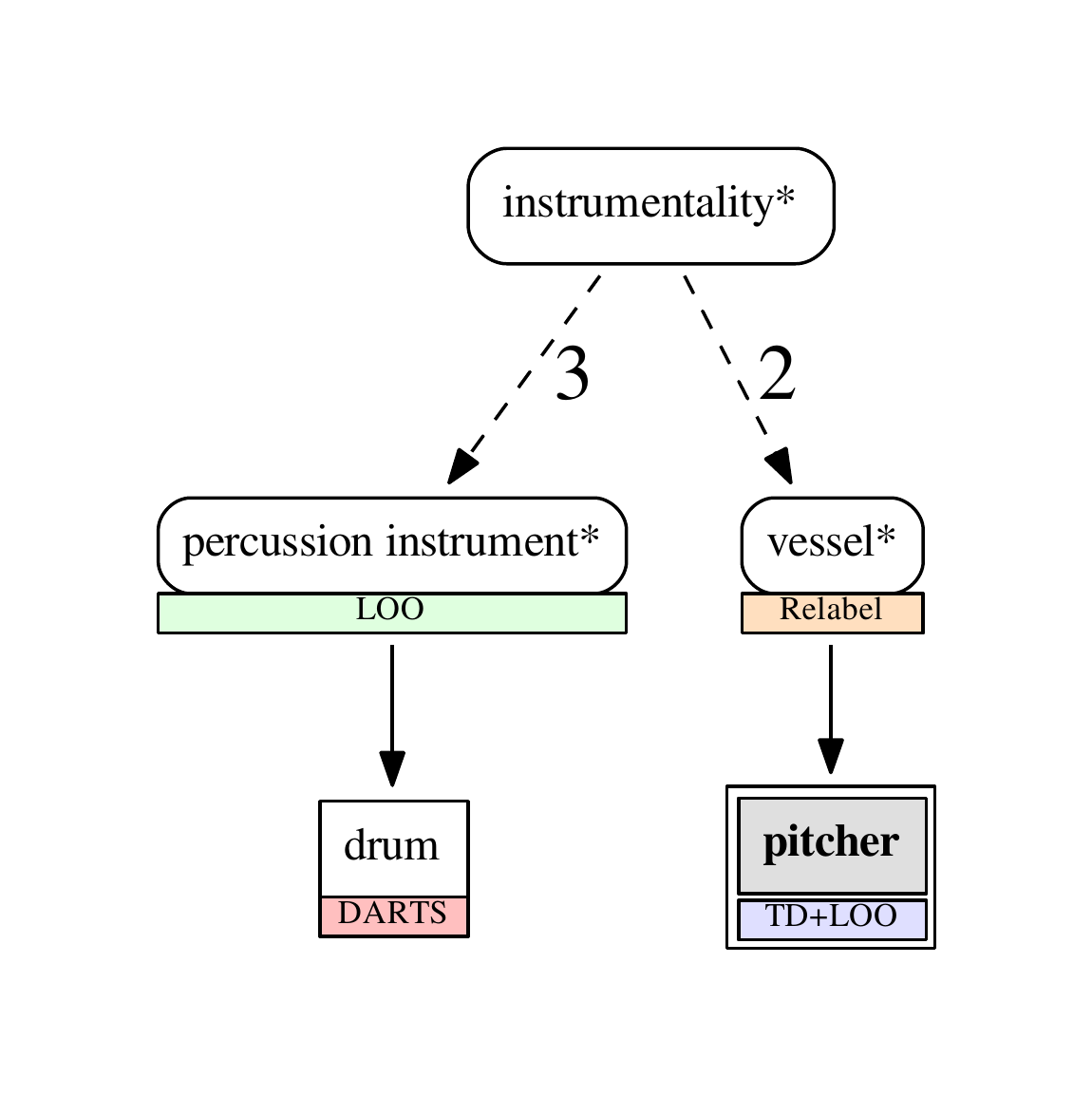}} & 
\multicolumn{4}{c}{\includegraphics[width=4.24cm, height=3.6cm, keepaspectratio]{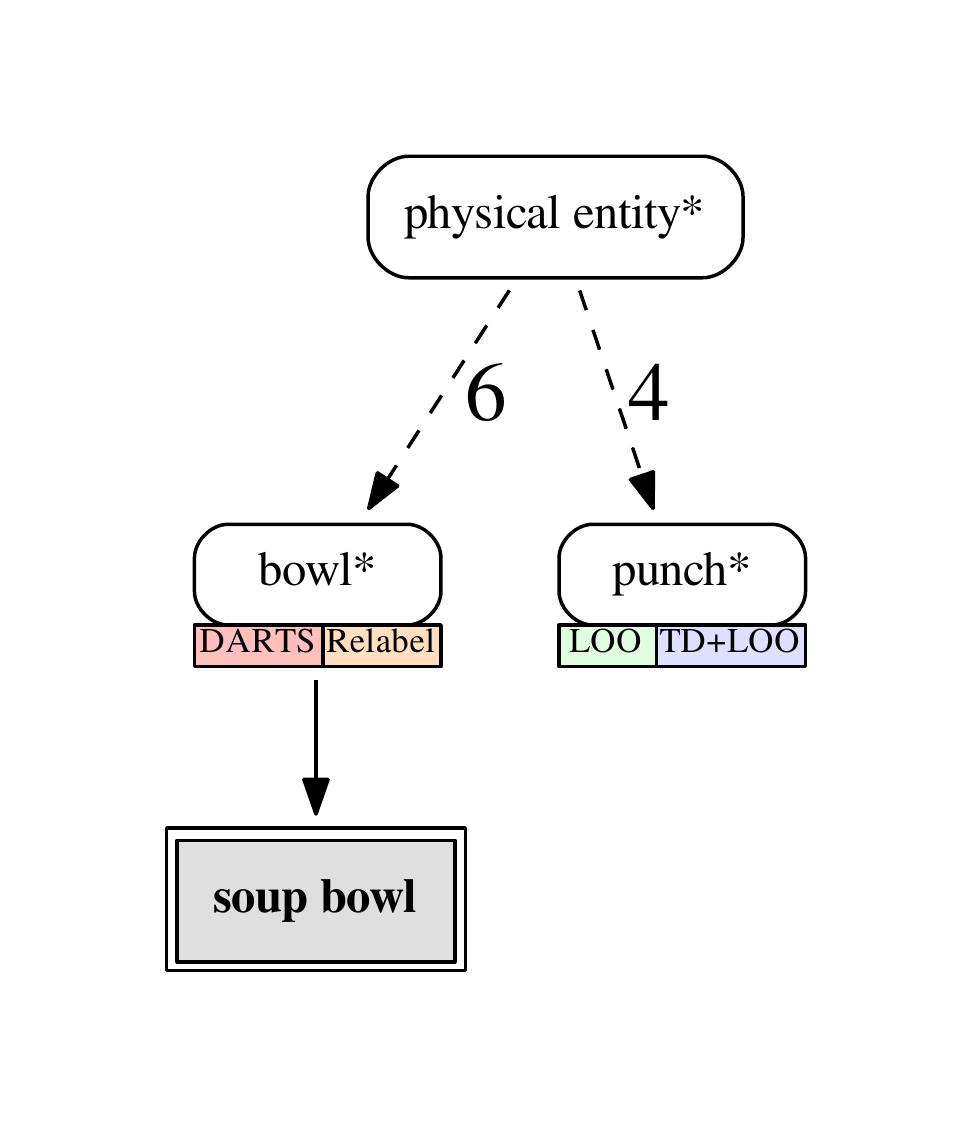}} & 
\multicolumn{4}{c}{\includegraphics[width=4.24cm, height=3.6cm, keepaspectratio]{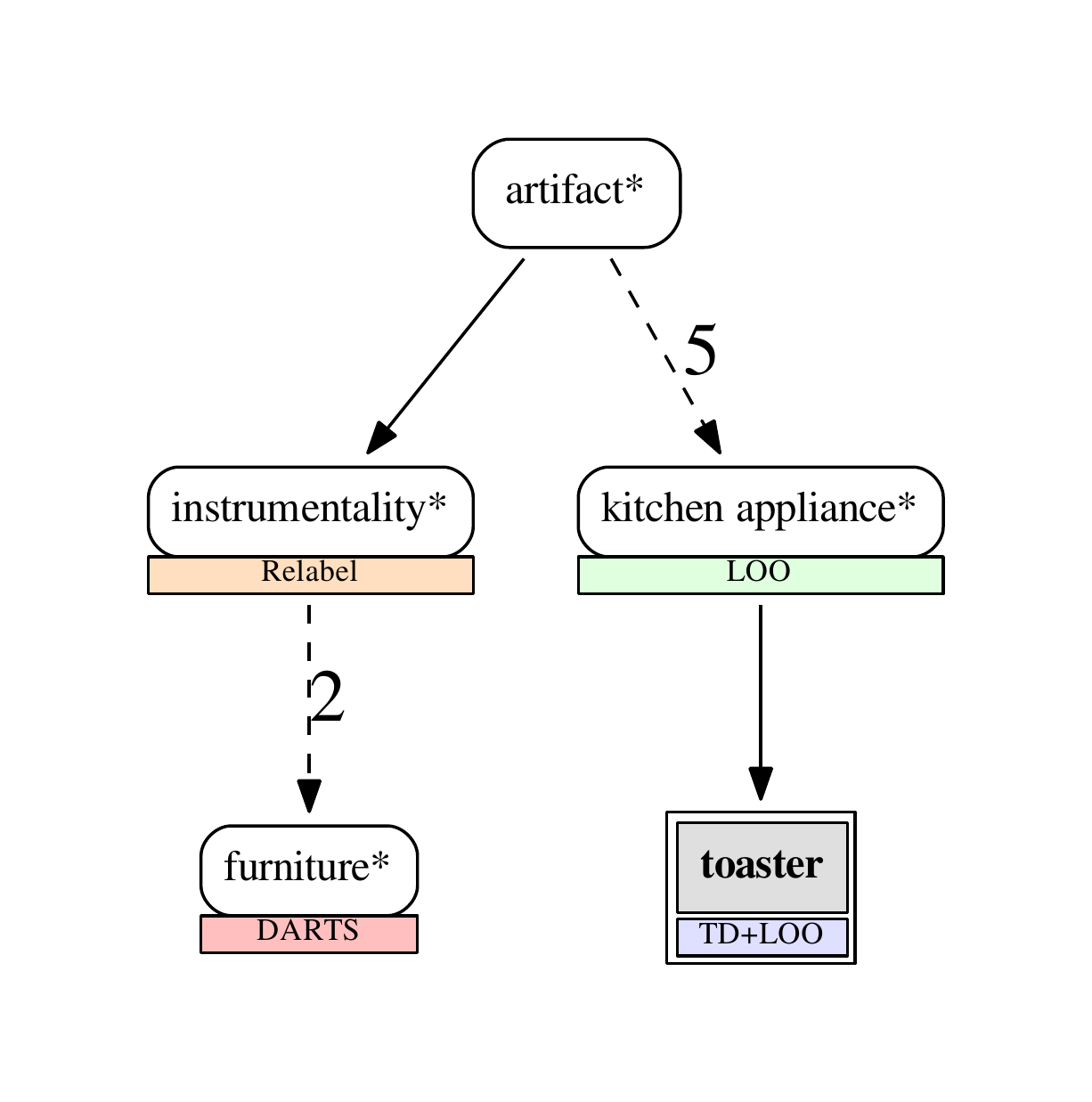}} \cr
\end{tabular}
\caption{Qualitative results of hierarchical novelty detection on ImageNet.
``GT'' is the true known leaf class, which is the expected prediction,
``DARTS'' is the baseline method proposed in \cite{deng2012hedging} where we modify the method for our purpose, and the others are our proposed methods.
``$\epsilon$'' is the distance between the prediction and GT,
``A'' indicates whether the prediction is an ancestor of GT, and
``Word'' is the English word of the predicted label.
Dashed edges represent multi-hop connection, where the number indicates the number of edges between classes.
If the prediction is on a super class (marked with * and rounded), then the test image is classified as a novel class whose closest class in the taxonomy is the super class.
}
\label{fig:qual_smp_2}
\end{figure*}

%% file: qual_smp/qual_smp_3.tex
\begin{figure*}[t]
\footnotesize\centering\setlength{\tabcolsep}{0cm}
\begin{tabular}{
>{\centering}m{1.12cm}>{\centering}m{0.4cm}>{\centering}m{0.4cm}m{2.32cm}
>{\centering}m{1.12cm}>{\centering}m{0.4cm}>{\centering}m{0.4cm}m{2.32cm}
>{\centering}m{1.12cm}>{\centering}m{0.4cm}>{\centering}m{0.4cm}m{2.32cm}
>{\centering}m{1.12cm}>{\centering}m{0.4cm}>{\centering}m{0.4cm}m{2.32cm}
}
\multicolumn{4}{c}{(a)} & 
\multicolumn{4}{c}{(b)} & 
\multicolumn{4}{c}{(c)} & 
\multicolumn{4}{c}{(d)} \cr
\multicolumn{4}{c}{\includegraphics[width=4.24cm, height=3.18cm, keepaspectratio]{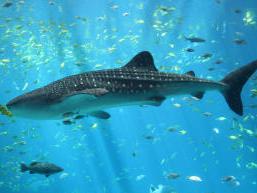}} & 
\multicolumn{4}{c}{\includegraphics[width=4.24cm, height=3.18cm, keepaspectratio]{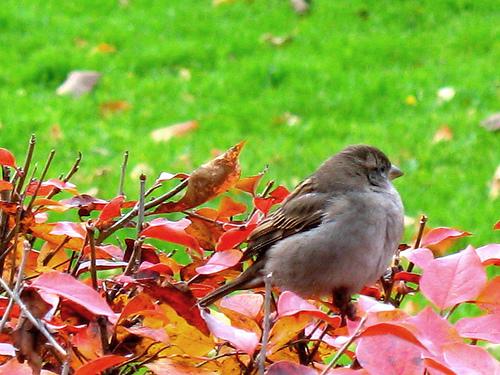}} & 
\multicolumn{4}{c}{\includegraphics[width=4.24cm, height=3.18cm, keepaspectratio]{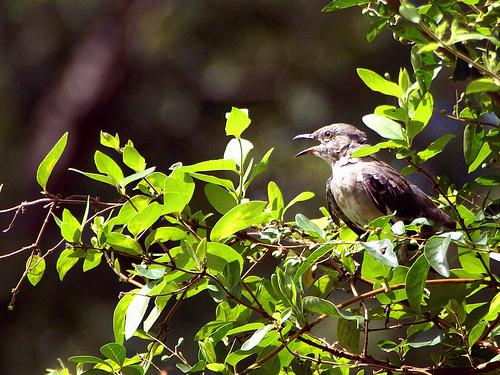}} & 
\multicolumn{4}{c}{\includegraphics[width=4.24cm, height=3.18cm, keepaspectratio]{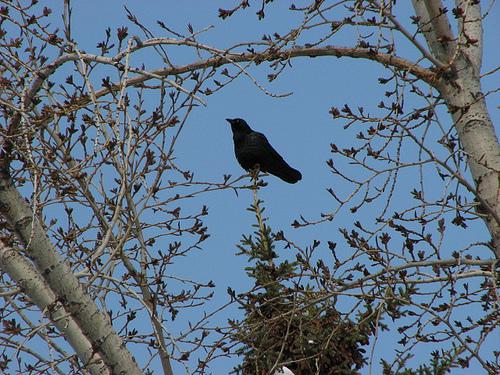}} \cr
\multicolumn{4}{l}{\cellcolor{ColorA} Novel class: whale shark} & 
\multicolumn{4}{l}{\cellcolor{ColorA} Novel class: dickeybird} & 
\multicolumn{4}{l}{\cellcolor{ColorA} Novel class: songbird} & 
\multicolumn{4}{l}{\cellcolor{ColorA} Novel class: American crow} \cr
\cellcolor{ColorB} Method & \cellcolor{ColorB} $\epsilon$ & \cellcolor{ColorB} A & \multicolumn{1}{c}{\cellcolor{ColorB} Word} & 
\cellcolor{ColorB} Method & \cellcolor{ColorB} $\epsilon$ & \cellcolor{ColorB} A & \multicolumn{1}{c}{\cellcolor{ColorB} Word} & 
\cellcolor{ColorB} Method & \cellcolor{ColorB} $\epsilon$ & \cellcolor{ColorB} A & \multicolumn{1}{c}{\cellcolor{ColorB} Word} & 
\cellcolor{ColorB} Method & \cellcolor{ColorB} $\epsilon$ & \cellcolor{ColorB} A & \multicolumn{1}{c}{\cellcolor{ColorB} Word} \cr
\cellcolor{Color0} GT & \cellcolor{Color0} & \cellcolor{Color0} & \cellcolor{Color0} shark & 
\cellcolor{Color0} GT & \cellcolor{Color0} & \cellcolor{Color0} & \cellcolor{Color0} bird & 
\cellcolor{Color0} GT & \cellcolor{Color0} & \cellcolor{Color0} & \cellcolor{Color0} oscine bird & 
\cellcolor{Color0} GT & \cellcolor{Color0} & \cellcolor{Color0} & \cellcolor{Color0} corvine bird \cr
\cellcolor{Color1} DARTS & \cellcolor{Color1} 1 & \cellcolor{Color1} N & \cellcolor{Color1} tiger shark & 
\cellcolor{Color1} DARTS & \cellcolor{Color1} 3 & \cellcolor{Color1} N & \cellcolor{Color1} junco & 
\cellcolor{Color1} DARTS & \cellcolor{Color1} 1 & \cellcolor{Color1} N & \cellcolor{Color1} thrush & 
\cellcolor{Color1} DARTS & \cellcolor{Color1} 2 & \cellcolor{Color1} Y & \cellcolor{Color1} bird \cr
\cellcolor{Color2} Relabel & \cellcolor{Color2} 0 & \cellcolor{Color2} Y & \cellcolor{Color2} shark & 
\cellcolor{Color2} Relabel & \cellcolor{Color2} 2 & \cellcolor{Color2} N & \cellcolor{Color2} finch & 
\cellcolor{Color2} Relabel & \cellcolor{Color2} 1 & \cellcolor{Color2} Y & \cellcolor{Color2} bird & 
\cellcolor{Color2} Relabel & \cellcolor{Color2} 3 & \cellcolor{Color2} N & \cellcolor{Color2} bird of prey \cr
\cellcolor{Color3} LOO & \cellcolor{Color3} 2 & \cellcolor{Color3} Y & \cellcolor{Color3} fish & 
\cellcolor{Color3} LOO & \cellcolor{Color3} 2 & \cellcolor{Color3} N & \cellcolor{Color3} thrush & 
\cellcolor{Color3} LOO & \cellcolor{Color3} 0 & \cellcolor{Color3} Y & \cellcolor{Color3} oscine bird & 
\cellcolor{Color3} LOO & \cellcolor{Color3} 1 & \cellcolor{Color3} Y & \cellcolor{Color3} oscine bird \cr
\cellcolor{Color4} TD+LOO & \cellcolor{Color4} 0 & \cellcolor{Color4} Y & \cellcolor{Color4} shark & 
\cellcolor{Color4} TD+LOO & \cellcolor{Color4} 1 & \cellcolor{Color4} N & \cellcolor{Color4} oscine bird & 
\cellcolor{Color4} TD+LOO & \cellcolor{Color4} 1 & \cellcolor{Color4} N & \cellcolor{Color4} corvine bird & 
\cellcolor{Color4} TD+LOO & \cellcolor{Color4} 0 & \cellcolor{Color4} Y & \cellcolor{Color4} corvine bird \cr
\multicolumn{4}{c}{\includegraphics[width=4.24cm, height=3.6cm, keepaspectratio]{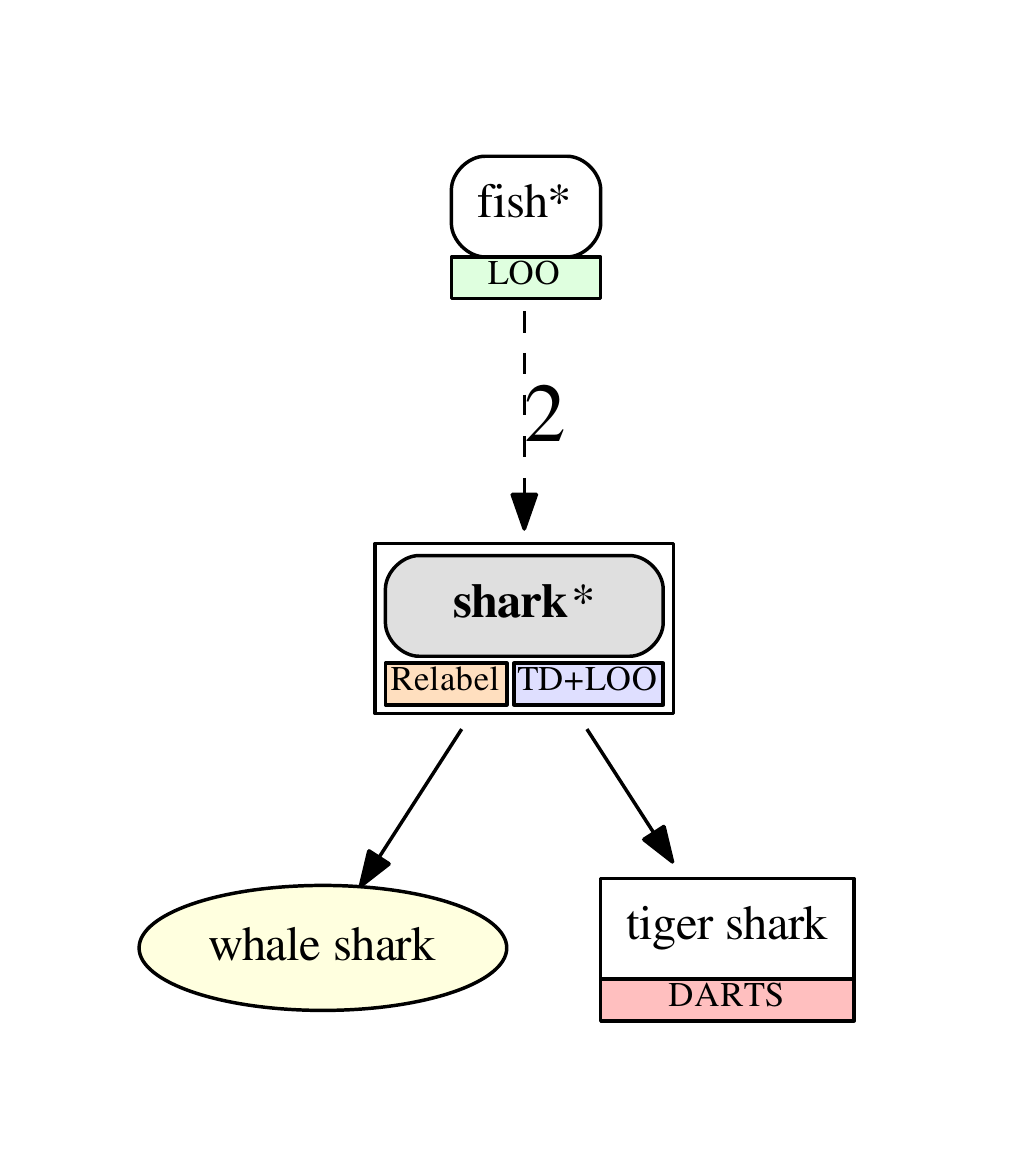}} & 
\multicolumn{4}{c}{\includegraphics[width=4.24cm, height=3.6cm, keepaspectratio]{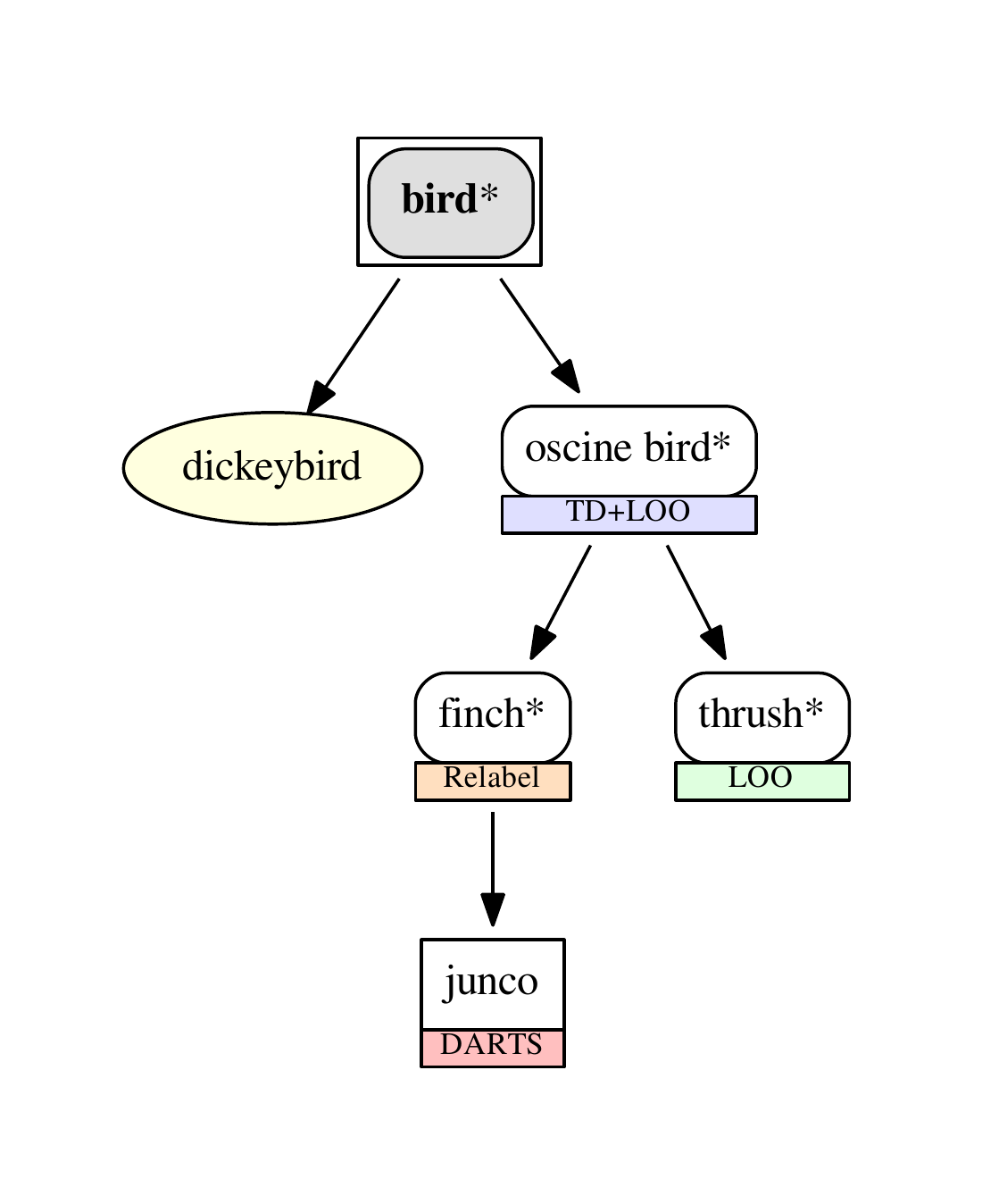}} & 
\multicolumn{4}{c}{\includegraphics[width=4.24cm, height=3.6cm, keepaspectratio]{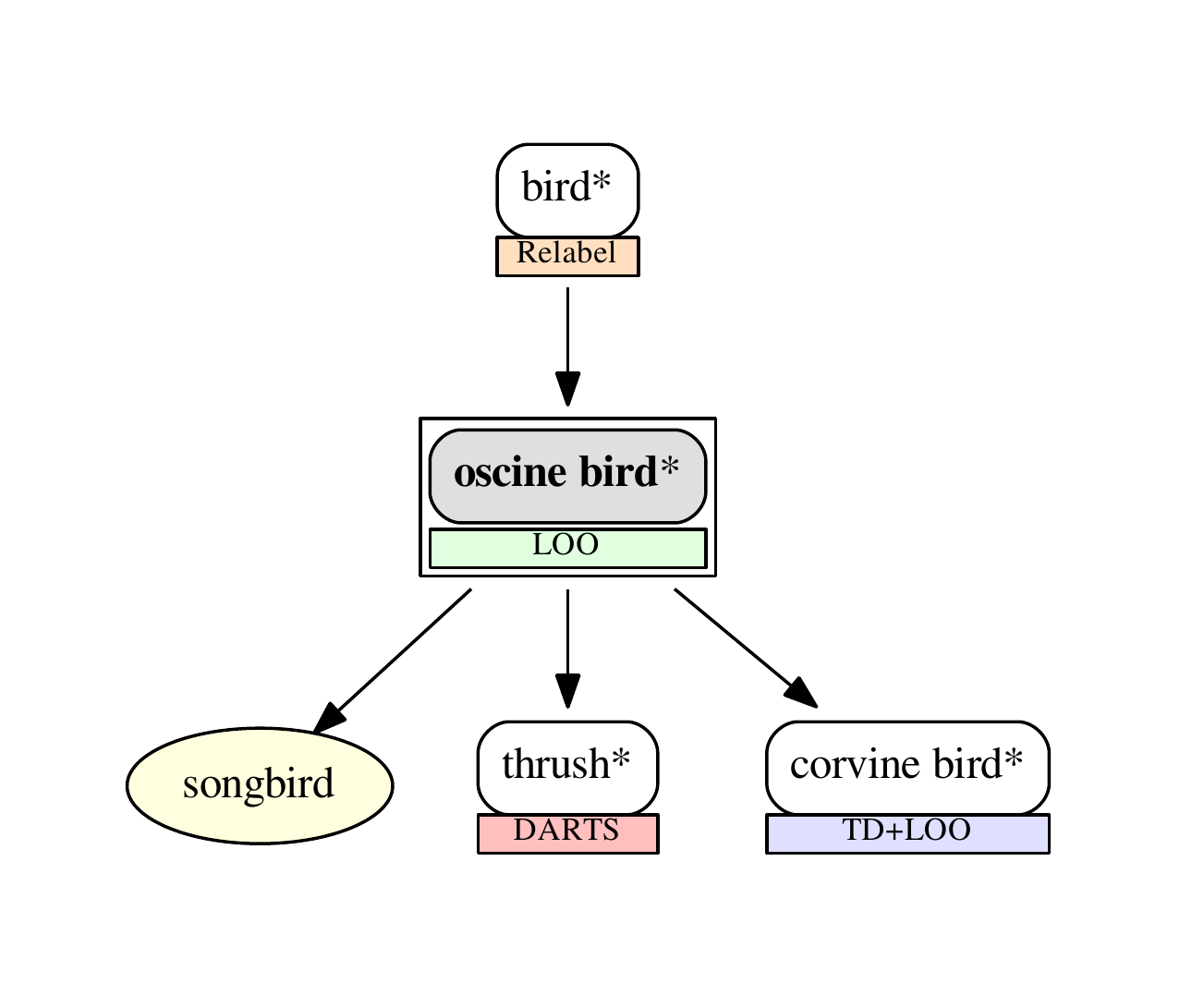}} & 
\multicolumn{4}{c}{\includegraphics[width=4.24cm, height=3.6cm, keepaspectratio]{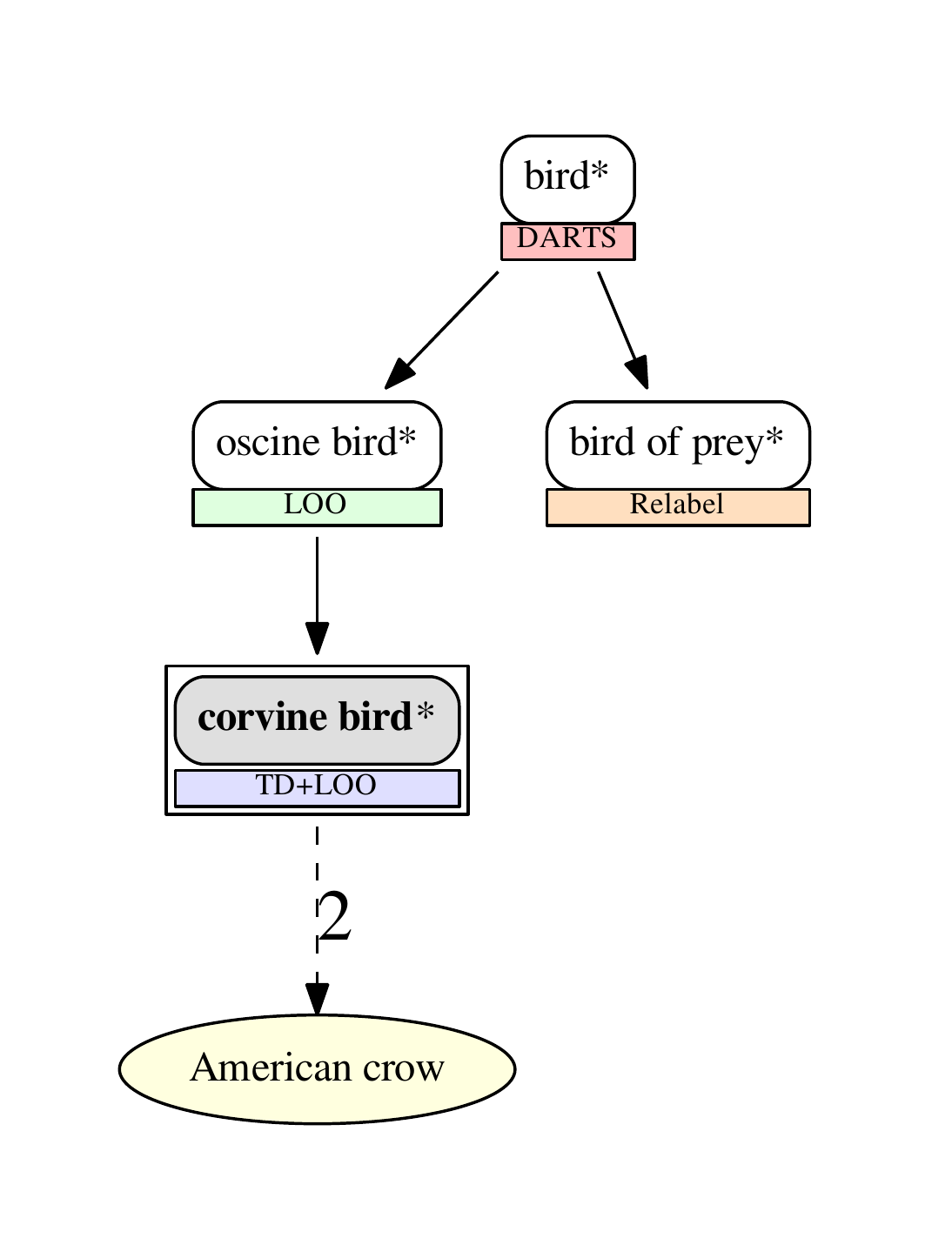}} \cr
\multicolumn{4}{c}{(e)} & 
\multicolumn{4}{c}{(f)} & 
\multicolumn{4}{c}{(g)} & 
\multicolumn{4}{c}{(h)} \cr
\multicolumn{4}{c}{\includegraphics[width=4.24cm, height=3.18cm, keepaspectratio]{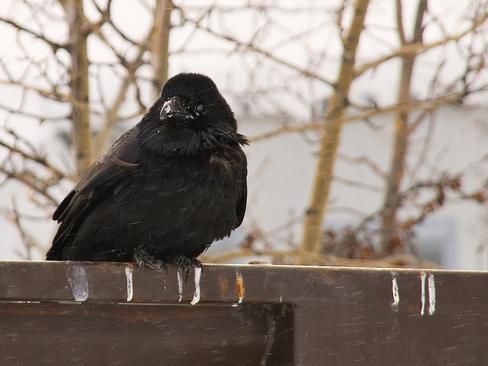}} & 
\multicolumn{4}{c}{\includegraphics[width=4.24cm, height=3.18cm, keepaspectratio]{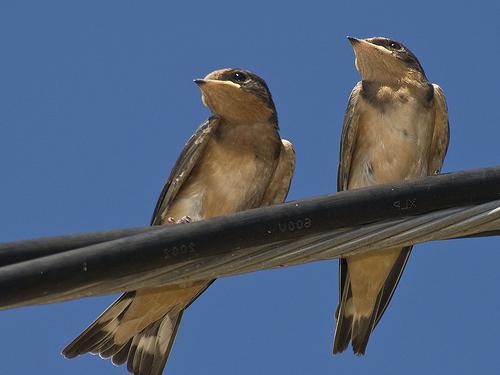}} & 
\multicolumn{4}{c}{\includegraphics[width=4.24cm, height=3.18cm, keepaspectratio]{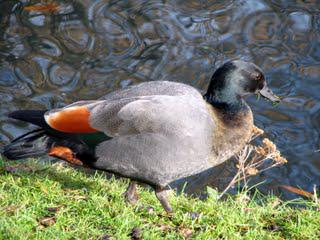}} & 
\multicolumn{4}{c}{\includegraphics[width=4.24cm, height=3.18cm, keepaspectratio]{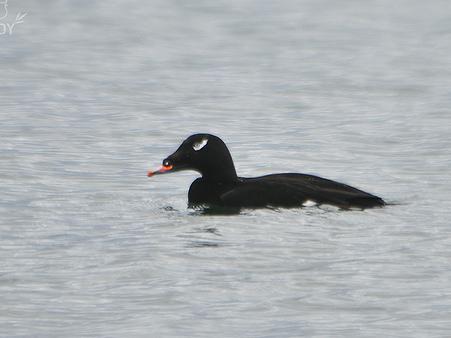}} \cr
\multicolumn{4}{l}{\cellcolor{ColorA} Novel class: raven} & 
\multicolumn{4}{l}{\cellcolor{ColorA} Novel class: swallow} & 
\multicolumn{4}{l}{\cellcolor{ColorA} Novel class: sheldrake} & 
\multicolumn{4}{l}{\cellcolor{ColorA} Novel class: scoter} \cr
\cellcolor{ColorB} Method & \cellcolor{ColorB} $\epsilon$ & \cellcolor{ColorB} A & \multicolumn{1}{c}{\cellcolor{ColorB} Word} & 
\cellcolor{ColorB} Method & \cellcolor{ColorB} $\epsilon$ & \cellcolor{ColorB} A & \multicolumn{1}{c}{\cellcolor{ColorB} Word} & 
\cellcolor{ColorB} Method & \cellcolor{ColorB} $\epsilon$ & \cellcolor{ColorB} A & \multicolumn{1}{c}{\cellcolor{ColorB} Word} & 
\cellcolor{ColorB} Method & \cellcolor{ColorB} $\epsilon$ & \cellcolor{ColorB} A & \multicolumn{1}{c}{\cellcolor{ColorB} Word} \cr
\cellcolor{Color0} GT & \cellcolor{Color0} & \cellcolor{Color0} & \cellcolor{Color0} corvine bird & 
\cellcolor{Color0} GT & \cellcolor{Color0} & \cellcolor{Color0} & \cellcolor{Color0} oscine bird & 
\cellcolor{Color0} GT & \cellcolor{Color0} & \cellcolor{Color0} & \cellcolor{Color0} duck & 
\cellcolor{Color0} GT & \cellcolor{Color0} & \cellcolor{Color0} & \cellcolor{Color0} duck \cr
\cellcolor{Color1} DARTS & \cellcolor{Color1} 0 & \cellcolor{Color1} Y & \cellcolor{Color1} corvine bird & 
\cellcolor{Color1} DARTS & \cellcolor{Color1} 0 & \cellcolor{Color1} Y & \cellcolor{Color1} oscine bird & 
\cellcolor{Color1} DARTS & \cellcolor{Color1} 4 & \cellcolor{Color1} N & \cellcolor{Color1} American coot & 
\cellcolor{Color1} DARTS & \cellcolor{Color1} 4 & \cellcolor{Color1} N & \cellcolor{Color1} American coot \cr
\cellcolor{Color2} Relabel & \cellcolor{Color2} 2 & \cellcolor{Color2} Y & \cellcolor{Color2} bird & 
\cellcolor{Color2} Relabel & \cellcolor{Color2} 1 & \cellcolor{Color2} Y & \cellcolor{Color2} bird & 
\cellcolor{Color2} Relabel & \cellcolor{Color2} 2 & \cellcolor{Color2} Y & \cellcolor{Color2} aquatic bird & 
\cellcolor{Color2} Relabel & \cellcolor{Color2} 2 & \cellcolor{Color2} Y & \cellcolor{Color2} aquatic bird \cr
\cellcolor{Color3} LOO & \cellcolor{Color3} 1 & \cellcolor{Color3} Y & \cellcolor{Color3} oscine bird & 
\cellcolor{Color3} LOO & \cellcolor{Color3} 1 & \cellcolor{Color3} N & \cellcolor{Color3} finch & 
\cellcolor{Color3} LOO & \cellcolor{Color3} 1 & \cellcolor{Color3} Y & \cellcolor{Color3} anseriform bird & 
\cellcolor{Color3} LOO & \cellcolor{Color3} 1 & \cellcolor{Color3} Y & \cellcolor{Color3} anseriform bird \cr
\cellcolor{Color4} TD+LOO & \cellcolor{Color4} 2 & \cellcolor{Color4} N & \cellcolor{Color4} thrush & 
\cellcolor{Color4} TD+LOO & \cellcolor{Color4} 3 & \cellcolor{Color4} N & \cellcolor{Color4} kite & 
\cellcolor{Color4} TD+LOO & \cellcolor{Color4} 0 & \cellcolor{Color4} Y & \cellcolor{Color4} duck & 
\cellcolor{Color4} TD+LOO & \cellcolor{Color4} 0 & \cellcolor{Color4} Y & \cellcolor{Color4} duck \cr
\multicolumn{4}{c}{\includegraphics[width=4.24cm, height=3.6cm, keepaspectratio]{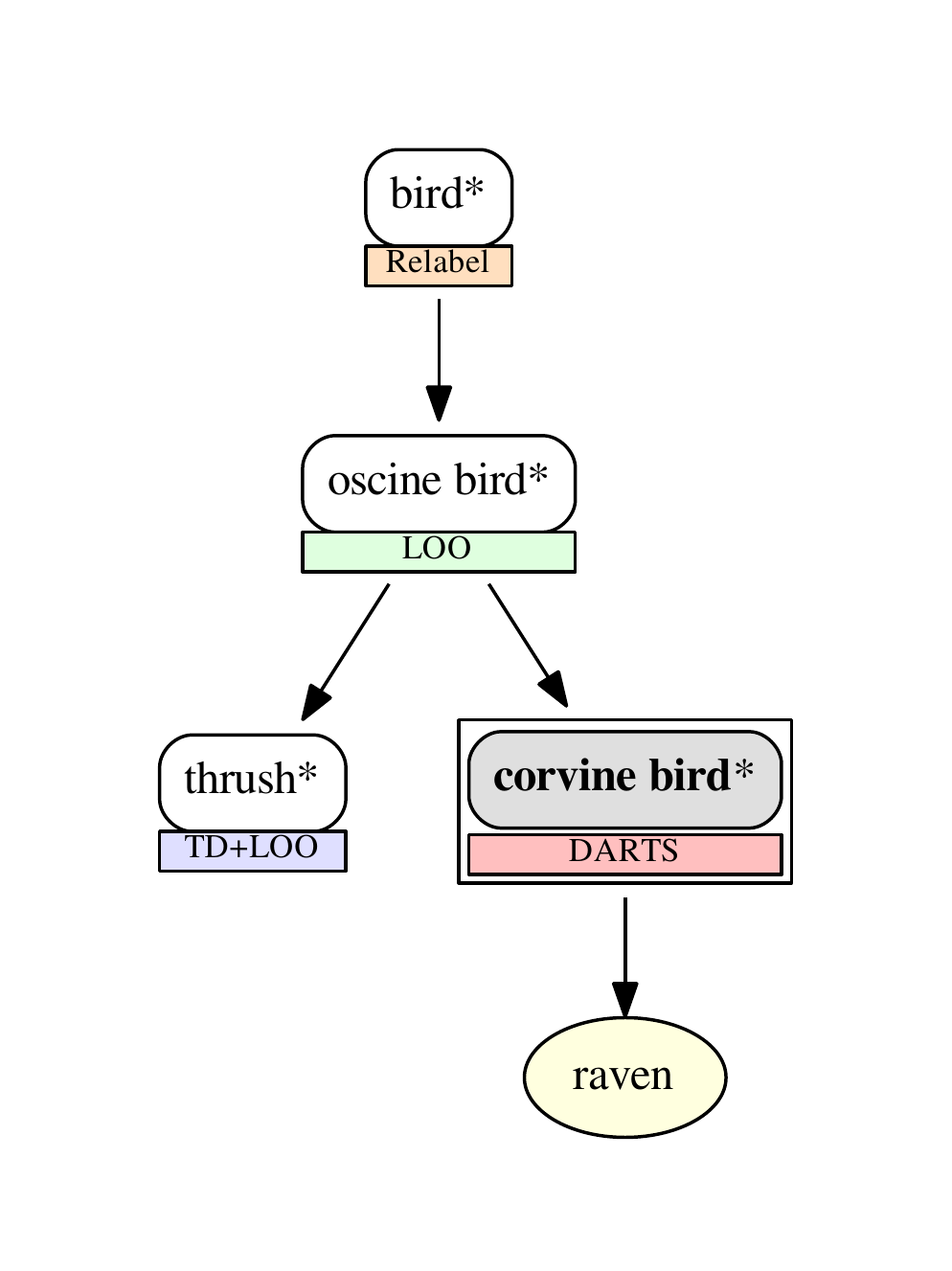}} & 
\multicolumn{4}{c}{\includegraphics[width=4.24cm, height=3.6cm, keepaspectratio]{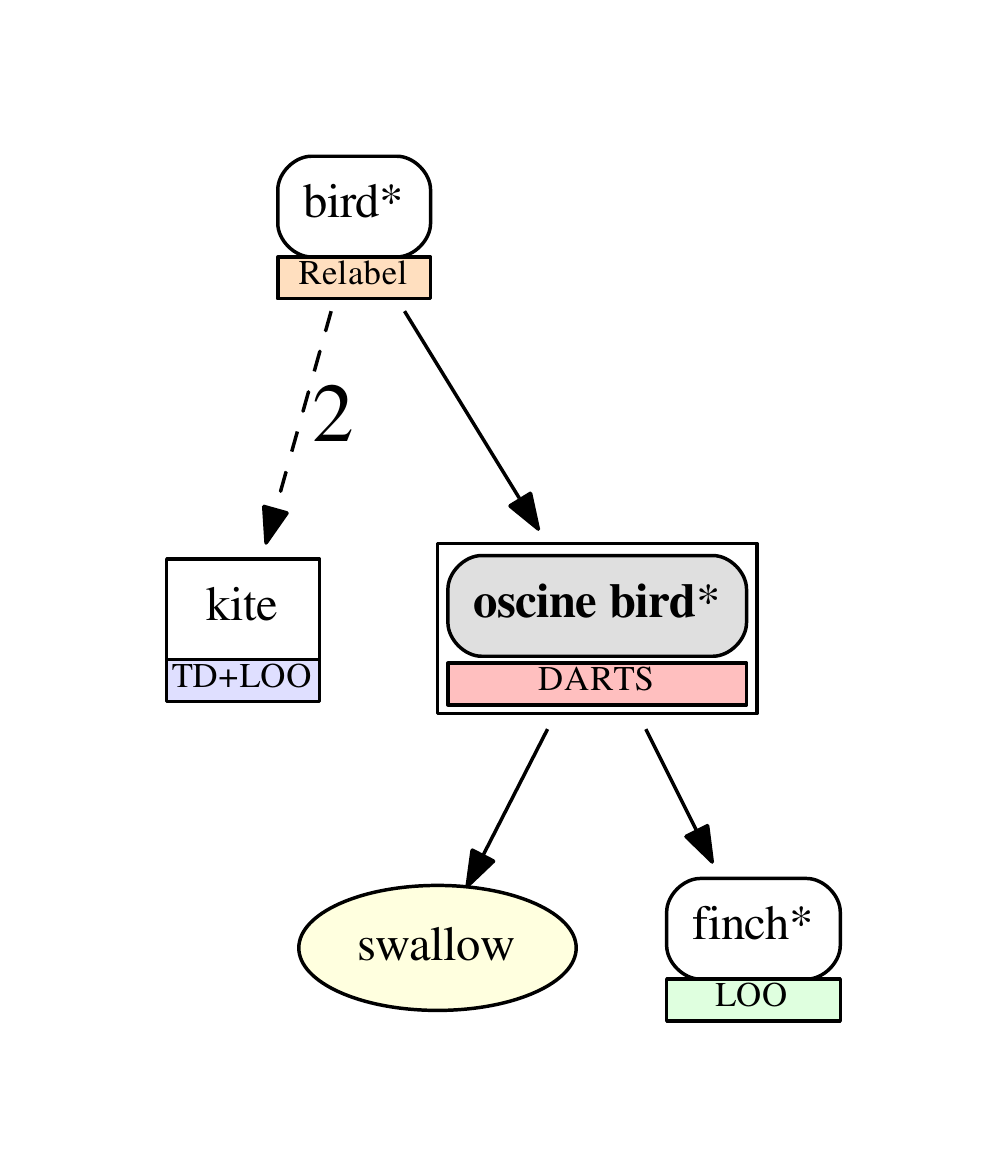}} & 
\multicolumn{4}{c}{\includegraphics[width=4.24cm, height=3.6cm, keepaspectratio]{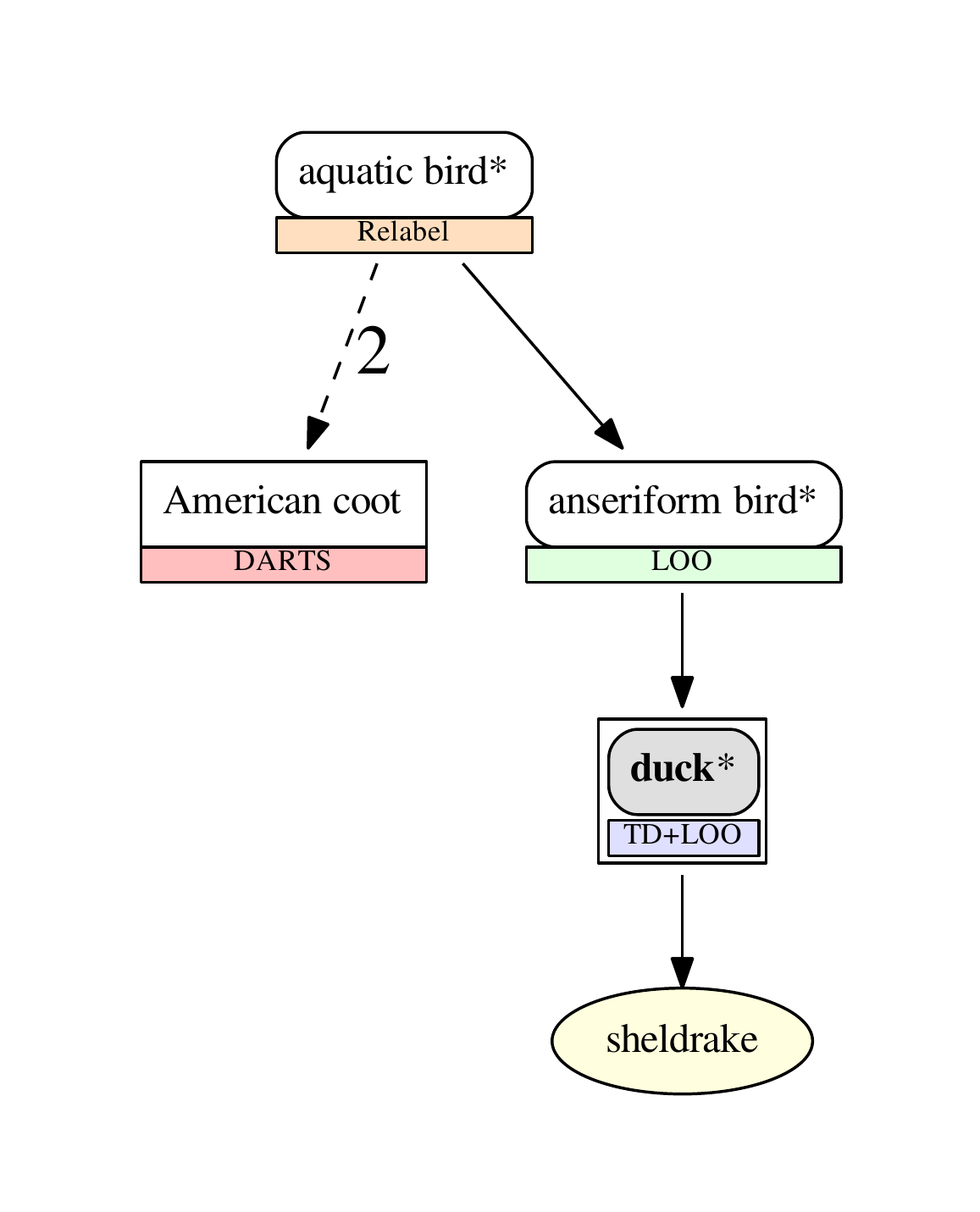}} & 
\multicolumn{4}{c}{\includegraphics[width=4.24cm, height=3.6cm, keepaspectratio]{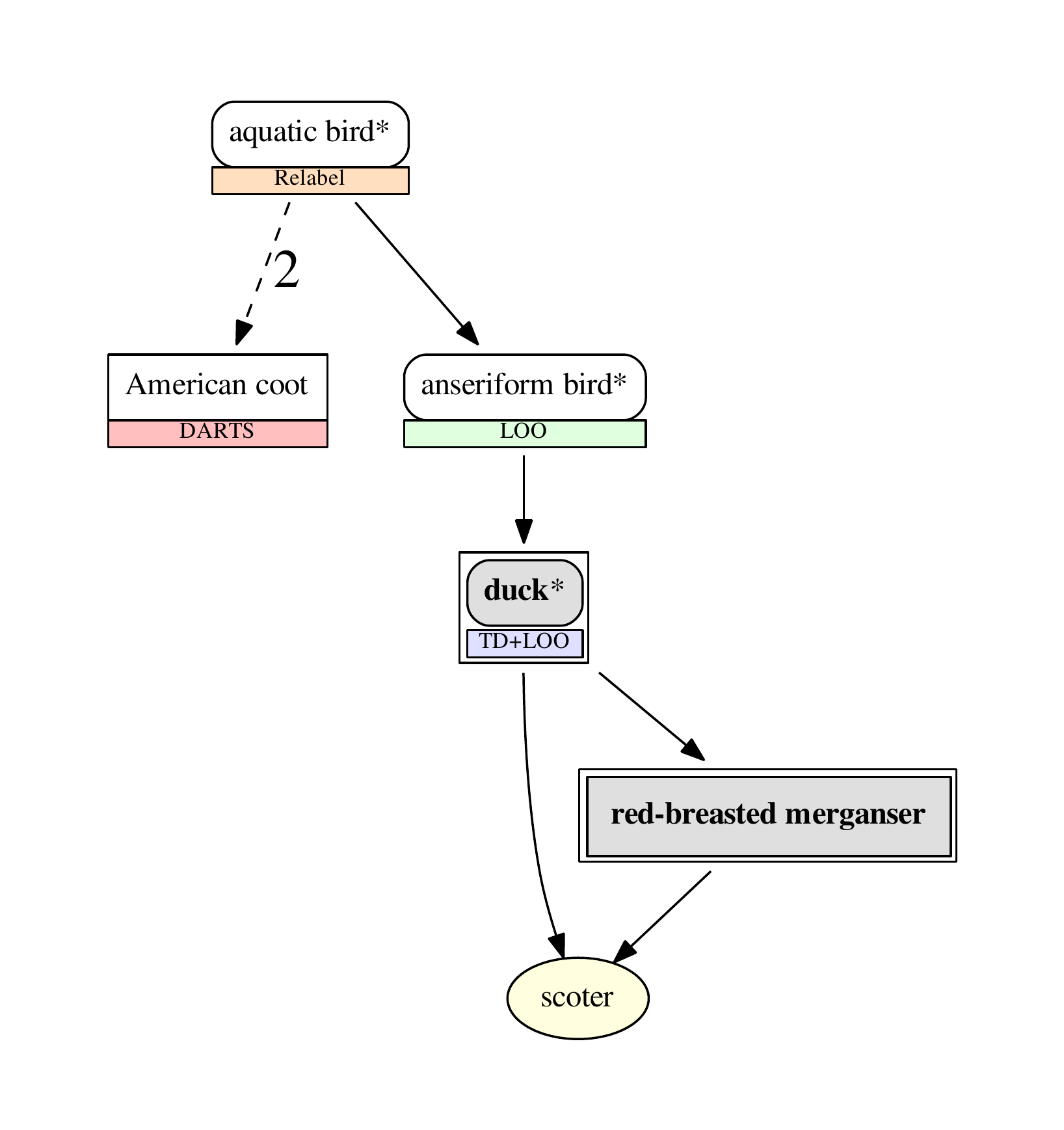}} \cr
\end{tabular}
\caption{Qualitative results of hierarchical novelty detection on ImageNet.
``GT'' is the closest known ancestor of the novel class, which is the expected prediction,
``DARTS'' is the baseline method proposed in \cite{deng2012hedging} where we modify the method for our purpose, and the others are our proposed methods.
``$\epsilon$'' is the distance between the prediction and GT,
``A'' indicates whether the prediction is an ancestor of GT, and
``Word'' is the English word of the predicted label.
Dashed edges represent multi-hop connection, where the number indicates the number of edges between classes.
If the prediction is on a super class (marked with * and rounded), then the test image is classified as a novel class whose closest class in the taxonomy is the super class.
}
\label{fig:qual_smp_3}
\end{figure*}

%% file: qual_smp/qual_smp_4.tex
\begin{figure*}[t]
\footnotesize\centering\setlength{\tabcolsep}{0cm}
\begin{tabular}{
>{\centering}m{1.12cm}>{\centering}m{0.4cm}>{\centering}m{0.4cm}m{2.32cm}
>{\centering}m{1.12cm}>{\centering}m{0.4cm}>{\centering}m{0.4cm}m{2.32cm}
>{\centering}m{1.12cm}>{\centering}m{0.4cm}>{\centering}m{0.4cm}m{2.32cm}
>{\centering}m{1.12cm}>{\centering}m{0.4cm}>{\centering}m{0.4cm}m{2.32cm}
}
\multicolumn{4}{c}{(a)} & 
\multicolumn{4}{c}{(b)} & 
\multicolumn{4}{c}{(c)} & 
\multicolumn{4}{c}{(d)} \cr
\multicolumn{4}{c}{\includegraphics[width=4.24cm, height=3.18cm, keepaspectratio]{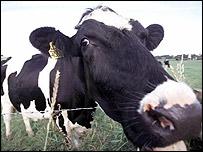}} & 
\multicolumn{4}{c}{\includegraphics[width=4.24cm, height=3.18cm, keepaspectratio]{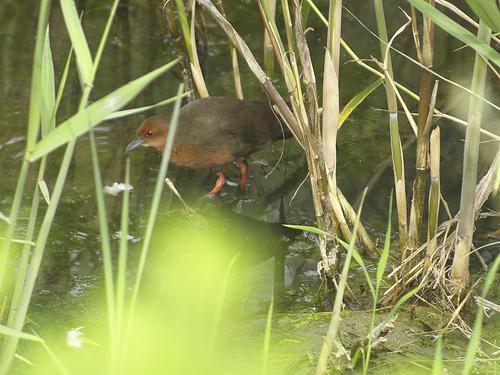}} & 
\multicolumn{4}{c}{\includegraphics[width=4.24cm, height=3.18cm, keepaspectratio]{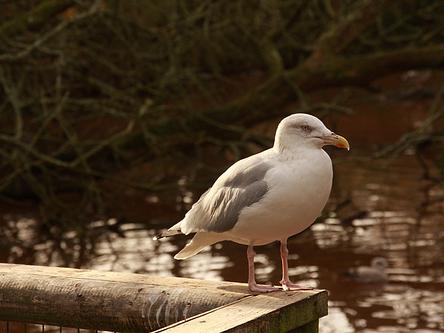}} & 
\multicolumn{4}{c}{\includegraphics[width=4.24cm, height=3.18cm, keepaspectratio]{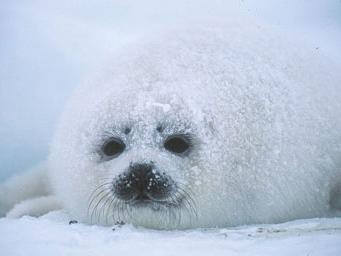}} \cr
\multicolumn{4}{l}{\cellcolor{ColorA} Novel class: cow} & 
\multicolumn{4}{l}{\cellcolor{ColorA} Novel class: crake} & 
\multicolumn{4}{l}{\cellcolor{ColorA} Novel class: gull} & 
\multicolumn{4}{l}{\cellcolor{ColorA} Novel class: harp seal} \cr
\cellcolor{ColorB} Method & \cellcolor{ColorB} $\epsilon$ & \cellcolor{ColorB} A & \multicolumn{1}{c}{\cellcolor{ColorB} Word} & 
\cellcolor{ColorB} Method & \cellcolor{ColorB} $\epsilon$ & \cellcolor{ColorB} A & \multicolumn{1}{c}{\cellcolor{ColorB} Word} & 
\cellcolor{ColorB} Method & \cellcolor{ColorB} $\epsilon$ & \cellcolor{ColorB} A & \multicolumn{1}{c}{\cellcolor{ColorB} Word} & 
\cellcolor{ColorB} Method & \cellcolor{ColorB} $\epsilon$ & \cellcolor{ColorB} A & \multicolumn{1}{c}{\cellcolor{ColorB} Word} \cr
\cellcolor{Color0} GT & \cellcolor{Color0} & \cellcolor{Color0} & \cellcolor{Color0} placental mammal & 
\cellcolor{Color0} GT & \cellcolor{Color0} & \cellcolor{Color0} & \cellcolor{Color0} wading bird & 
\cellcolor{Color0} GT & \cellcolor{Color0} & \cellcolor{Color0} & \cellcolor{Color0} seabird & 
\cellcolor{Color0} GT & \cellcolor{Color0} & \cellcolor{Color0} & \cellcolor{Color0} aquatic mammal \cr
\cellcolor{Color1} DARTS & \cellcolor{Color1} 4 & \cellcolor{Color1} N & \cellcolor{Color1} ox & 
\cellcolor{Color1} DARTS & \cellcolor{Color1} 2 & \cellcolor{Color1} N & \cellcolor{Color1} European gallinule & 
\cellcolor{Color1} DARTS & \cellcolor{Color1} 1 & \cellcolor{Color1} Y & \cellcolor{Color1} aquatic bird & 
\cellcolor{Color1} DARTS & \cellcolor{Color1} 3 & \cellcolor{Color1} N & \cellcolor{Color1} bear \cr
\cellcolor{Color2} Relabel & \cellcolor{Color2} 3 & \cellcolor{Color2} N & \cellcolor{Color2} bovid & 
\cellcolor{Color2} Relabel & \cellcolor{Color2} 3 & \cellcolor{Color2} Y & \cellcolor{Color2} vertebrate & 
\cellcolor{Color2} Relabel & \cellcolor{Color2} 2 & \cellcolor{Color2} N & \cellcolor{Color2} wading bird & 
\cellcolor{Color2} Relabel & \cellcolor{Color2} 1 & \cellcolor{Color2} Y & \cellcolor{Color2} placental mammal \cr
\cellcolor{Color3} LOO & \cellcolor{Color3} 1 & \cellcolor{Color3} N & \cellcolor{Color3} ungulate & 
\cellcolor{Color3} LOO & \cellcolor{Color3} 0 & \cellcolor{Color3} Y & \cellcolor{Color3} wading bird & 
\cellcolor{Color3} LOO & \cellcolor{Color3} 0 & \cellcolor{Color3} Y & \cellcolor{Color3} seabird & 
\cellcolor{Color3} LOO & \cellcolor{Color3} 2 & \cellcolor{Color3} N & \cellcolor{Color3} carnivore \cr
\cellcolor{Color4} TD+LOO & \cellcolor{Color4} 2 & \cellcolor{Color4} N & \cellcolor{Color4} equine & 
\cellcolor{Color4} TD+LOO & \cellcolor{Color4} 1 & \cellcolor{Color4} Y & \cellcolor{Color4} aquatic bird & 
\cellcolor{Color4} TD+LOO & \cellcolor{Color4} 1 & \cellcolor{Color4} N & \cellcolor{Color4} albatross & 
\cellcolor{Color4} TD+LOO & \cellcolor{Color4} 0 & \cellcolor{Color4} Y & \cellcolor{Color4} aquatic mammal \cr
\multicolumn{4}{c}{\includegraphics[width=4.24cm, height=3.6cm, keepaspectratio]{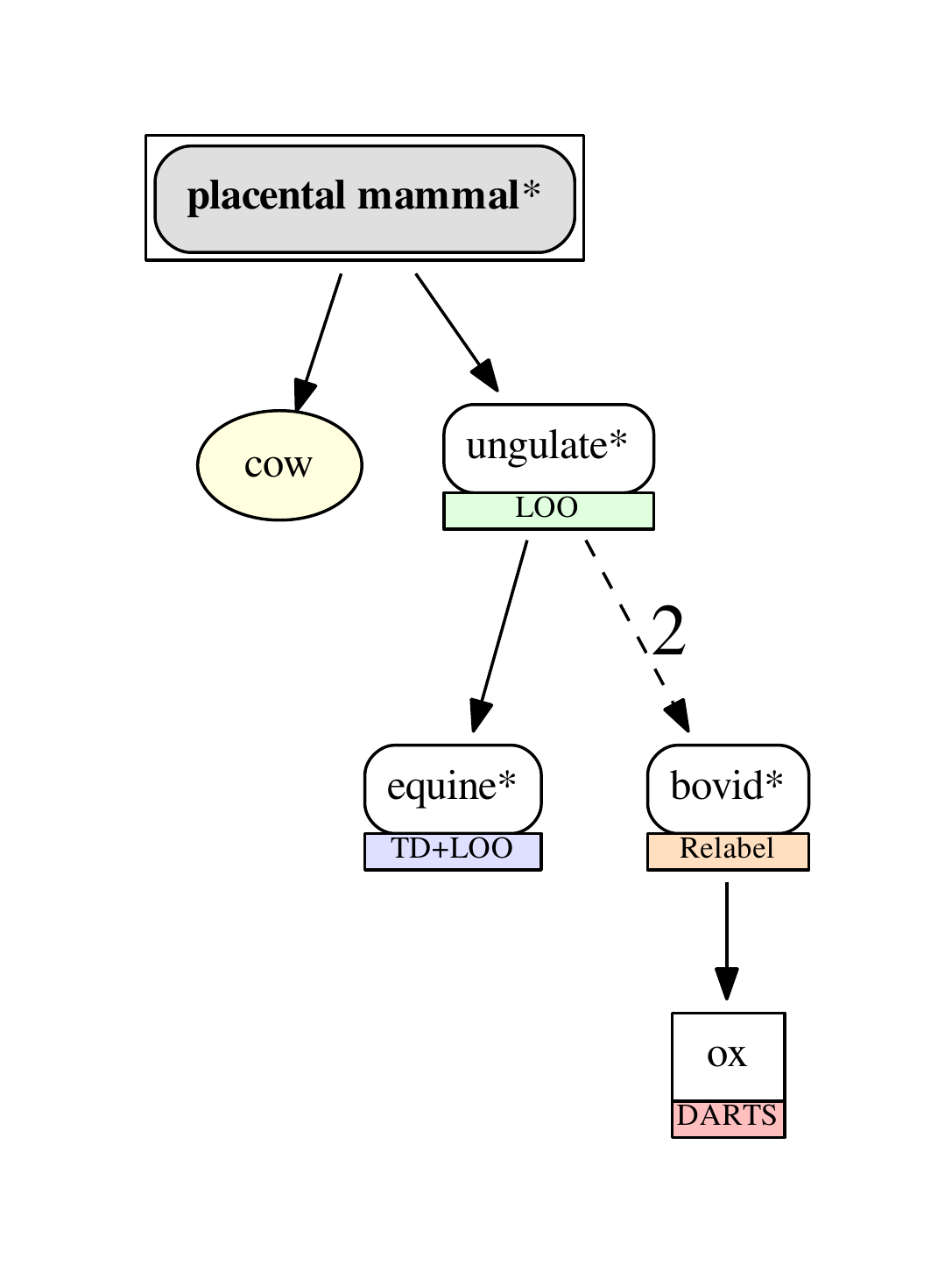}} & 
\multicolumn{4}{c}{\includegraphics[width=4.24cm, height=3.6cm, keepaspectratio]{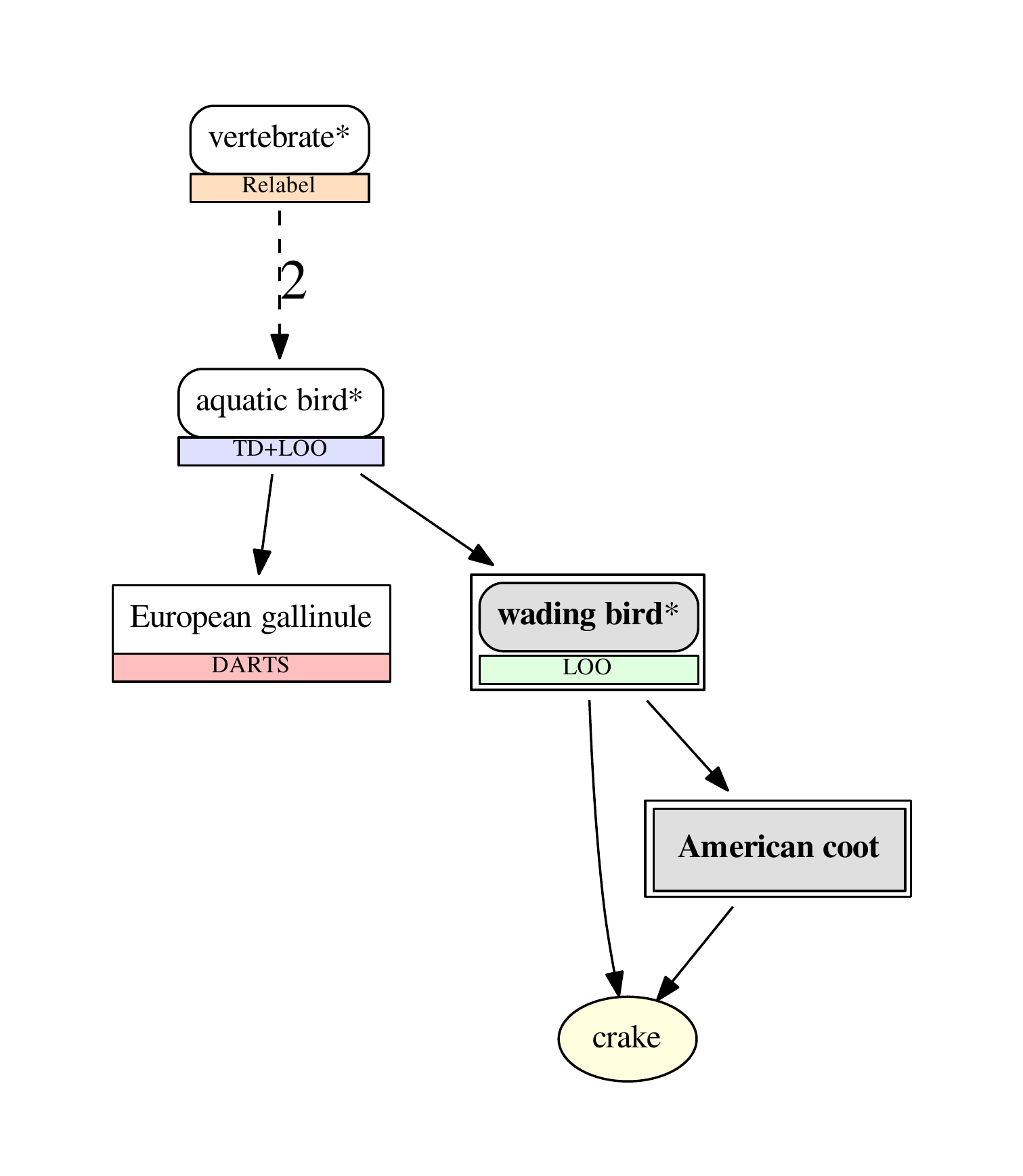}} & 
\multicolumn{4}{c}{\includegraphics[width=4.24cm, height=3.6cm, keepaspectratio]{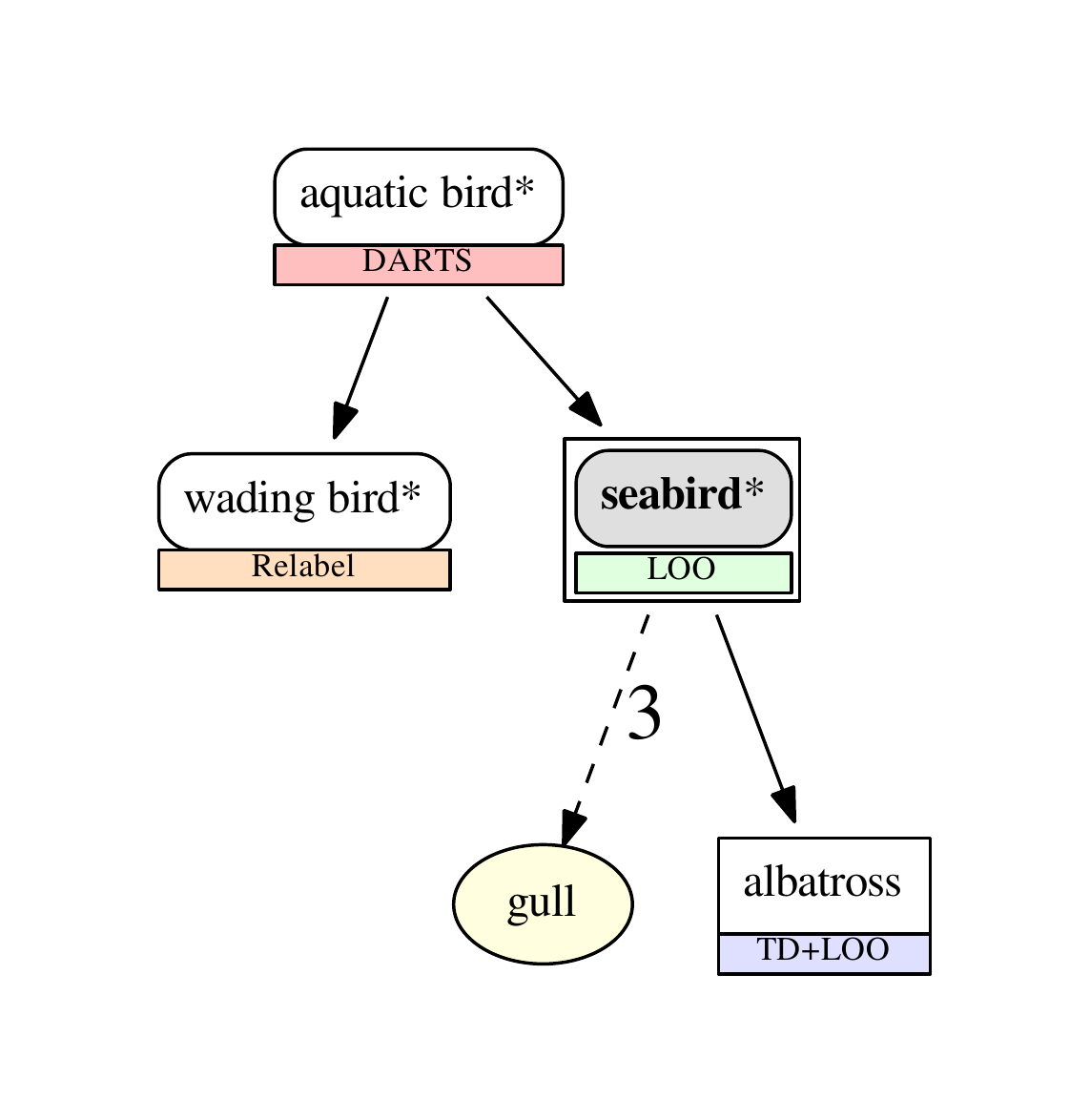}} & 
\multicolumn{4}{c}{\includegraphics[width=4.24cm, height=3.6cm, keepaspectratio]{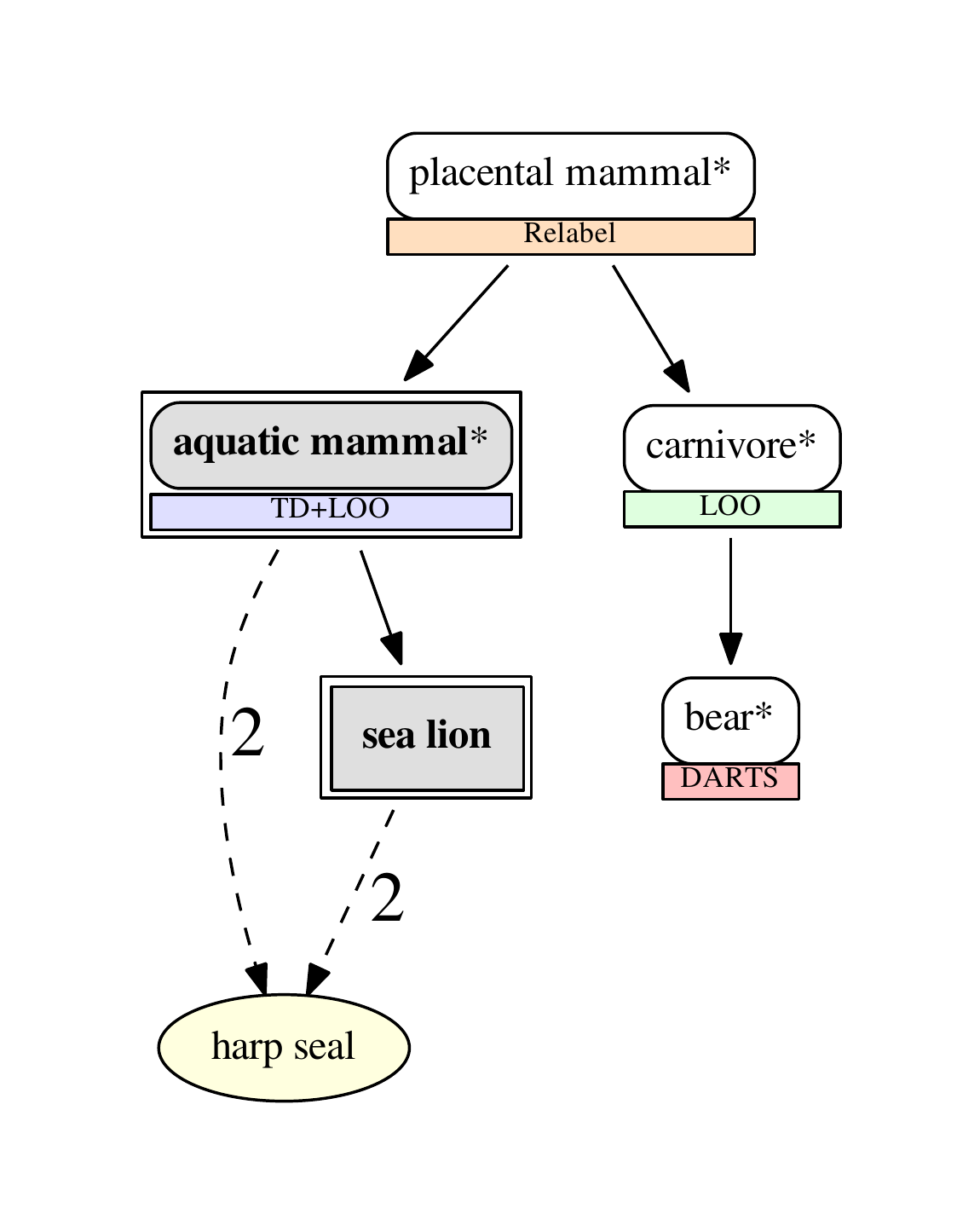}} \cr
\multicolumn{4}{c}{(e)} & 
\multicolumn{4}{c}{(f)} & 
\multicolumn{4}{c}{(g)} & 
\multicolumn{4}{c}{(h)} \cr
\multicolumn{4}{c}{\includegraphics[width=4.24cm, height=3.18cm, keepaspectratio]{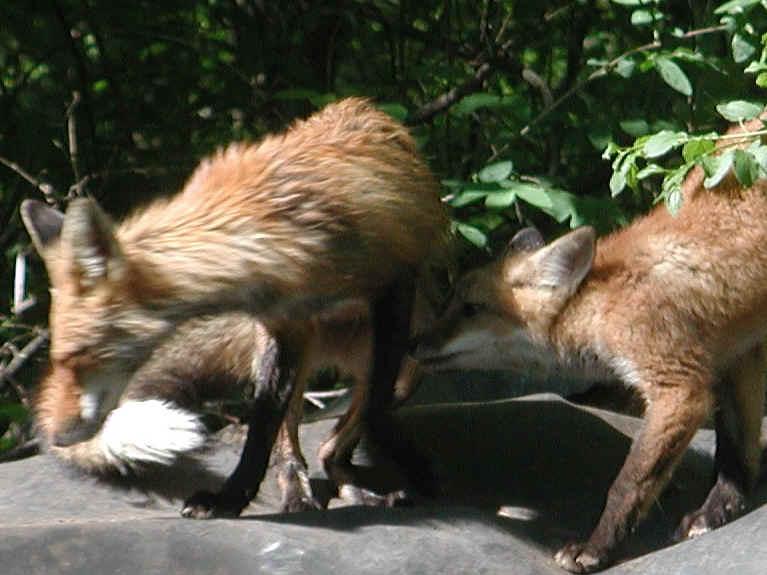}} & 
\multicolumn{4}{c}{\includegraphics[width=4.24cm, height=3.18cm, keepaspectratio]{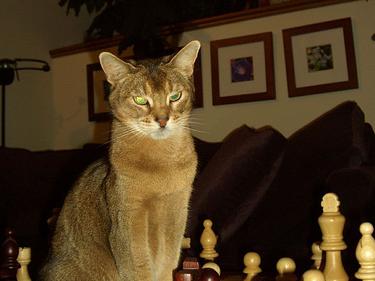}} & 
\multicolumn{4}{c}{\includegraphics[width=4.24cm, height=3.18cm, keepaspectratio]{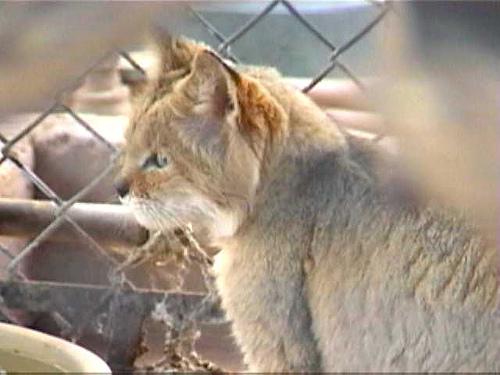}} & 
\multicolumn{4}{c}{\includegraphics[width=4.24cm, height=3.18cm, keepaspectratio]{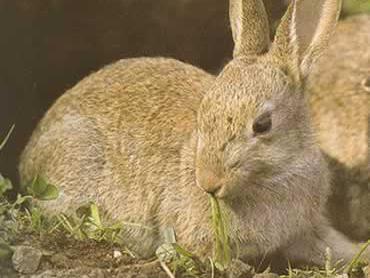}} \cr
\multicolumn{4}{l}{\cellcolor{ColorA} Novel class: {\scriptsize red fox, Vulpes fulva}} & 
\multicolumn{4}{l}{\cellcolor{ColorA} Novel class: Abyssinian cat} & 
\multicolumn{4}{l}{\cellcolor{ColorA} Novel class: sand cat} & 
\multicolumn{4}{l}{\cellcolor{ColorA} Novel class: European rabbit} \cr
\cellcolor{ColorB} Method & \cellcolor{ColorB} $\epsilon$ & \cellcolor{ColorB} A & \multicolumn{1}{c}{\cellcolor{ColorB} Word} & 
\cellcolor{ColorB} Method & \cellcolor{ColorB} $\epsilon$ & \cellcolor{ColorB} A & \multicolumn{1}{c}{\cellcolor{ColorB} Word} & 
\cellcolor{ColorB} Method & \cellcolor{ColorB} $\epsilon$ & \cellcolor{ColorB} A & \multicolumn{1}{c}{\cellcolor{ColorB} Word} & 
\cellcolor{ColorB} Method & \cellcolor{ColorB} $\epsilon$ & \cellcolor{ColorB} A & \multicolumn{1}{c}{\cellcolor{ColorB} Word} \cr
\cellcolor{Color0} GT & \cellcolor{Color0} & \cellcolor{Color0} & \cellcolor{Color0} fox & 
\cellcolor{Color0} GT & \cellcolor{Color0} & \cellcolor{Color0} & \cellcolor{Color0} domestic cat & 
\cellcolor{Color0} GT & \cellcolor{Color0} & \cellcolor{Color0} & \cellcolor{Color0} wildcat & 
\cellcolor{Color0} GT & \cellcolor{Color0} & \cellcolor{Color0} & \cellcolor{Color0} rabbit \cr
\cellcolor{Color1} DARTS & \cellcolor{Color1} 1 & \cellcolor{Color1} N & \cellcolor{Color1} {\scriptsize red fox, Vulpes vulpes} & 
\cellcolor{Color1} DARTS & \cellcolor{Color1} 1 & \cellcolor{Color1} N & \cellcolor{Color1} Egyptian cat & 
\cellcolor{Color1} DARTS & \cellcolor{Color1} 2 & \cellcolor{Color1} Y & \cellcolor{Color1} feline & 
\cellcolor{Color1} DARTS & \cellcolor{Color1} 1 & \cellcolor{Color1} Y & \cellcolor{Color1} leporid mammal \cr
\cellcolor{Color2} Relabel & \cellcolor{Color2} 1 & \cellcolor{Color2} Y & \cellcolor{Color2} canine & 
\cellcolor{Color2} Relabel & \cellcolor{Color2} 0 & \cellcolor{Color2} Y & \cellcolor{Color2} domestic cat & 
\cellcolor{Color2} Relabel & \cellcolor{Color2} 2 & \cellcolor{Color2} N & \cellcolor{Color2} domestic cat & 
\cellcolor{Color2} Relabel & \cellcolor{Color2} 1 & \cellcolor{Color2} N & \cellcolor{Color2} wood rabbit \cr
\cellcolor{Color3} LOO & \cellcolor{Color3} 0 & \cellcolor{Color3} Y & \cellcolor{Color3} fox & 
\cellcolor{Color3} LOO & \cellcolor{Color3} 1 & \cellcolor{Color3} Y & \cellcolor{Color3} cat & 
\cellcolor{Color3} LOO & \cellcolor{Color3} 1 & \cellcolor{Color3} Y & \cellcolor{Color3} cat & 
\cellcolor{Color3} LOO & \cellcolor{Color3} 0 & \cellcolor{Color3} Y & \cellcolor{Color3} rabbit \cr
\cellcolor{Color4} TD+LOO & \cellcolor{Color4} 0 & \cellcolor{Color4} Y & \cellcolor{Color4} fox & 
\cellcolor{Color4} TD+LOO & \cellcolor{Color4} 0 & \cellcolor{Color4} Y & \cellcolor{Color4} domestic cat & 
\cellcolor{Color4} TD+LOO & \cellcolor{Color4} 0 & \cellcolor{Color4} Y & \cellcolor{Color4} wildcat & 
\cellcolor{Color4} TD+LOO & \cellcolor{Color4} 0 & \cellcolor{Color4} Y & \cellcolor{Color4} rabbit \cr
\multicolumn{4}{c}{\includegraphics[width=4.24cm, height=3.6cm, keepaspectratio]{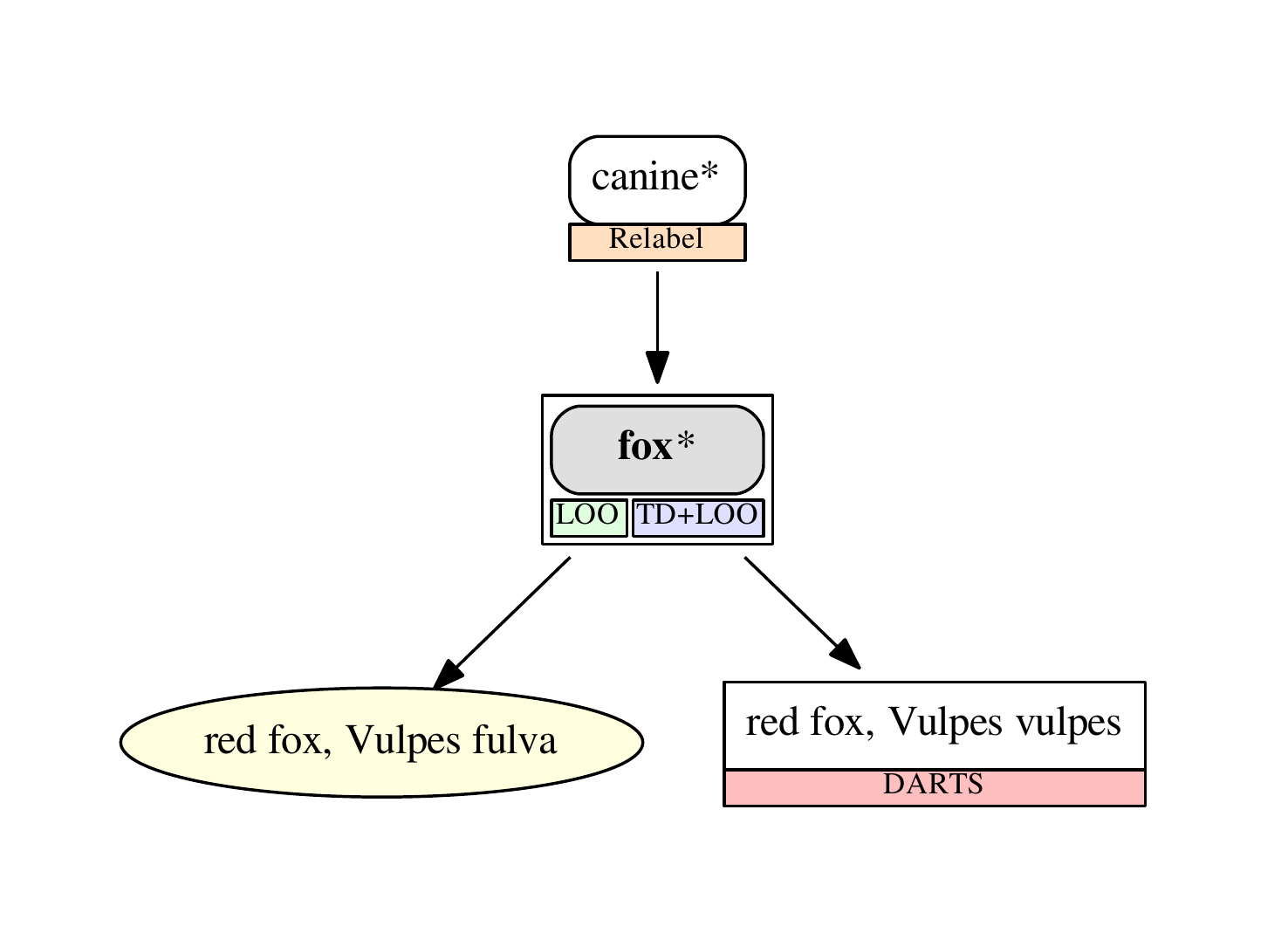}} & 
\multicolumn{4}{c}{\includegraphics[width=4.24cm, height=3.6cm, keepaspectratio]{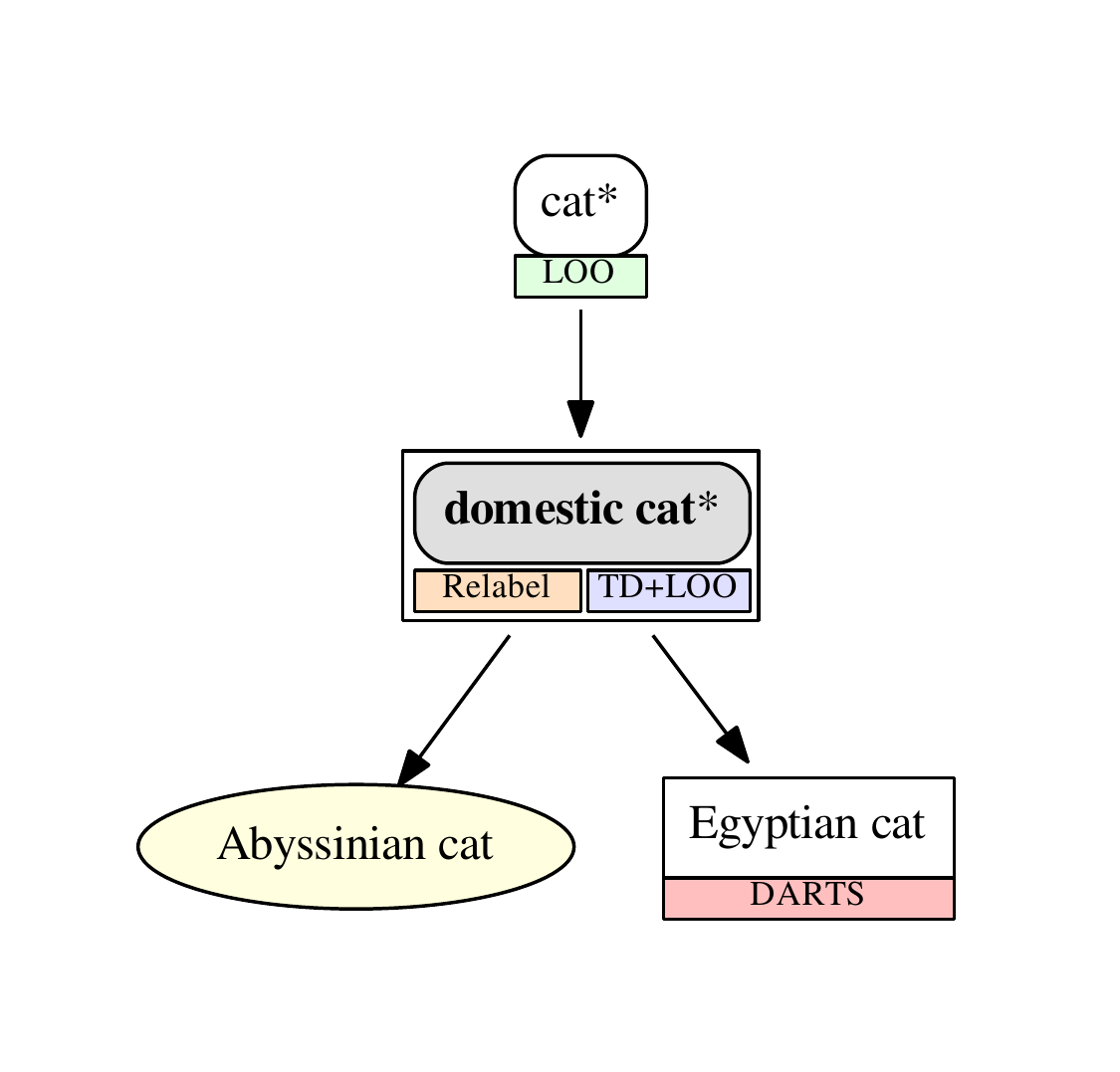}} & 
\multicolumn{4}{c}{\includegraphics[width=4.24cm, height=3.6cm, keepaspectratio]{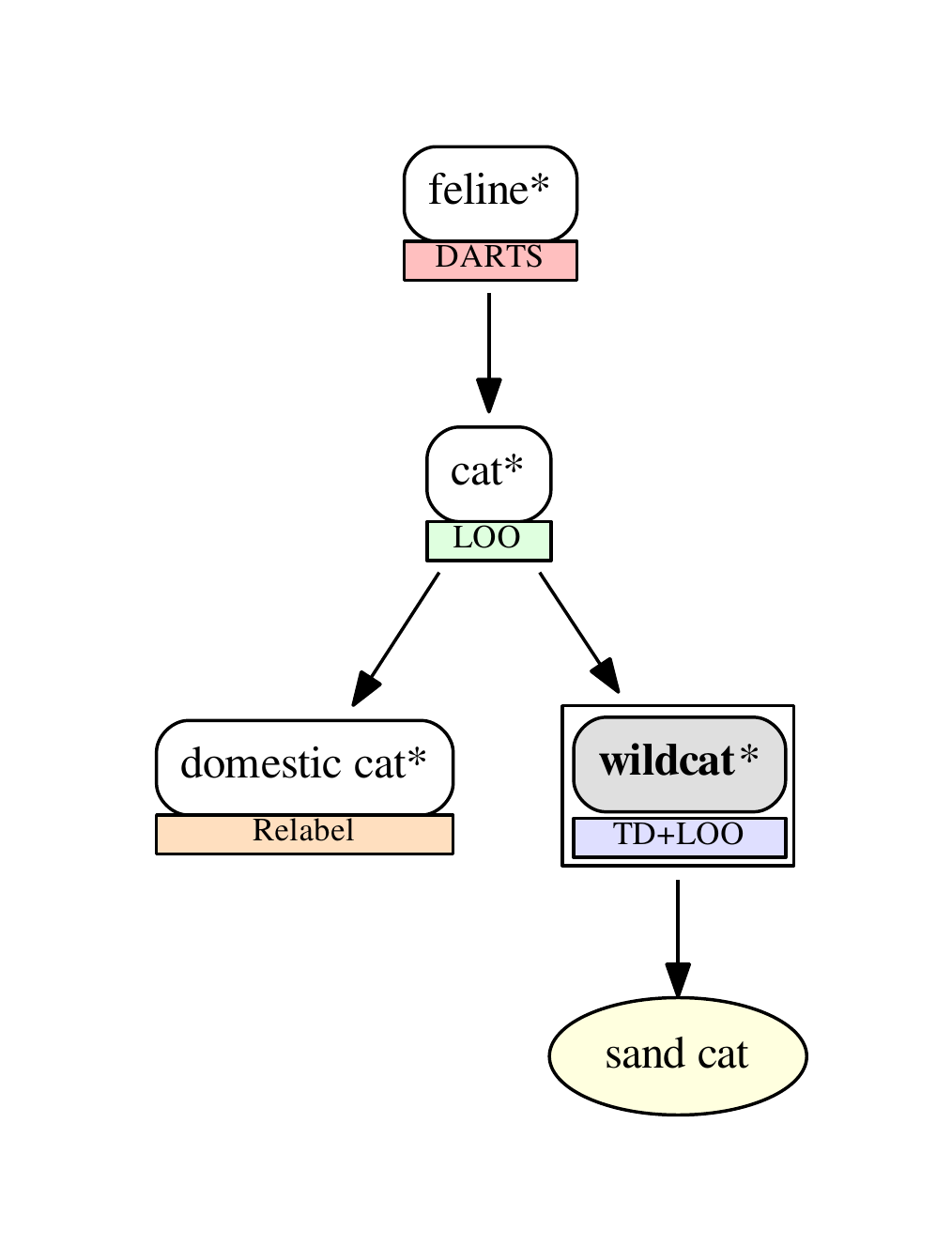}} & 
\multicolumn{4}{c}{\includegraphics[width=4.24cm, height=3.6cm, keepaspectratio]{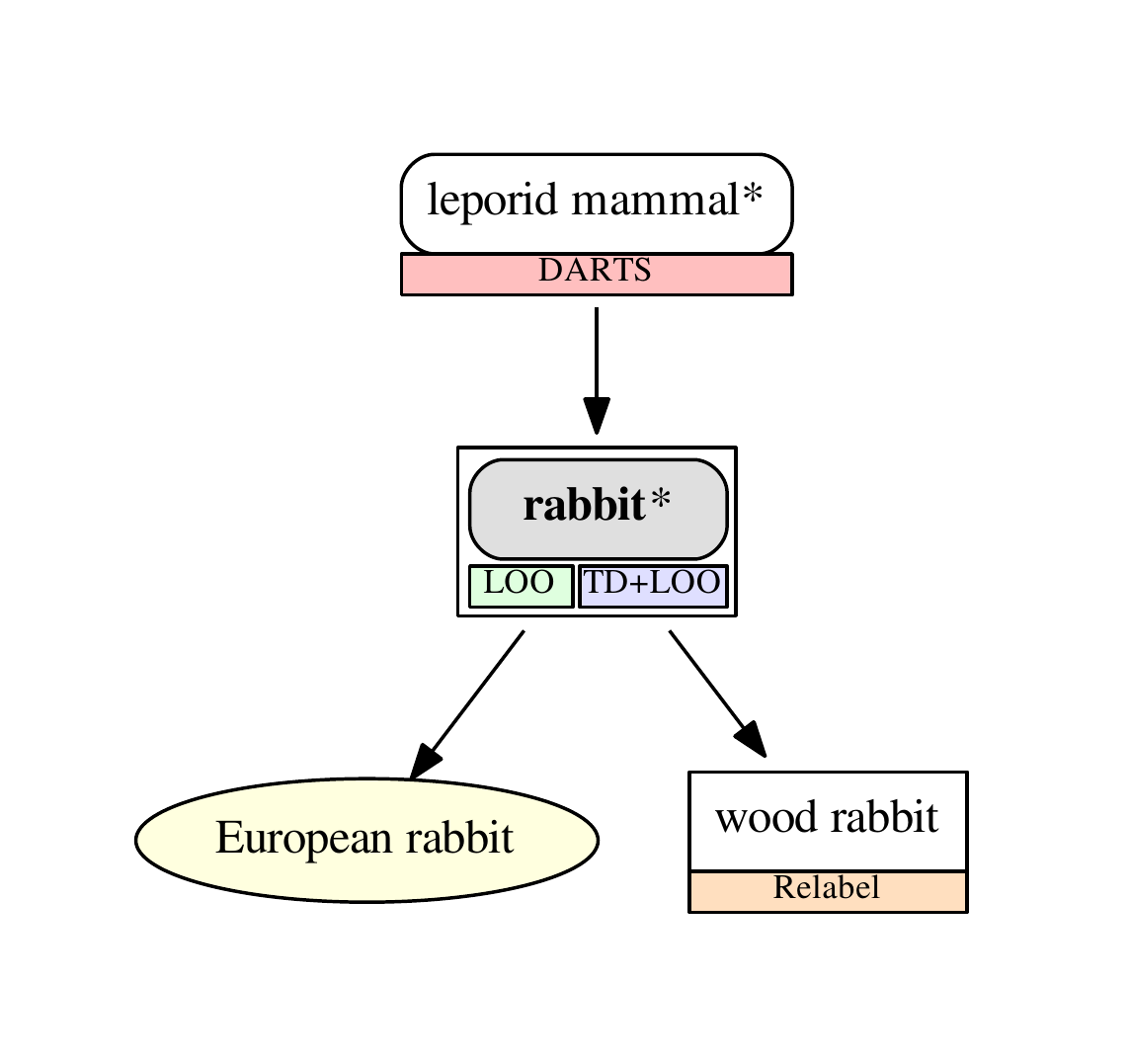}} \cr
\end{tabular}
\caption{Qualitative results of hierarchical novelty detection on ImageNet.
``GT'' is the closest known ancestor of the novel class, which is the expected prediction,
``DARTS'' is the baseline method proposed in \cite{deng2012hedging} where we modify the method for our purpose, and the others are our proposed methods.
``$\epsilon$'' is the distance between the prediction and GT,
``A'' indicates whether the prediction is an ancestor of GT, and
``Word'' is the English word of the predicted label.
Dashed edges represent multi-hop connection, where the number indicates the number of edges between classes.
If the prediction is on a super class (marked with * and rounded), then the test image is classified as a novel class whose closest class in the taxonomy is the super class.
}
\label{fig:qual_smp_4}
\end{figure*}

%% file: qual_smp/qual_smp_5.tex
\begin{figure*}[t]
\footnotesize\centering\setlength{\tabcolsep}{0cm}
\begin{tabular}{
>{\centering}m{1.12cm}>{\centering}m{0.4cm}>{\centering}m{0.4cm}m{2.32cm}
>{\centering}m{1.12cm}>{\centering}m{0.4cm}>{\centering}m{0.4cm}m{2.32cm}
>{\centering}m{1.12cm}>{\centering}m{0.4cm}>{\centering}m{0.4cm}m{2.32cm}
>{\centering}m{1.12cm}>{\centering}m{0.4cm}>{\centering}m{0.4cm}m{2.32cm}
}
\multicolumn{4}{c}{(a)} & 
\multicolumn{4}{c}{(b)} & 
\multicolumn{4}{c}{(c)} & 
\multicolumn{4}{c}{(d)} \cr
\multicolumn{4}{c}{\includegraphics[width=4.24cm, height=3.18cm, keepaspectratio]{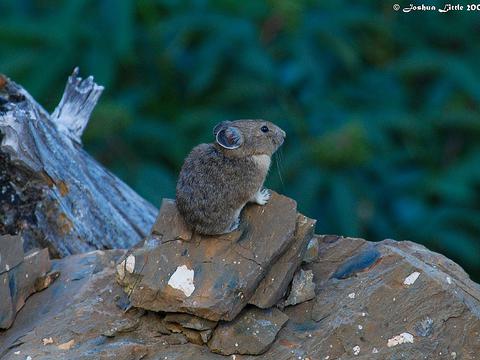}} & 
\multicolumn{4}{c}{\includegraphics[width=4.24cm, height=3.18cm, keepaspectratio]{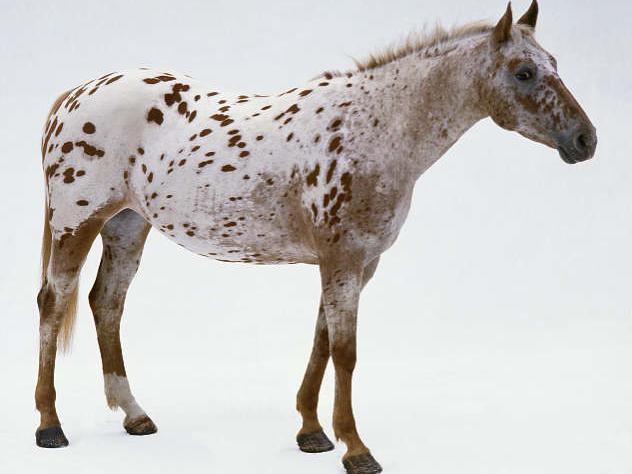}} & 
\multicolumn{4}{c}{\includegraphics[width=4.24cm, height=3.18cm, keepaspectratio]{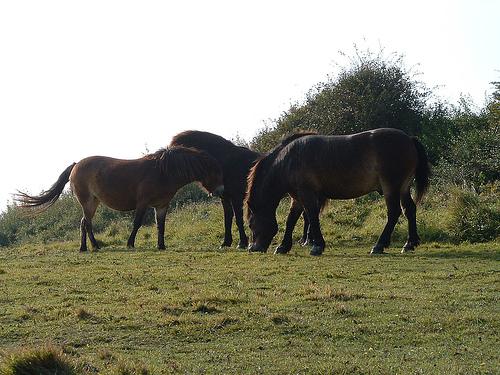}} & 
\multicolumn{4}{c}{\includegraphics[width=4.24cm, height=3.18cm, keepaspectratio]{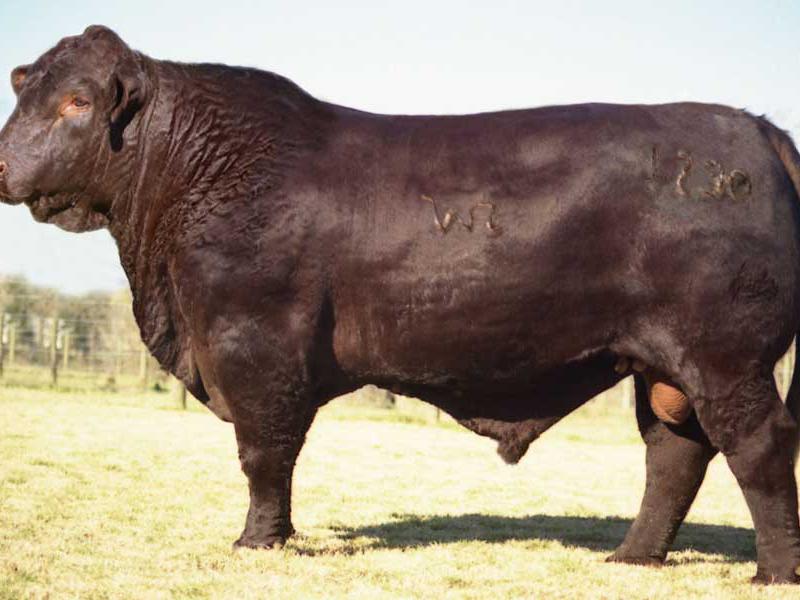}} \cr
\multicolumn{4}{l}{\cellcolor{ColorA} Novel class: pika} & 
\multicolumn{4}{l}{\cellcolor{ColorA} Novel class: Appaloosa} & 
\multicolumn{4}{l}{\cellcolor{ColorA} Novel class: Exmoor} & 
\multicolumn{4}{l}{\cellcolor{ColorA} Novel class: bull} \cr
\cellcolor{ColorB} Method & \cellcolor{ColorB} $\epsilon$ & \cellcolor{ColorB} A & \multicolumn{1}{c}{\cellcolor{ColorB} Word} & 
\cellcolor{ColorB} Method & \cellcolor{ColorB} $\epsilon$ & \cellcolor{ColorB} A & \multicolumn{1}{c}{\cellcolor{ColorB} Word} & 
\cellcolor{ColorB} Method & \cellcolor{ColorB} $\epsilon$ & \cellcolor{ColorB} A & \multicolumn{1}{c}{\cellcolor{ColorB} Word} & 
\cellcolor{ColorB} Method & \cellcolor{ColorB} $\epsilon$ & \cellcolor{ColorB} A & \multicolumn{1}{c}{\cellcolor{ColorB} Word} \cr
\cellcolor{Color0} GT & \cellcolor{Color0} & \cellcolor{Color0} & \cellcolor{Color0} placental mammal & 
\cellcolor{Color0} GT & \cellcolor{Color0} & \cellcolor{Color0} & \cellcolor{Color0} equine & 
\cellcolor{Color0} GT & \cellcolor{Color0} & \cellcolor{Color0} & \cellcolor{Color0} equine & 
\cellcolor{Color0} GT & \cellcolor{Color0} & \cellcolor{Color0} & \cellcolor{Color0} bovid \cr
\cellcolor{Color1} DARTS & \cellcolor{Color1} 2 & \cellcolor{Color1} N & \cellcolor{Color1} marmot & 
\cellcolor{Color1} DARTS & \cellcolor{Color1} 3 & \cellcolor{Color1} N & \cellcolor{Color1} bovid & 
\cellcolor{Color1} DARTS & \cellcolor{Color1} 4 & \cellcolor{Color1} N & \cellcolor{Color1} warthog & 
\cellcolor{Color1} DARTS & \cellcolor{Color1} 0 & \cellcolor{Color1} Y & \cellcolor{Color1} ox \cr
\cellcolor{Color2} Relabel & \cellcolor{Color2} 1 & \cellcolor{Color2} N & \cellcolor{Color2} rodent & 
\cellcolor{Color2} Relabel & \cellcolor{Color2} 2 & \cellcolor{Color2} N & \cellcolor{Color2} even-toed ungulate & 
\cellcolor{Color2} Relabel & \cellcolor{Color2} 2 & \cellcolor{Color2} N & \cellcolor{Color2} even-toed ungulate & 
\cellcolor{Color2} Relabel & \cellcolor{Color2} 0 & \cellcolor{Color2} Y & \cellcolor{Color2} bovid \cr
\cellcolor{Color3} LOO & \cellcolor{Color3} 1 & \cellcolor{Color3} N & \cellcolor{Color3} rodent & 
\cellcolor{Color3} LOO & \cellcolor{Color3} 1 & \cellcolor{Color3} Y & \cellcolor{Color3} ungulate & 
\cellcolor{Color3} LOO & \cellcolor{Color3} 1 & \cellcolor{Color3} Y & \cellcolor{Color3} ungulate & 
\cellcolor{Color3} LOO & \cellcolor{Color3} 2 & \cellcolor{Color3} Y & \cellcolor{Color3} ungulate \cr
\cellcolor{Color4} TD+LOO & \cellcolor{Color4} 0 & \cellcolor{Color4} Y & \cellcolor{Color4} leporid mammal & 
\cellcolor{Color4} TD+LOO & \cellcolor{Color4} 0 & \cellcolor{Color4} Y & \cellcolor{Color4} equine & 
\cellcolor{Color4} TD+LOO & \cellcolor{Color4} 0 & \cellcolor{Color4} Y & \cellcolor{Color4} equine & 
\cellcolor{Color4} TD+LOO & \cellcolor{Color4} 1 & \cellcolor{Color4} N & \cellcolor{Color4} bison \cr
\multicolumn{4}{c}{\includegraphics[width=4.24cm, height=3.6cm, keepaspectratio]{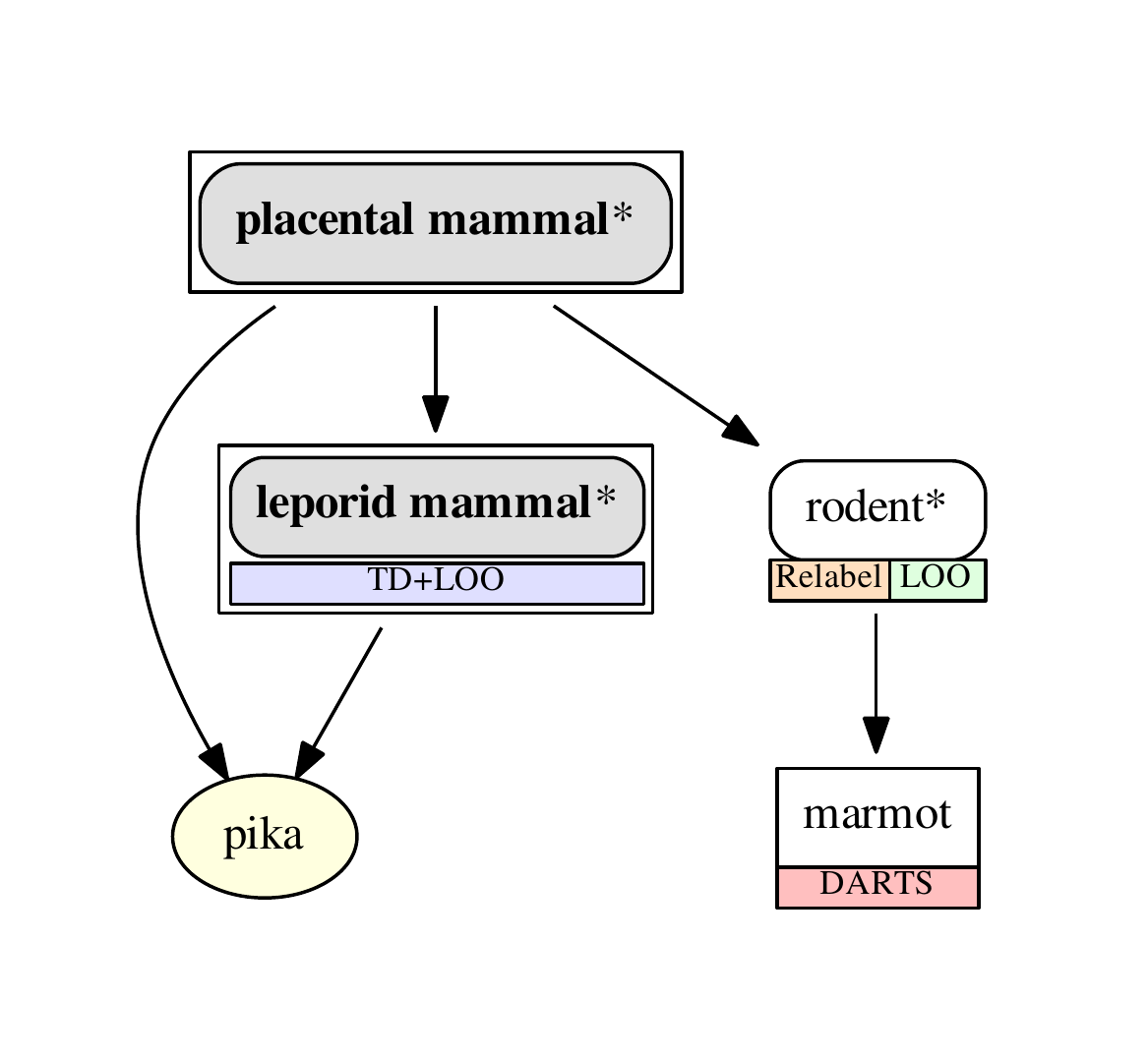}} & 
\multicolumn{4}{c}{\includegraphics[width=4.24cm, height=3.6cm, keepaspectratio]{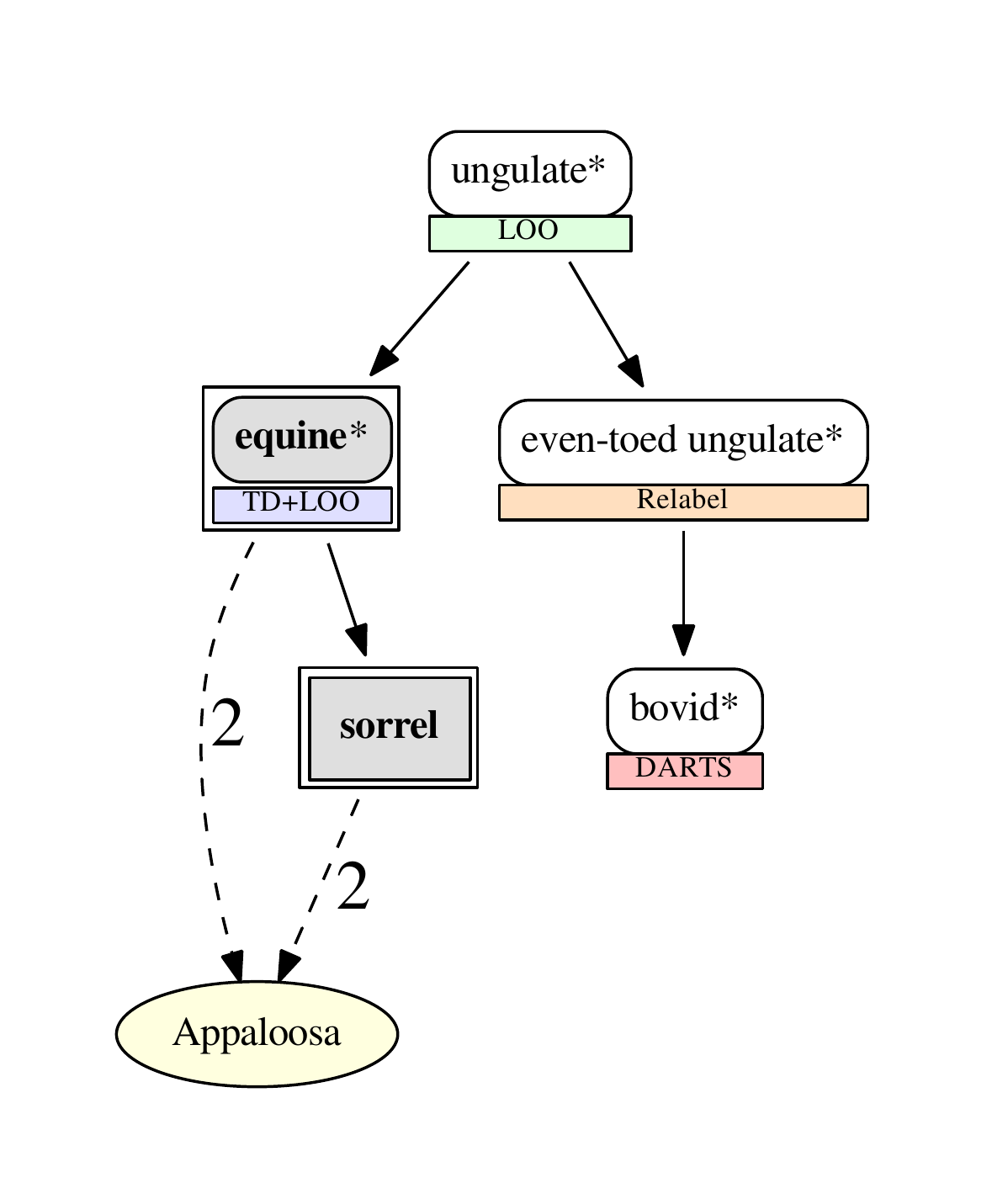}} & 
\multicolumn{4}{c}{\includegraphics[width=4.24cm, height=3.6cm, keepaspectratio]{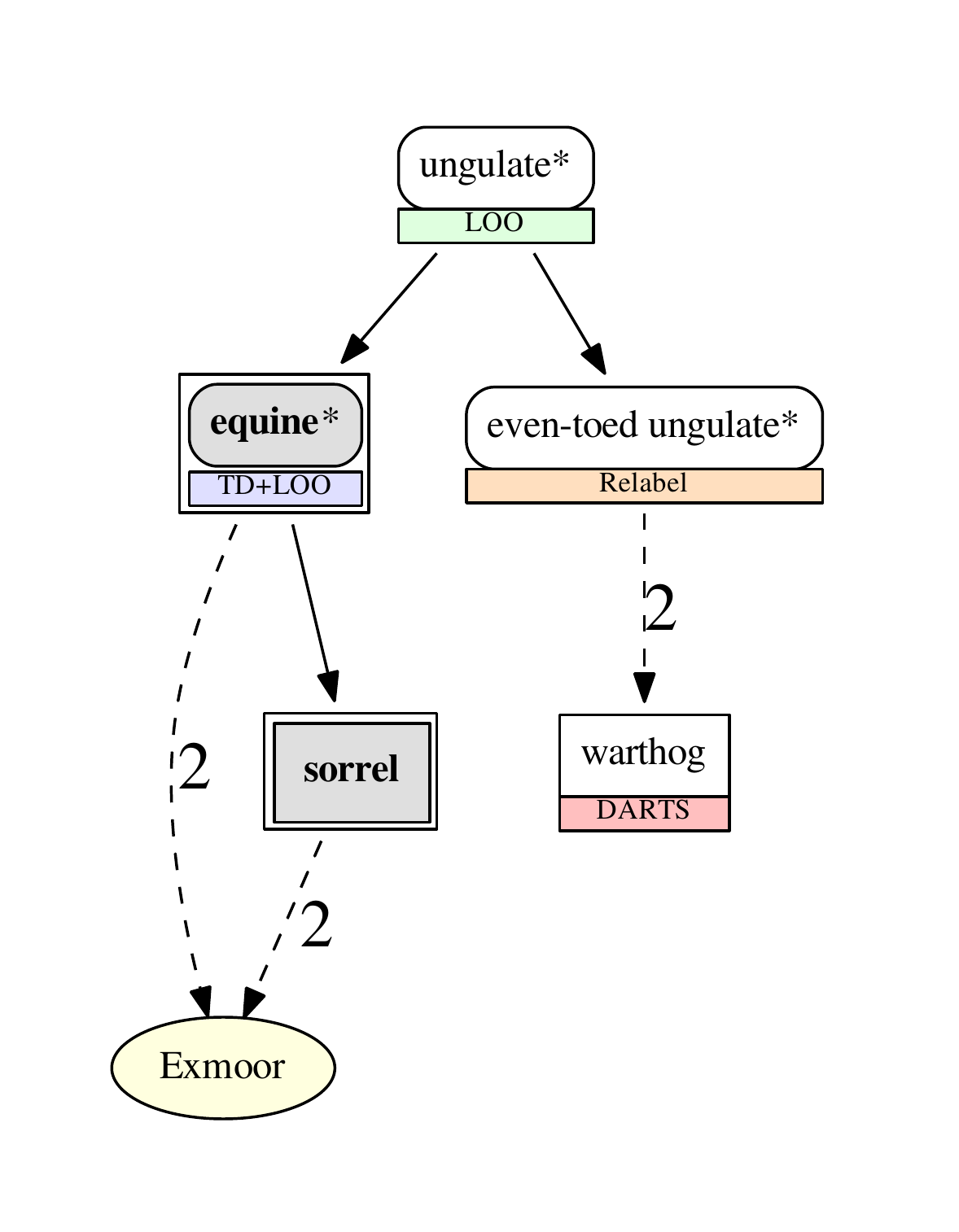}} & 
\multicolumn{4}{c}{\includegraphics[width=4.24cm, height=3.6cm, keepaspectratio]{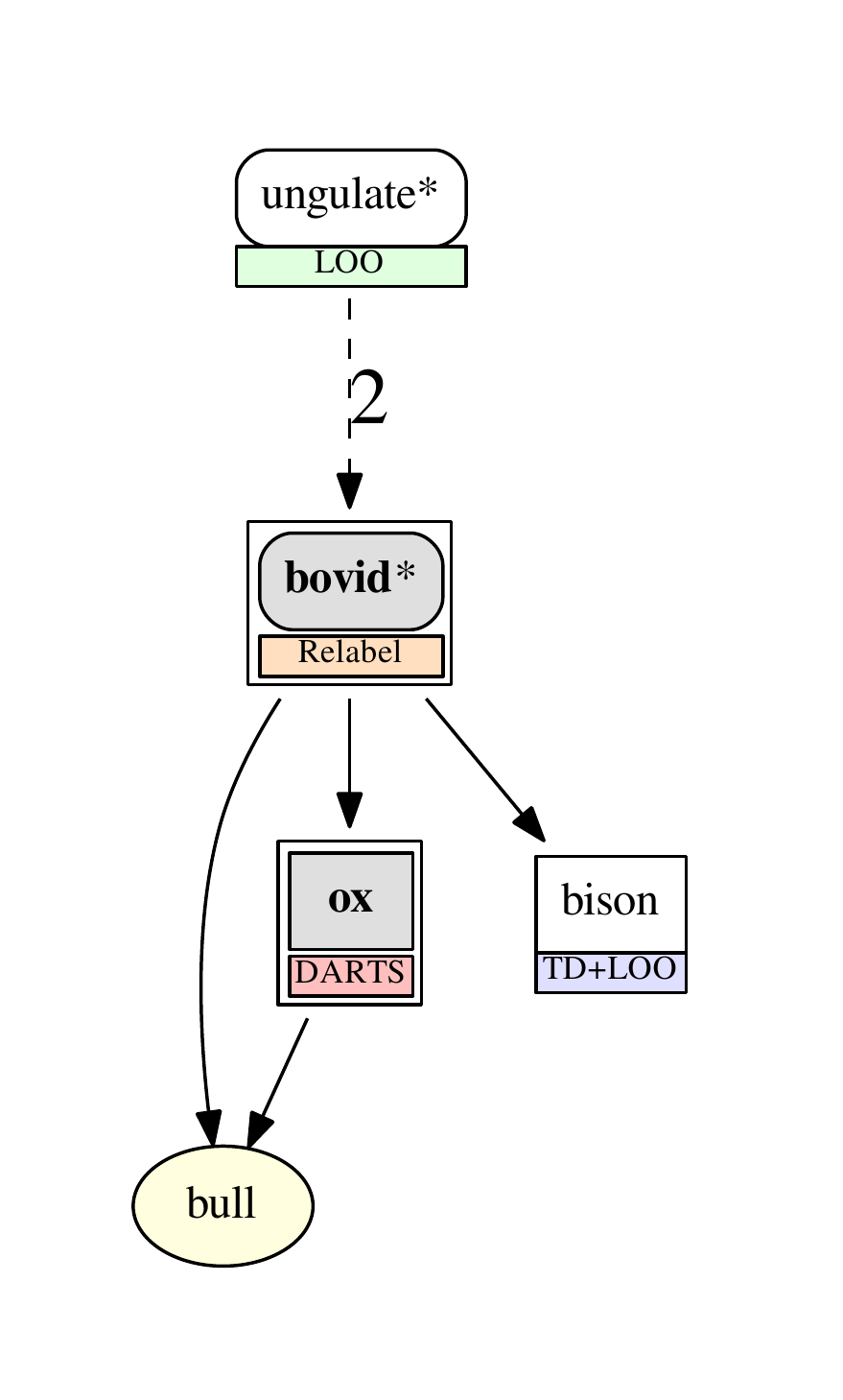}} \cr
\multicolumn{4}{c}{(e)} & 
\multicolumn{4}{c}{(f)} & 
\multicolumn{4}{c}{(g)} & 
\multicolumn{4}{c}{(h)} \cr
\multicolumn{4}{c}{\includegraphics[width=4.24cm, height=3.18cm, keepaspectratio]{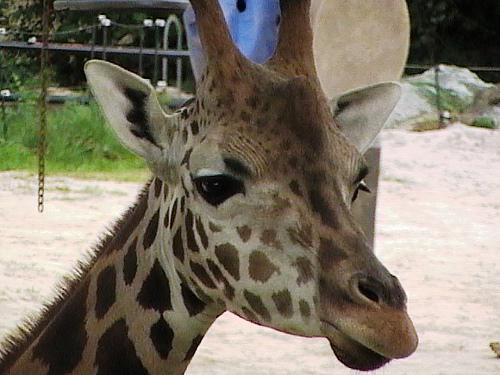}} & 
\multicolumn{4}{c}{\includegraphics[width=4.24cm, height=3.18cm, keepaspectratio]{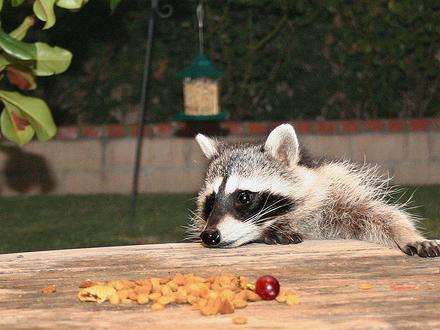}} & 
\multicolumn{4}{c}{\includegraphics[width=4.24cm, height=3.18cm, keepaspectratio]{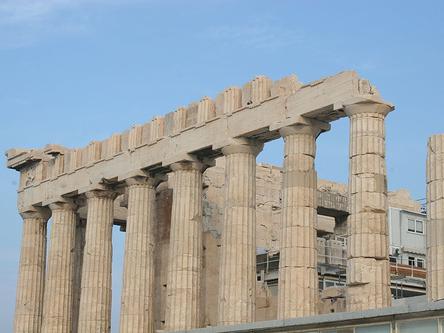}} & 
\multicolumn{4}{c}{\includegraphics[width=4.24cm, height=3.18cm, keepaspectratio]{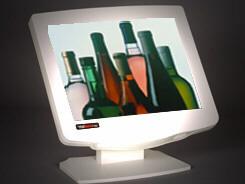}} \cr
\multicolumn{4}{l}{\cellcolor{ColorA} Novel class: giraffe} & 
\multicolumn{4}{l}{\cellcolor{ColorA} Novel class: raccoon} & 
\multicolumn{4}{l}{\cellcolor{ColorA} Novel class: acropolis} & 
\multicolumn{4}{l}{\cellcolor{ColorA} Novel class: {\scriptsize active matrix screen}} \cr
\cellcolor{ColorB} Method & \cellcolor{ColorB} $\epsilon$ & \cellcolor{ColorB} A & \multicolumn{1}{c}{\cellcolor{ColorB} Word} & 
\cellcolor{ColorB} Method & \cellcolor{ColorB} $\epsilon$ & \cellcolor{ColorB} A & \multicolumn{1}{c}{\cellcolor{ColorB} Word} & 
\cellcolor{ColorB} Method & \cellcolor{ColorB} $\epsilon$ & \cellcolor{ColorB} A & \multicolumn{1}{c}{\cellcolor{ColorB} Word} & 
\cellcolor{ColorB} Method & \cellcolor{ColorB} $\epsilon$ & \cellcolor{ColorB} A & \multicolumn{1}{c}{\cellcolor{ColorB} Word} \cr
\cellcolor{Color0} GT & \cellcolor{Color0} & \cellcolor{Color0} & \cellcolor{Color0} even-toed ungulate & 
\cellcolor{Color0} GT & \cellcolor{Color0} & \cellcolor{Color0} & \cellcolor{Color0} procyonid & 
\cellcolor{Color0} GT & \cellcolor{Color0} & \cellcolor{Color0} & \cellcolor{Color0} castle & 
\cellcolor{Color0} GT & \cellcolor{Color0} & \cellcolor{Color0} & \cellcolor{Color0} electronic device \cr
\cellcolor{Color1} DARTS & \cellcolor{Color1} 1 & \cellcolor{Color1} N & \cellcolor{Color1} antelope & 
\cellcolor{Color1} DARTS & \cellcolor{Color1} 2 & \cellcolor{Color1} N & \cellcolor{Color1} musteline mammal & 
\cellcolor{Color1} DARTS & \cellcolor{Color1} 2 & \cellcolor{Color1} N & \cellcolor{Color1} dam & 
\cellcolor{Color1} DARTS & \cellcolor{Color1} 4 & \cellcolor{Color1} N & \cellcolor{Color1} personal computer \cr
\cellcolor{Color2} Relabel & \cellcolor{Color2} 0 & \cellcolor{Color2} Y & \cellcolor{Color2} even-toed ungulate & 
\cellcolor{Color2} Relabel & \cellcolor{Color2} 1 & \cellcolor{Color2} Y & \cellcolor{Color2} carnivore & 
\cellcolor{Color2} Relabel & \cellcolor{Color2} 0 & \cellcolor{Color2} Y & \cellcolor{Color2} {\scriptsize structure, construction} & 
\cellcolor{Color2} Relabel & \cellcolor{Color2} 2 & \cellcolor{Color2} Y & \cellcolor{Color2} instrumentality \cr
\cellcolor{Color3} LOO & \cellcolor{Color3} 1 & \cellcolor{Color3} Y & \cellcolor{Color3} ungulate & 
\cellcolor{Color3} LOO & \cellcolor{Color3} 1 & \cellcolor{Color3} Y & \cellcolor{Color3} carnivore & 
\cellcolor{Color3} LOO & \cellcolor{Color3} 2 & \cellcolor{Color3} N & \cellcolor{Color3} residence & 
\cellcolor{Color3} LOO & \cellcolor{Color3} 5 & \cellcolor{Color3} N & \cellcolor{Color3} peripheral \cr
\cellcolor{Color4} TD+LOO & \cellcolor{Color4} 2 & \cellcolor{Color4} N & \cellcolor{Color4} equine & 
\cellcolor{Color4} TD+LOO & \cellcolor{Color4} 0 & \cellcolor{Color4} Y & \cellcolor{Color4} procyonid & 
\cellcolor{Color4} TD+LOO & \cellcolor{Color4} 1 & \cellcolor{Color4} N & \cellcolor{Color4} triumphal arch & 
\cellcolor{Color4} TD+LOO & \cellcolor{Color4} 0 & \cellcolor{Color4} Y & \cellcolor{Color4} electronic device \cr
\multicolumn{4}{c}{\includegraphics[width=4.24cm, height=3.6cm, keepaspectratio]{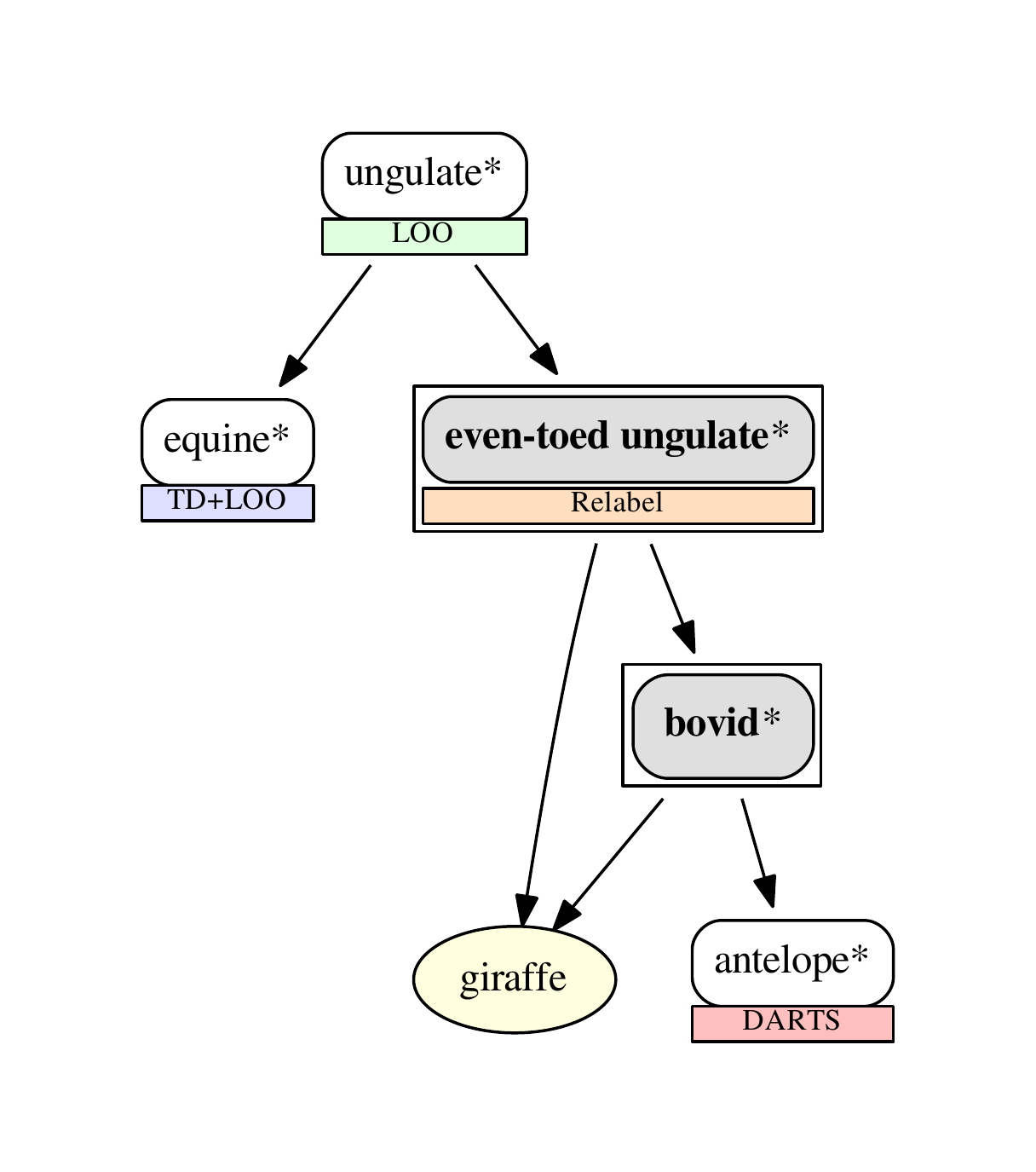}} & 
\multicolumn{4}{c}{\includegraphics[width=4.24cm, height=3.6cm, keepaspectratio]{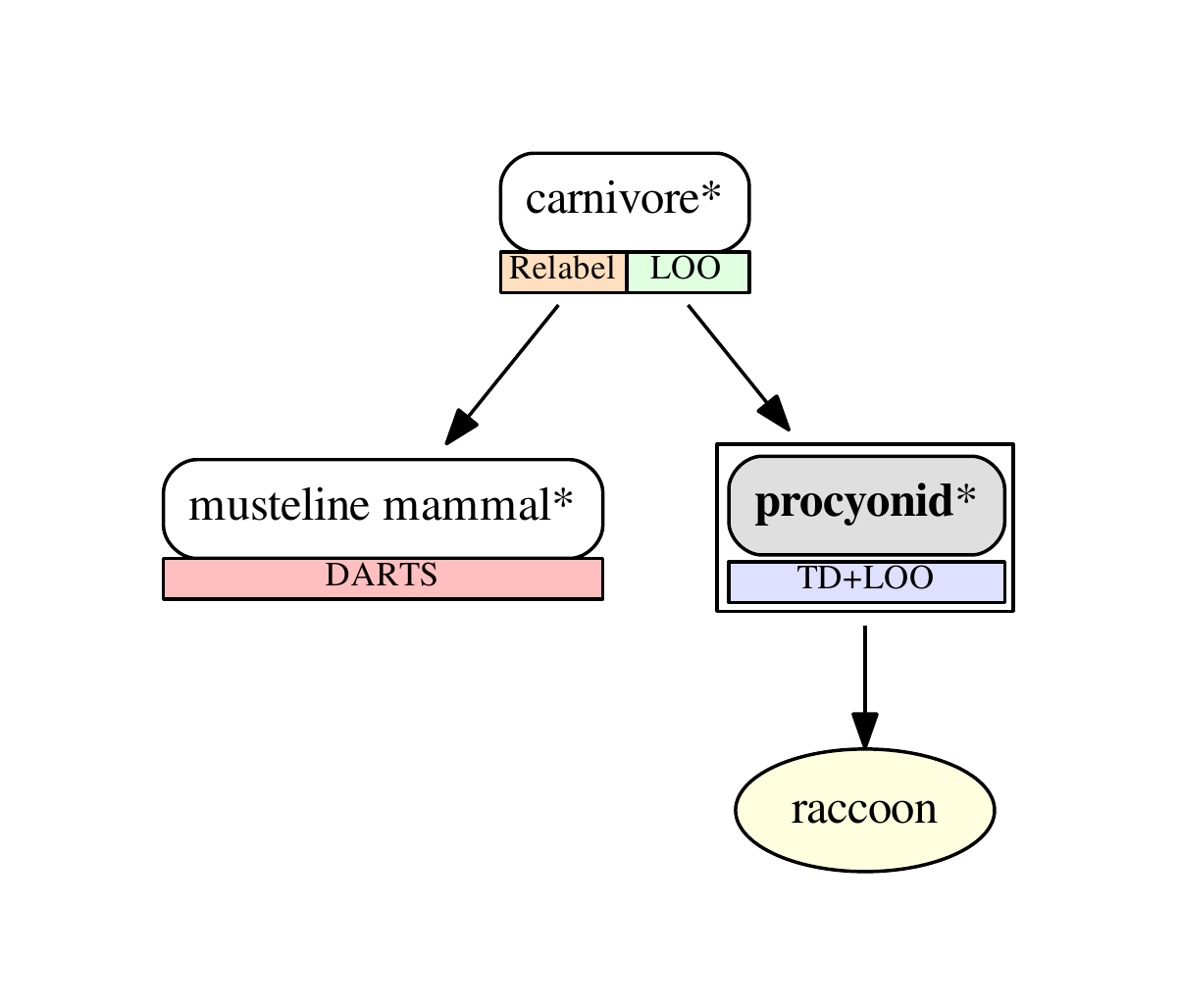}} & 
\multicolumn{4}{c}{\includegraphics[width=4.24cm, height=3.6cm, keepaspectratio]{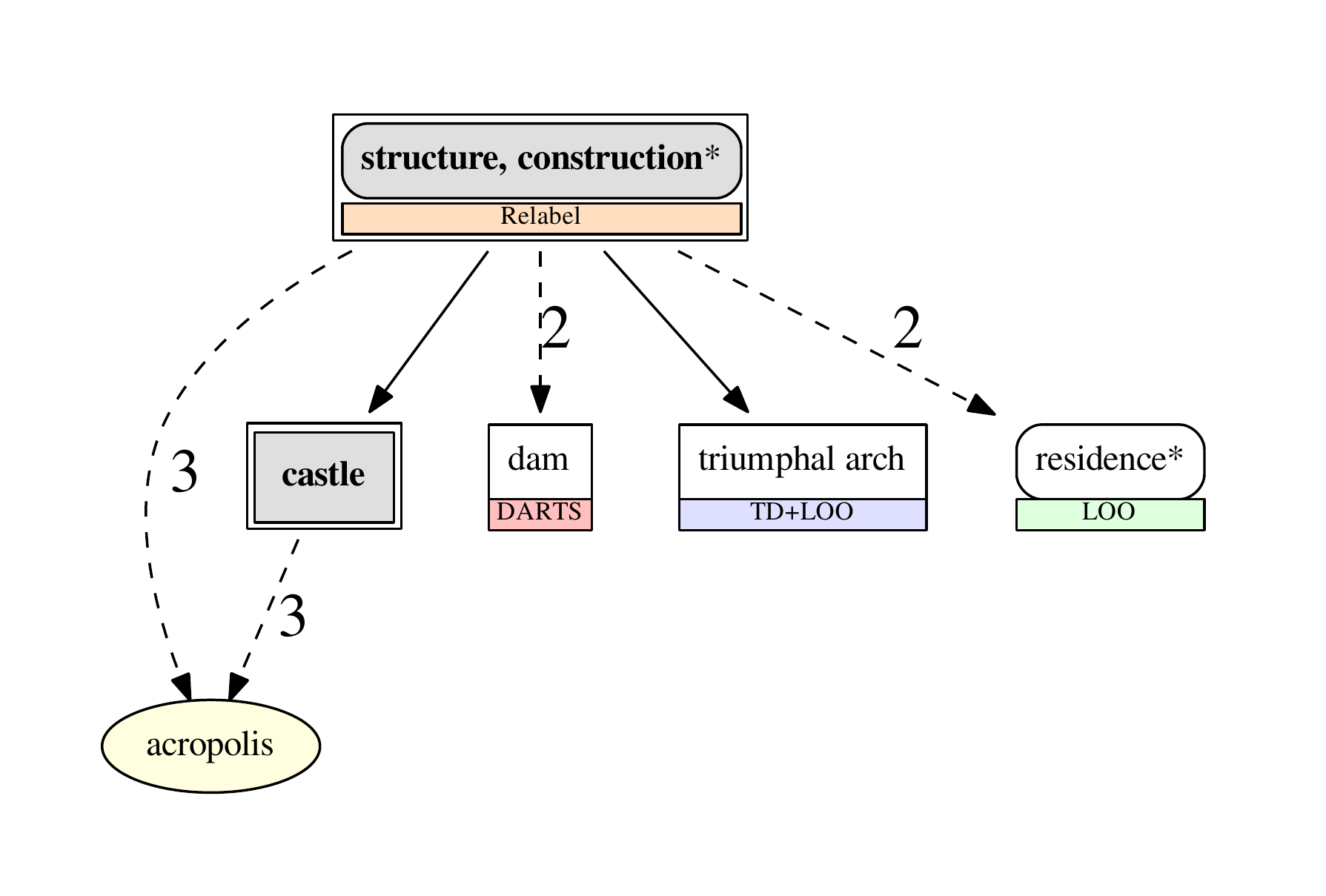}} & 
\multicolumn{4}{c}{\includegraphics[width=4.24cm, height=3.6cm, keepaspectratio]{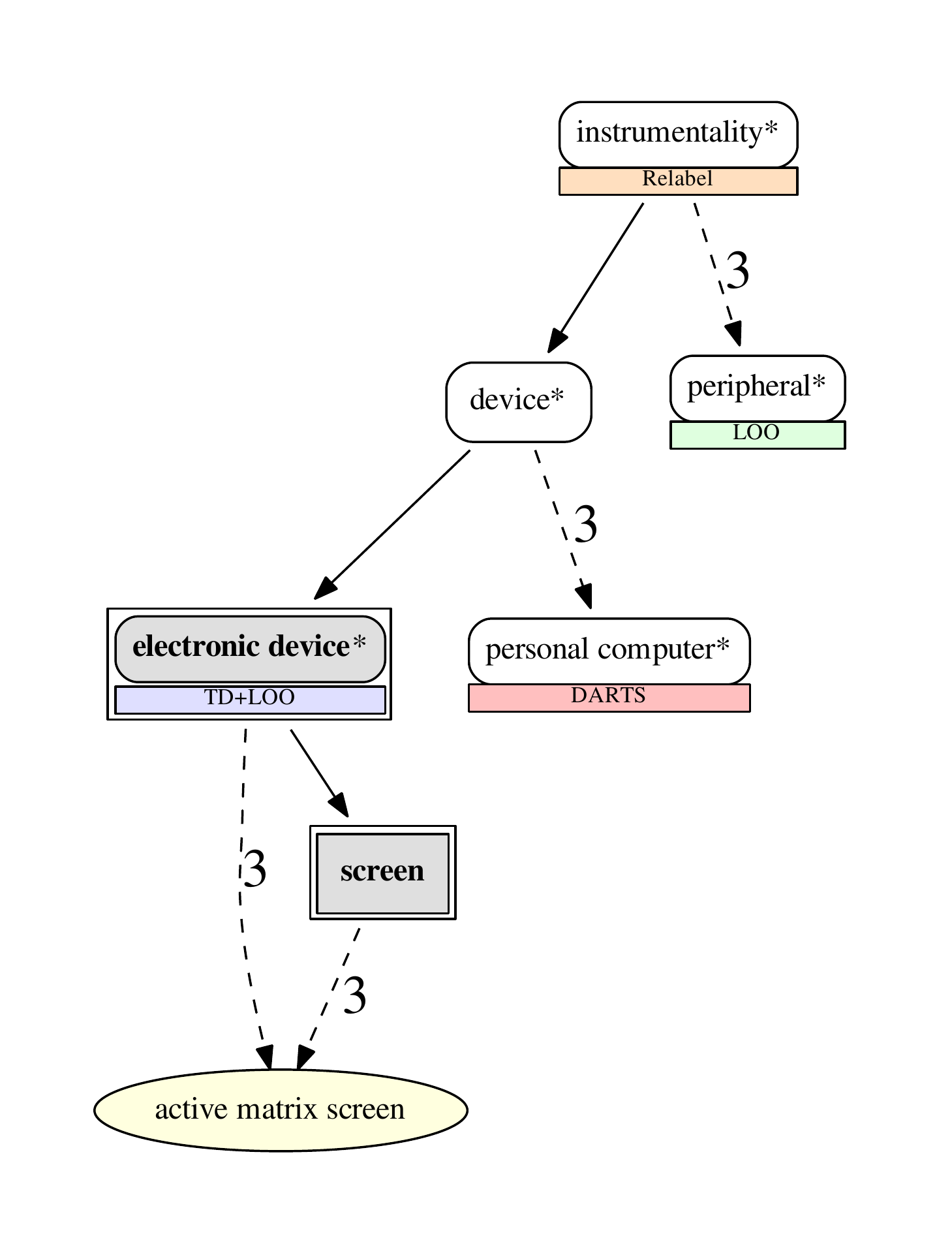}} \cr
\end{tabular}
\caption{Qualitative results of hierarchical novelty detection on ImageNet.
``GT'' is the closest known ancestor of the novel class, which is the expected prediction,
``DARTS'' is the baseline method proposed in \cite{deng2012hedging} where we modify the method for our purpose, and the others are our proposed methods.
``$\epsilon$'' is the distance between the prediction and GT,
``A'' indicates whether the prediction is an ancestor of GT, and
``Word'' is the English word of the predicted label.
Dashed edges represent multi-hop connection, where the number indicates the number of edges between classes.
If the prediction is on a super class (marked with * and rounded), then the test image is classified as a novel class whose closest class in the taxonomy is the super class.
}
\label{fig:qual_smp_5}
\end{figure*}

%% file: qual_smp/qual_smp_6.tex
\begin{figure*}[t]
\footnotesize\centering\setlength{\tabcolsep}{0cm}
\begin{tabular}{
>{\centering}m{1.12cm}>{\centering}m{0.4cm}>{\centering}m{0.4cm}m{2.32cm}
>{\centering}m{1.12cm}>{\centering}m{0.4cm}>{\centering}m{0.4cm}m{2.32cm}
>{\centering}m{1.12cm}>{\centering}m{0.4cm}>{\centering}m{0.4cm}m{2.32cm}
>{\centering}m{1.12cm}>{\centering}m{0.4cm}>{\centering}m{0.4cm}m{2.32cm}
}
\multicolumn{4}{c}{(a)} & 
\multicolumn{4}{c}{(b)} & 
\multicolumn{4}{c}{(c)} & 
\multicolumn{4}{c}{(d)} \cr
\multicolumn{4}{c}{\includegraphics[width=4.24cm, height=3.18cm, keepaspectratio]{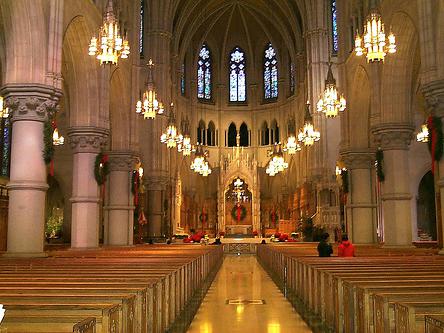}} & 
\multicolumn{4}{c}{\includegraphics[width=4.24cm, height=3.18cm, keepaspectratio]{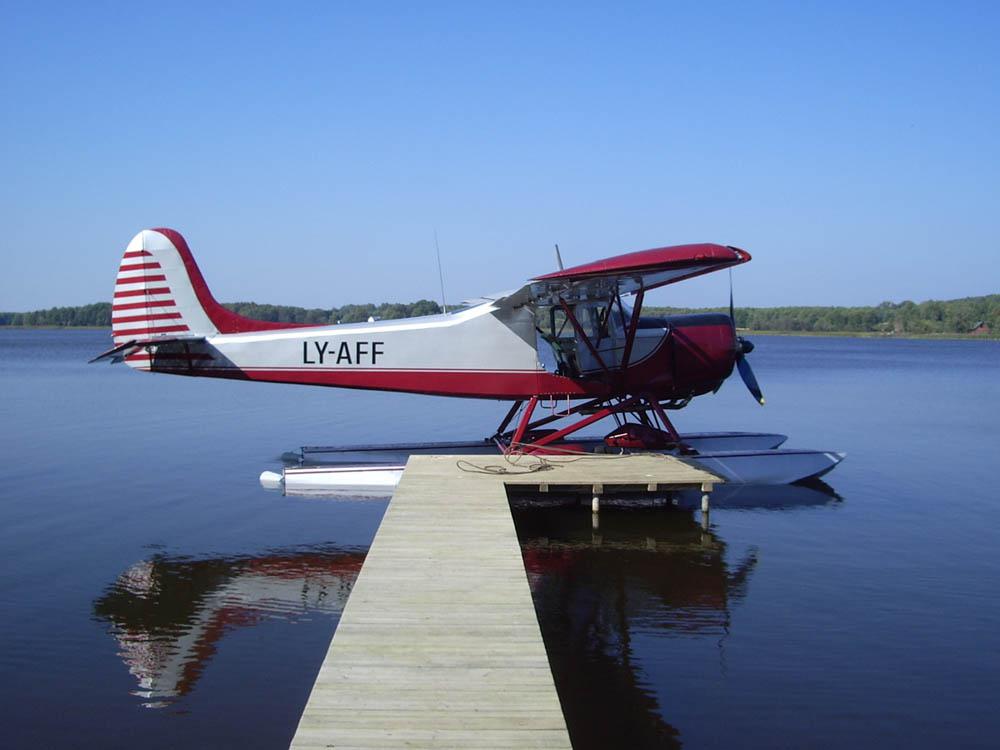}} & 
\multicolumn{4}{c}{\includegraphics[width=4.24cm, height=3.18cm, keepaspectratio]{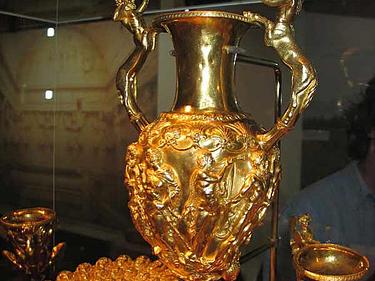}} & 
\multicolumn{4}{c}{\includegraphics[width=4.24cm, height=3.18cm, keepaspectratio]{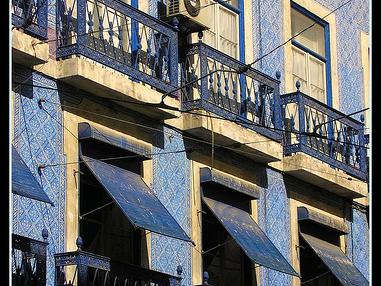}} \cr
\multicolumn{4}{l}{\cellcolor{ColorA} Novel class: aisle} & 
\multicolumn{4}{l}{\cellcolor{ColorA} Novel class: amphibian} & 
\multicolumn{4}{l}{\cellcolor{ColorA} Novel class: amphora} & 
\multicolumn{4}{l}{\cellcolor{ColorA} Novel class: balcony} \cr
\cellcolor{ColorB} Method & \cellcolor{ColorB} $\epsilon$ & \cellcolor{ColorB} A & \multicolumn{1}{c}{\cellcolor{ColorB} Word} & 
\cellcolor{ColorB} Method & \cellcolor{ColorB} $\epsilon$ & \cellcolor{ColorB} A & \multicolumn{1}{c}{\cellcolor{ColorB} Word} & 
\cellcolor{ColorB} Method & \cellcolor{ColorB} $\epsilon$ & \cellcolor{ColorB} A & \multicolumn{1}{c}{\cellcolor{ColorB} Word} & 
\cellcolor{ColorB} Method & \cellcolor{ColorB} $\epsilon$ & \cellcolor{ColorB} A & \multicolumn{1}{c}{\cellcolor{ColorB} Word} \cr
\cellcolor{Color0} GT & \cellcolor{Color0} & \cellcolor{Color0} & \cellcolor{Color0} patio & 
\cellcolor{Color0} GT & \cellcolor{Color0} & \cellcolor{Color0} & \cellcolor{Color0} airliner & 
\cellcolor{Color0} GT & \cellcolor{Color0} & \cellcolor{Color0} & \cellcolor{Color0} jar & 
\cellcolor{Color0} GT & \cellcolor{Color0} & \cellcolor{Color0} & \cellcolor{Color0} {\scriptsize structure, construction} \cr
\cellcolor{Color1} DARTS & \cellcolor{Color1} 2 & \cellcolor{Color1} N & \cellcolor{Color1} place of worship & 
\cellcolor{Color1} DARTS & \cellcolor{Color1} 7 & \cellcolor{Color1} N & \cellcolor{Color1} wing & 
\cellcolor{Color1} DARTS & \cellcolor{Color1} 1 & \cellcolor{Color1} Y & \cellcolor{Color1} vessel & 
\cellcolor{Color1} DARTS & \cellcolor{Color1} 2 & \cellcolor{Color1} N & \cellcolor{Color1} prison \cr
\cellcolor{Color2} Relabel & \cellcolor{Color2} 0 & \cellcolor{Color2} Y & \cellcolor{Color2} {\scriptsize structure, construction} & 
\cellcolor{Color2} Relabel & \cellcolor{Color2} 5 & \cellcolor{Color2} N & \cellcolor{Color2} sailboat & 
\cellcolor{Color2} Relabel & \cellcolor{Color2} 1 & \cellcolor{Color2} N & \cellcolor{Color2} vase & 
\cellcolor{Color2} Relabel & \cellcolor{Color2} 0 & \cellcolor{Color2} Y & \cellcolor{Color2} {\scriptsize structure, construction} \cr
\cellcolor{Color3} LOO & \cellcolor{Color3} 3 & \cellcolor{Color3} N & \cellcolor{Color3} church & 
\cellcolor{Color3} LOO & \cellcolor{Color3} 2 & \cellcolor{Color3} Y & \cellcolor{Color3} craft & 
\cellcolor{Color3} LOO & \cellcolor{Color3} 0 & \cellcolor{Color3} Y & \cellcolor{Color3} jar & 
\cellcolor{Color3} LOO & \cellcolor{Color3} 1 & \cellcolor{Color3} N & \cellcolor{Color3} building \cr
\cellcolor{Color4} TD+LOO & \cellcolor{Color4} 1 & \cellcolor{Color4} N & \cellcolor{Color4} altar & 
\cellcolor{Color4} TD+LOO & \cellcolor{Color4} 0 & \cellcolor{Color4} Y & \cellcolor{Color4} {\scriptsize heavier-than-air craft} & 
\cellcolor{Color4} TD+LOO & \cellcolor{Color4} 3 & \cellcolor{Color4} N & \cellcolor{Color4} jug & 
\cellcolor{Color4} TD+LOO & \cellcolor{Color4} 1 & \cellcolor{Color4} N & \cellcolor{Color4} establishment \cr
\multicolumn{4}{c}{\includegraphics[width=4.24cm, height=3.6cm, keepaspectratio]{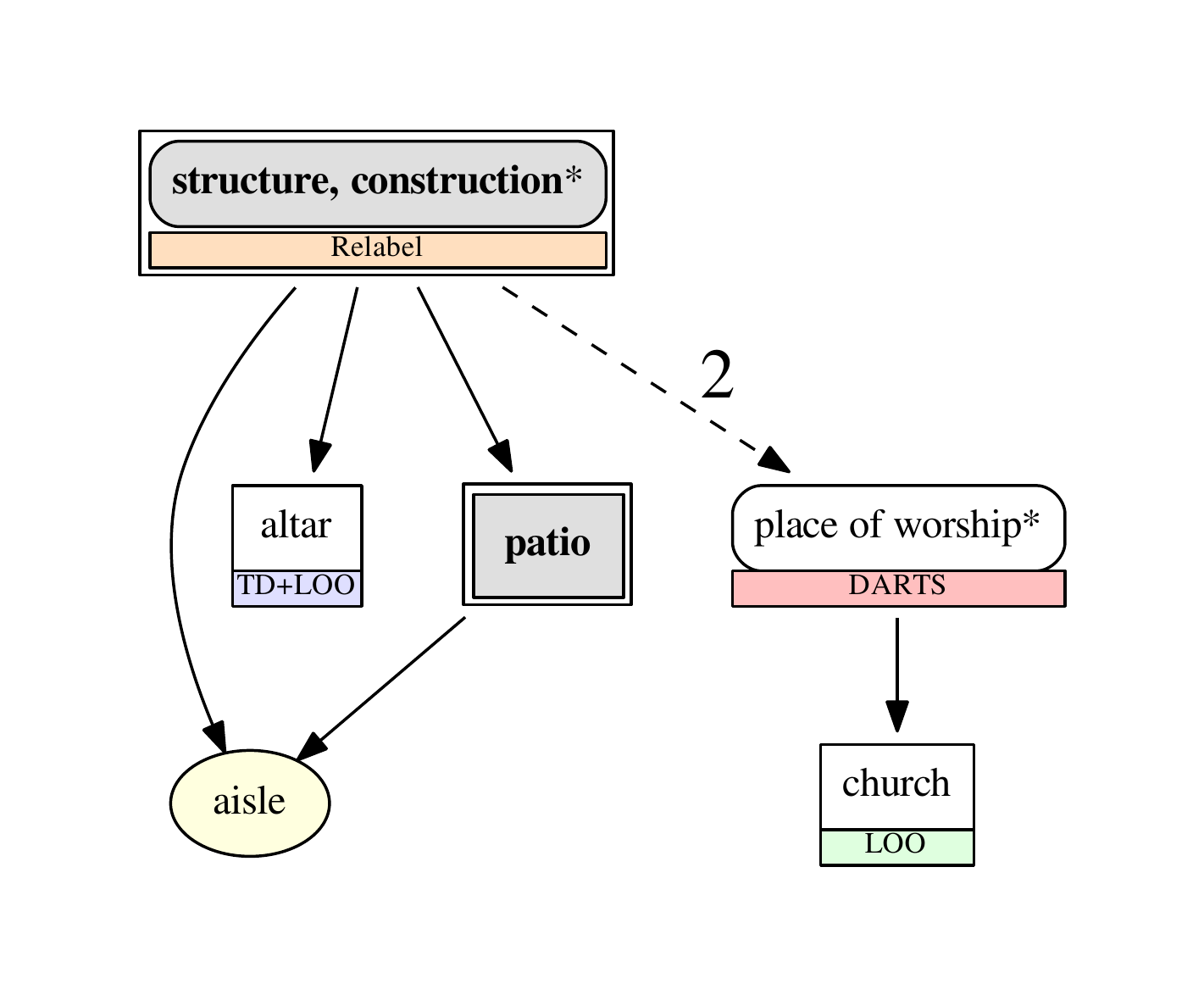}} & 
\multicolumn{4}{c}{\includegraphics[width=4.24cm, height=3.6cm, keepaspectratio]{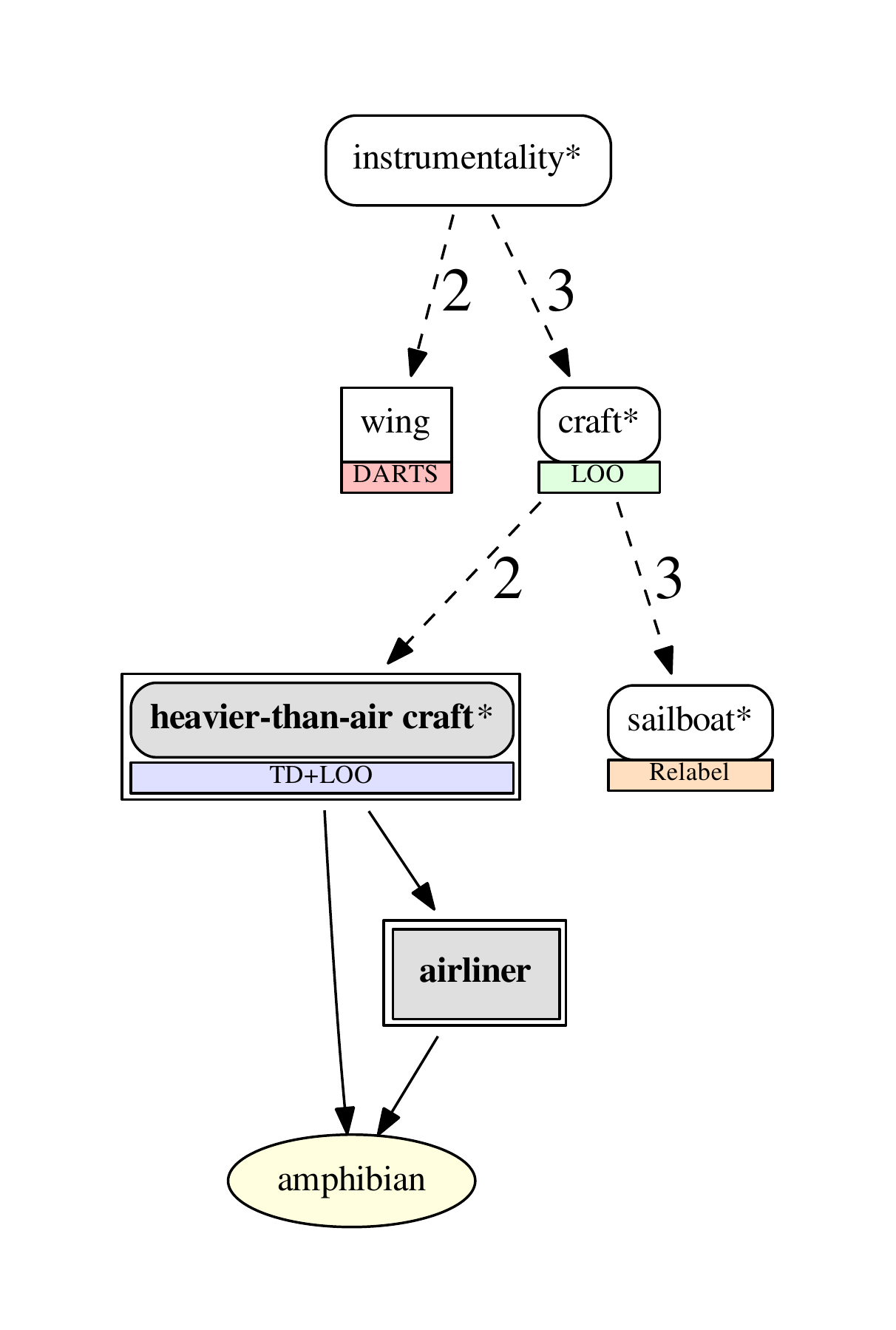}} & 
\multicolumn{4}{c}{\includegraphics[width=4.24cm, height=3.6cm, keepaspectratio]{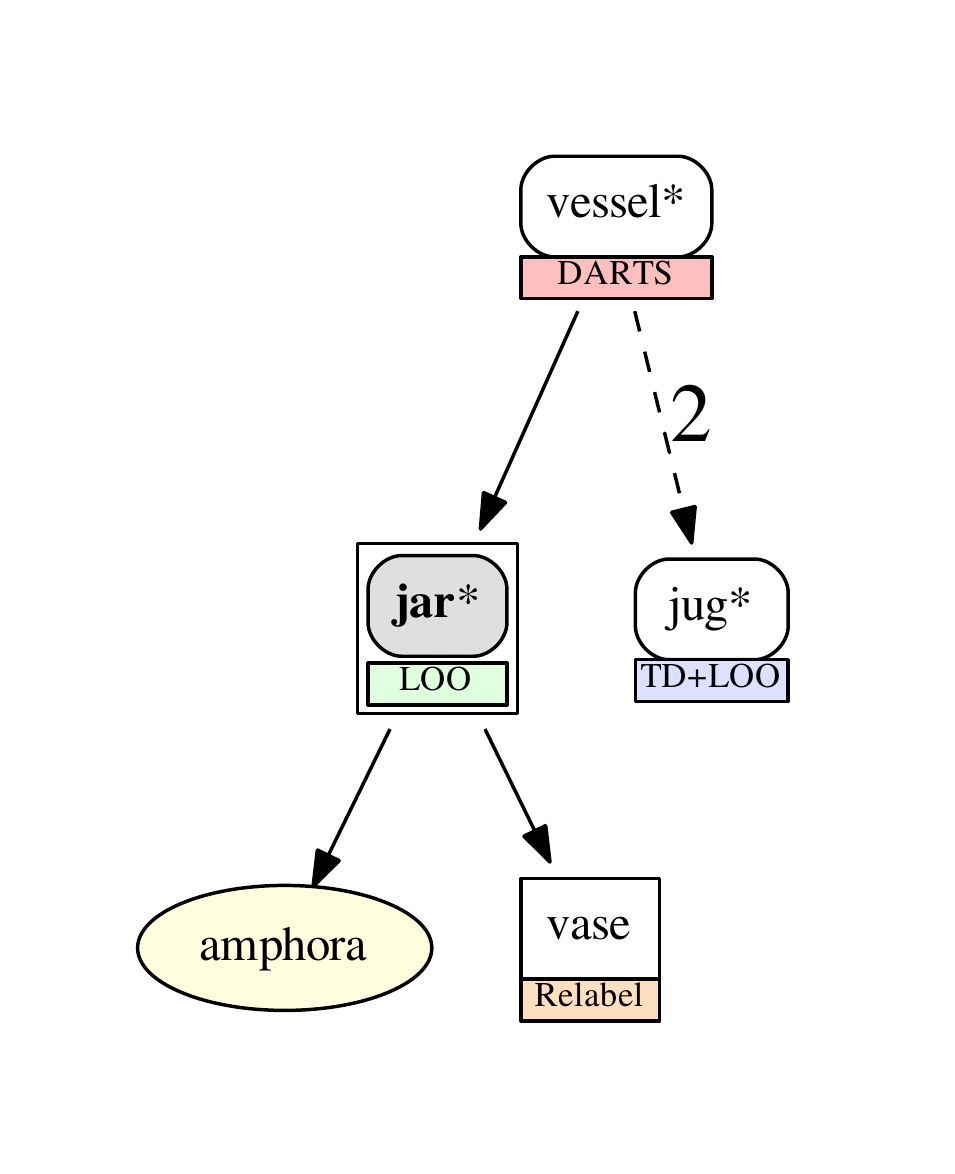}} & 
\multicolumn{4}{c}{\includegraphics[width=4.24cm, height=3.6cm, keepaspectratio]{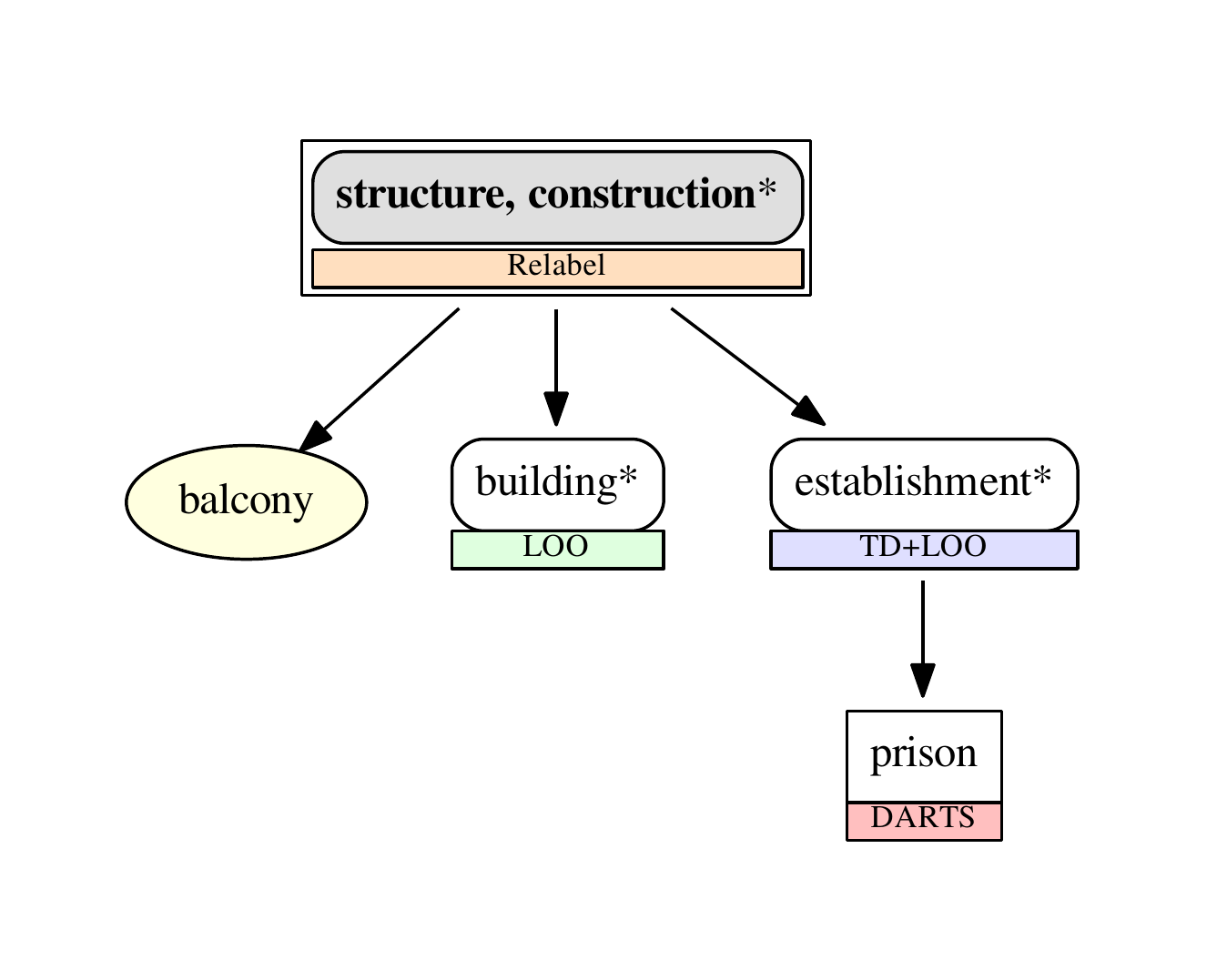}} \cr
\multicolumn{4}{c}{(e)} & 
\multicolumn{4}{c}{(f)} & 
\multicolumn{4}{c}{(g)} & 
\multicolumn{4}{c}{(h)} \cr
\multicolumn{4}{c}{\includegraphics[width=4.24cm, height=3.18cm, keepaspectratio]{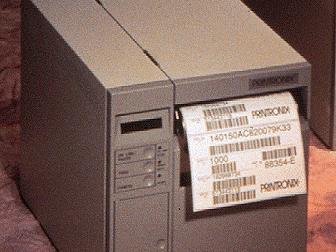}} & 
\multicolumn{4}{c}{\includegraphics[width=4.24cm, height=3.18cm, keepaspectratio]{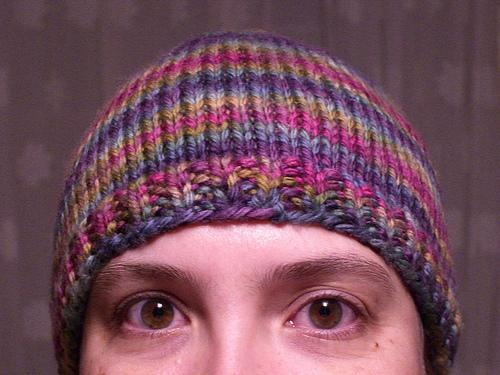}} & 
\multicolumn{4}{c}{\includegraphics[width=4.24cm, height=3.18cm, keepaspectratio]{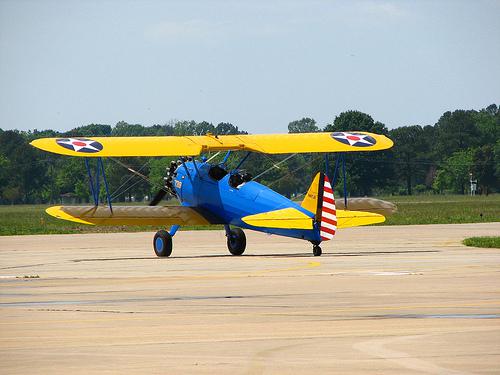}} & 
\multicolumn{4}{c}{\includegraphics[width=4.24cm, height=3.18cm, keepaspectratio]{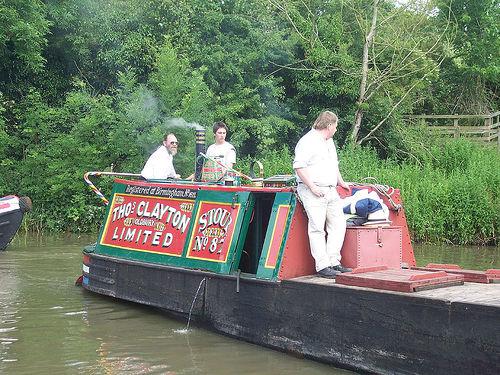}} \cr
\multicolumn{4}{l}{\cellcolor{ColorA} Novel class: bar printer} & 
\multicolumn{4}{l}{\cellcolor{ColorA} Novel class: beanie} & 
\multicolumn{4}{l}{\cellcolor{ColorA} Novel class: biplane} & 
\multicolumn{4}{l}{\cellcolor{ColorA} Novel class: canal boat} \cr
\cellcolor{ColorB} Method & \cellcolor{ColorB} $\epsilon$ & \cellcolor{ColorB} A & \multicolumn{1}{c}{\cellcolor{ColorB} Word} & 
\cellcolor{ColorB} Method & \cellcolor{ColorB} $\epsilon$ & \cellcolor{ColorB} A & \multicolumn{1}{c}{\cellcolor{ColorB} Word} & 
\cellcolor{ColorB} Method & \cellcolor{ColorB} $\epsilon$ & \cellcolor{ColorB} A & \multicolumn{1}{c}{\cellcolor{ColorB} Word} & 
\cellcolor{ColorB} Method & \cellcolor{ColorB} $\epsilon$ & \cellcolor{ColorB} A & \multicolumn{1}{c}{\cellcolor{ColorB} Word} \cr
\cellcolor{Color0} GT & \cellcolor{Color0} & \cellcolor{Color0} & \cellcolor{Color0} machine & 
\cellcolor{Color0} GT & \cellcolor{Color0} & \cellcolor{Color0} & \cellcolor{Color0} cap & 
\cellcolor{Color0} GT & \cellcolor{Color0} & \cellcolor{Color0} & \cellcolor{Color0} airliner & 
\cellcolor{Color0} GT & \cellcolor{Color0} & \cellcolor{Color0} & \cellcolor{Color0} boat \cr
\cellcolor{Color1} DARTS & \cellcolor{Color1} 1 & \cellcolor{Color1} Y & \cellcolor{Color1} peripheral & 
\cellcolor{Color1} DARTS & \cellcolor{Color1} 6 & \cellcolor{Color1} N & \cellcolor{Color1} wool & 
\cellcolor{Color1} DARTS & \cellcolor{Color1} 7 & \cellcolor{Color1} N & \cellcolor{Color1} wing & 
\cellcolor{Color1} DARTS & \cellcolor{Color1} 3 & \cellcolor{Color1} Y & \cellcolor{Color1} vehicle \cr
\cellcolor{Color2} Relabel & \cellcolor{Color2} 2 & \cellcolor{Color2} Y & \cellcolor{Color2} {\scriptsize electronic equipment} & 
\cellcolor{Color2} Relabel & \cellcolor{Color2} 2 & \cellcolor{Color2} N & \cellcolor{Color2} hat & 
\cellcolor{Color2} Relabel & \cellcolor{Color2} 7 & \cellcolor{Color2} N & \cellcolor{Color2} parachute & 
\cellcolor{Color2} Relabel & \cellcolor{Color2} 7 & \cellcolor{Color2} N & \cellcolor{Color2} {\scriptsize structure, construction} \cr
\cellcolor{Color3} LOO & \cellcolor{Color3} 0 & \cellcolor{Color3} Y & \cellcolor{Color3} machine & 
\cellcolor{Color3} LOO & \cellcolor{Color3} 5 & \cellcolor{Color3} N & \cellcolor{Color3} mask & 
\cellcolor{Color3} LOO & \cellcolor{Color3} 1 & \cellcolor{Color3} Y & \cellcolor{Color3} aircraft & 
\cellcolor{Color3} LOO & \cellcolor{Color3} 9 & \cellcolor{Color3} N & \cellcolor{Color3} shed \cr
\cellcolor{Color4} TD+LOO & \cellcolor{Color4} 0 & \cellcolor{Color4} Y & \cellcolor{Color4} printer & 
\cellcolor{Color4} TD+LOO & \cellcolor{Color4} 6 & \cellcolor{Color4} N & \cellcolor{Color4} ski mask & 
\cellcolor{Color4} TD+LOO & \cellcolor{Color4} 0 & \cellcolor{Color4} Y & \cellcolor{Color4} {\scriptsize heavier-than-air craft} & 
\cellcolor{Color4} TD+LOO & \cellcolor{Color4} 0 & \cellcolor{Color4} Y & \cellcolor{Color4} boat \cr
\multicolumn{4}{c}{\includegraphics[width=4.24cm, height=3.6cm, keepaspectratio]{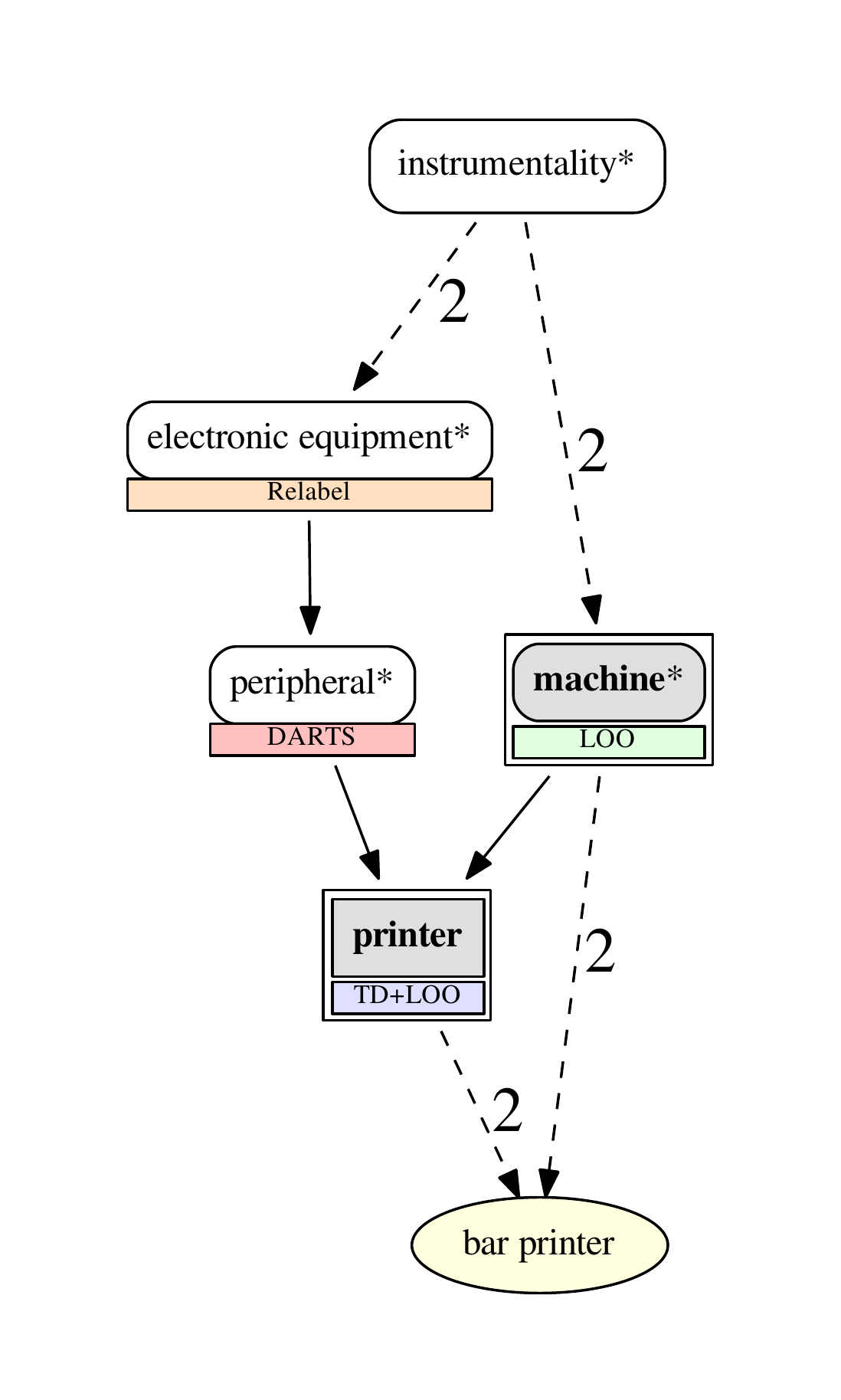}} & 
\multicolumn{4}{c}{\includegraphics[width=4.24cm, height=3.6cm, keepaspectratio]{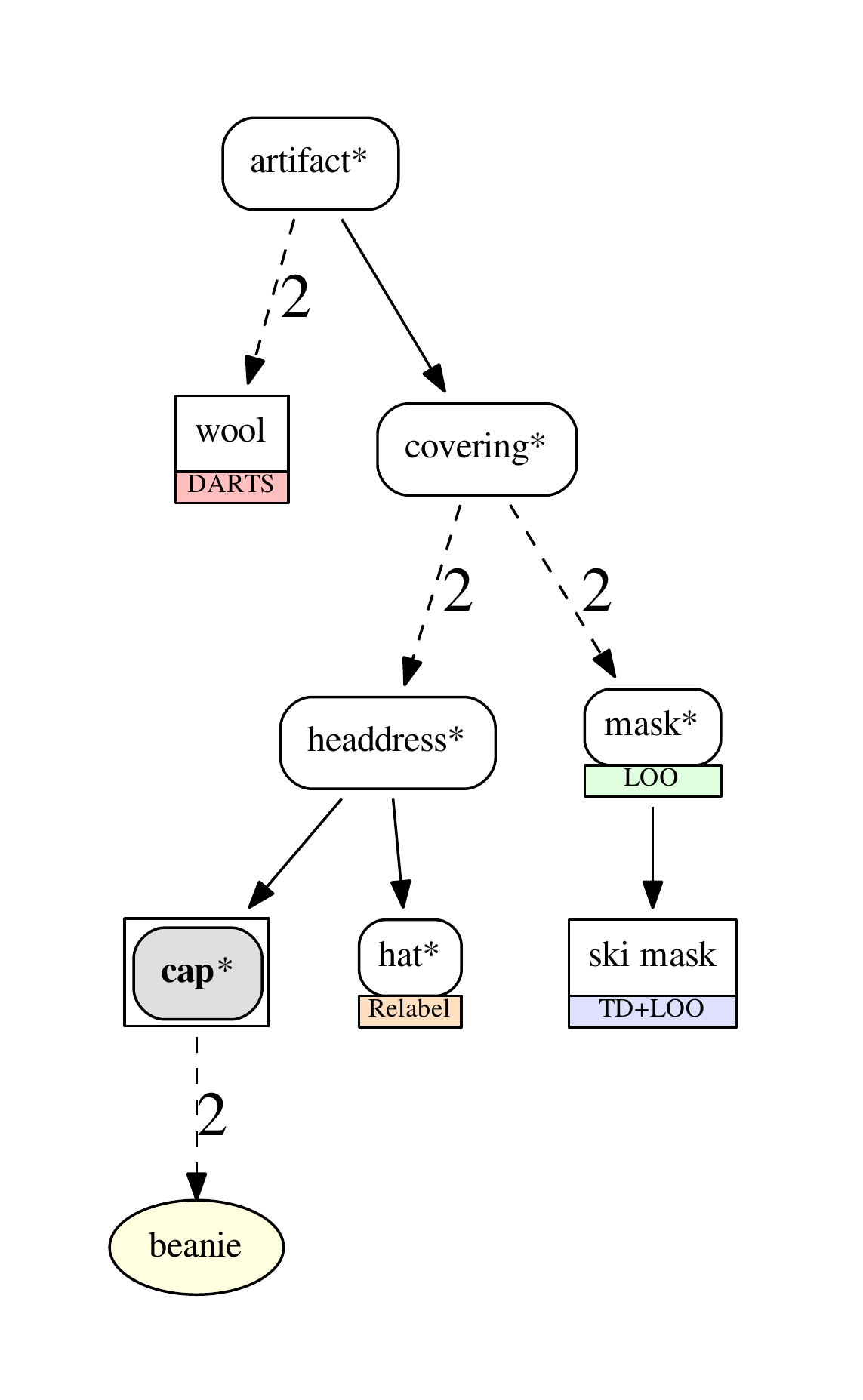}} & 
\multicolumn{4}{c}{\includegraphics[width=4.24cm, height=3.6cm, keepaspectratio]{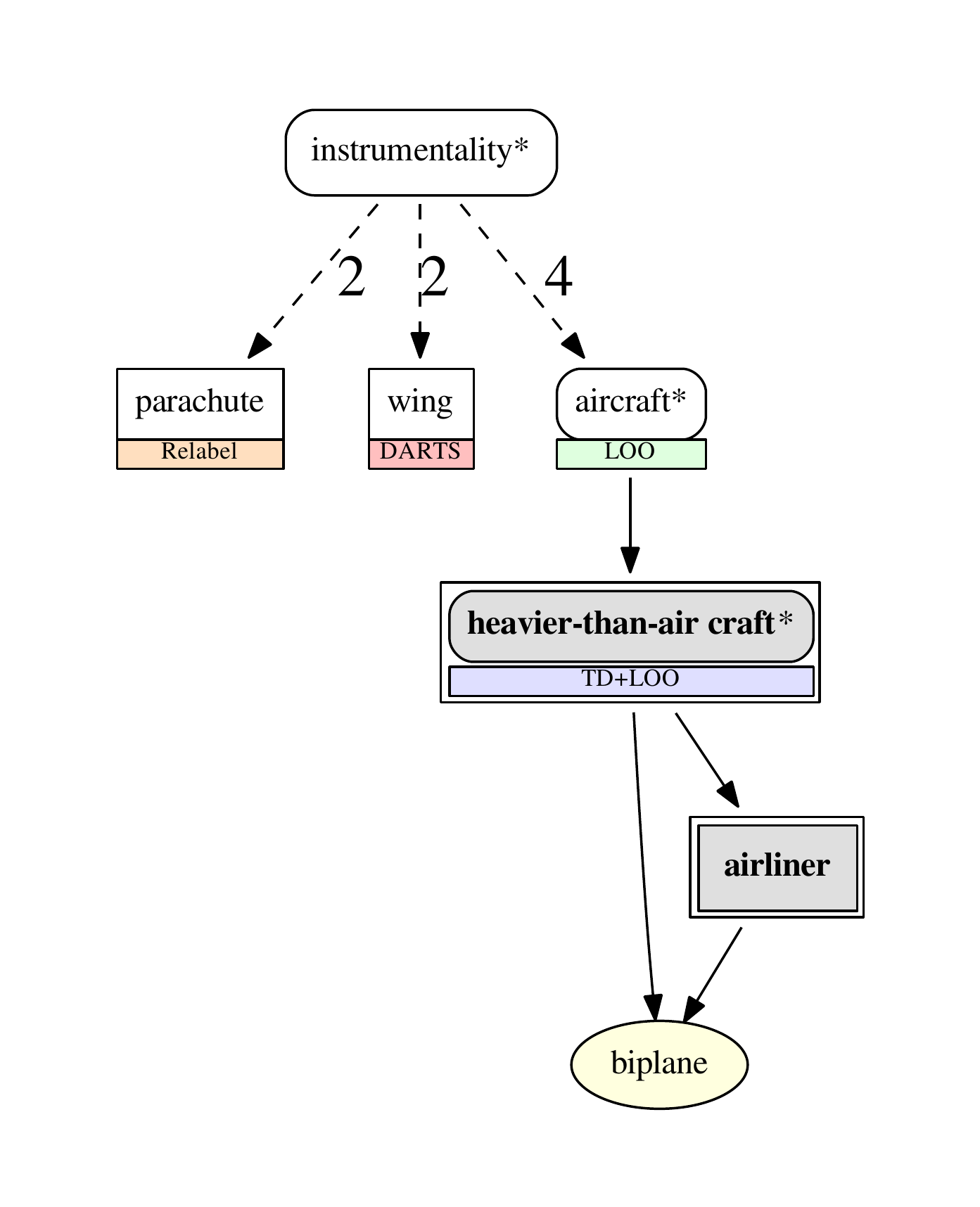}} & 
\multicolumn{4}{c}{\includegraphics[width=4.24cm, height=3.6cm, keepaspectratio]{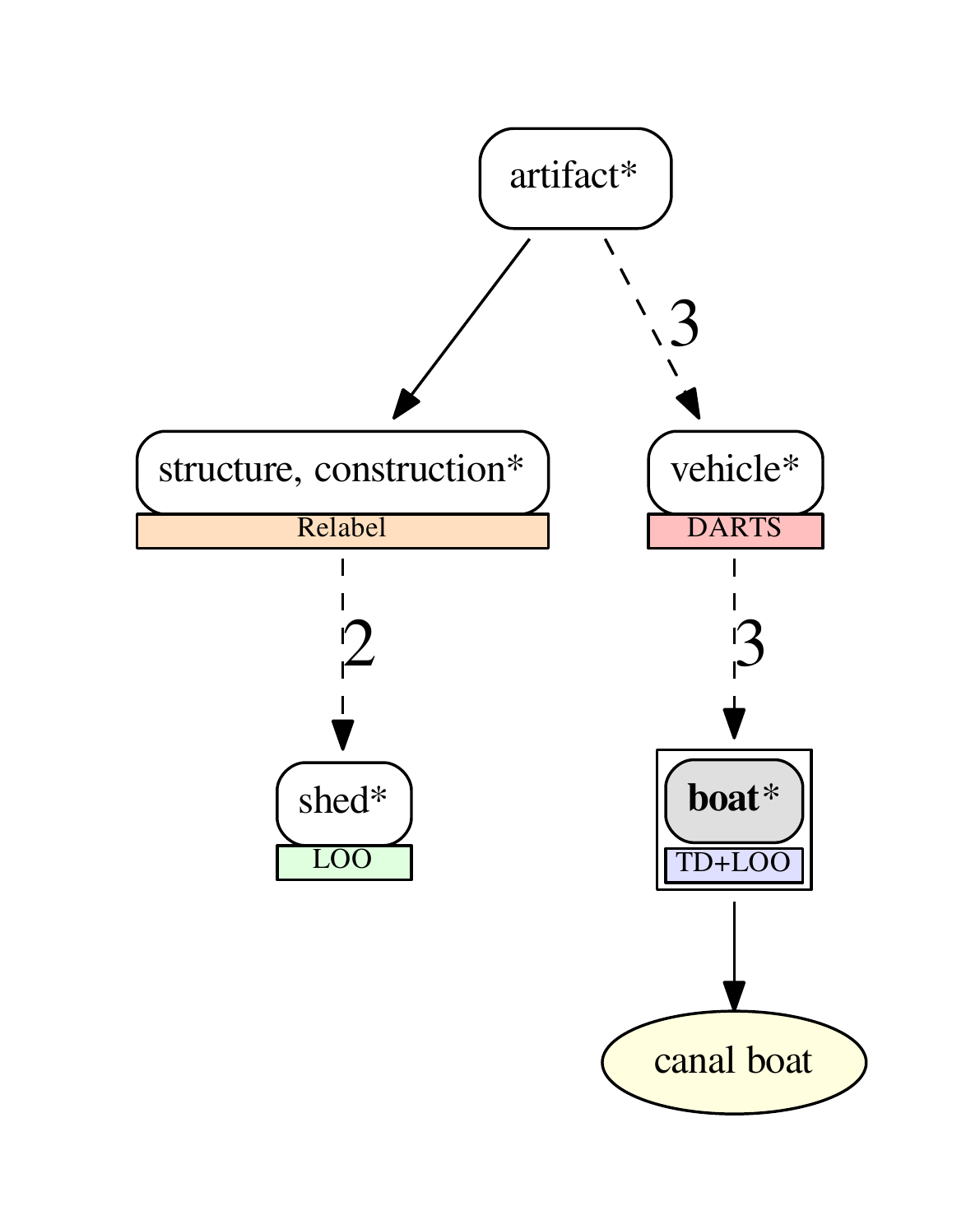}} \cr
\end{tabular}
\caption{Qualitative results of hierarchical novelty detection on ImageNet.
``GT'' is the closest known ancestor of the novel class, which is the expected prediction,
``DARTS'' is the baseline method proposed in \cite{deng2012hedging} where we modify the method for our purpose, and the others are our proposed methods.
``$\epsilon$'' is the distance between the prediction and GT,
``A'' indicates whether the prediction is an ancestor of GT, and
``Word'' is the English word of the predicted label.
Dashed edges represent multi-hop connection, where the number indicates the number of edges between classes.
If the prediction is on a super class (marked with * and rounded), then the test image is classified as a novel class whose closest class in the taxonomy is the super class.
}
\label{fig:qual_smp_6}
\end{figure*}

%% file: qual_smp/qual_smp_7.tex
\begin{figure*}[t]
\footnotesize\centering\setlength{\tabcolsep}{0cm}
\begin{tabular}{
>{\centering}m{1.12cm}>{\centering}m{0.4cm}>{\centering}m{0.4cm}m{2.32cm}
>{\centering}m{1.12cm}>{\centering}m{0.4cm}>{\centering}m{0.4cm}m{2.32cm}
>{\centering}m{1.12cm}>{\centering}m{0.4cm}>{\centering}m{0.4cm}m{2.32cm}
>{\centering}m{1.12cm}>{\centering}m{0.4cm}>{\centering}m{0.4cm}m{2.32cm}
}
\multicolumn{4}{c}{(a)} & 
\multicolumn{4}{c}{(b)} & 
\multicolumn{4}{c}{(c)} & 
\multicolumn{4}{c}{(d)} \cr
\multicolumn{4}{c}{\includegraphics[width=4.24cm, height=3.18cm, keepaspectratio]{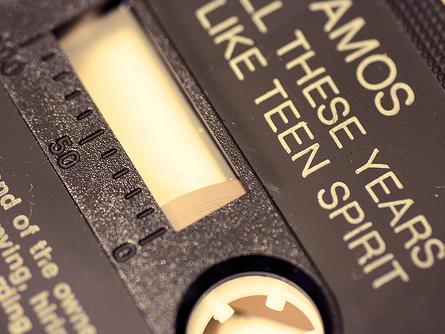}} & 
\multicolumn{4}{c}{\includegraphics[width=4.24cm, height=3.18cm, keepaspectratio]{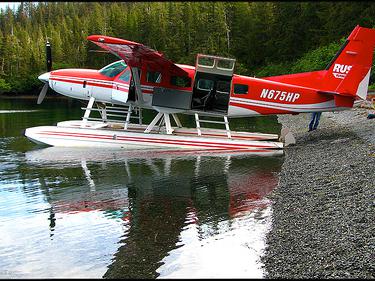}} & 
\multicolumn{4}{c}{\includegraphics[width=4.24cm, height=3.18cm, keepaspectratio]{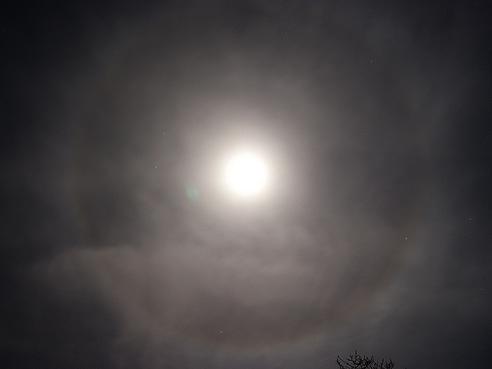}} & 
\multicolumn{4}{c}{\includegraphics[width=4.24cm, height=3.18cm, keepaspectratio]{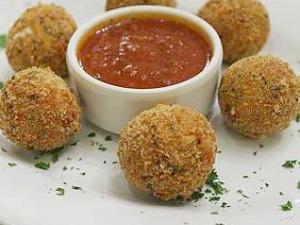}} \cr
\multicolumn{4}{l}{\cellcolor{ColorA} Novel class: cassette tape} & 
\multicolumn{4}{l}{\cellcolor{ColorA} Novel class: floatplane} & 
\multicolumn{4}{l}{\cellcolor{ColorA} Novel class: aura} & 
\multicolumn{4}{l}{\cellcolor{ColorA} Novel class: appetizer} \cr
\cellcolor{ColorB} Method & \cellcolor{ColorB} $\epsilon$ & \cellcolor{ColorB} A & \multicolumn{1}{c}{\cellcolor{ColorB} Word} & 
\cellcolor{ColorB} Method & \cellcolor{ColorB} $\epsilon$ & \cellcolor{ColorB} A & \multicolumn{1}{c}{\cellcolor{ColorB} Word} & 
\cellcolor{ColorB} Method & \cellcolor{ColorB} $\epsilon$ & \cellcolor{ColorB} A & \multicolumn{1}{c}{\cellcolor{ColorB} Word} & 
\cellcolor{ColorB} Method & \cellcolor{ColorB} $\epsilon$ & \cellcolor{ColorB} A & \multicolumn{1}{c}{\cellcolor{ColorB} Word} \cr
\cellcolor{Color0} GT & \cellcolor{Color0} & \cellcolor{Color0} & \cellcolor{Color0} device & 
\cellcolor{Color0} GT & \cellcolor{Color0} & \cellcolor{Color0} & \cellcolor{Color0} airliner & 
\cellcolor{Color0} GT & \cellcolor{Color0} & \cellcolor{Color0} & \cellcolor{Color0} abstraction & 
\cellcolor{Color0} GT & \cellcolor{Color0} & \cellcolor{Color0} & \cellcolor{Color0} course \cr
\cellcolor{Color1} DARTS & \cellcolor{Color1} 3 & \cellcolor{Color1} N & \cellcolor{Color1} cassette & 
\cellcolor{Color1} DARTS & \cellcolor{Color1} 7 & \cellcolor{Color1} N & \cellcolor{Color1} wing & 
\cellcolor{Color1} DARTS & \cellcolor{Color1} 9 & \cellcolor{Color1} N & \cellcolor{Color1} lamp & 
\cellcolor{Color1} DARTS & \cellcolor{Color1} 1 & \cellcolor{Color1} Y & \cellcolor{Color1} nutriment \cr
\cellcolor{Color2} Relabel & \cellcolor{Color2} 1 & \cellcolor{Color2} Y & \cellcolor{Color2} instrumentality & 
\cellcolor{Color2} Relabel & \cellcolor{Color2} 4 & \cellcolor{Color2} N & \cellcolor{Color2} boat & 
\cellcolor{Color2} Relabel & \cellcolor{Color2} 7 & \cellcolor{Color2} N & \cellcolor{Color2} device & 
\cellcolor{Color2} Relabel & \cellcolor{Color2} 2 & \cellcolor{Color2} N & \cellcolor{Color2} dish \cr
\cellcolor{Color3} LOO & \cellcolor{Color3} 2 & \cellcolor{Color3} N & \cellcolor{Color3} {\scriptsize measuring instrument} & 
\cellcolor{Color3} LOO & \cellcolor{Color3} 2 & \cellcolor{Color3} Y & \cellcolor{Color3} craft & 
\cellcolor{Color3} LOO & \cellcolor{Color3} 6 & \cellcolor{Color3} N & \cellcolor{Color3} mountain & 
\cellcolor{Color3} LOO & \cellcolor{Color3} 1 & \cellcolor{Color3} N & \cellcolor{Color3} plate \cr
\cellcolor{Color4} TD+LOO & \cellcolor{Color4} 0 & \cellcolor{Color4} Y & \cellcolor{Color4} hard disc & 
\cellcolor{Color4} TD+LOO & \cellcolor{Color4} 0 & \cellcolor{Color4} Y & \cellcolor{Color4} {\scriptsize heavier-than-air craft} & 
\cellcolor{Color4} TD+LOO & \cellcolor{Color4} 0 & \cellcolor{Color4} Y & \cellcolor{Color4} abstraction & 
\cellcolor{Color4} TD+LOO & \cellcolor{Color4} 0 & \cellcolor{Color4} Y & \cellcolor{Color4} course \cr
\multicolumn{4}{c}{\includegraphics[width=4.24cm, height=3.6cm, keepaspectratio]{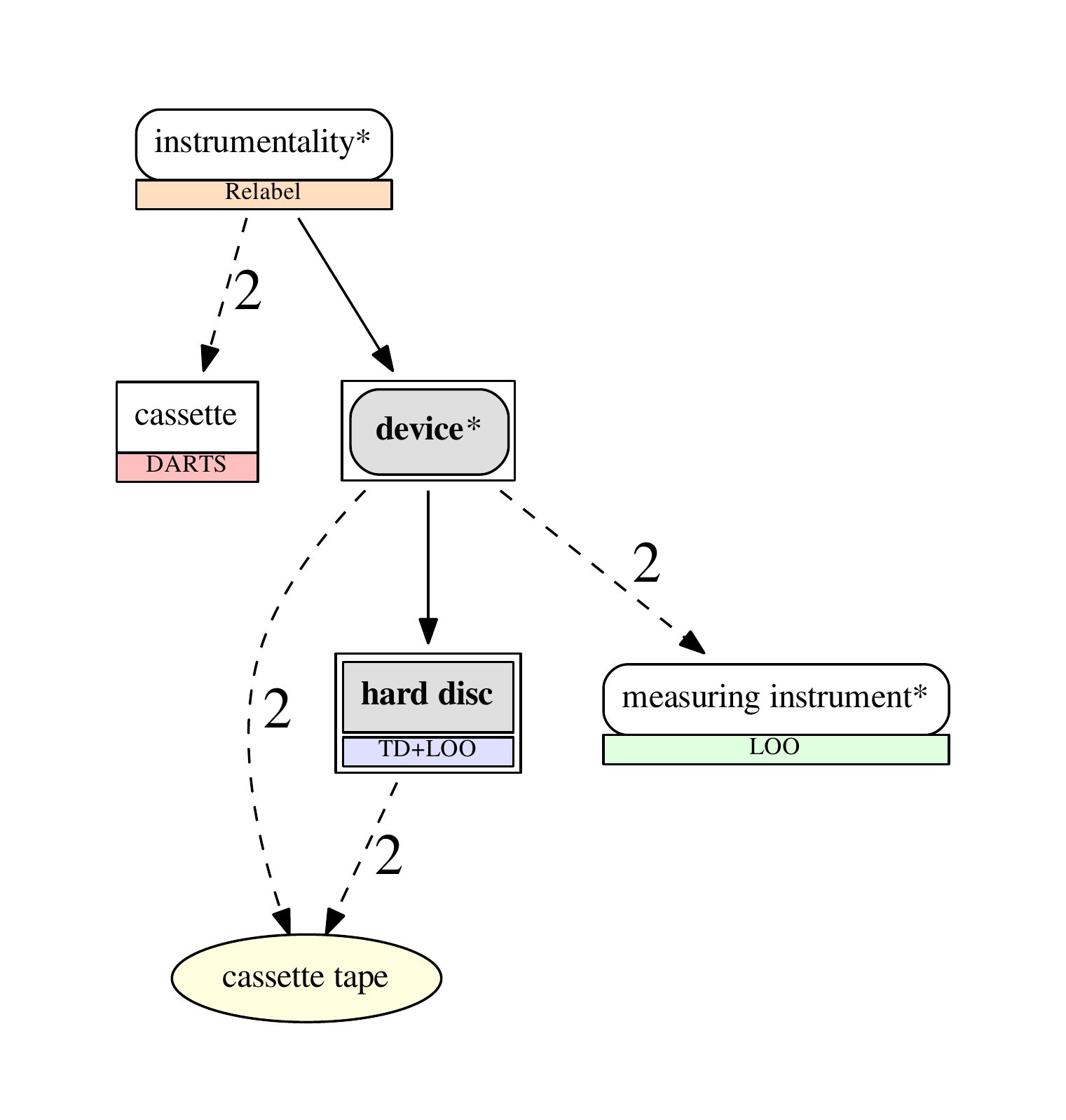}} & 
\multicolumn{4}{c}{\includegraphics[width=4.24cm, height=3.6cm, keepaspectratio]{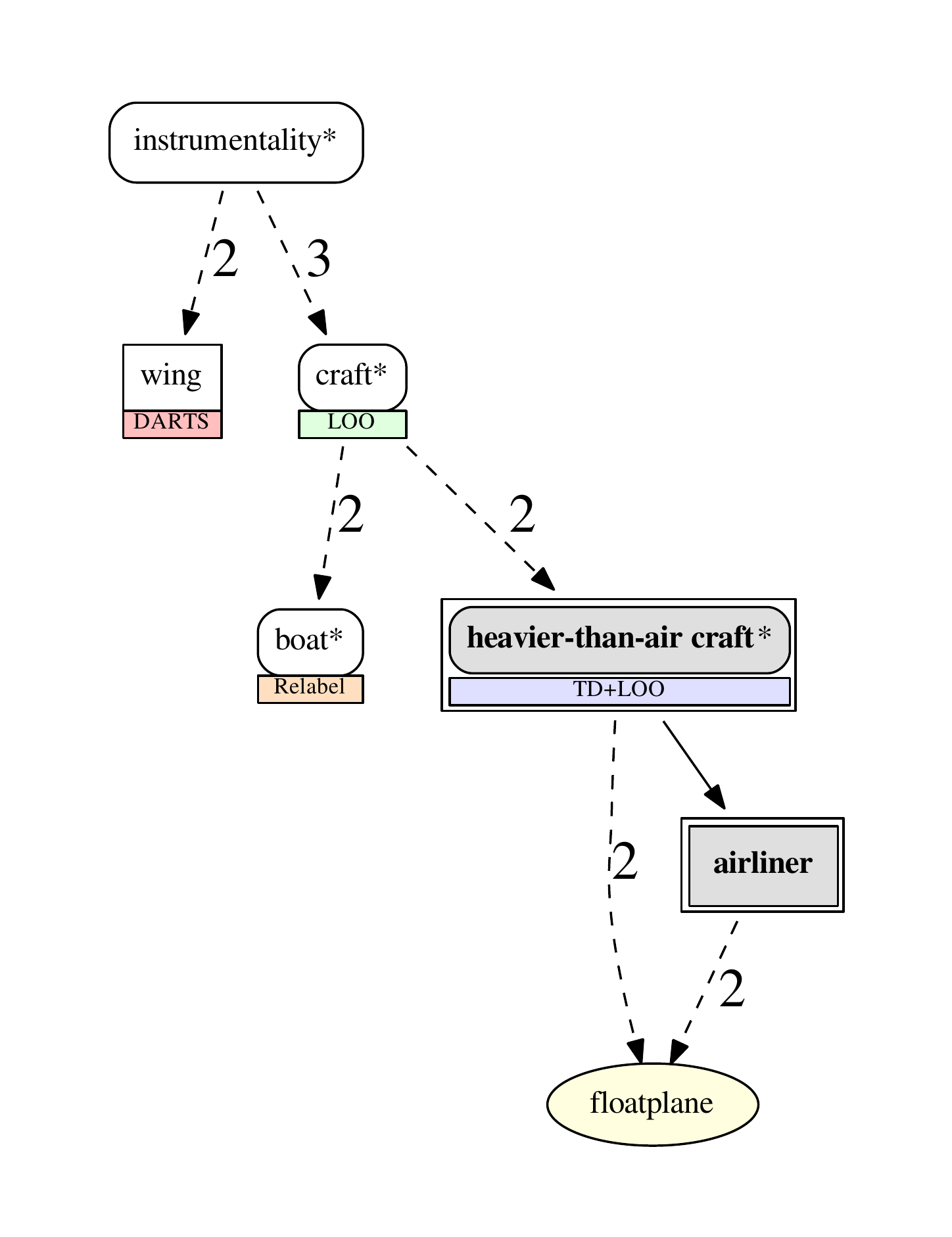}} & 
\multicolumn{4}{c}{\includegraphics[width=4.24cm, height=3.6cm, keepaspectratio]{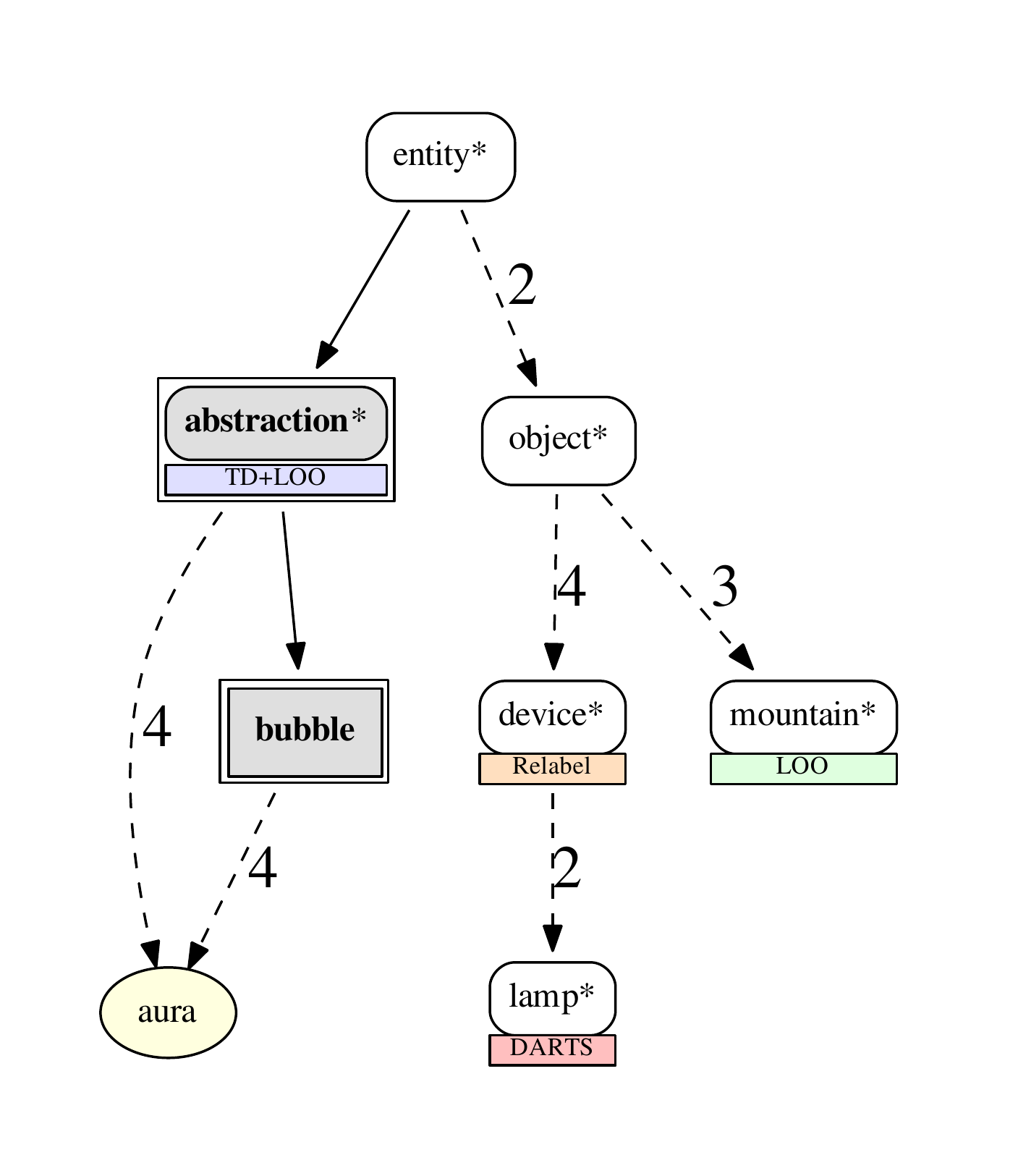}} & 
\multicolumn{4}{c}{\includegraphics[width=4.24cm, height=3.6cm, keepaspectratio]{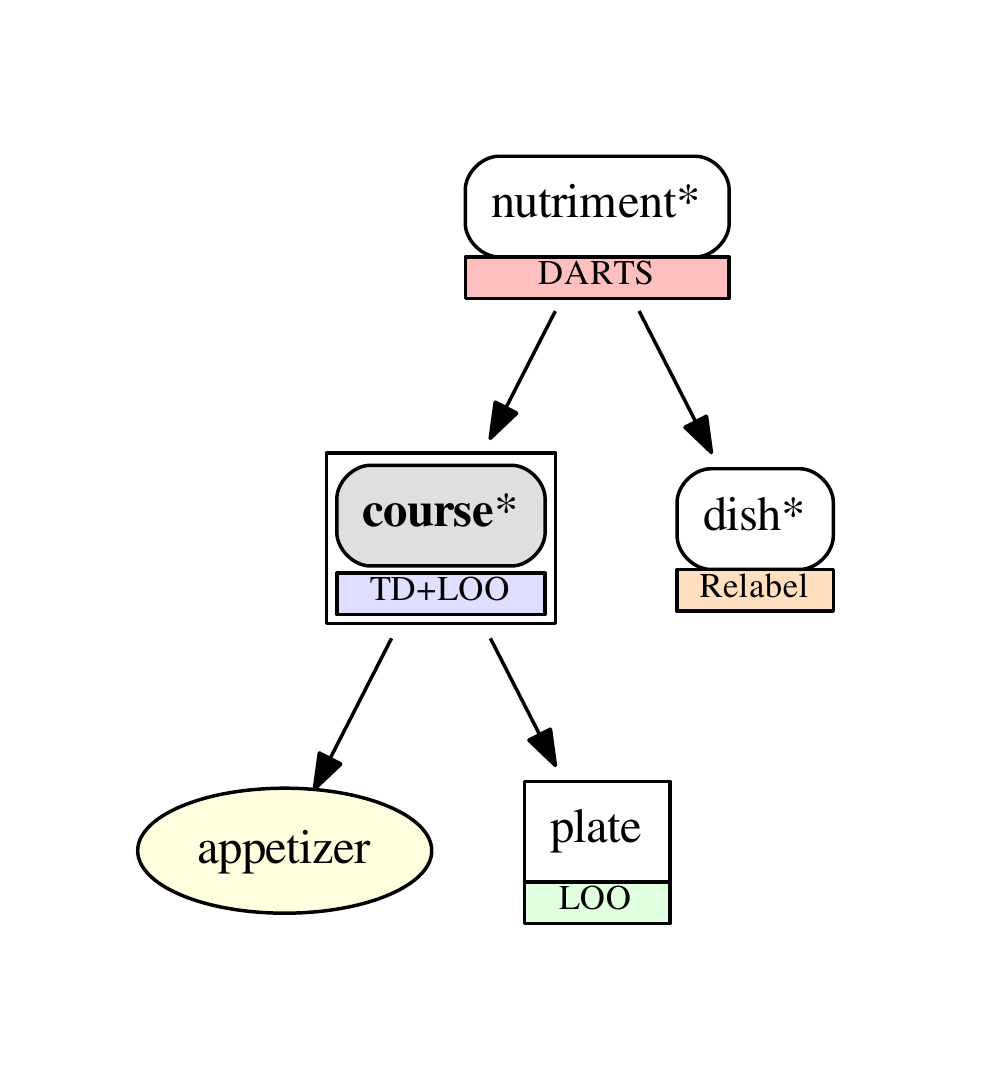}} \cr
\multicolumn{4}{c}{(e)} & 
\multicolumn{4}{c}{(f)} & 
\multicolumn{4}{c}{(g)} & 
\multicolumn{4}{c}{(h)} \cr
\multicolumn{4}{c}{\includegraphics[width=4.24cm, height=3.18cm, keepaspectratio]{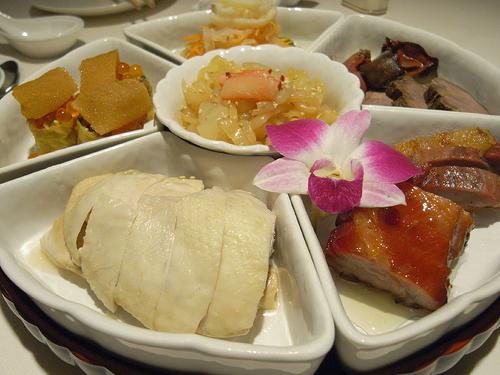}} & 
\multicolumn{4}{c}{\includegraphics[width=4.24cm, height=3.18cm, keepaspectratio]{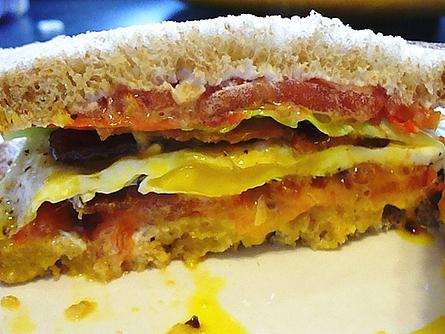}} & 
\multicolumn{4}{c}{\includegraphics[width=4.24cm, height=3.18cm, keepaspectratio]{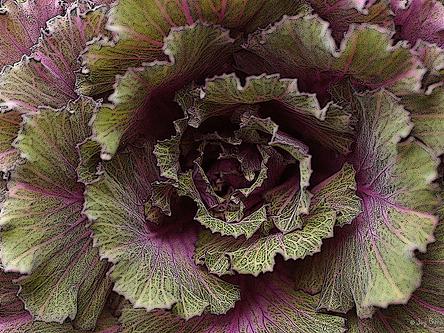}} & 
\multicolumn{4}{c}{\includegraphics[width=4.24cm, height=3.18cm, keepaspectratio]{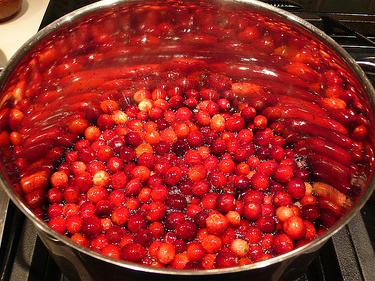}} \cr
\multicolumn{4}{l}{\cellcolor{ColorA} Novel class: hors d'oeuvre} & 
\multicolumn{4}{l}{\cellcolor{ColorA} Novel class: BLT sandwich} & 
\multicolumn{4}{l}{\cellcolor{ColorA} Novel class: kale} & 
\multicolumn{4}{l}{\cellcolor{ColorA} Novel class: cranberry} \cr
\cellcolor{ColorB} Method & \cellcolor{ColorB} $\epsilon$ & \cellcolor{ColorB} A & \multicolumn{1}{c}{\cellcolor{ColorB} Word} & 
\cellcolor{ColorB} Method & \cellcolor{ColorB} $\epsilon$ & \cellcolor{ColorB} A & \multicolumn{1}{c}{\cellcolor{ColorB} Word} & 
\cellcolor{ColorB} Method & \cellcolor{ColorB} $\epsilon$ & \cellcolor{ColorB} A & \multicolumn{1}{c}{\cellcolor{ColorB} Word} & 
\cellcolor{ColorB} Method & \cellcolor{ColorB} $\epsilon$ & \cellcolor{ColorB} A & \multicolumn{1}{c}{\cellcolor{ColorB} Word} \cr
\cellcolor{Color0} GT & \cellcolor{Color0} & \cellcolor{Color0} & \cellcolor{Color0} course & 
\cellcolor{Color0} GT & \cellcolor{Color0} & \cellcolor{Color0} & \cellcolor{Color0} sandwich & 
\cellcolor{Color0} GT & \cellcolor{Color0} & \cellcolor{Color0} & \cellcolor{Color0} {\scriptsize cruciferous vegetable} & 
\cellcolor{Color0} GT & \cellcolor{Color0} & \cellcolor{Color0} & \cellcolor{Color0} edible fruit \cr
\cellcolor{Color1} DARTS & \cellcolor{Color1} 1 & \cellcolor{Color1} N & \cellcolor{Color1} plate & 
\cellcolor{Color1} DARTS & \cellcolor{Color1} 2 & \cellcolor{Color1} Y & \cellcolor{Color1} nutriment & 
\cellcolor{Color1} DARTS & \cellcolor{Color1} 0 & \cellcolor{Color1} Y & \cellcolor{Color1} {\scriptsize cruciferous vegetable} & 
\cellcolor{Color1} DARTS & \cellcolor{Color1} 0 & \cellcolor{Color1} Y & \cellcolor{Color1} fruit \cr
\cellcolor{Color2} Relabel & \cellcolor{Color2} 2 & \cellcolor{Color2} N & \cellcolor{Color2} dish & 
\cellcolor{Color2} Relabel & \cellcolor{Color2} 1 & \cellcolor{Color2} N & \cellcolor{Color2} cheeseburger & 
\cellcolor{Color2} Relabel & \cellcolor{Color2} 1 & \cellcolor{Color2} Y & \cellcolor{Color2} vegetable & 
\cellcolor{Color2} Relabel & \cellcolor{Color2} 0 & \cellcolor{Color2} Y & \cellcolor{Color2} edible fruit \cr
\cellcolor{Color3} LOO & \cellcolor{Color3} 1 & \cellcolor{Color3} Y & \cellcolor{Color3} nutriment & 
\cellcolor{Color3} LOO & \cellcolor{Color3} 0 & \cellcolor{Color3} Y & \cellcolor{Color3} sandwich & 
\cellcolor{Color3} LOO & \cellcolor{Color3} 1 & \cellcolor{Color3} Y & \cellcolor{Color3} vegetable & 
\cellcolor{Color3} LOO & \cellcolor{Color3} 1 & \cellcolor{Color3} N & \cellcolor{Color3} pomegranate \cr
\cellcolor{Color4} TD+LOO & \cellcolor{Color4} 0 & \cellcolor{Color4} Y & \cellcolor{Color4} course & 
\cellcolor{Color4} TD+LOO & \cellcolor{Color4} 0 & \cellcolor{Color4} Y & \cellcolor{Color4} sandwich & 
\cellcolor{Color4} TD+LOO & \cellcolor{Color4} 0 & \cellcolor{Color4} Y & \cellcolor{Color4} head cabbage & 
\cellcolor{Color4} TD+LOO & \cellcolor{Color4} 0 & \cellcolor{Color4} Y & \cellcolor{Color4} strawberry \cr
\multicolumn{4}{c}{\includegraphics[width=4.24cm, height=3.6cm, keepaspectratio]{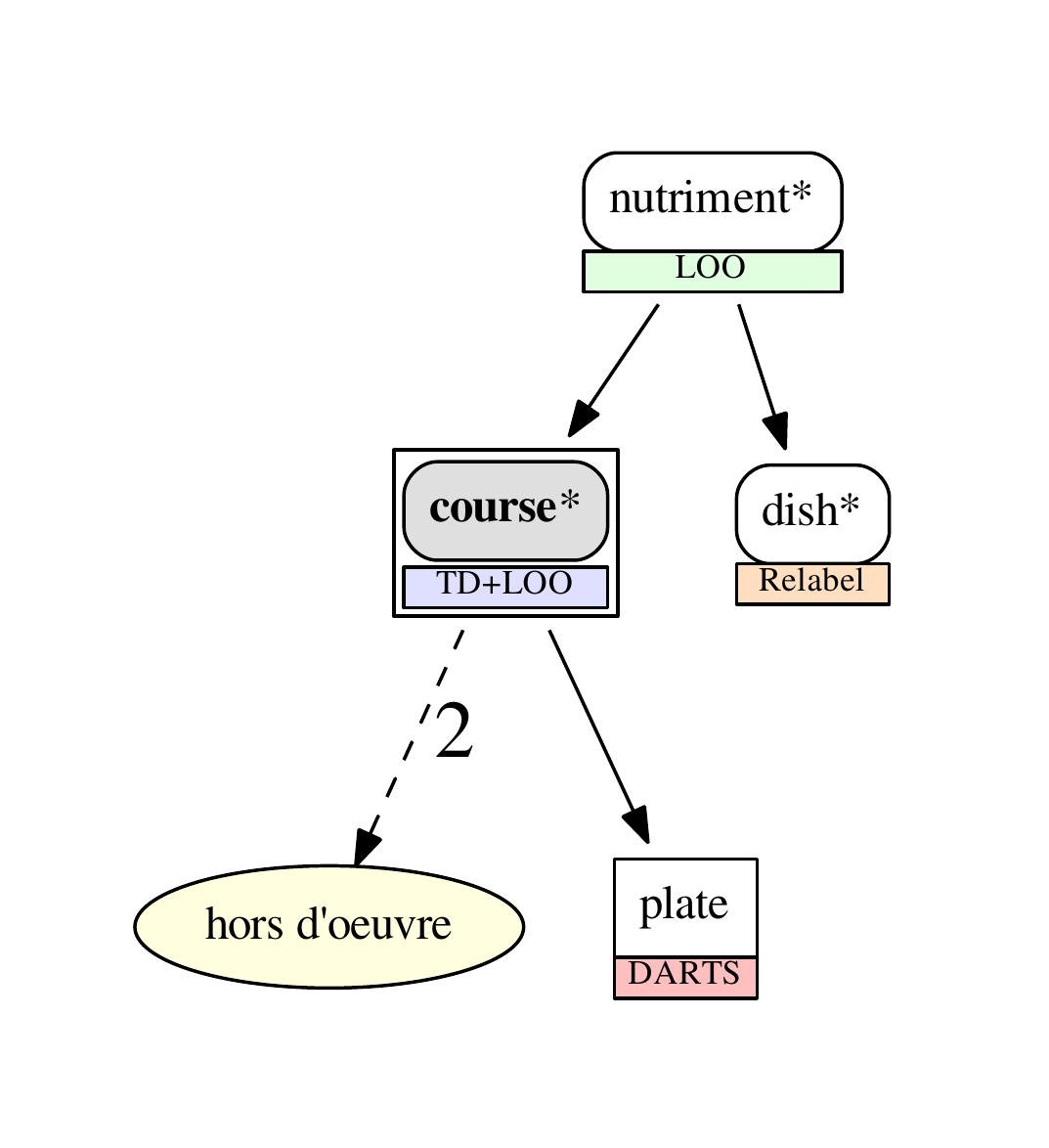}} & 
\multicolumn{4}{c}{\includegraphics[width=4.24cm, height=3.6cm, keepaspectratio]{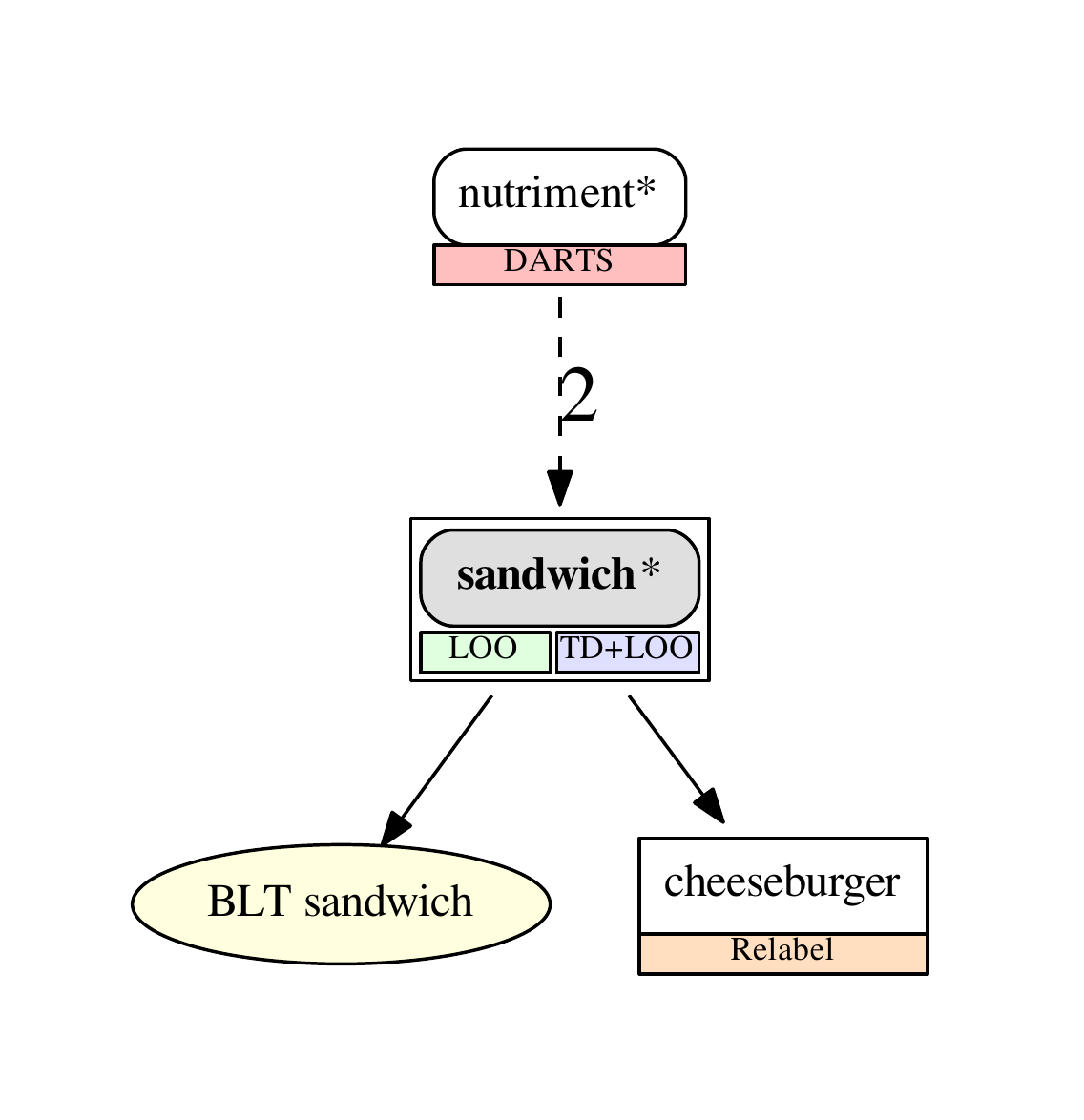}} & 
\multicolumn{4}{c}{\includegraphics[width=4.24cm, height=3.6cm, keepaspectratio]{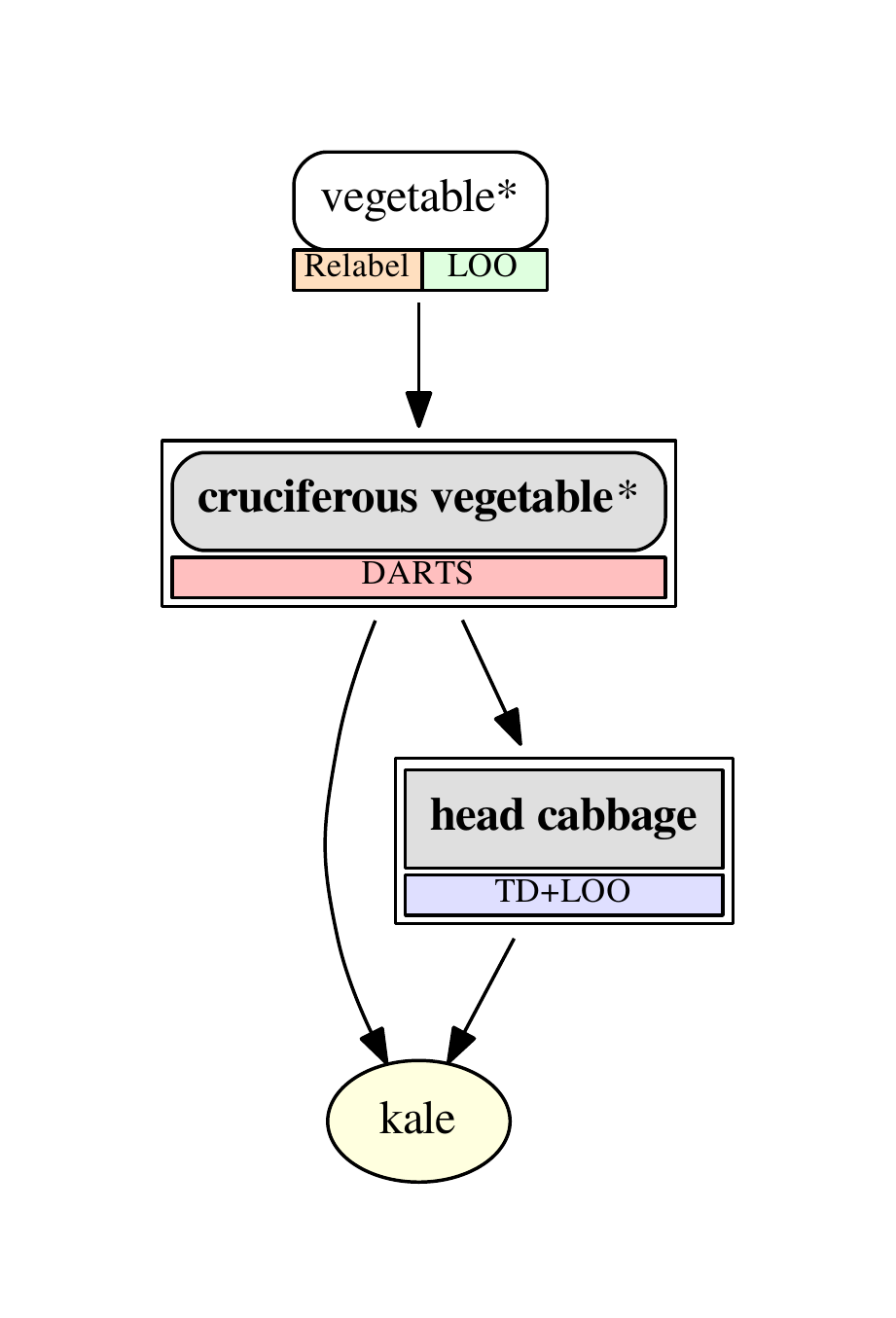}} & 
\multicolumn{4}{c}{\includegraphics[width=4.24cm, height=3.6cm, keepaspectratio]{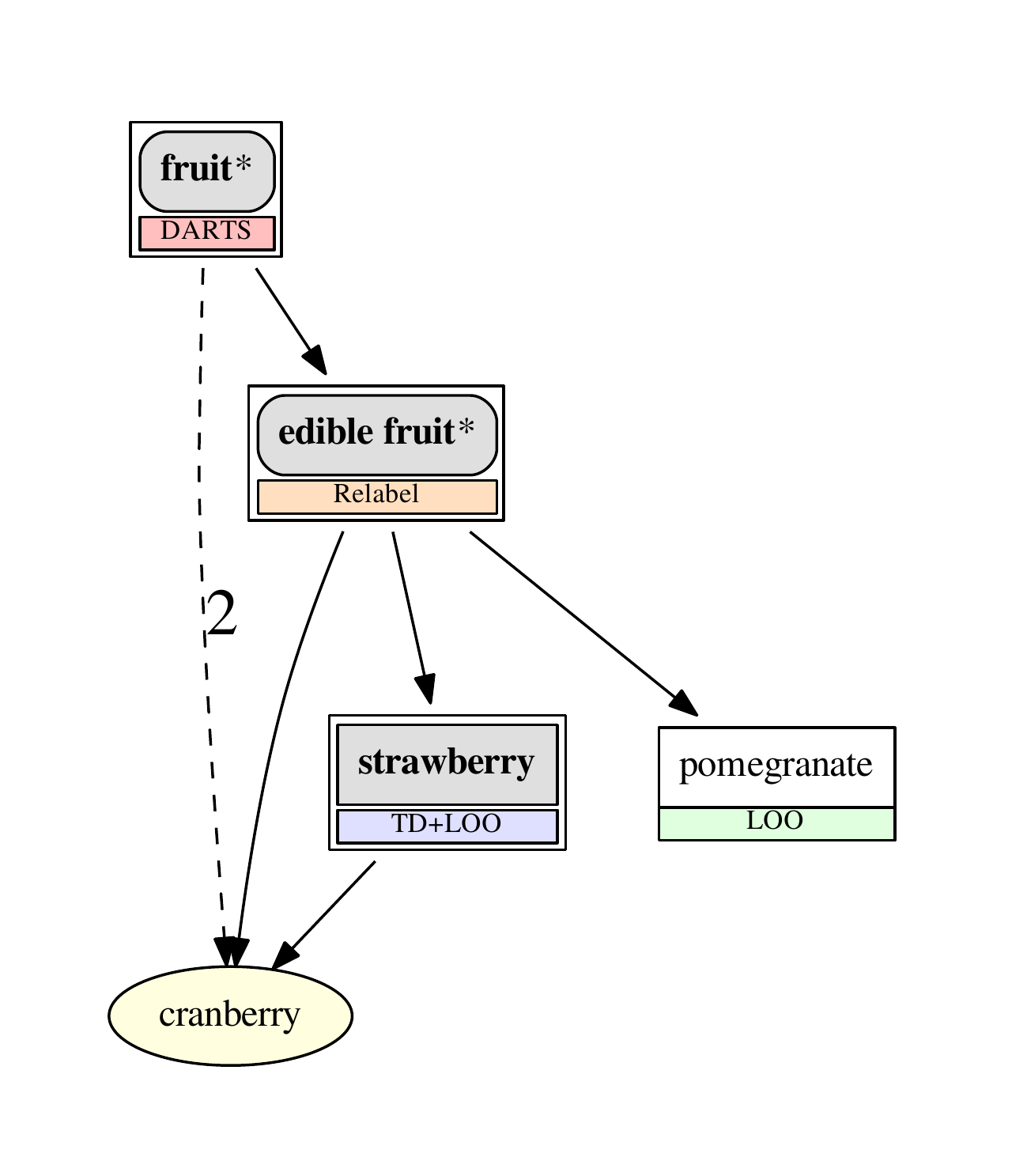}} \cr
\end{tabular}
\caption{Qualitative results of hierarchical novelty detection on ImageNet.
``GT'' is the closest known ancestor of the novel class, which is the expected prediction,
``DARTS'' is the baseline method proposed in \cite{deng2012hedging} where we modify the method for our purpose, and the others are our proposed methods.
``$\epsilon$'' is the distance between the prediction and GT,
``A'' indicates whether the prediction is an ancestor of GT, and
``Word'' is the English word of the predicted label.
Dashed edges represent multi-hop connection, where the number indicates the number of edges between classes.
If the prediction is on a super class (marked with * and rounded), then the test image is classified as a novel class whose closest class in the taxonomy is the super class.
}
\label{fig:qual_smp_7}
\end{figure*}

%% file: qual_smp/qual_smp_8.tex
\begin{figure*}[t]
\footnotesize\centering\setlength{\tabcolsep}{0cm}
\begin{tabular}{
>{\centering}m{1.12cm}>{\centering}m{0.4cm}>{\centering}m{0.4cm}m{2.32cm}
>{\centering}m{1.12cm}>{\centering}m{0.4cm}>{\centering}m{0.4cm}m{2.32cm}
>{\centering}m{1.12cm}>{\centering}m{0.4cm}>{\centering}m{0.4cm}m{2.32cm}
>{\centering}m{1.12cm}>{\centering}m{0.4cm}>{\centering}m{0.4cm}m{2.32cm}
}
\multicolumn{4}{c}{(a)} & 
\multicolumn{4}{c}{(b)} & 
\multicolumn{4}{c}{(c)} & 
\multicolumn{4}{c}{(d)} \cr
\multicolumn{4}{c}{\includegraphics[width=4.24cm, height=3.18cm, keepaspectratio]{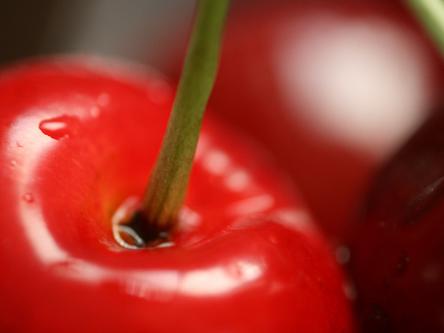}} & 
\multicolumn{4}{c}{\includegraphics[width=4.24cm, height=3.18cm, keepaspectratio]{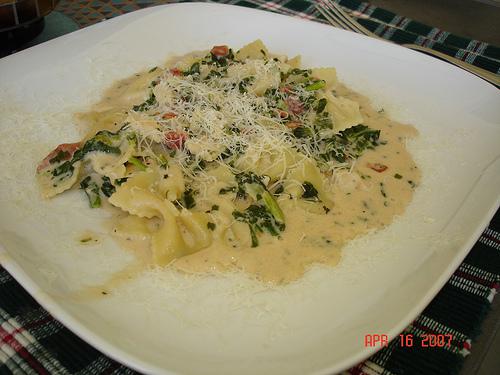}} & 
\multicolumn{4}{c}{\includegraphics[width=4.24cm, height=3.18cm, keepaspectratio]{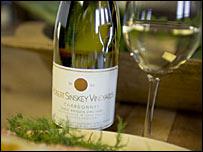}} & 
\multicolumn{4}{c}{\includegraphics[width=4.24cm, height=3.18cm, keepaspectratio]{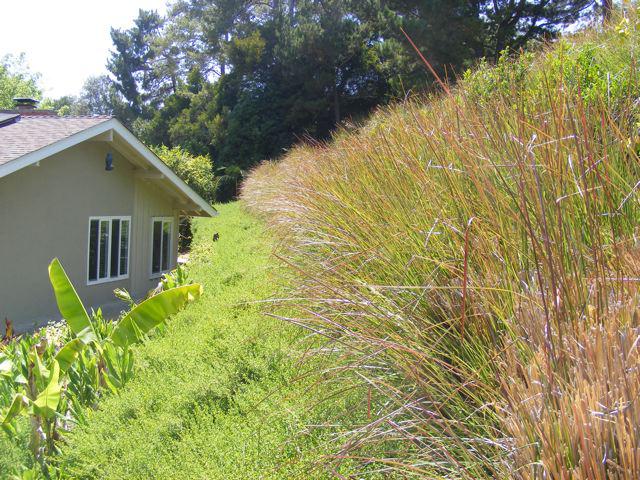}} \cr
\multicolumn{4}{l}{\cellcolor{ColorA} Novel class: cherry} & 
\multicolumn{4}{l}{\cellcolor{ColorA} Novel class: cream sauce} & 
\multicolumn{4}{l}{\cellcolor{ColorA} Novel class: Chardonnay} & 
\multicolumn{4}{l}{\cellcolor{ColorA} Novel class: hillside} \cr
\cellcolor{ColorB} Method & \cellcolor{ColorB} $\epsilon$ & \cellcolor{ColorB} A & \multicolumn{1}{c}{\cellcolor{ColorB} Word} & 
\cellcolor{ColorB} Method & \cellcolor{ColorB} $\epsilon$ & \cellcolor{ColorB} A & \multicolumn{1}{c}{\cellcolor{ColorB} Word} & 
\cellcolor{ColorB} Method & \cellcolor{ColorB} $\epsilon$ & \cellcolor{ColorB} A & \multicolumn{1}{c}{\cellcolor{ColorB} Word} & 
\cellcolor{ColorB} Method & \cellcolor{ColorB} $\epsilon$ & \cellcolor{ColorB} A & \multicolumn{1}{c}{\cellcolor{ColorB} Word} \cr
\cellcolor{Color0} GT & \cellcolor{Color0} & \cellcolor{Color0} & \cellcolor{Color0} edible fruit & 
\cellcolor{Color0} GT & \cellcolor{Color0} & \cellcolor{Color0} & \cellcolor{Color0} sauce & 
\cellcolor{Color0} GT & \cellcolor{Color0} & \cellcolor{Color0} & \cellcolor{Color0} alcohol & 
\cellcolor{Color0} GT & \cellcolor{Color0} & \cellcolor{Color0} & \cellcolor{Color0} {\scriptsize geological formation} \cr
\cellcolor{Color1} DARTS & \cellcolor{Color1} 3 & \cellcolor{Color1} N & \cellcolor{Color1} {\scriptsize solanaceous vegetable} & 
\cellcolor{Color1} DARTS & \cellcolor{Color1} 3 & \cellcolor{Color1} Y & \cellcolor{Color1} food, nutrient & 
\cellcolor{Color1} DARTS & \cellcolor{Color1} 11 & \cellcolor{Color1} N & \cellcolor{Color1} wine bottle & 
\cellcolor{Color1} DARTS & \cellcolor{Color1} 6 & \cellcolor{Color1} N & \cellcolor{Color1} roof \cr
\cellcolor{Color2} Relabel & \cellcolor{Color2} 1 & \cellcolor{Color2} N & \cellcolor{Color2} Granny Smith & 
\cellcolor{Color2} Relabel & \cellcolor{Color2} 5 & \cellcolor{Color2} N & \cellcolor{Color2} dish & 
\cellcolor{Color2} Relabel & \cellcolor{Color2} 10 & \cellcolor{Color2} N & \cellcolor{Color2} bottle & 
\cellcolor{Color2} Relabel & \cellcolor{Color2} 6 & \cellcolor{Color2} N & \cellcolor{Color2} fence \cr
\cellcolor{Color3} LOO & \cellcolor{Color3} 0 & \cellcolor{Color3} Y & \cellcolor{Color3} fruit & 
\cellcolor{Color3} LOO & \cellcolor{Color3} 1 & \cellcolor{Color3} N & \cellcolor{Color3} carbonara & 
\cellcolor{Color3} LOO & \cellcolor{Color3} 0 & \cellcolor{Color3} Y & \cellcolor{Color3} alcohol & 
\cellcolor{Color3} LOO & \cellcolor{Color3} 5 & \cellcolor{Color3} N & \cellcolor{Color3} housing \cr
\cellcolor{Color4} TD+LOO & \cellcolor{Color4} 4 & \cellcolor{Color4} N & \cellcolor{Color4} bell pepper & 
\cellcolor{Color4} TD+LOO & \cellcolor{Color4} 0 & \cellcolor{Color4} Y & \cellcolor{Color4} sauce & 
\cellcolor{Color4} TD+LOO & \cellcolor{Color4} 0 & \cellcolor{Color4} Y & \cellcolor{Color4} red wine & 
\cellcolor{Color4} TD+LOO & \cellcolor{Color4} 0 & \cellcolor{Color4} Y & \cellcolor{Color4} {\scriptsize geological formation} \cr
\multicolumn{4}{c}{\includegraphics[width=4.24cm, height=3.6cm, keepaspectratio]{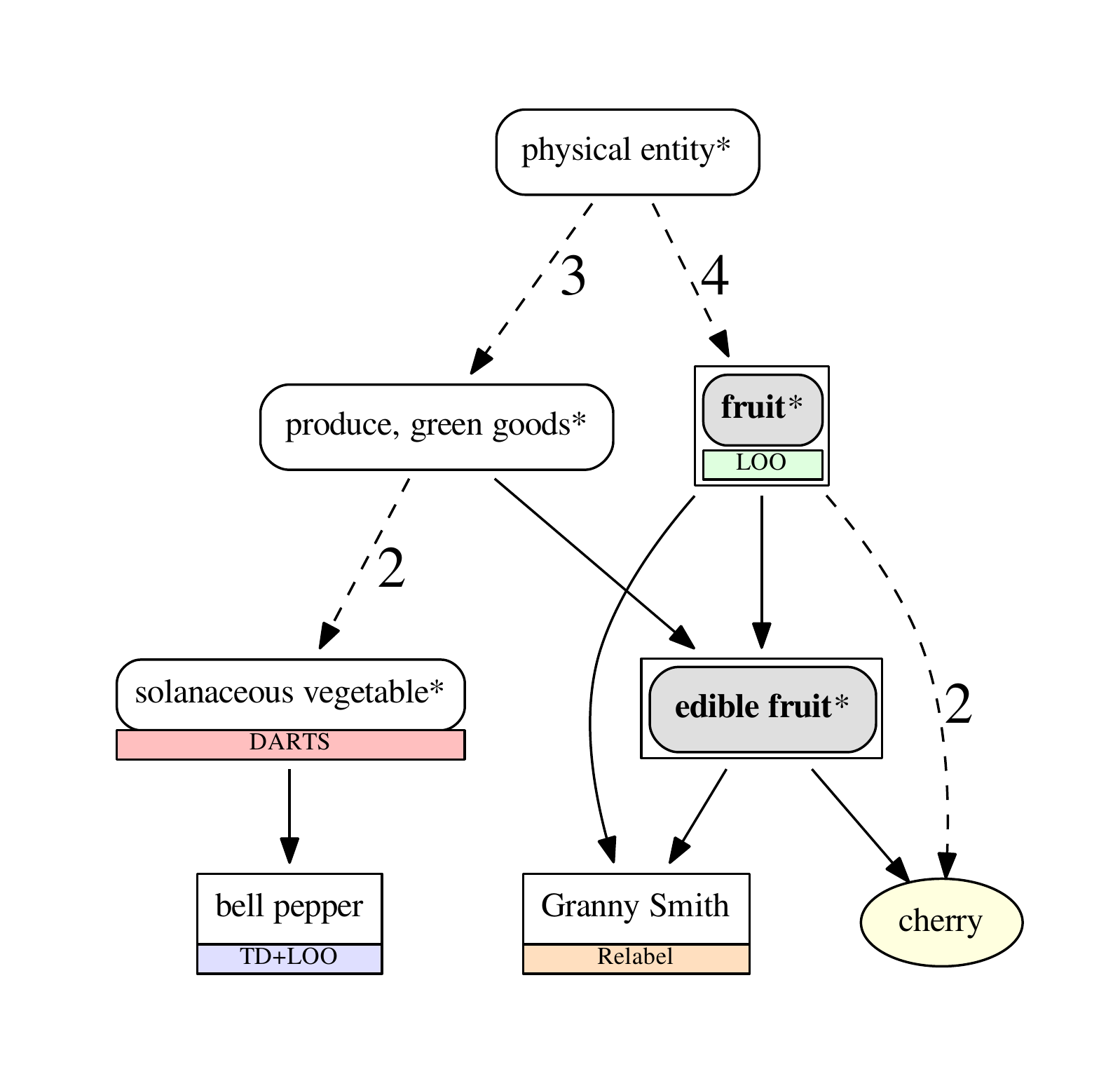}} & 
\multicolumn{4}{c}{\includegraphics[width=4.24cm, height=3.6cm, keepaspectratio]{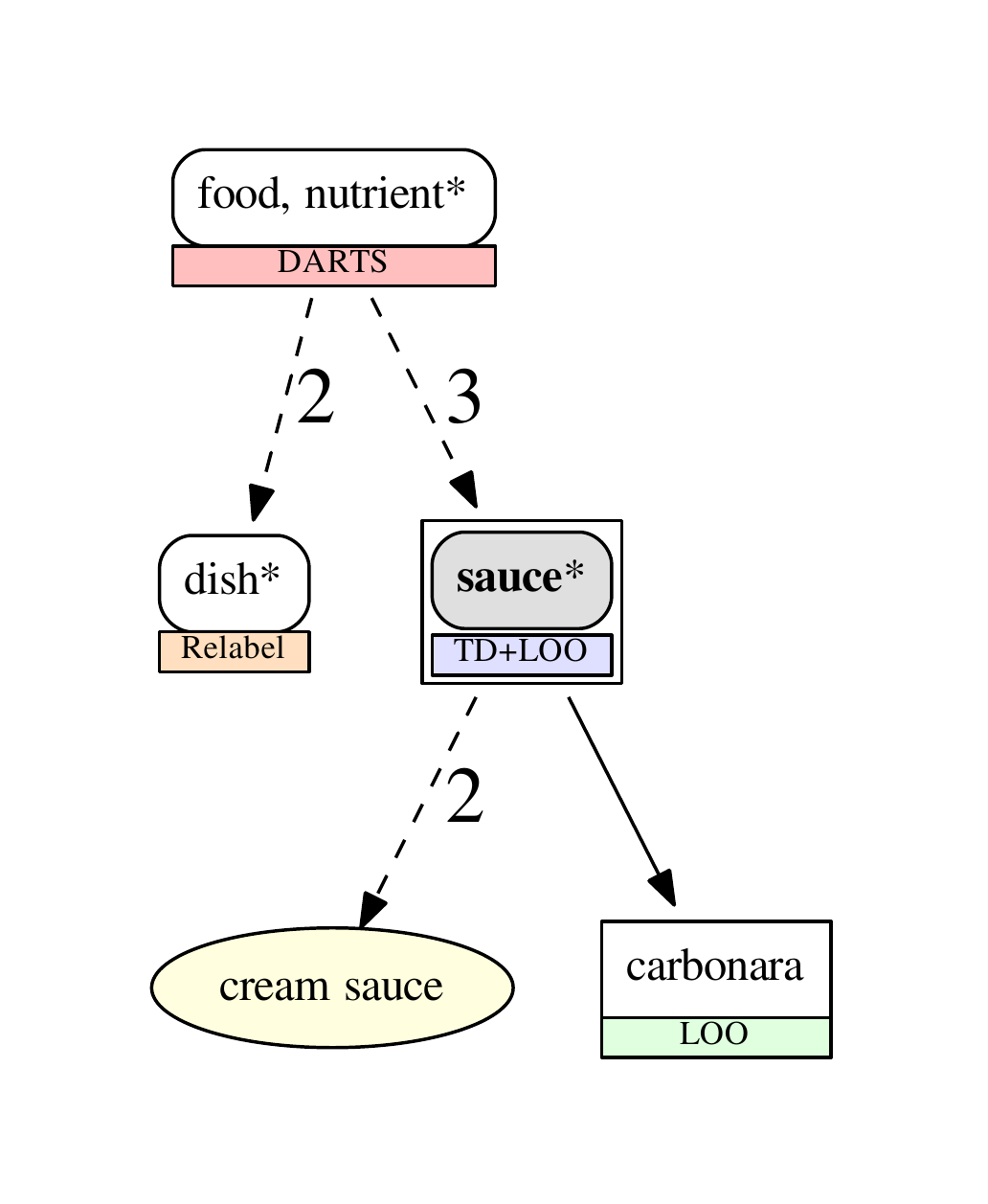}} & 
\multicolumn{4}{c}{\includegraphics[width=4.24cm, height=3.6cm, keepaspectratio]{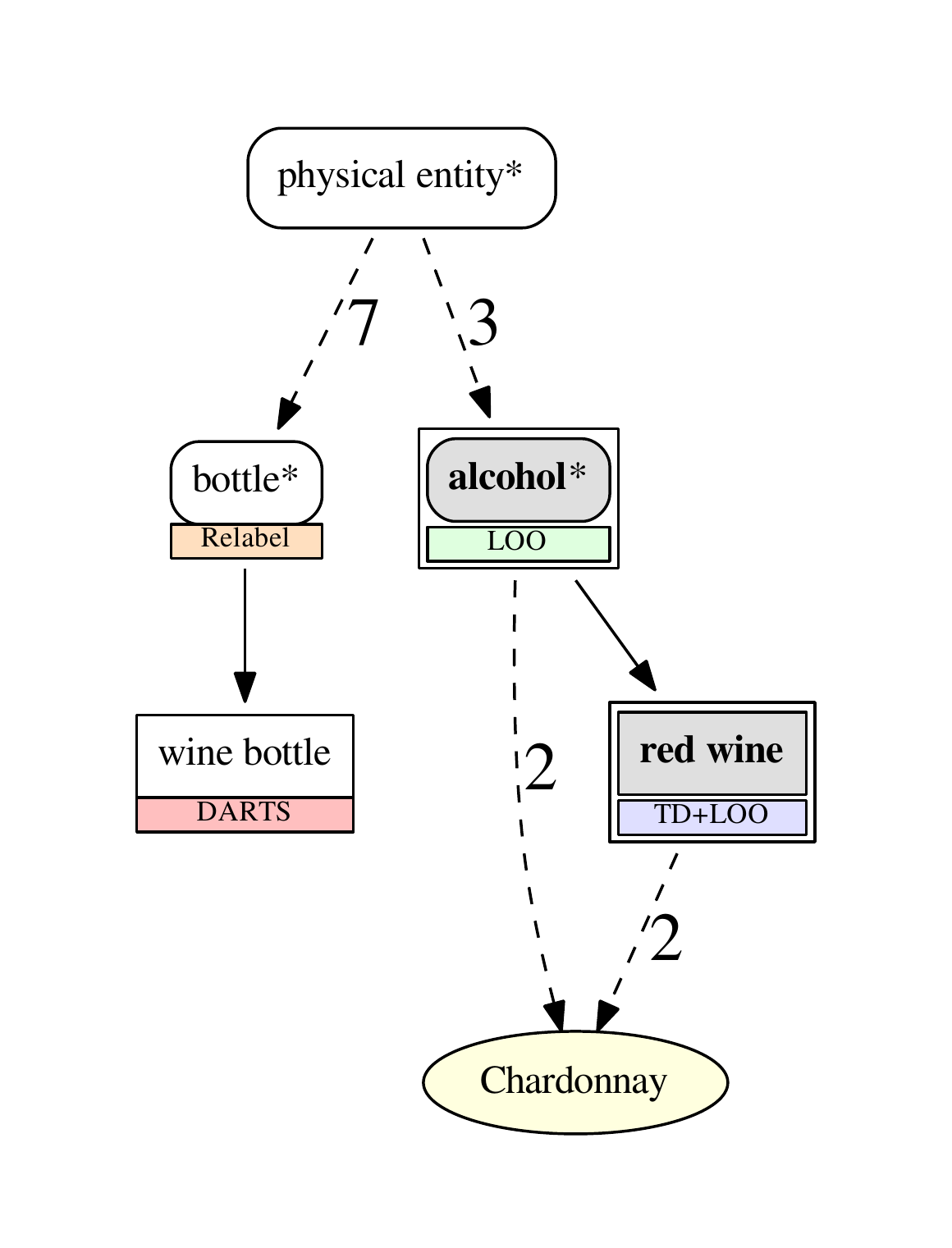}} & 
\multicolumn{4}{c}{\includegraphics[width=4.24cm, height=3.6cm, keepaspectratio]{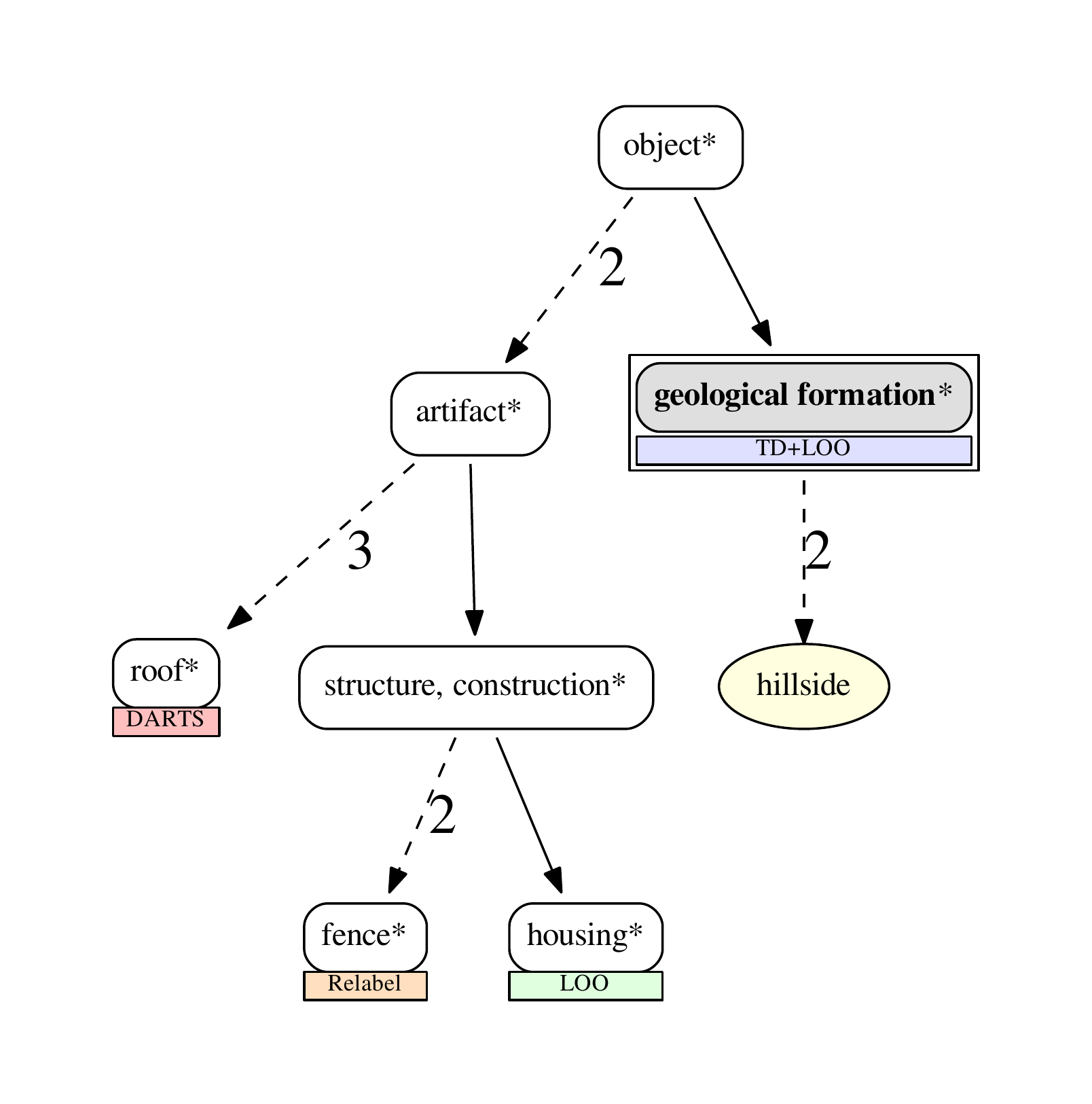}} \cr
\multicolumn{4}{c}{(e)} & 
\multicolumn{4}{c}{(f)} & 
\multicolumn{4}{c}{(g)} & 
\multicolumn{4}{c}{(h)} \cr
\multicolumn{4}{c}{\includegraphics[width=4.24cm, height=3.18cm, keepaspectratio]{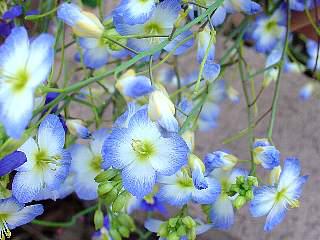}} & 
\multicolumn{4}{c}{\includegraphics[width=4.24cm, height=3.18cm, keepaspectratio]{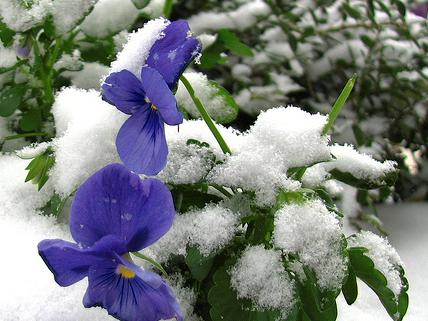}} & 
\multicolumn{4}{c}{\includegraphics[width=4.24cm, height=3.18cm, keepaspectratio]{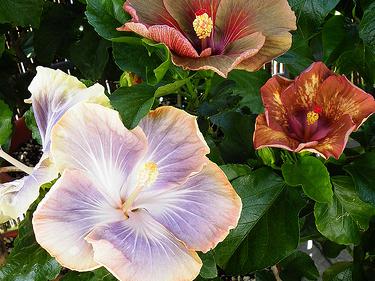}} & 
\multicolumn{4}{c}{\includegraphics[width=4.24cm, height=3.18cm, keepaspectratio]{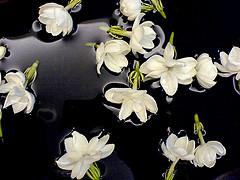}} \cr
\multicolumn{4}{l}{\cellcolor{ColorA} Novel class: heliophila} & 
\multicolumn{4}{l}{\cellcolor{ColorA} Novel class: tangle orchid} & 
\multicolumn{4}{l}{\cellcolor{ColorA} Novel class: rose mallow} & 
\multicolumn{4}{l}{\cellcolor{ColorA} Novel class: jasmine} \cr
\cellcolor{ColorB} Method & \cellcolor{ColorB} $\epsilon$ & \cellcolor{ColorB} A & \multicolumn{1}{c}{\cellcolor{ColorB} Word} & 
\cellcolor{ColorB} Method & \cellcolor{ColorB} $\epsilon$ & \cellcolor{ColorB} A & \multicolumn{1}{c}{\cellcolor{ColorB} Word} & 
\cellcolor{ColorB} Method & \cellcolor{ColorB} $\epsilon$ & \cellcolor{ColorB} A & \multicolumn{1}{c}{\cellcolor{ColorB} Word} & 
\cellcolor{ColorB} Method & \cellcolor{ColorB} $\epsilon$ & \cellcolor{ColorB} A & \multicolumn{1}{c}{\cellcolor{ColorB} Word} \cr
\cellcolor{Color0} GT & \cellcolor{Color0} & \cellcolor{Color0} & \cellcolor{Color0} flower & 
\cellcolor{Color0} GT & \cellcolor{Color0} & \cellcolor{Color0} & \cellcolor{Color0} flower & 
\cellcolor{Color0} GT & \cellcolor{Color0} & \cellcolor{Color0} & \cellcolor{Color0} organism, being & 
\cellcolor{Color0} GT & \cellcolor{Color0} & \cellcolor{Color0} & \cellcolor{Color0} organism, being \cr
\cellcolor{Color1} DARTS & \cellcolor{Color1} 3 & \cellcolor{Color1} N & \cellcolor{Color1} earthstar & 
\cellcolor{Color1} DARTS & \cellcolor{Color1} 6 & \cellcolor{Color1} N & \cellcolor{Color1} pot, flowerpot & 
\cellcolor{Color1} DARTS & \cellcolor{Color1} 5 & \cellcolor{Color1} N & \cellcolor{Color1} pot, flowerpot & 
\cellcolor{Color1} DARTS & \cellcolor{Color1} 6 & \cellcolor{Color1} N & \cellcolor{Color1} jar \cr
\cellcolor{Color2} Relabel & \cellcolor{Color2} 8 & \cellcolor{Color2} N & \cellcolor{Color2} vegetable & 
\cellcolor{Color2} Relabel & \cellcolor{Color2} 1 & \cellcolor{Color2} N & \cellcolor{Color2} daisy & 
\cellcolor{Color2} Relabel & \cellcolor{Color2} 7 & \cellcolor{Color2} N & \cellcolor{Color2} vegetable & 
\cellcolor{Color2} Relabel & \cellcolor{Color2} 1 & \cellcolor{Color2} N & \cellcolor{Color2} daisy \cr
\cellcolor{Color3} LOO & \cellcolor{Color3} 1 & \cellcolor{Color3} Y & \cellcolor{Color3} organism, being & 
\cellcolor{Color3} LOO & \cellcolor{Color3} 8 & \cellcolor{Color3} N & \cellcolor{Color3} vegetable & 
\cellcolor{Color3} LOO & \cellcolor{Color3} 0 & \cellcolor{Color3} Y & \cellcolor{Color3} organism, being & 
\cellcolor{Color3} LOO & \cellcolor{Color3} 0 & \cellcolor{Color3} Y & \cellcolor{Color3} organism, being \cr
\cellcolor{Color4} TD+LOO & \cellcolor{Color4} 0 & \cellcolor{Color4} Y & \cellcolor{Color4} flower & 
\cellcolor{Color4} TD+LOO & \cellcolor{Color4} 0 & \cellcolor{Color4} Y & \cellcolor{Color4} flower & 
\cellcolor{Color4} TD+LOO & \cellcolor{Color4} 0 & \cellcolor{Color4} Y & \cellcolor{Color4} flower & 
\cellcolor{Color4} TD+LOO & \cellcolor{Color4} 0 & \cellcolor{Color4} Y & \cellcolor{Color4} flower \cr
\multicolumn{4}{c}{\includegraphics[width=4.24cm, height=3.6cm, keepaspectratio]{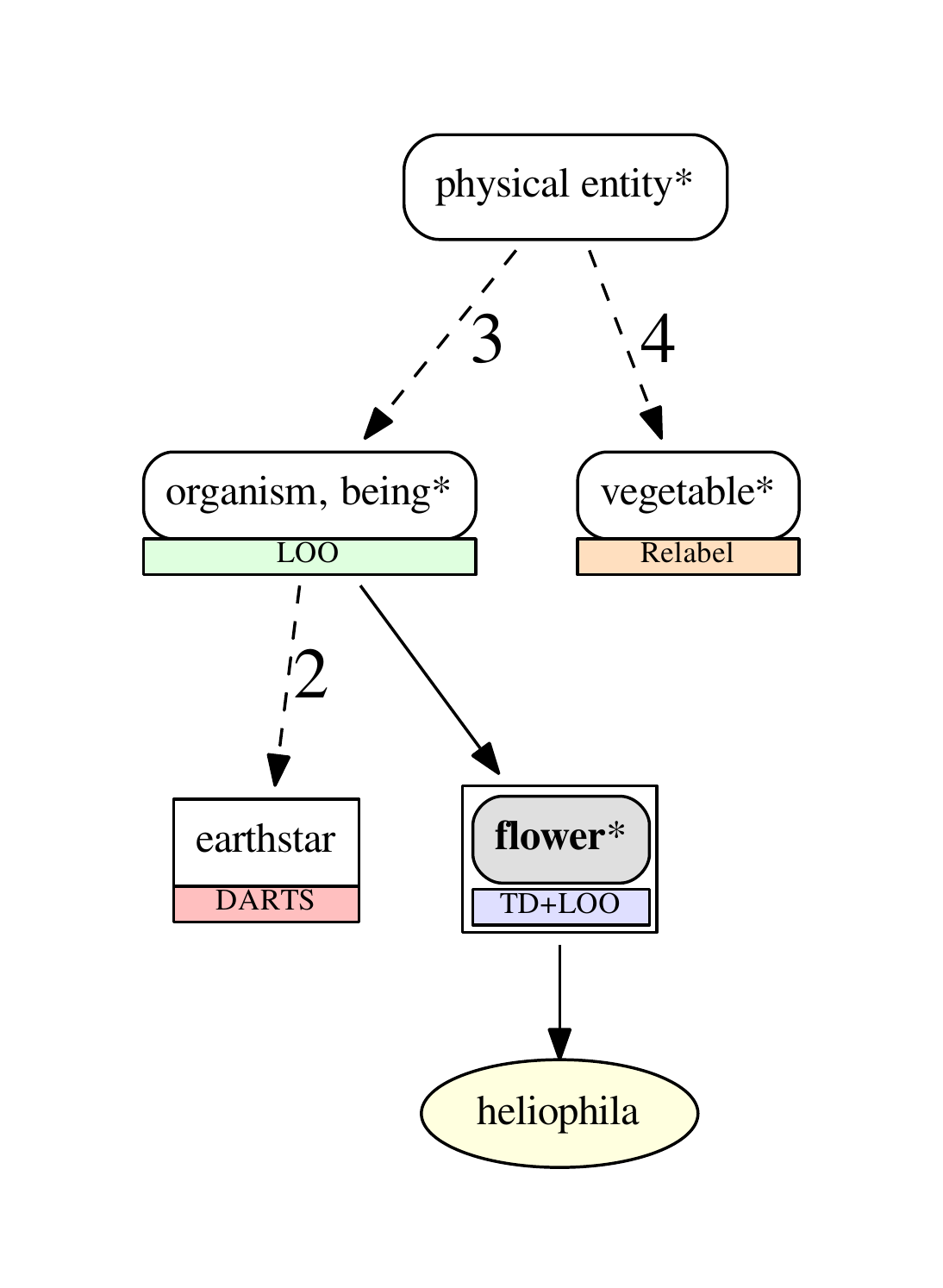}} & 
\multicolumn{4}{c}{\includegraphics[width=4.24cm, height=3.6cm, keepaspectratio]{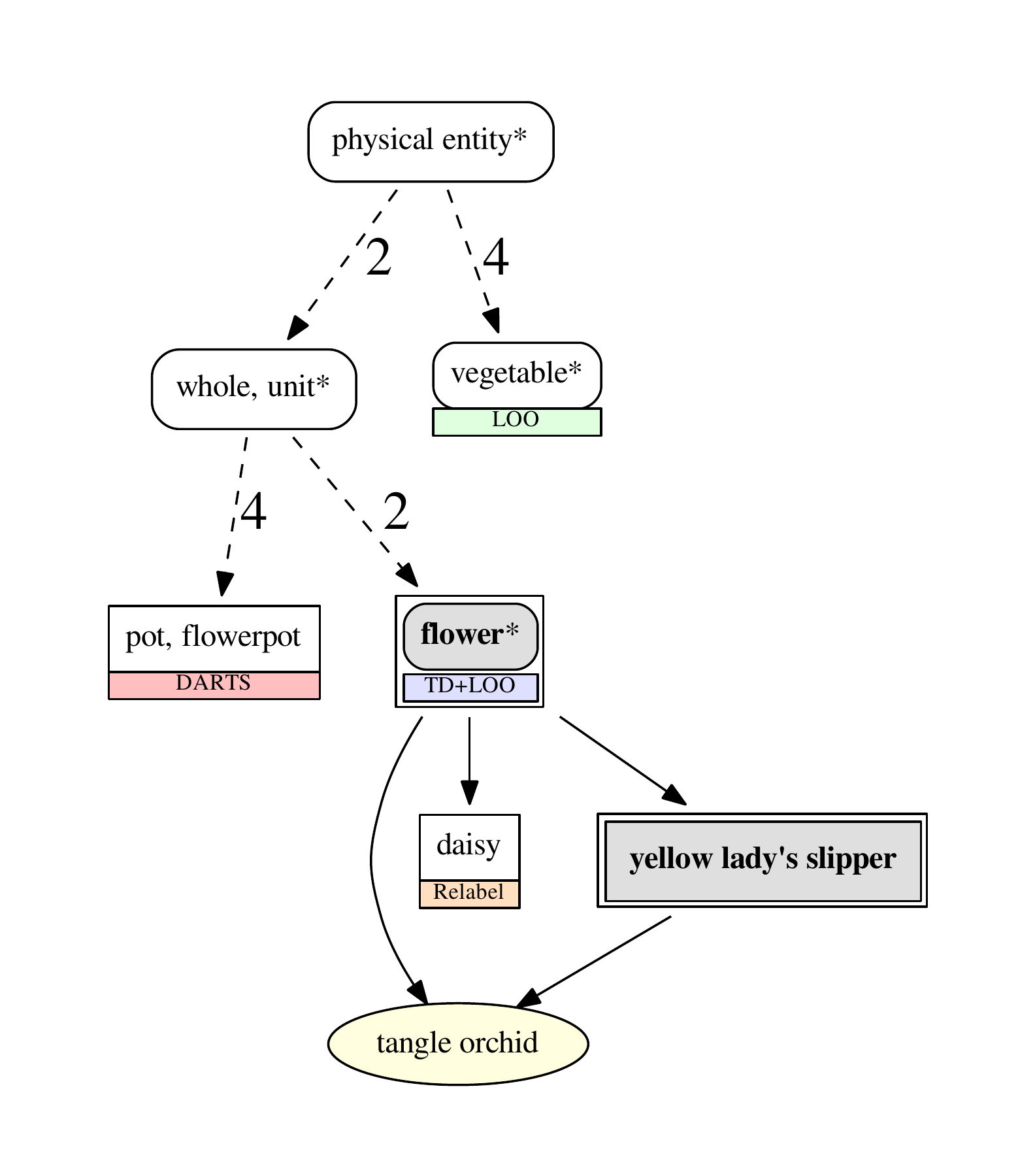}} & 
\multicolumn{4}{c}{\includegraphics[width=4.24cm, height=3.6cm, keepaspectratio]{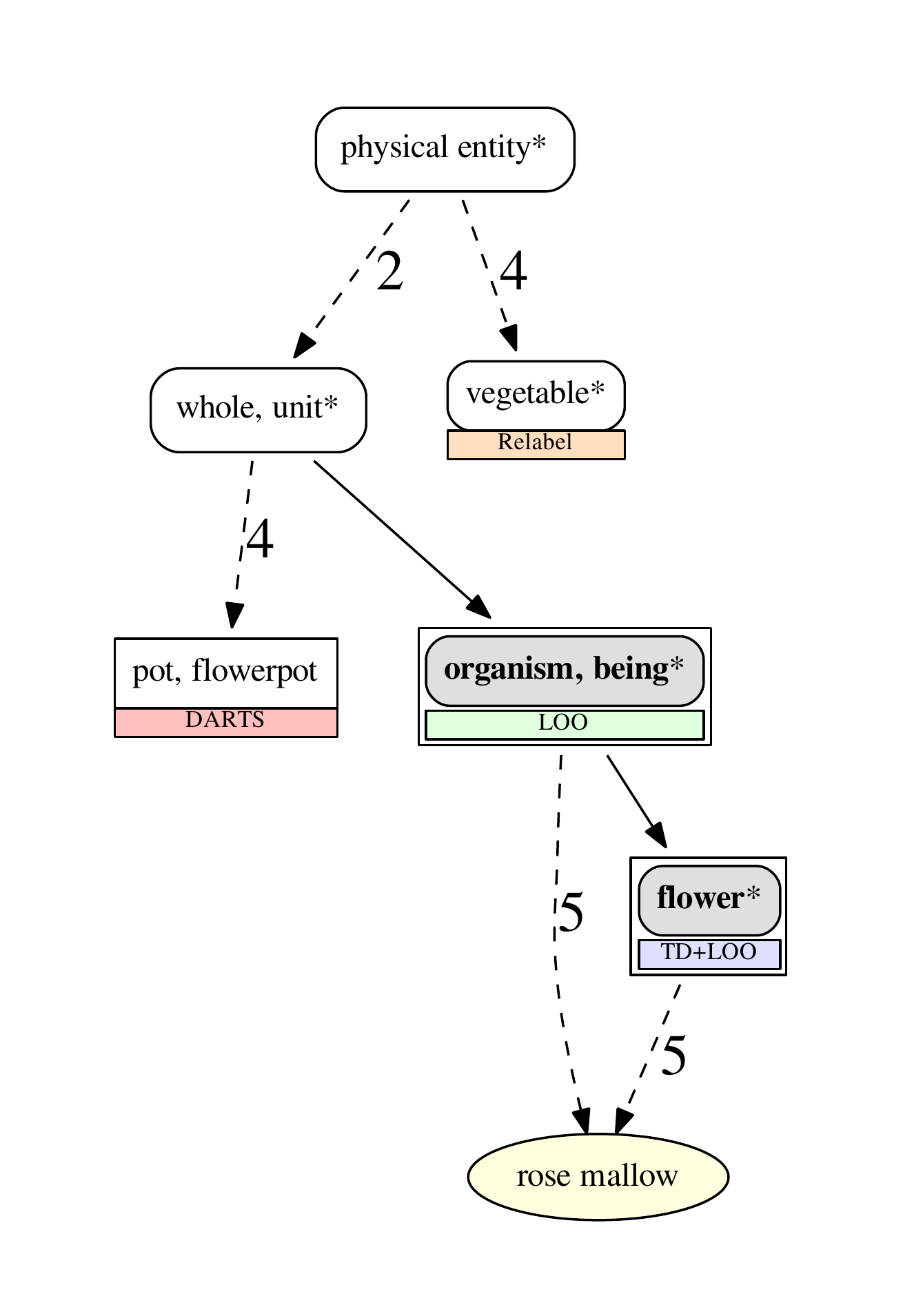}} & 
\multicolumn{4}{c}{\includegraphics[width=4.24cm, height=3.6cm, keepaspectratio]{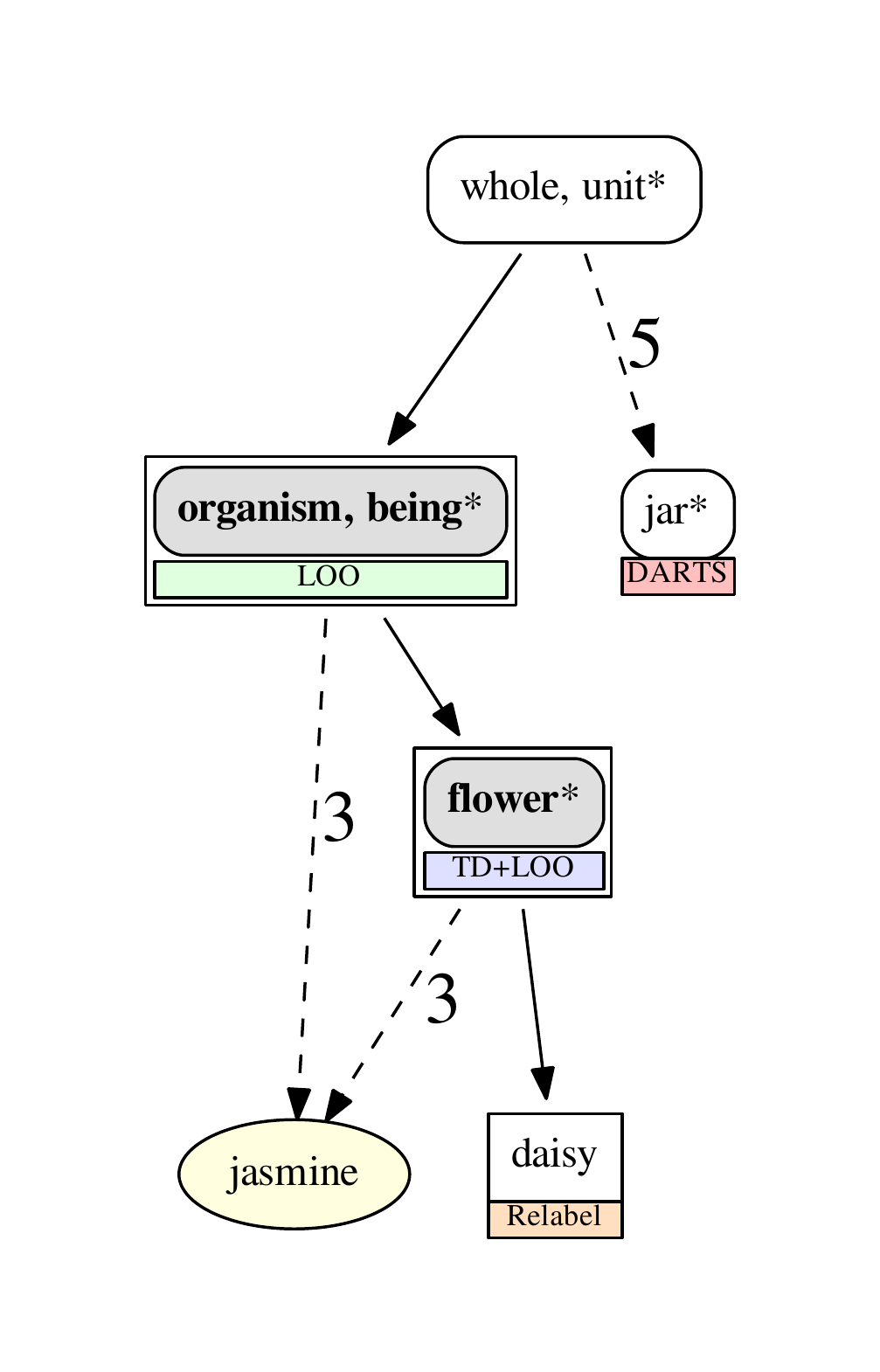}} \cr
\end{tabular}
\caption{Qualitative results of hierarchical novelty detection on ImageNet.
``GT'' is the closest known ancestor of the novel class, which is the expected prediction,
``DARTS'' is the baseline method proposed in \cite{deng2012hedging} where we modify the method for our purpose, and the others are our proposed methods.
``$\epsilon$'' is the distance between the prediction and GT,
``A'' indicates whether the prediction is an ancestor of GT, and
``Word'' is the English word of the predicted label.
Dashed edges represent multi-hop connection, where the number indicates the number of edges between classes.
If the prediction is on a super class (marked with * and rounded), then the test image is classified as a novel class whose closest class in the taxonomy is the super class.
}
\label{fig:qual_smp_8}
\end{figure*}

%% file: c_qual_cls.tex
\cutsectionup
\section{Class-wise qualitative results} \label{sec:qual_cls}
\cutsectiondown

\begin{figure*}[ht]
\centering\setlength{\tabcolsep}{0cm}
\begin{tabular}{p{1cm}p{1cm}p{1cm}p{1cm}p{1cm}p{1cm}p{1cm}p{1cm}p{1cm}p{1cm}p{1cm}}
\multicolumn{10}{c}{\includegraphics[width=10cm]{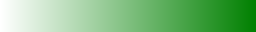}} & \\
0.0 & 0.1 & 0.2 & 0.3 & 0.4 & 0.5 & 0.6 & 0.7 & 0.8 & 0.9 & 1.0 \\
\end{tabular}
\\
\includegraphics[width=\clswidth, height=\clsheight, keepaspectratio]{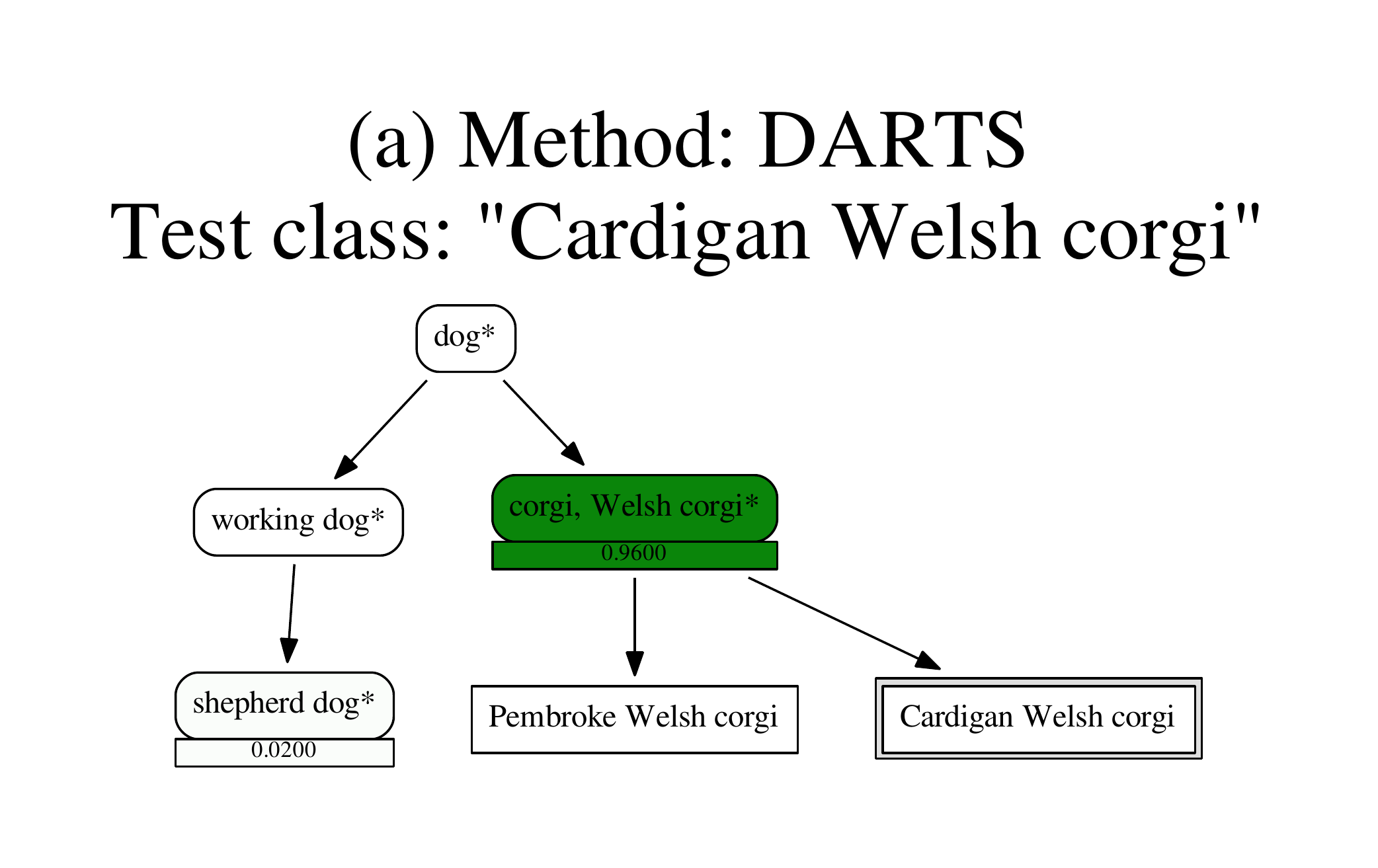}
\includegraphics[width=\clswidth, height=\clsheight, keepaspectratio]{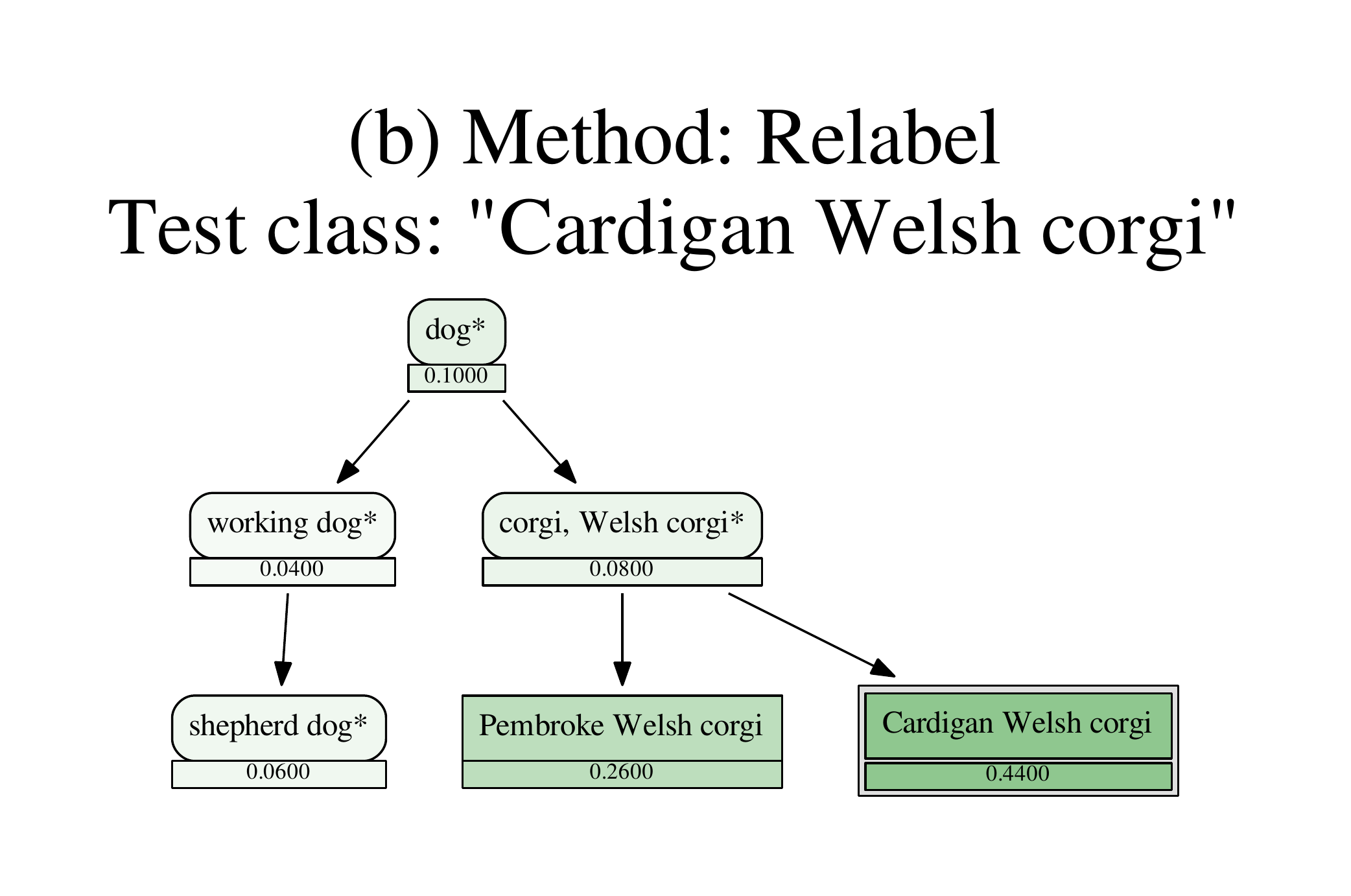}
\includegraphics[width=\clswidth, height=\clsheight, keepaspectratio]{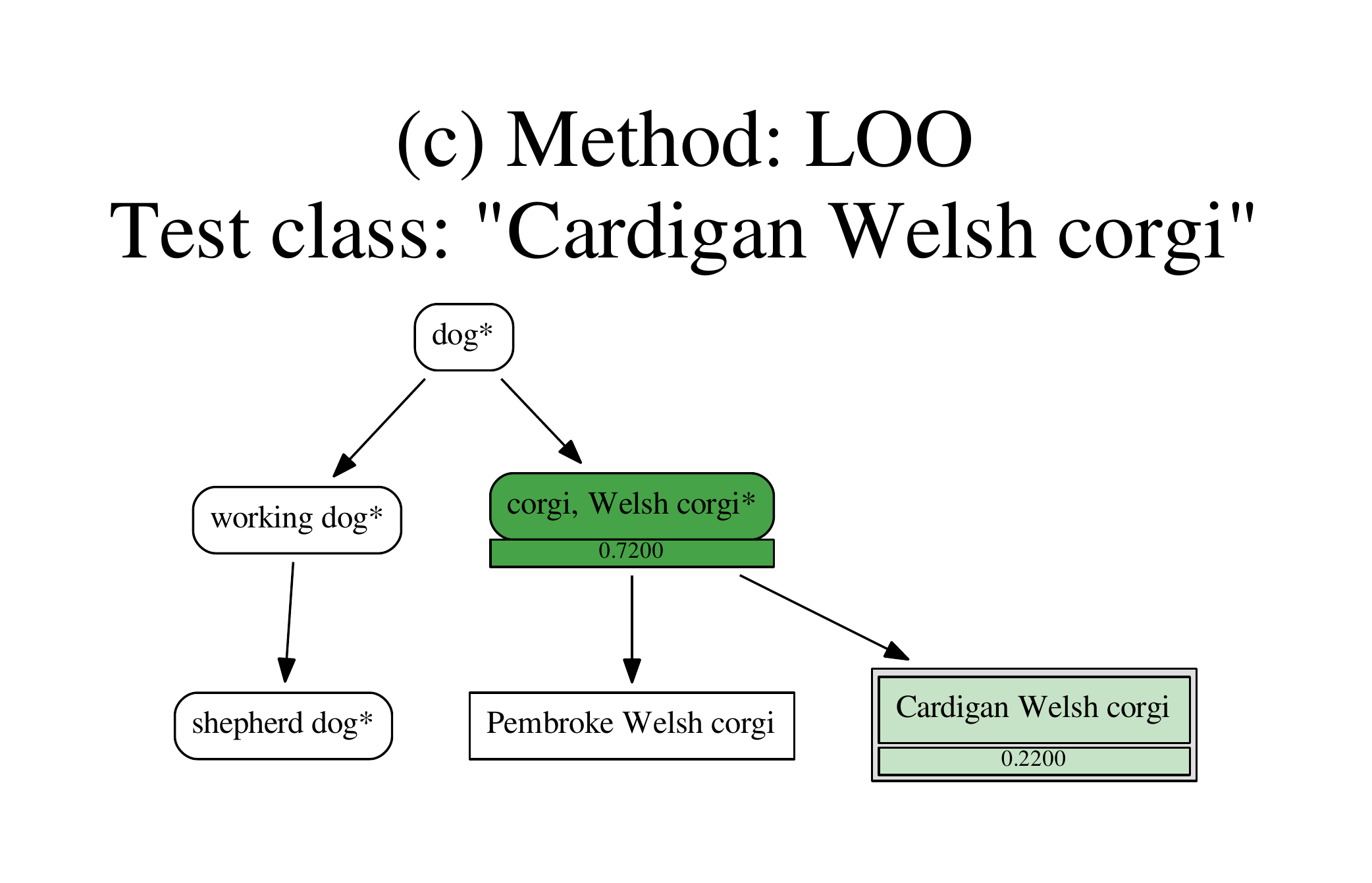}
\includegraphics[width=\clswidth, height=\clsheight, keepaspectratio]{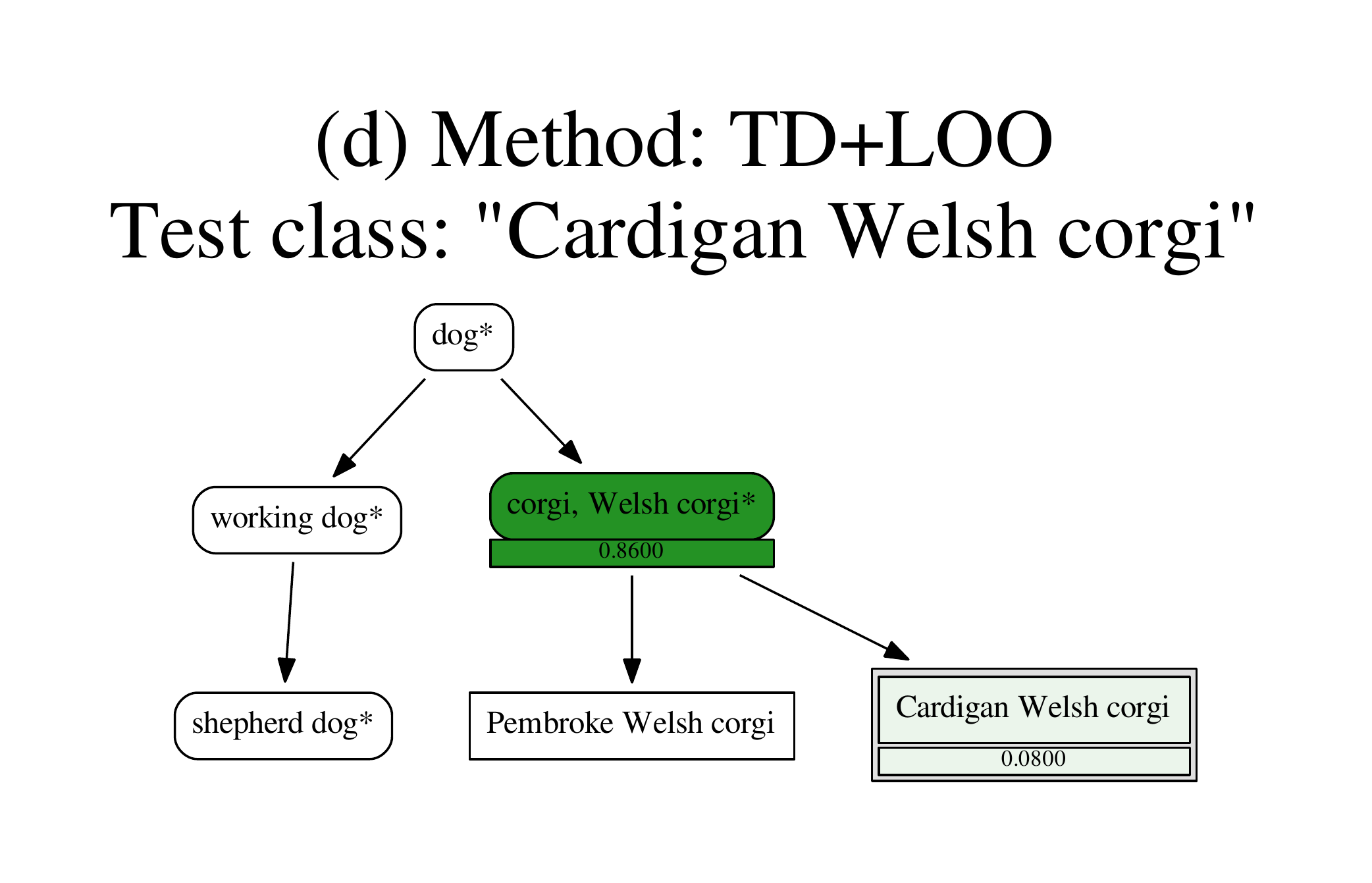}
\vspace{-0.1in}
\caption{Sub-taxonomies of the hierarchical novelty detection results of a known leaf class ``\catwordone.''
(Best viewed when zoomed in on a screen.)
}
\vspace{-0.2in}
\label{fig:qual_cls_1}
\end{figure*}

\begin{figure*}[ht]
\centering\setlength{\tabcolsep}{0cm}
\includegraphics[width=\clswidth, height=\clsheight, keepaspectratio]{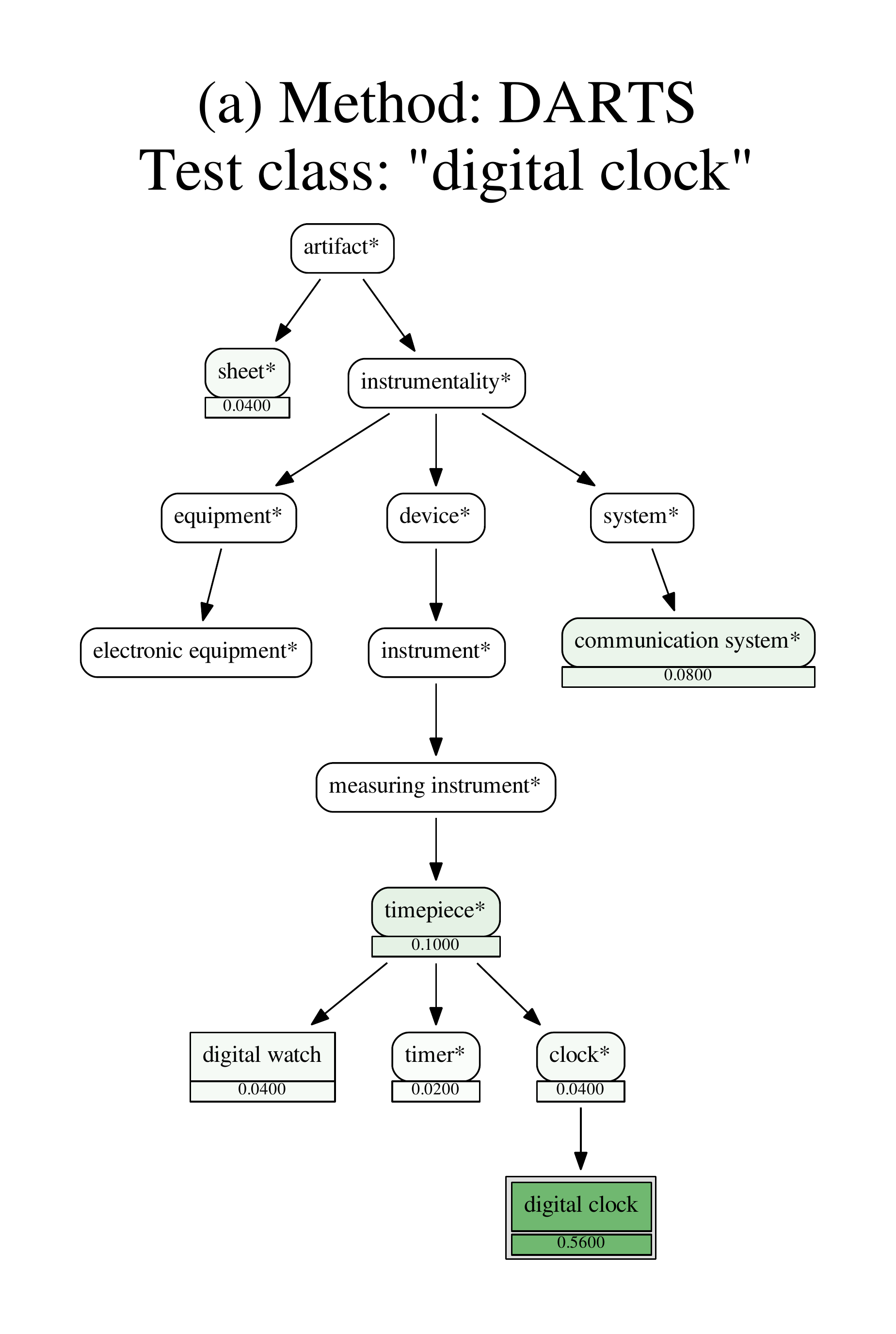}
\includegraphics[width=\clswidth, height=\clsheight, keepaspectratio]{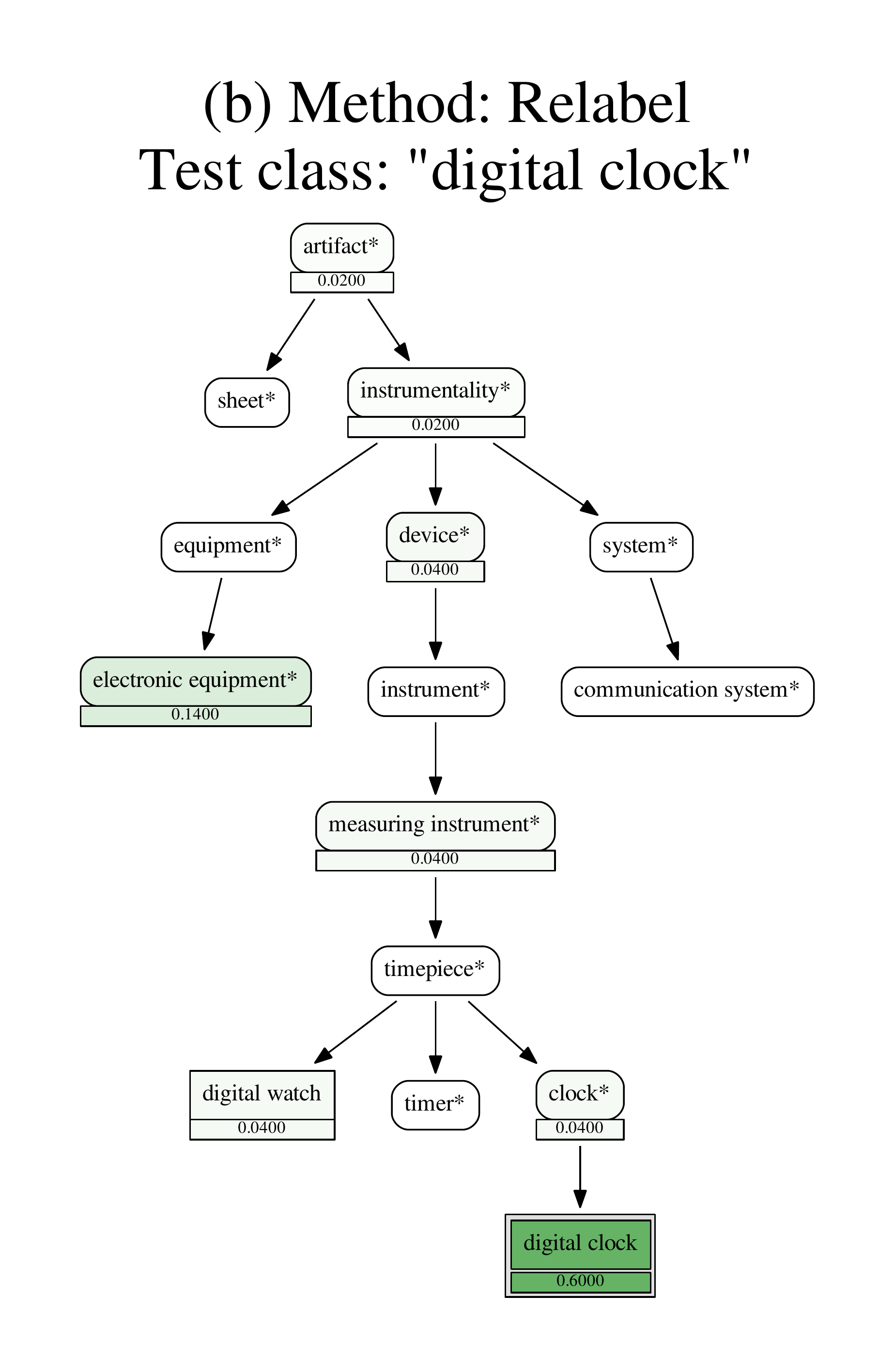}
\includegraphics[width=\clswidth, height=\clsheight, keepaspectratio]{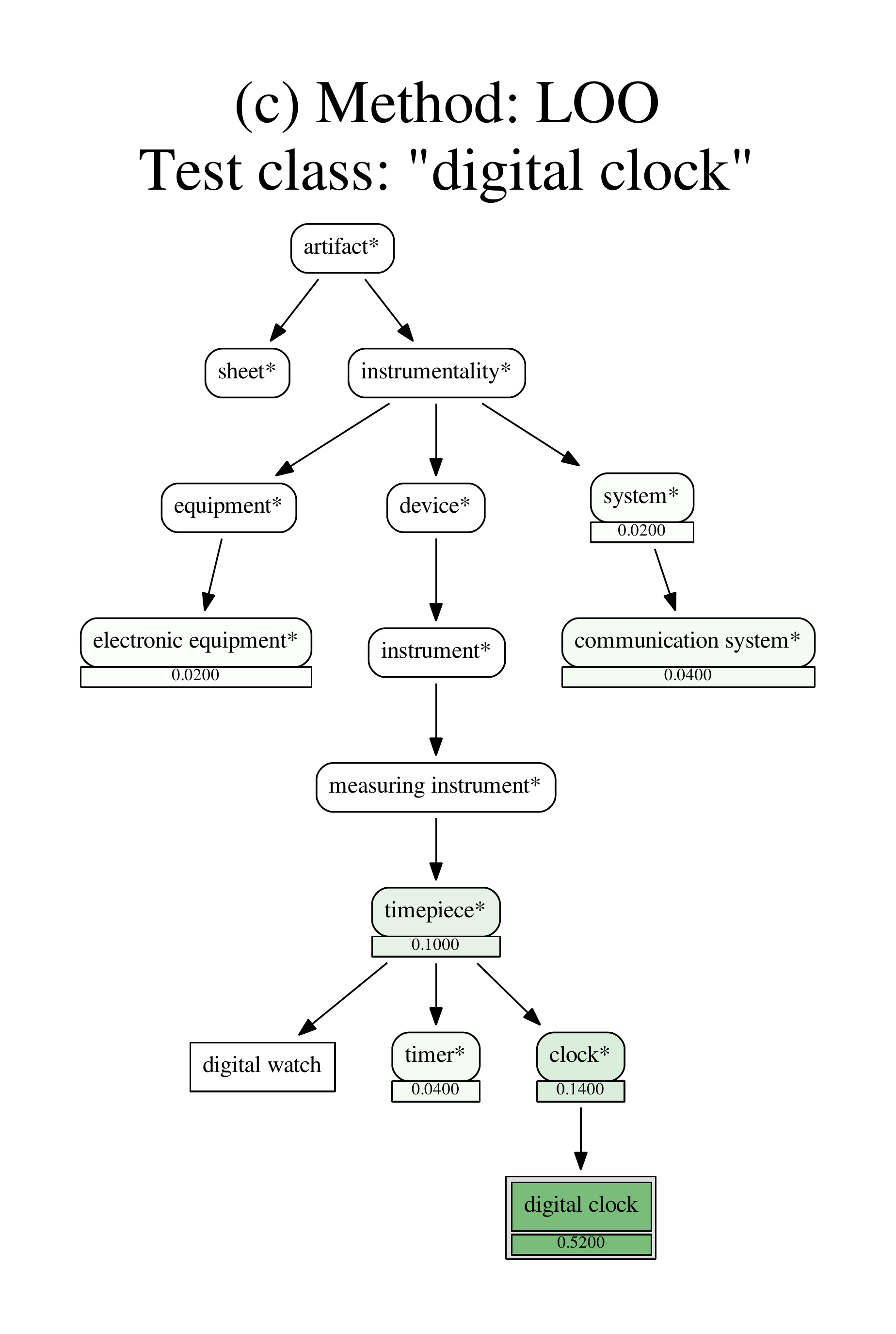}
\includegraphics[width=\clswidth, height=\clsheight, keepaspectratio]{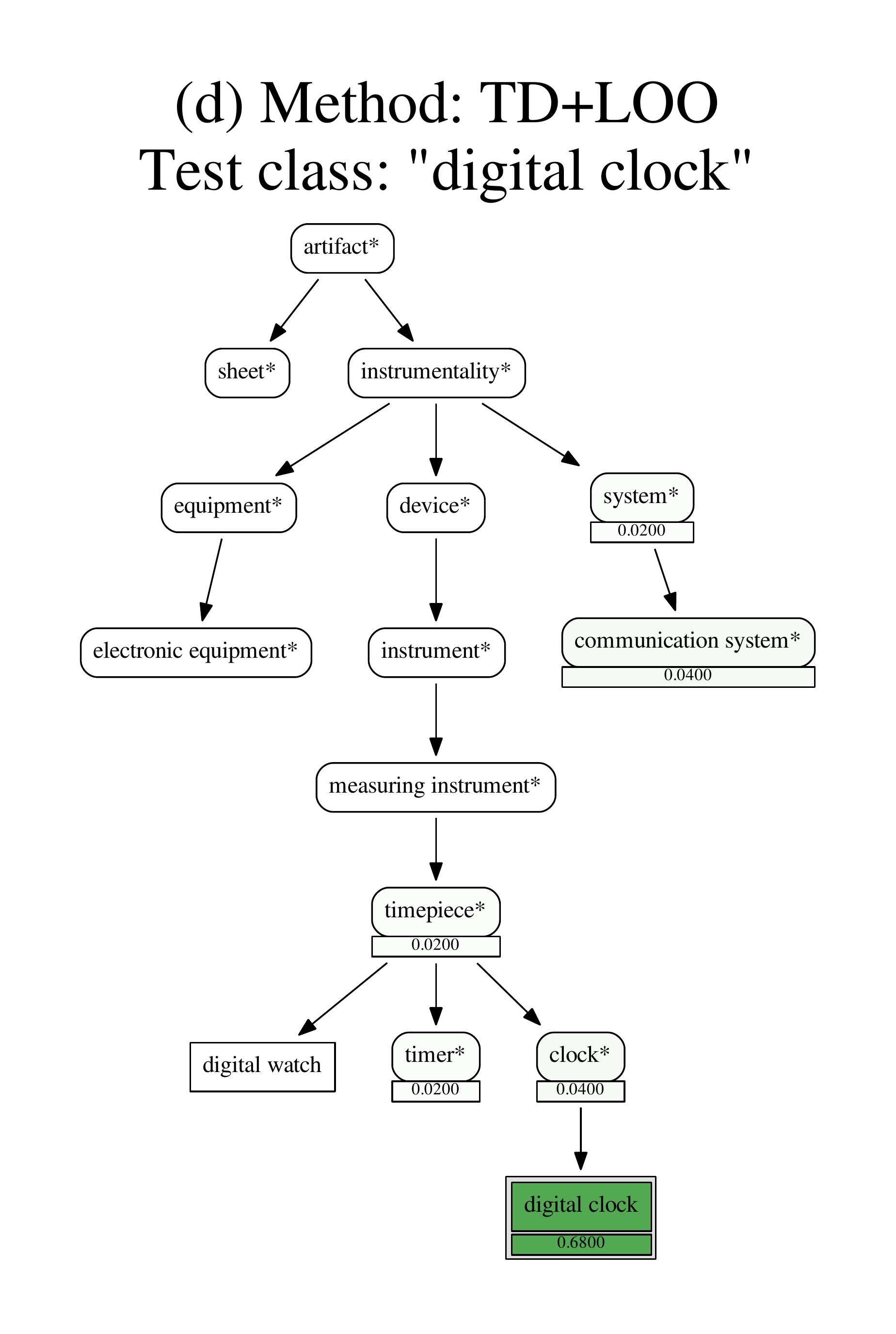}
\vspace{-0.1in}
\caption{Sub-taxonomies of the hierarchical novelty detection results of a known leaf class ``\catwordtwo.''
(Best viewed when zoomed in on a screen.)
}
\vspace{-0.2in}
\label{fig:qual_cls_2}
\end{figure*}

\begin{figure*}[ht]
\centering\setlength{\tabcolsep}{0cm}
\includegraphics[width=\clswidth, height=\clsheight, keepaspectratio]{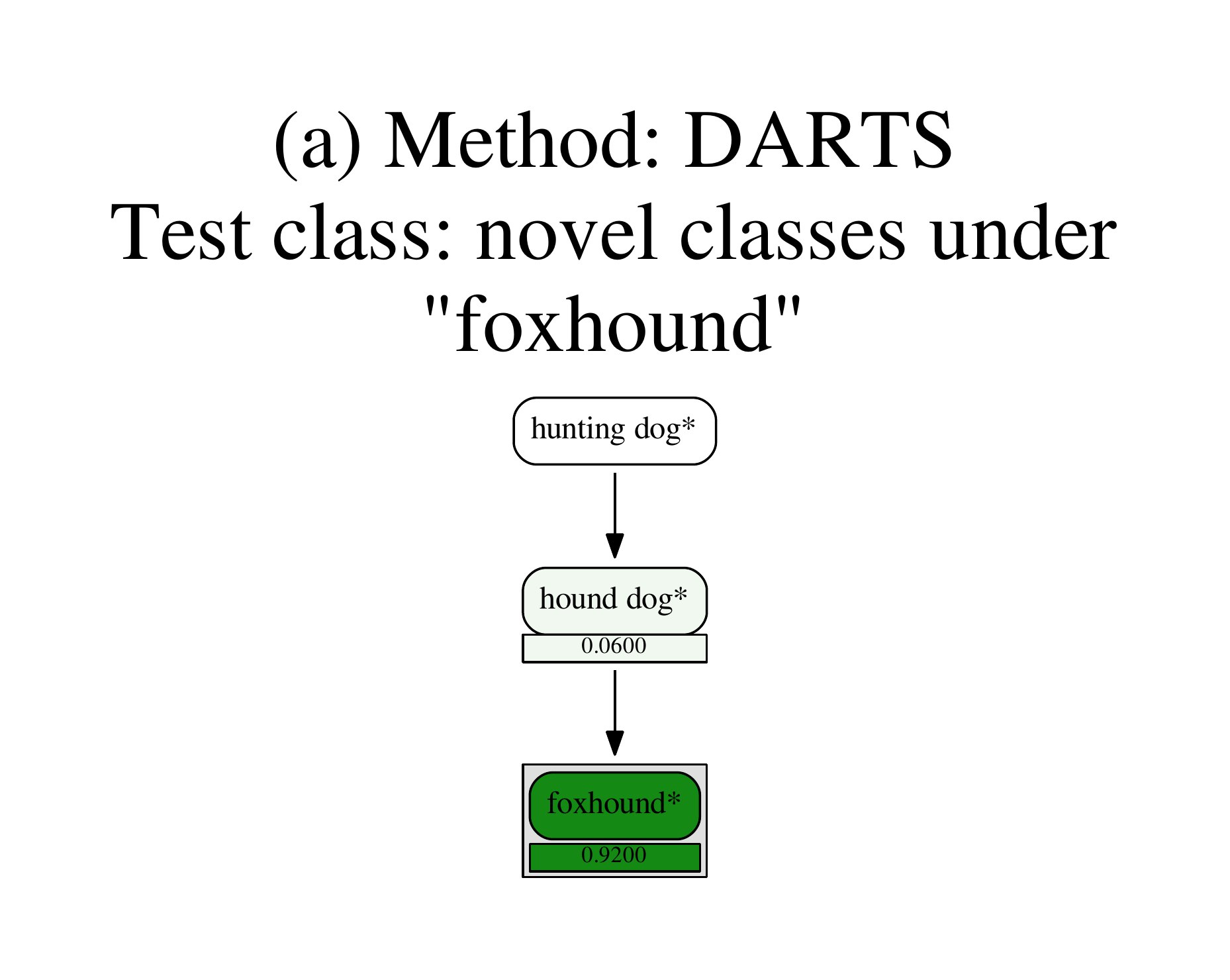}
\includegraphics[width=\clswidth, height=\clsheight, keepaspectratio]{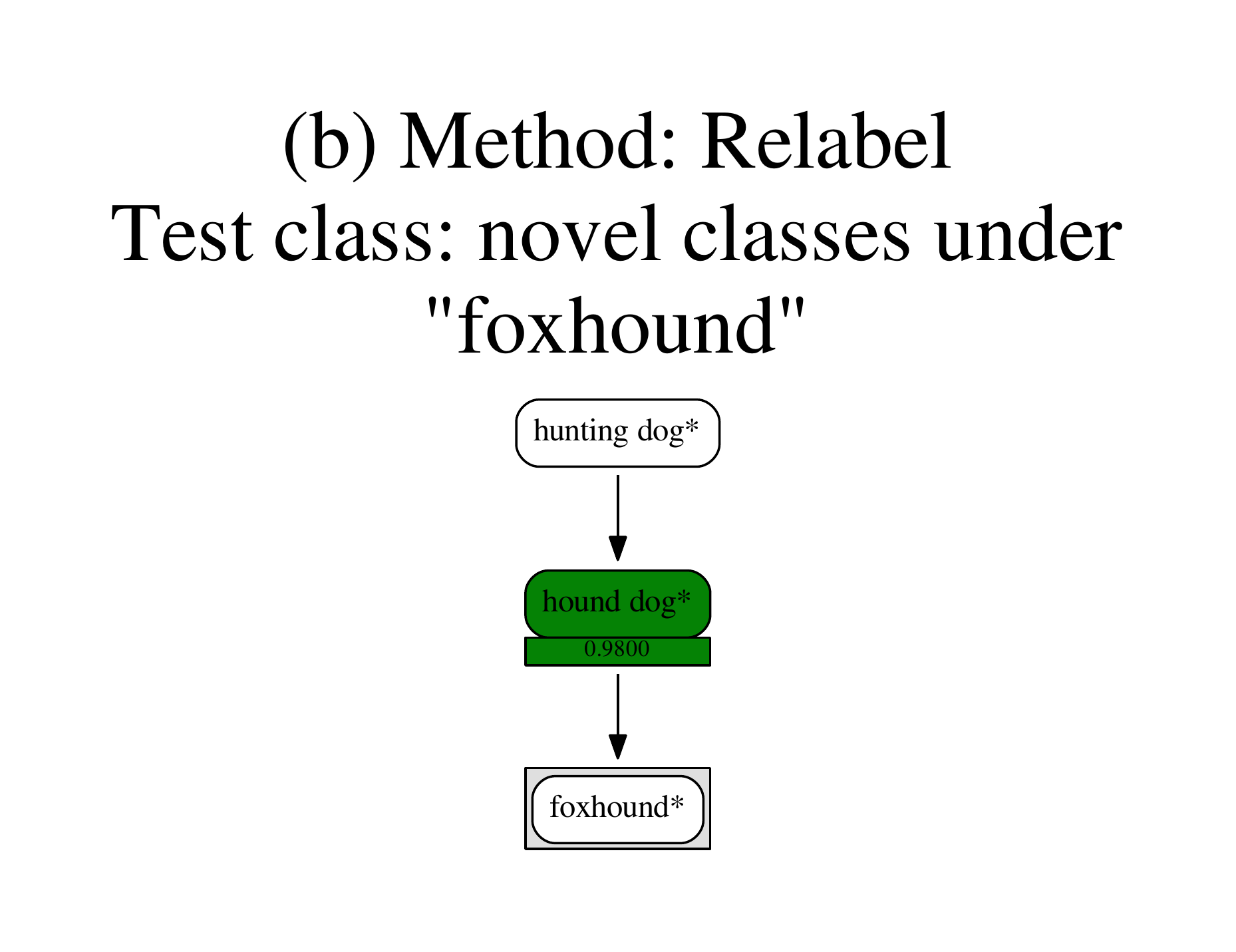}
\includegraphics[width=\clswidth, height=\clsheight, keepaspectratio]{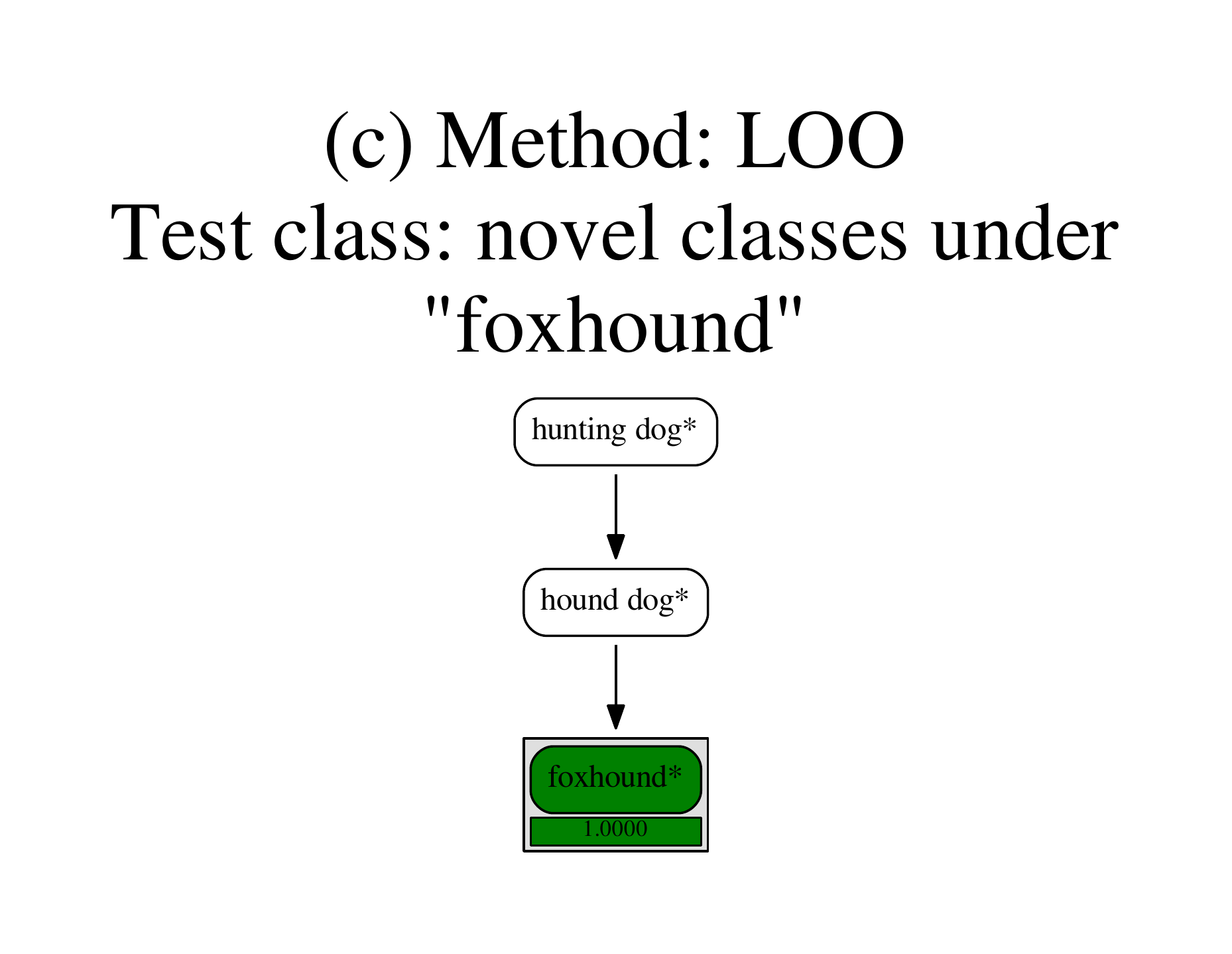}
\includegraphics[width=\clswidth, height=\clsheight, keepaspectratio]{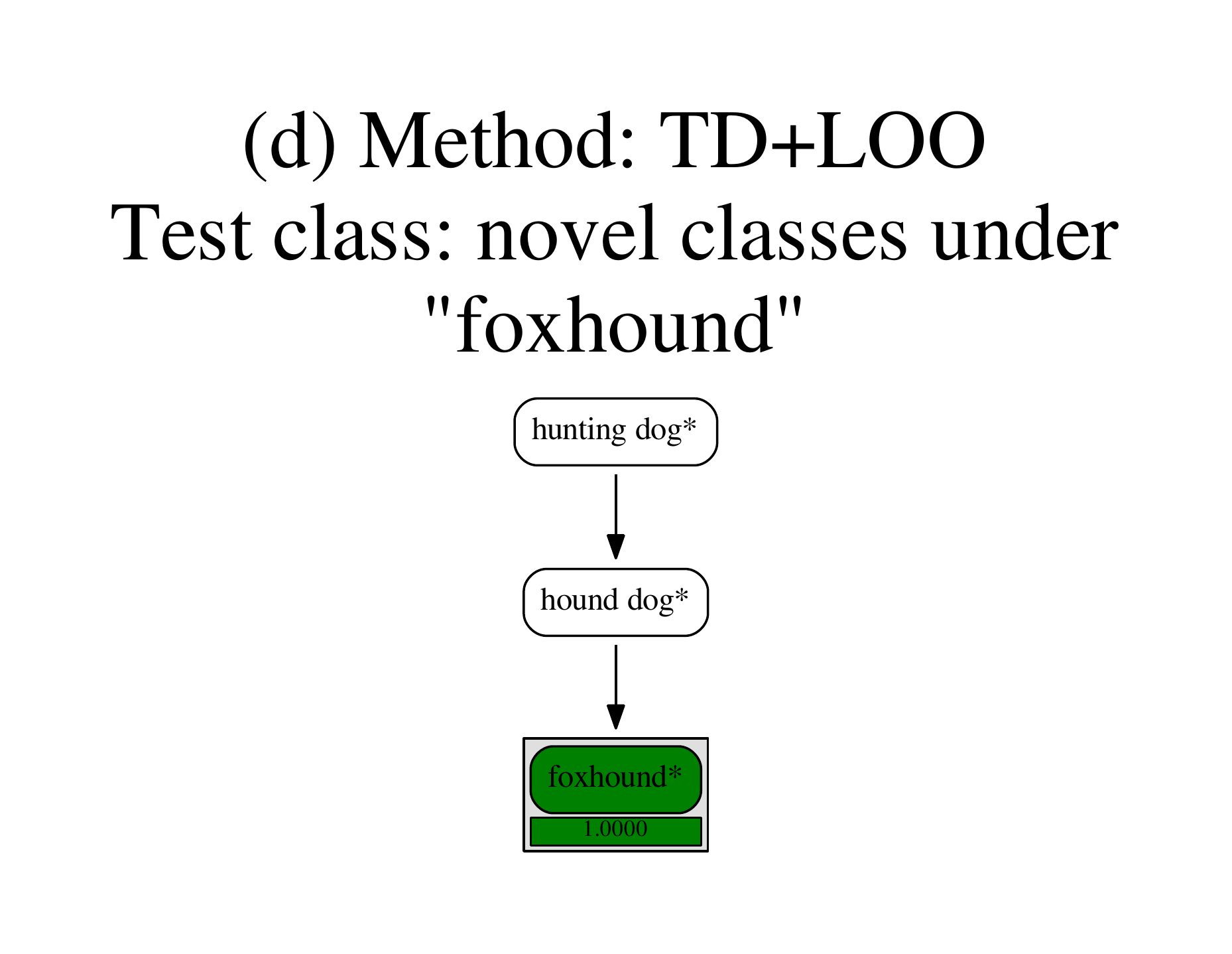}
\vspace{-0.1in}
\caption{Sub-taxonomies of the hierarchical novelty detection results of novel classes whose closest class in the taxonomy is ``\catwordthree.''
(Best viewed when zoomed in on a screen.)
}
\vspace{-0.2in}
\label{fig:qual_cls_3}
\end{figure*}

\begin{figure*}[ht]
\centering\setlength{\tabcolsep}{0cm}
\includegraphics[width=\clswidth, height=\clsheight, keepaspectratio]{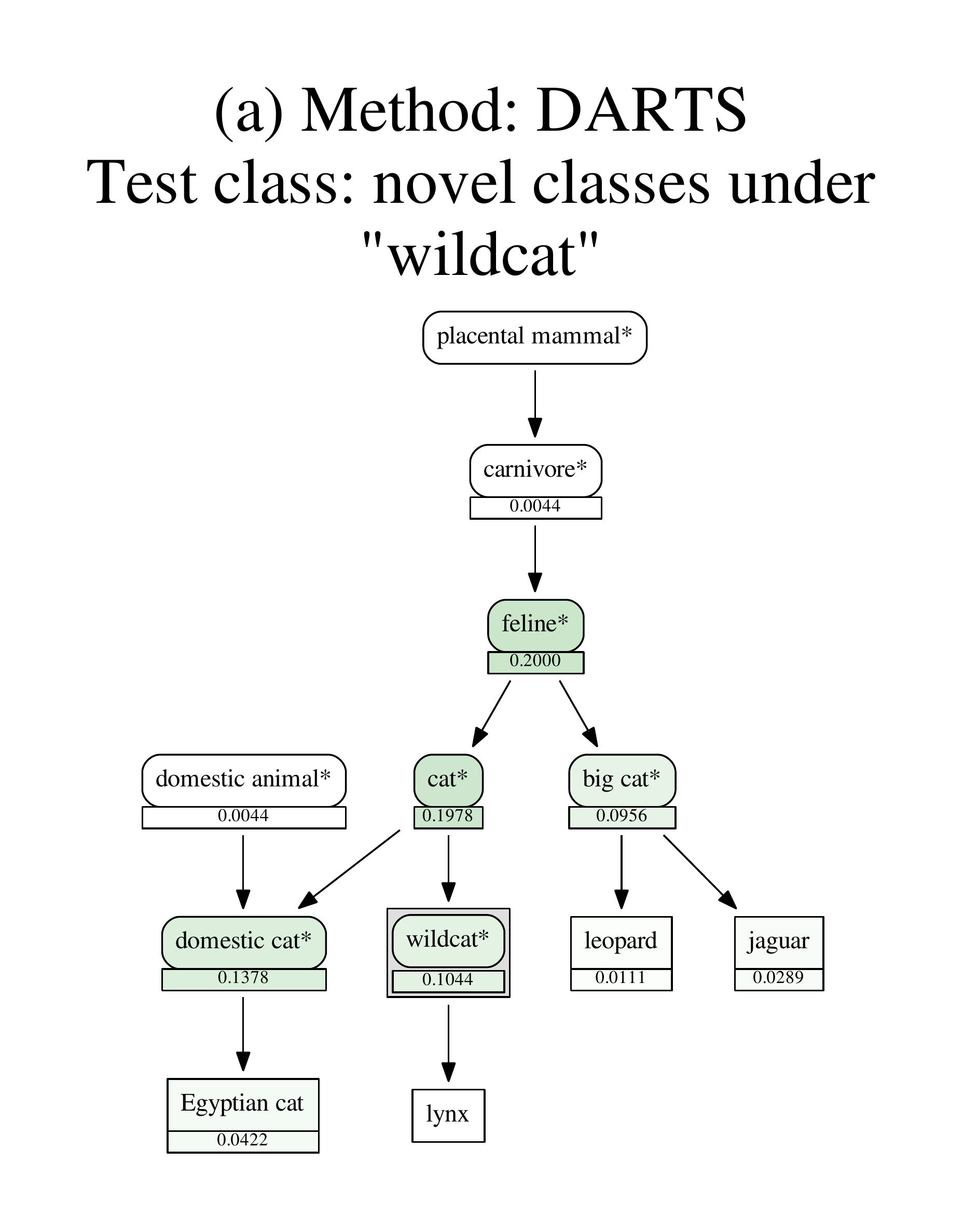}
\includegraphics[width=\clswidth, height=\clsheight, keepaspectratio]{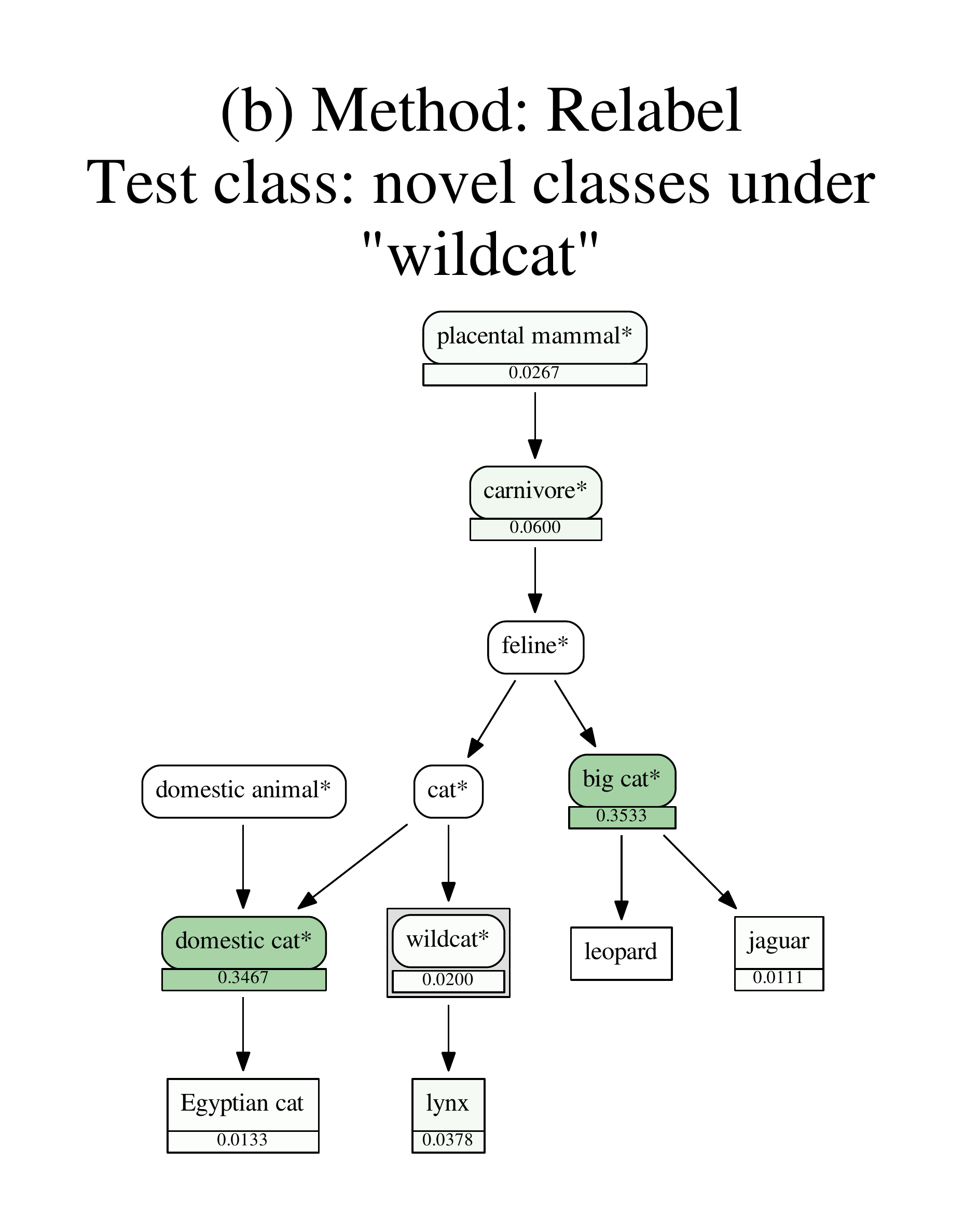}
\includegraphics[width=\clswidth, height=\clsheight, keepaspectratio]{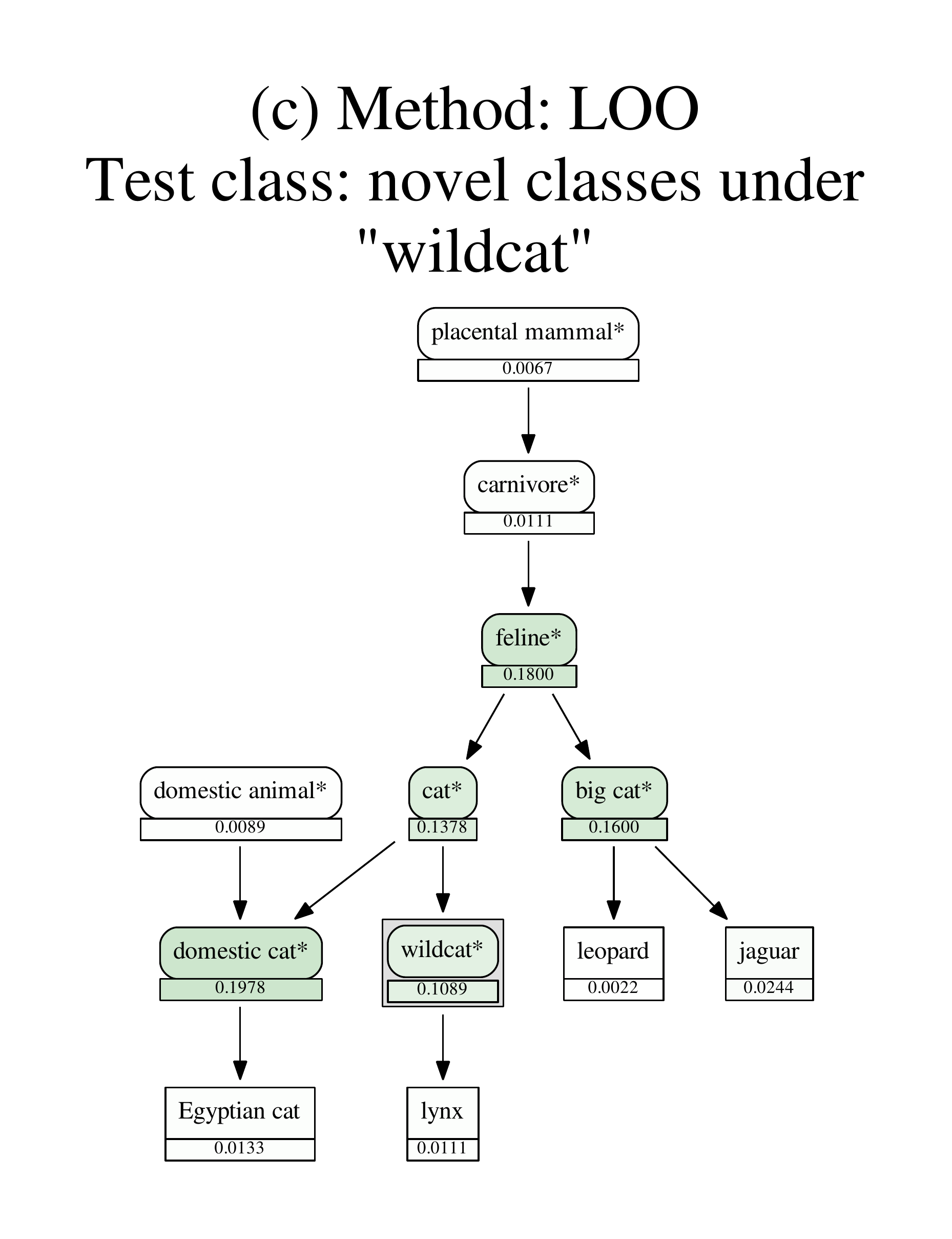}
\includegraphics[width=\clswidth, height=\clsheight, keepaspectratio]{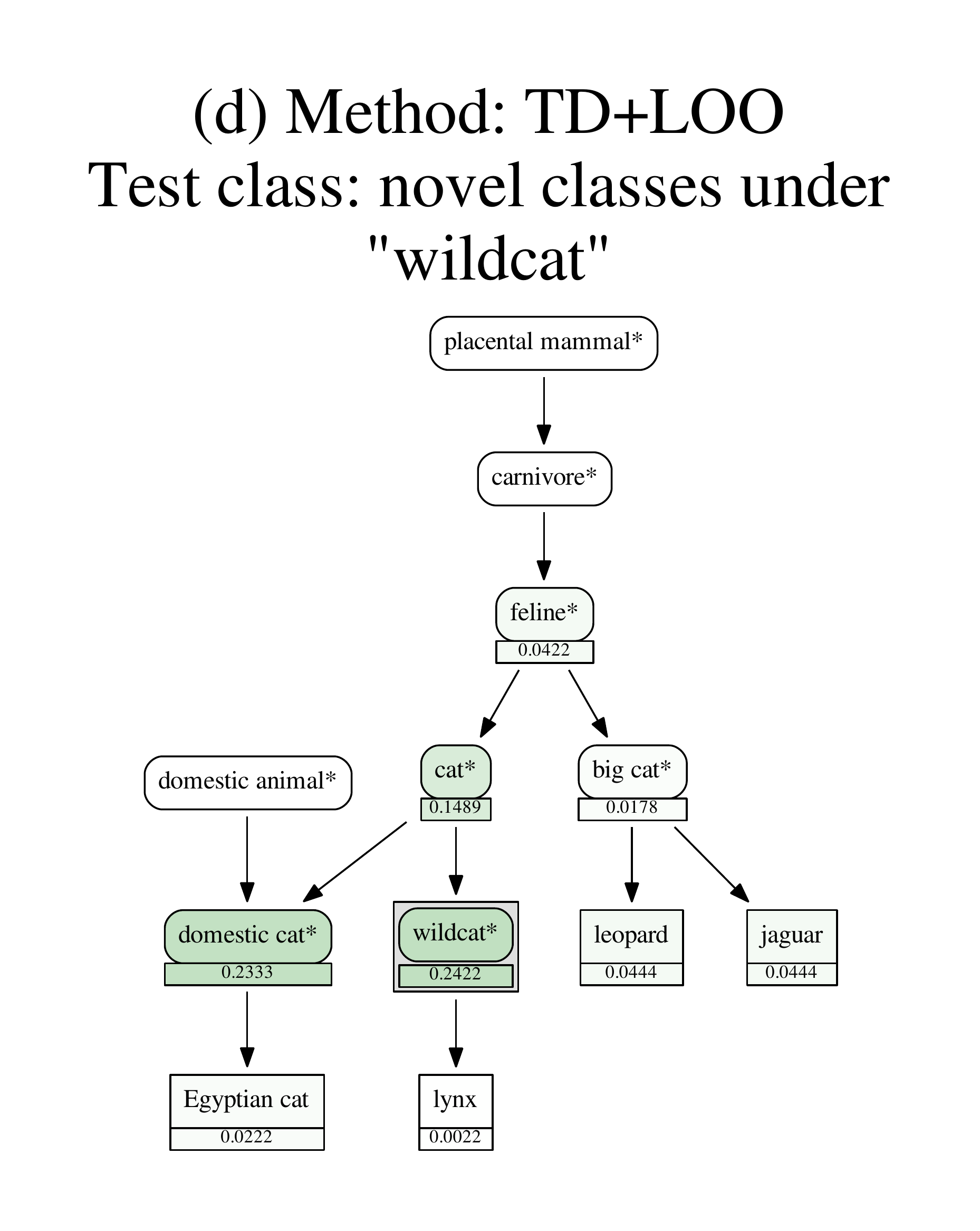}
\vspace{-0.1in}
\caption{Sub-taxonomies of the hierarchical novelty detection results of novel classes whose closest class in the taxonomy is ``\catwordfour.''
(Best viewed when zoomed in on a screen.)
}
\vspace{-0.2in}
\label{fig:qual_cls_4}
\end{figure*}

\begin{figure*}[ht]
\centering\setlength{\tabcolsep}{0cm}
\includegraphics[width=\clswidth, height=\clsheight, keepaspectratio]{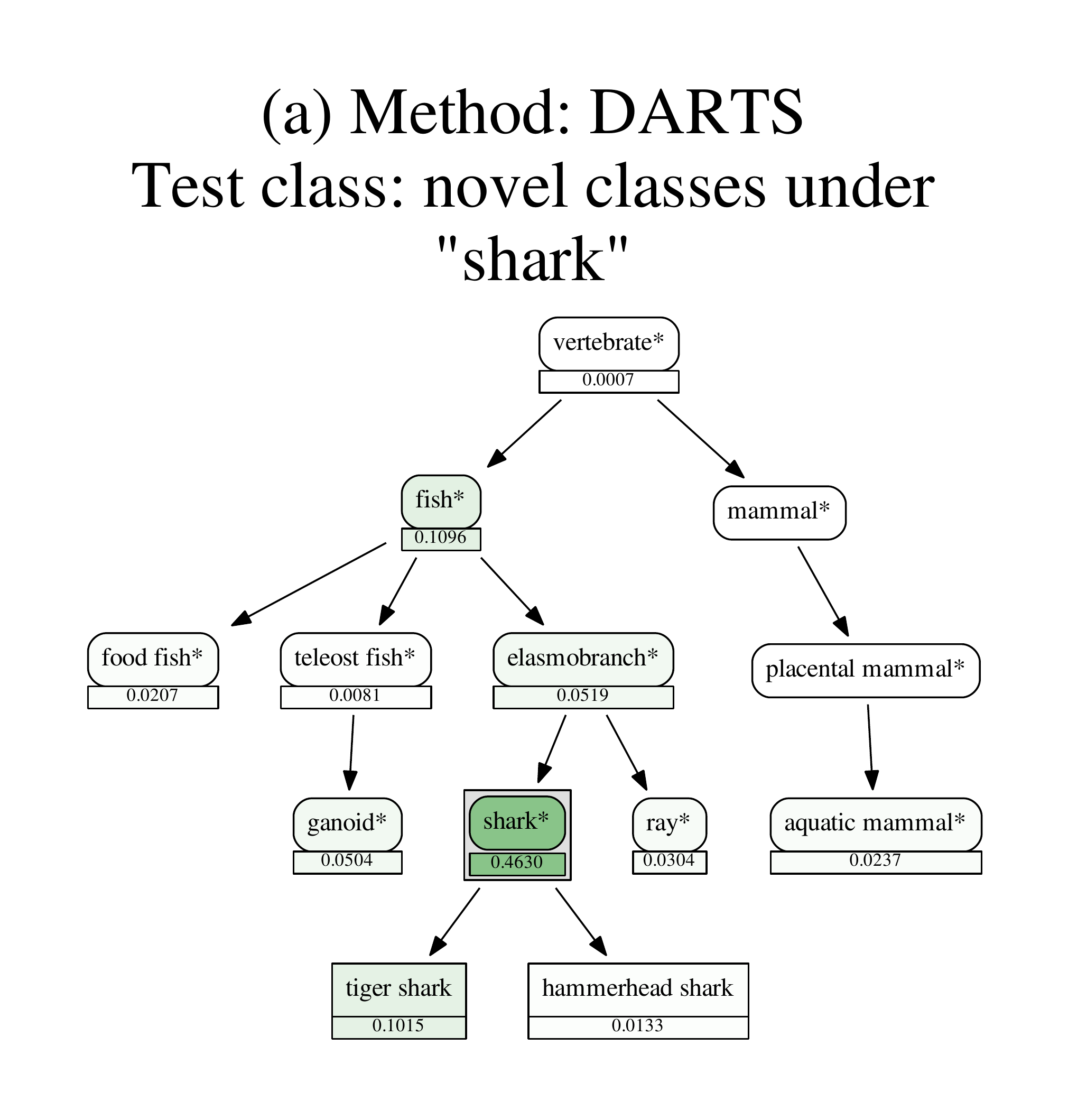}
\includegraphics[width=\clswidth, height=\clsheight, keepaspectratio]{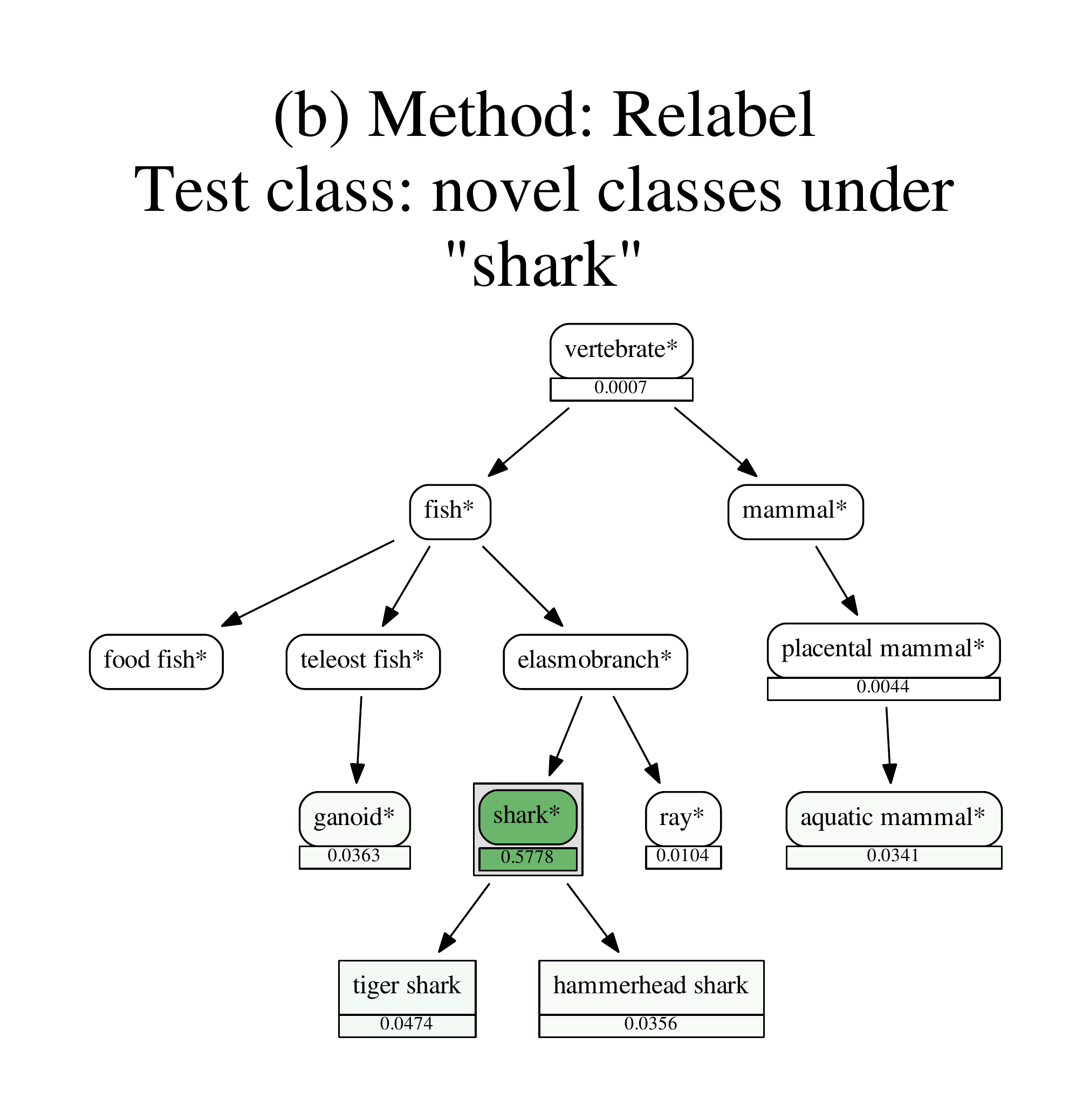}
\includegraphics[width=\clswidth, height=\clsheight, keepaspectratio]{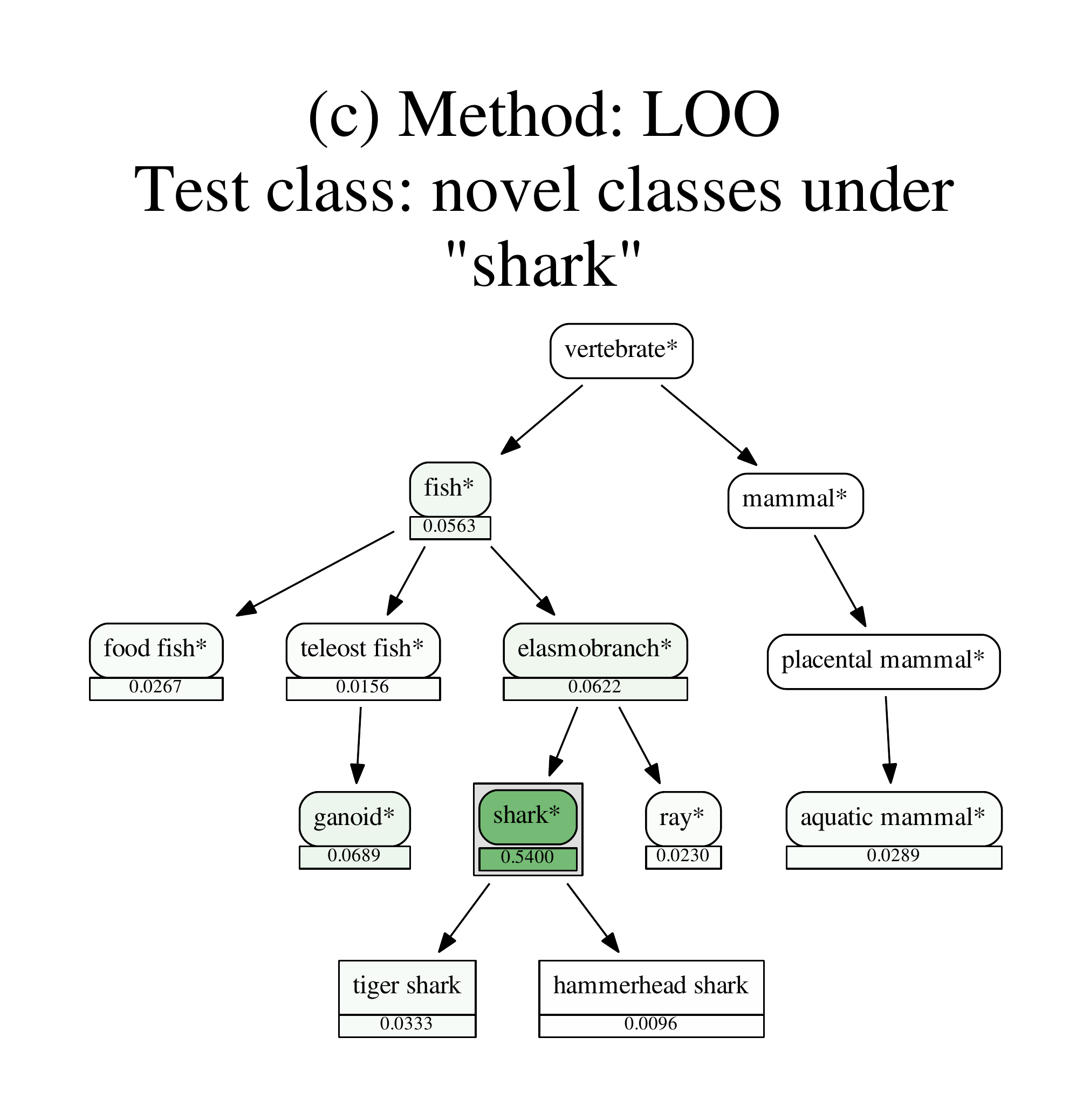}
\includegraphics[width=\clswidth, height=\clsheight, keepaspectratio]{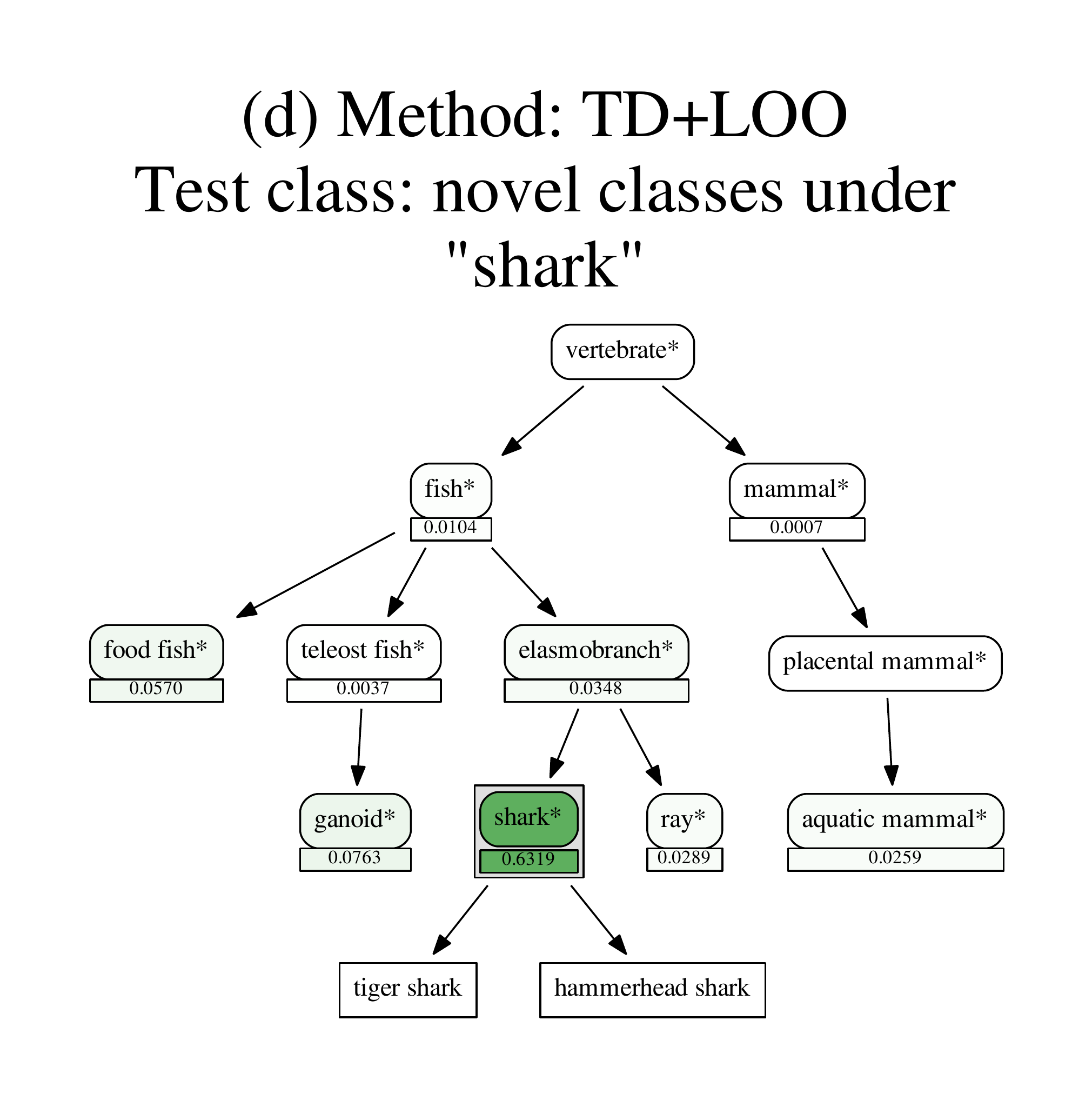}
\vspace{-0.1in}
\caption{Sub-taxonomies of the hierarchical novelty detection results of novel classes whose closest class in the taxonomy is ``\catwordfive.''
(Best viewed when zoomed in on a screen.)
}
\vspace{-0.2in}
\label{fig:qual_cls_5}
\end{figure*}

\begin{figure*}[ht]
\centering\setlength{\tabcolsep}{0cm}
\includegraphics[width=\clswidth, height=\clsheight, keepaspectratio]{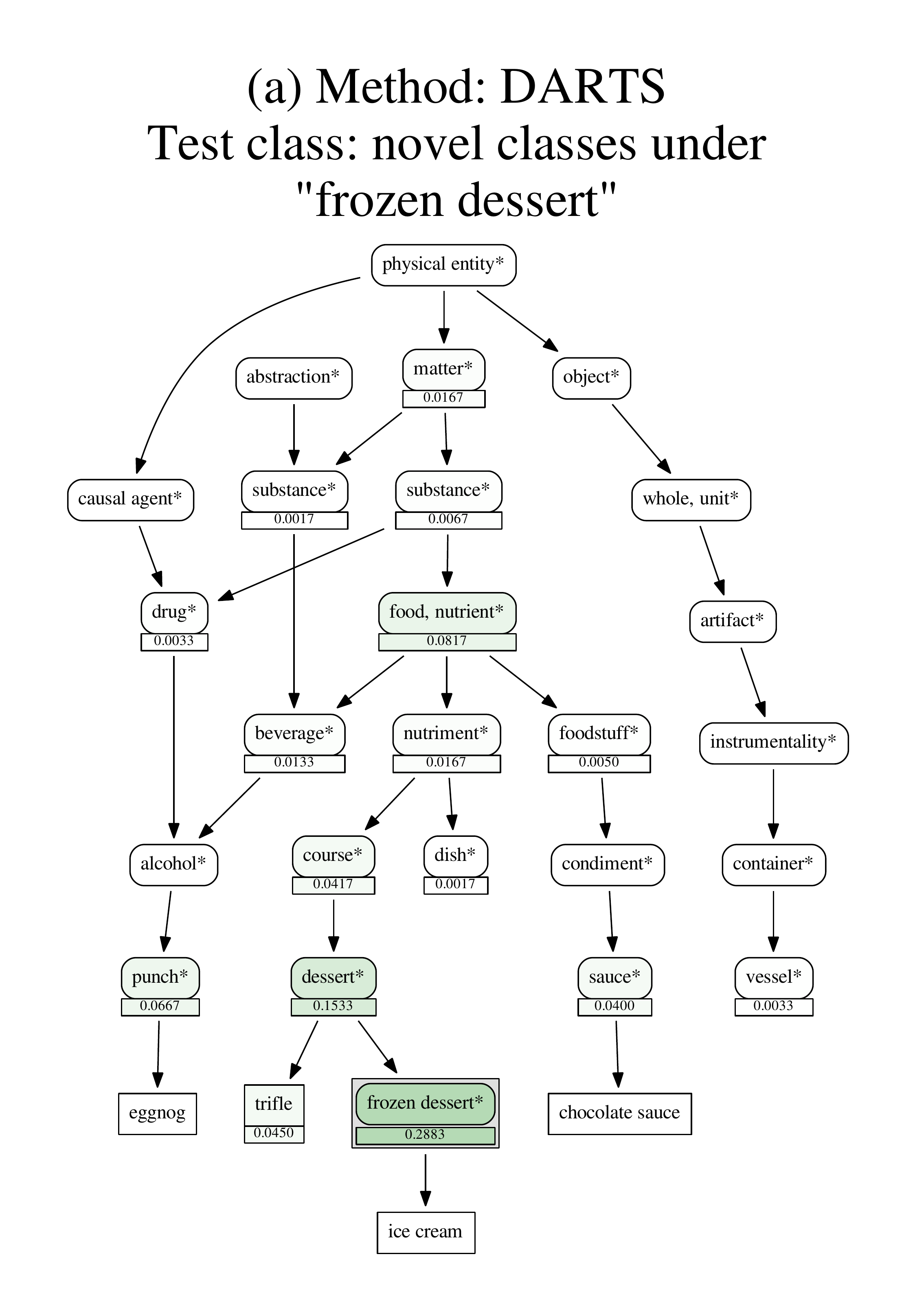}
\includegraphics[width=\clswidth, height=\clsheight, keepaspectratio]{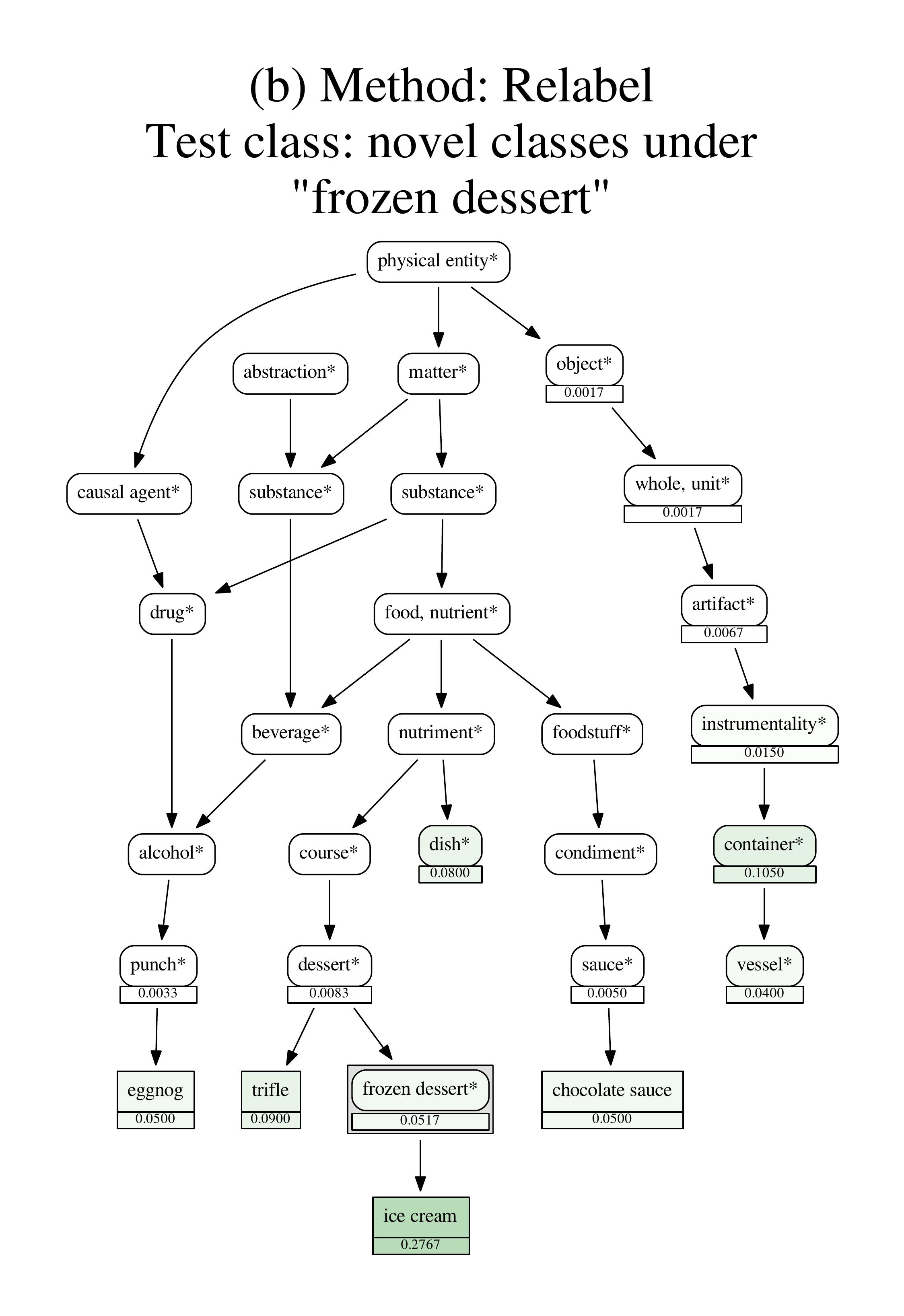}
\includegraphics[width=\clswidth, height=\clsheight, keepaspectratio]{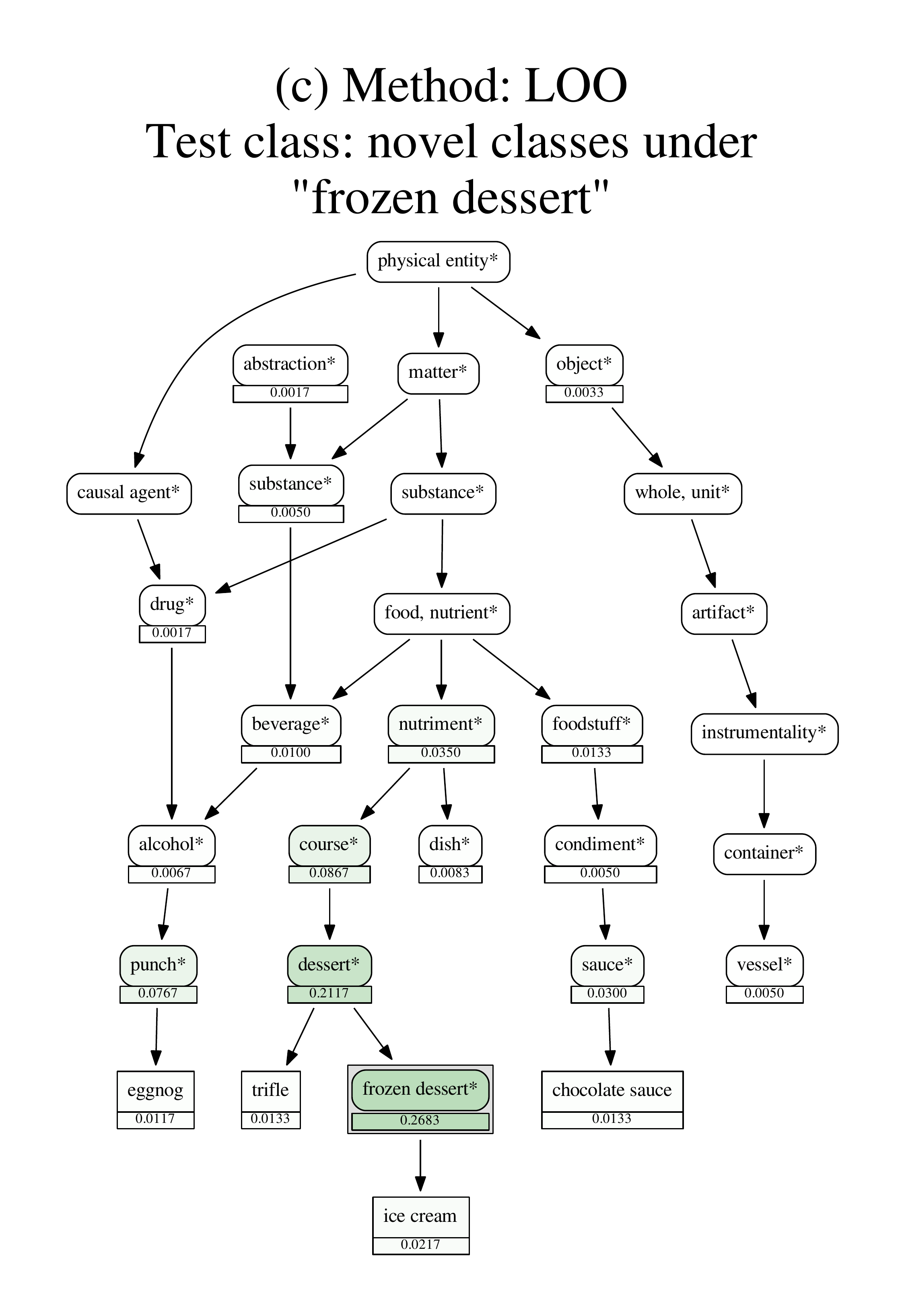}
\includegraphics[width=\clswidth, height=\clsheight, keepaspectratio]{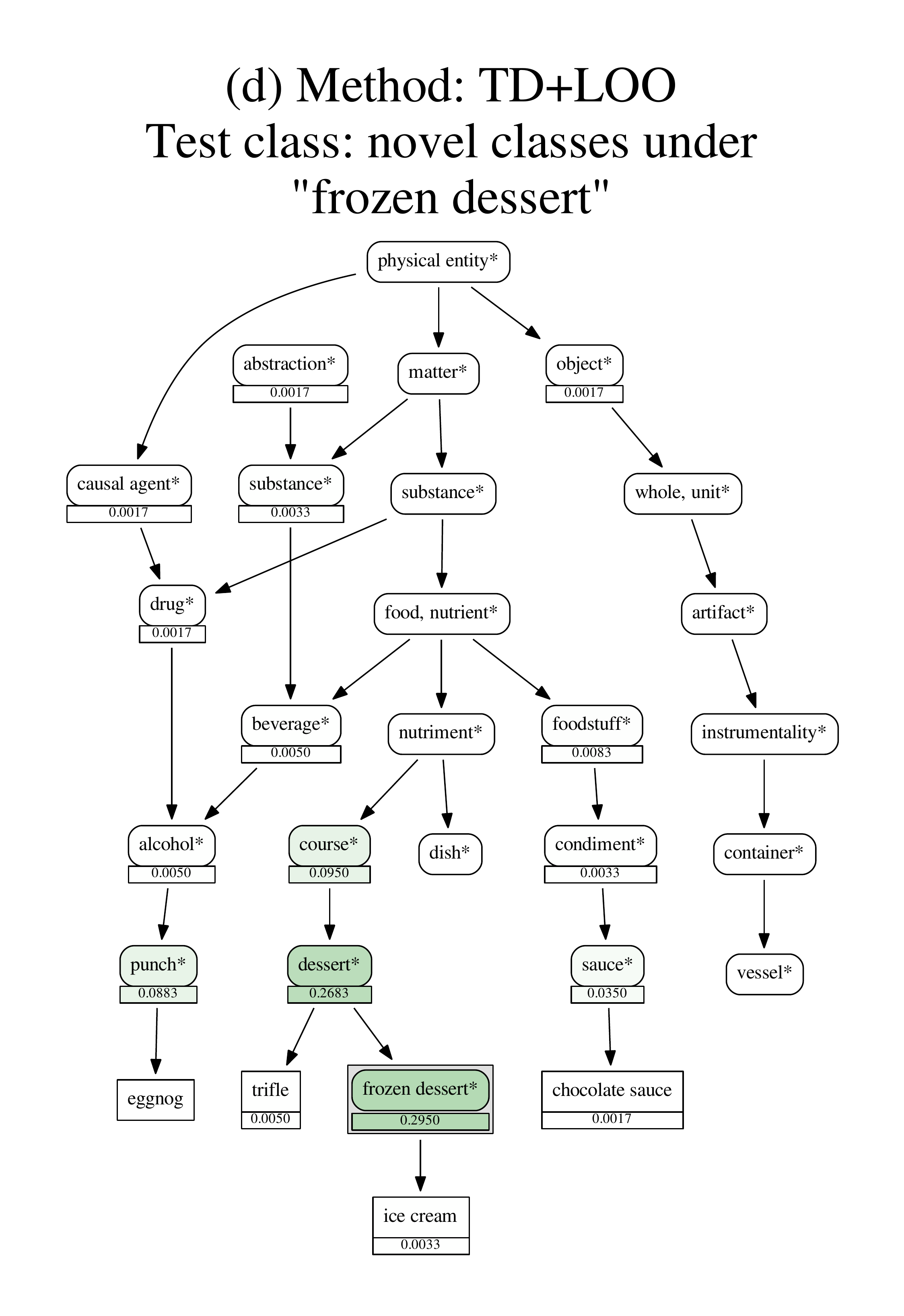}
\vspace{-0.1in}
\caption{Sub-taxonomies of the hierarchical novelty detection results of novel classes whose closest class in the taxonomy is ``\catwordsix.''
(Best viewed when zoomed in on a screen.)
}
\vspace{-0.2in}
\label{fig:qual_cls_6}
\end{figure*}

In this section, we show class-wise qualitative results on ImageNet.
We compared four different methods:
DARTS~\cite{deng2012hedging} is the baseline method where we modify the method for our purpose, and the others, Relabel, LOO, and TD+LOO, are our proposed methods.
%
In a sub-taxonomy, for each test class and method, we show the statistics of the hierarchical novelty detection results of known leaf classes in Figure~\ref{fig:qual_cls_1}--\ref{fig:qual_cls_2}, and that of novel classes in Figure~\ref{fig:qual_cls_3}--\ref{fig:qual_cls_6}.
Each sub-taxonomy is simplified by only showing test classes predicted with a probability greater than 0.03 in at least one method and their common ancestors.
The probability is represented in colored nodes as well as the number below the English word of the class, where the color scale is displayed below. 
Note that the summation of the probabilities shown in each sub-taxonomy may be less than 1, since some classes with a probability less than 0.03 are omitted.
In the graphs, known leaf classes are in rectangle, and super classes are rounded and starred.
If the prediction is on a super class, then the test image is classified as a novel class whose closest class in the taxonomy is the super class.
We remark that most of the incorrect prediction is in fact not very far from the ground truth, which means that the prediction still provides useful information.
While our proposed methods tend to find fine-grained classes, DARTS gives more coarse-grained classes, where one can find the trend clearly in deep sub-taxonomies.
Also, Relabel sometimes fails to predict the correct label but closer ones with a high probability which can be seen as the effect of relabeling.

%% file: d_zsl.tex
\cutsectionup
\section{More on generalized zero-shot learning}
\label{sec:a_zsl}

\cutsubsectionup
\subsection{Example of top-down embedding}
\cutsubsectiondown

Here we provide an example of the ideal output probability vector $t^{y}$ in a simple taxonomy, where $t^{y}$ corresponds to the concatenation of the ideal output of the top-down method when the input label is $y$.
\begin{figure*}[h]
\centering\setlength{\tabcolsep}{0cm}
\vspace*{-0.1in}
\begin{tabular}{c@{\hskip 0.3in}cc}
\multirow{9}{*}{\includegraphics[width=.34\linewidth]{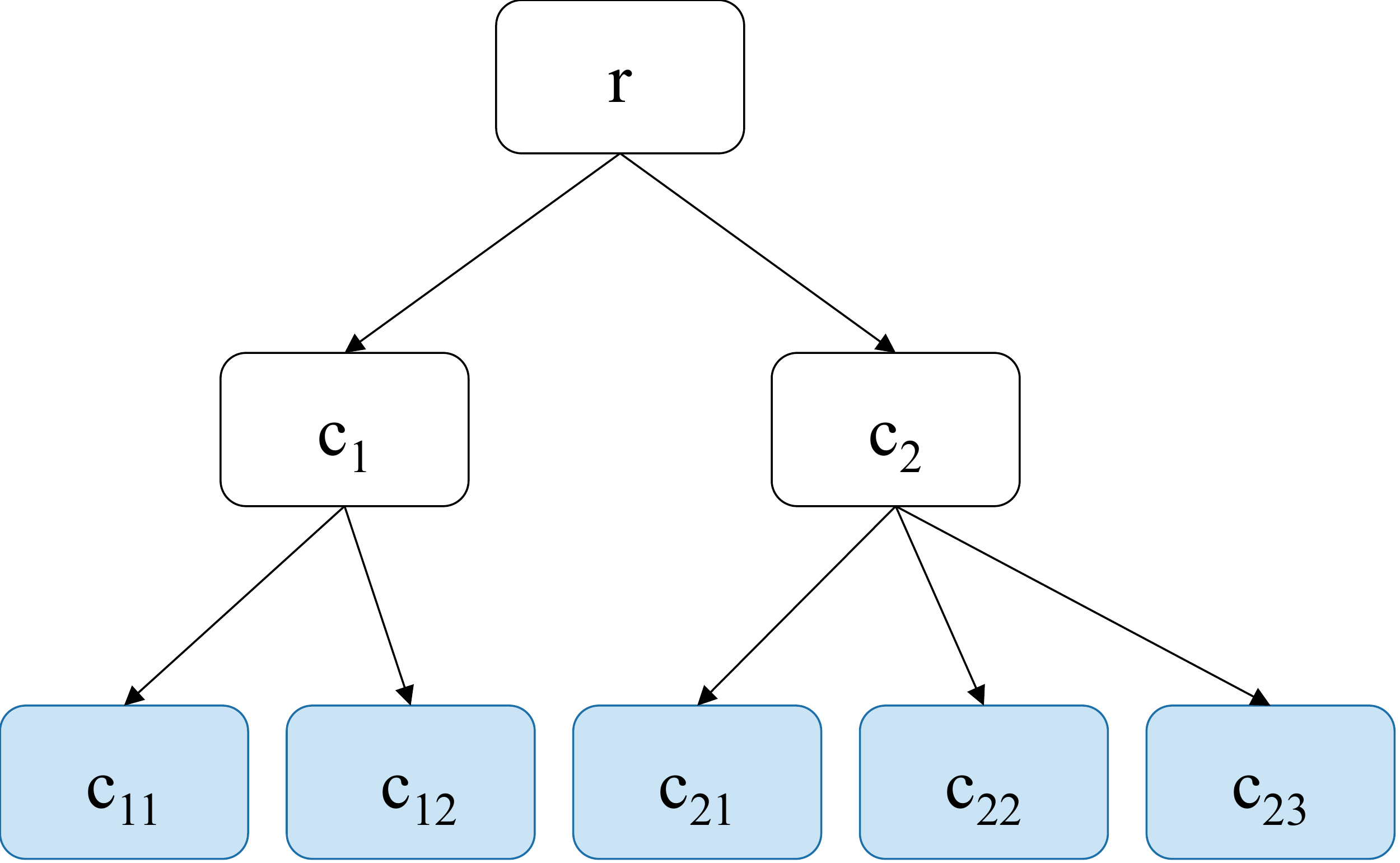}}
& $t^{y}$ & $= [ \hspace{0.22in} t^{(y,r)}, \hspace{0.44in} t^{(y,c_{1})}, \hspace{0.6in} t^{(y,c_{2})} \hspace{0.4in} ]$ \cr
\cline{2-3}
& $t^{r}$ & $= \left[ \text{\texttt{1/2,~1/2,~1/2,~1/2,~1/3,~1/3,~1/3}} \right]$ \cr
& $t^{c_{1}}$ & $= \left[ \text{\texttt{~1~,~~0~,~1/2,~1/2,~1/3,~1/3,~1/3}} \right]$ \cr
& $t^{c_{2}}$ & $= \left[ \text{\texttt{~0~,~~1~,~1/2,~1/2,~1/3,~1/3,~1/3}} \right]$ \cr
& $t^{c_{11}}$ & $= \left[ \text{\texttt{~1~,~~0~,~~1~,~~0~,~1/3,~1/3,~1/3}} \right]$ \cr
& $t^{c_{12}}$ & $= \left[ \text{\texttt{~1~,~~0~,~~0~,~~1~,~1/3,~1/3,~1/3}} \right]$ \cr
& $t^{c_{21}}$ & $= \left[ \text{\texttt{~0~,~~1~,~1/2,~1/2,~~1~,~~0~,~~0~}} \right]$ \cr
& $t^{c_{22}}$ & $= \left[ \text{\texttt{~0~,~~1~,~1/2,~1/2,~~0~,~~1~,~~0~}} \right]$ \cr
& $t^{c_{23}}$ & $= \left[ \text{\texttt{~0~,~~1~,~1/2,~1/2,~~0~,~~0~,~~1~}} \right]$ \cr
\end{tabular}
\caption{
An example of taxonomy and the corresponding $t^{y}$ values.
}
\cutcaptiondown
\label{fig:zsl_ex}
\end{figure*}

\cutsubsectionup
\subsection{Evaluation: Generalized zero-shot learning on different data splits}
\cutsubsectiondown

\begin{figure*}[h]
\centering
\vspace{-0.2in}
\includegraphics[width=.9\linewidth]{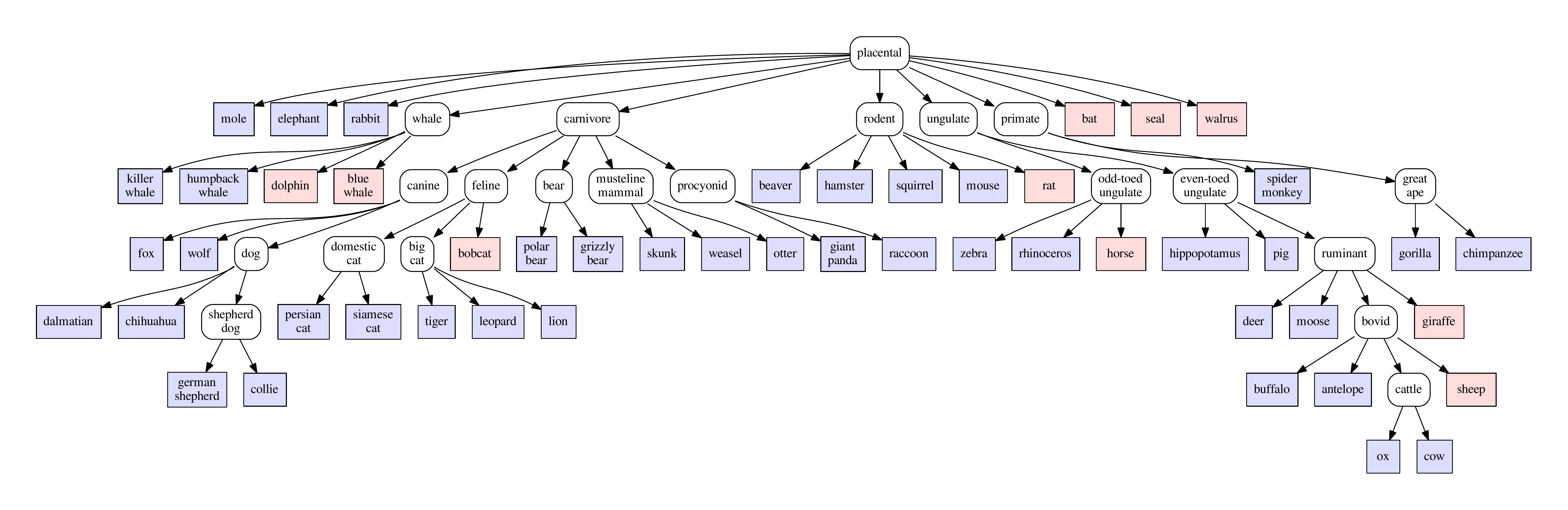} \\
\vspace{-0.27in}
\includegraphics[width=.9\linewidth]{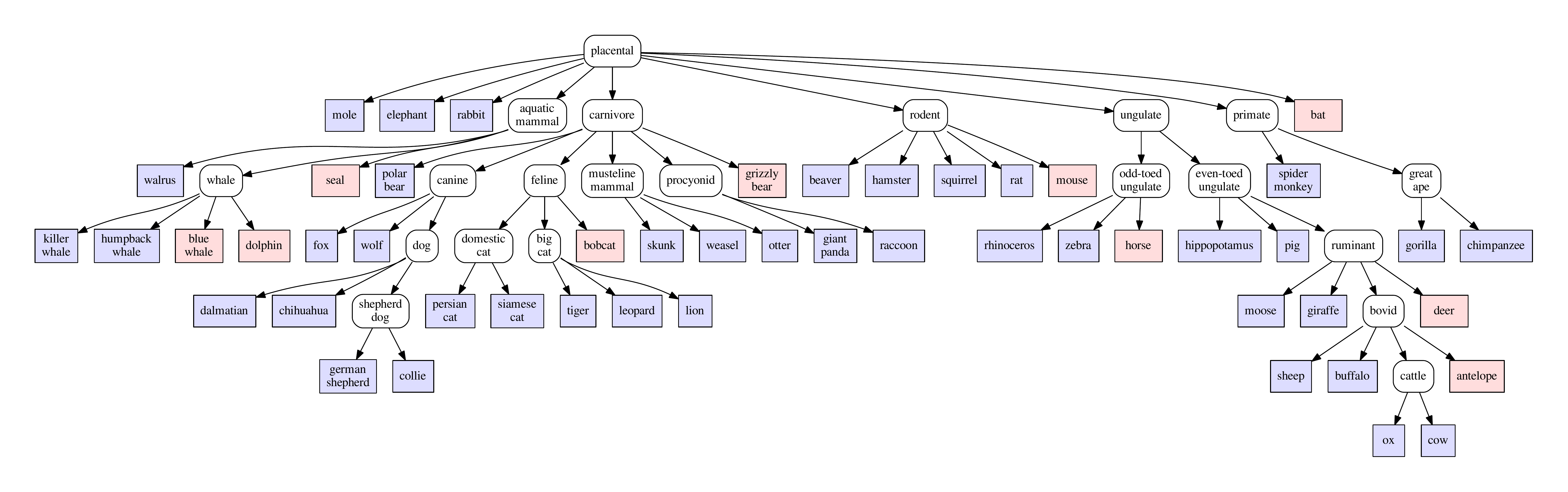}
\vspace{-0.15in}
\cutcaptionup
\caption{
Taxonomy of AwA built with the split proposed in \cite{xian2017zero} (top) and the split we propose for balanced taxonomy (bottom).
Taxonomy is built with known leaf classes (blue) by finding their super classes (white), and then novel classes (red) are attached for visualization.
}
\cutcaptiondown
\label{fig:awa}
\end{figure*}

We present the quantitative results on a different split of AwA1 and AwA2 in this section.
We note that the seen-unseen split of AwA proposed in \cite{xian2017zero} has an imbalanced taxonomy as shown in the top of Figure~\ref{fig:awa}.
Specifically, three classes belong to the root class, and another two classes belong to the same super class.
To show the importance of balanced taxonomy, we make another seen-unseen split for balancing taxonomy, while unseen classes are ensured not to be used for training the CNN feature extractor.
The taxonomy of new split is shown in the bottom of Figure~\ref{fig:awa}.

Table~\ref{tbl:a_zsl} shows the performance of the attribute and word embedding models, and two different hierarchical embedding models, i.e., Path and TD, and their combinations on AwA1 and AwA2 with the split of the imbalanced taxonomy and that of the balanced taxonomy.
Compared to the imbalanced taxonomy case, in the balanced taxonomy, the standalone performance of hierarchical embeddings has similar tendency, but the overall performance is better in all cases.
However, in the combined model, while Path does not improve the performance much, TD still shows improvement on both ZSL and GZSL tasks.
Note that the combination with TD has lower ZSL performance than the combination without TD in some cases, because only AUC is the criterion for optimization.
Compared to the best single semantic embedding model (with attributes), the combination with TD leads to absolute improvement of AUC by 1.66\% and 4.85\% in the split we propose for balanced taxonomy on AwA1 and AwA2, respectively.

These results imply that with more balanced taxonomy, the hierarchy of labels can be implicitly learned without a hierarchical embedding such that the performance is generally better, but yet the combination of an explicit hierarchical embedding improves the performance.

\begin{table}[h]
\centering\setlength{\tabcolsep}{0cm}
\cuttablecaptionup
\caption{
ZSL and GZSL performance of semantic embedding models and their combinations on AwA1 and AwA2 in the split of imbalanced taxonomy and that of balanced taxonomy.
``Att'' stands for continuous attributes labeled by human,
``Word'' stands for word embedding trained with the GloVe objective~\cite{pennington2014glove},
and ``Hier'' stands for the hierarchical embedding, where
``Path'' is proposed in \cite{akata2015evaluation}, and
``TD'' is output of the proposed top-down method.
``Unseen'' is the accuracy when only unseen classes are tested, and 
``AUC'' is the area under the seen-unseen curve 
where the unseen class score bias is varied for computation.
The curve used to obtain AUC is shown in Figure~\ref{fig:a_comparison_zsl}.
Values in bold indicate the best performance among the combined models.
}
\cuttablecaptiondown
\cuttableup
\begin{tabular}{
|>{\centering}m{1cm}|>{\centering}m{1cm}|>{\centering}m{1cm}|
|>{\centering}m{1.2cm}|>{\centering}m{1.2cm}|
|>{\centering}m{1.2cm}|>{\centering}m{1.2cm}|
}
\hline
\multicolumn{3}{|c||}{AwA1} & \multicolumn{2}{c||}{Imbalanced} & \multicolumn{2}{c|}{Balanced} \cr
\hline
Att & Word & Hier & Unseen & AUC & Unseen & AUC \cr
\hline
\hline
$\checkmark$ & & & 65.29 & 50.02 & 65.86 & 54.18 \cr
& $\checkmark$ & & 51.87 & 39.67 & 54.29 & 42.40 \cr
$\checkmark$ & $\checkmark$ & & 67.80 & 52.84 & {\bf 67.32} & 55.40 \cr
\hline
& & Path & 42.57 & 30.58 & 53.40 & 41.63 \cr
$\checkmark$ & & Path & 67.09 & 51.45 & 65.86 & 54.18 \cr
& $\checkmark$ & Path & 52.89 & 40.66 & 58.49 & 45.62 \cr
$\checkmark$ & $\checkmark$ & Path & 68.04 & 53.21 & {\bf 67.32} & 55.40 \cr
\hline
& & TD & 33.86 & 25.56 & 40.38 & 31.39 \cr
$\checkmark$ & & TD & 66.13 & 54.66 & 65.86 & 54.18 \cr
& $\checkmark$ & TD & 56.14 & 46.28 & 57.88 & 47.63 \cr
$\checkmark$ & $\checkmark$ & TD & {\bf 69.23} & {\bf 57.67} & 66.41 & {\bf 55.84} \cr
\hline
\end{tabular}
\quad
\begin{tabular}{
|>{\centering}m{1cm}|>{\centering}m{1cm}|>{\centering}m{1cm}|
|>{\centering}m{1.2cm}|>{\centering}m{1.2cm}|
|>{\centering}m{1.2cm}|>{\centering}m{1.2cm}|
}
\hline
\multicolumn{3}{|c||}{AwA2} & \multicolumn{2}{c||}{Imbalanced} & \multicolumn{2}{c|}{Balanced} \cr
\hline
Att & Word & Hier & Unseen & AUC & Unseen & AUC \cr
\hline
\hline
$\checkmark$ & & & 63.87 & 51.27 & 71.21 & 59.51 \cr
& $\checkmark$ & & 54.77 & 42.21 & 59.60 & 46.83 \cr
$\checkmark$ & $\checkmark$ & & 65.76 & 53.18 & 72.89 & 60.60 \cr
\hline
& & Path & 44.34 & 33.44 & 60.45 & 48.13 \cr
$\checkmark$ & & Path & 66.58 & 53.50 & 71.87 & 60.08 \cr
& $\checkmark$ & Path & 55.28 & 42.86 & 66.83 & 53.05 \cr
$\checkmark$ & $\checkmark$ & Path & 67.28 & 54.31 & 73.04 & 60.89 \cr
\hline
& & TD & 31.84 & 24.97 & 45.33 & 36.76 \cr
$\checkmark$ & & TD & 66.86 & 57.49 & 72.75 & 62.79 \cr
& $\checkmark$ & TD & 59.67 & 49.39 & 65.29 & 53.40 \cr
$\checkmark$ & $\checkmark$ & TD & {\bf 68.80} & {\bf 59.24} & {\bf 75.09} & {\bf 64.36} \cr
\hline
\end{tabular}
\cuttabledown
\label{tbl:a_zsl}
\end{table}

\begin{figure}[h]
\centering\setlength{\tabcolsep}{0cm}
\begin{tabular}{cc}
(a) AwA1 & (b) AwA2 \cr
\includegraphics[width=.49\linewidth]{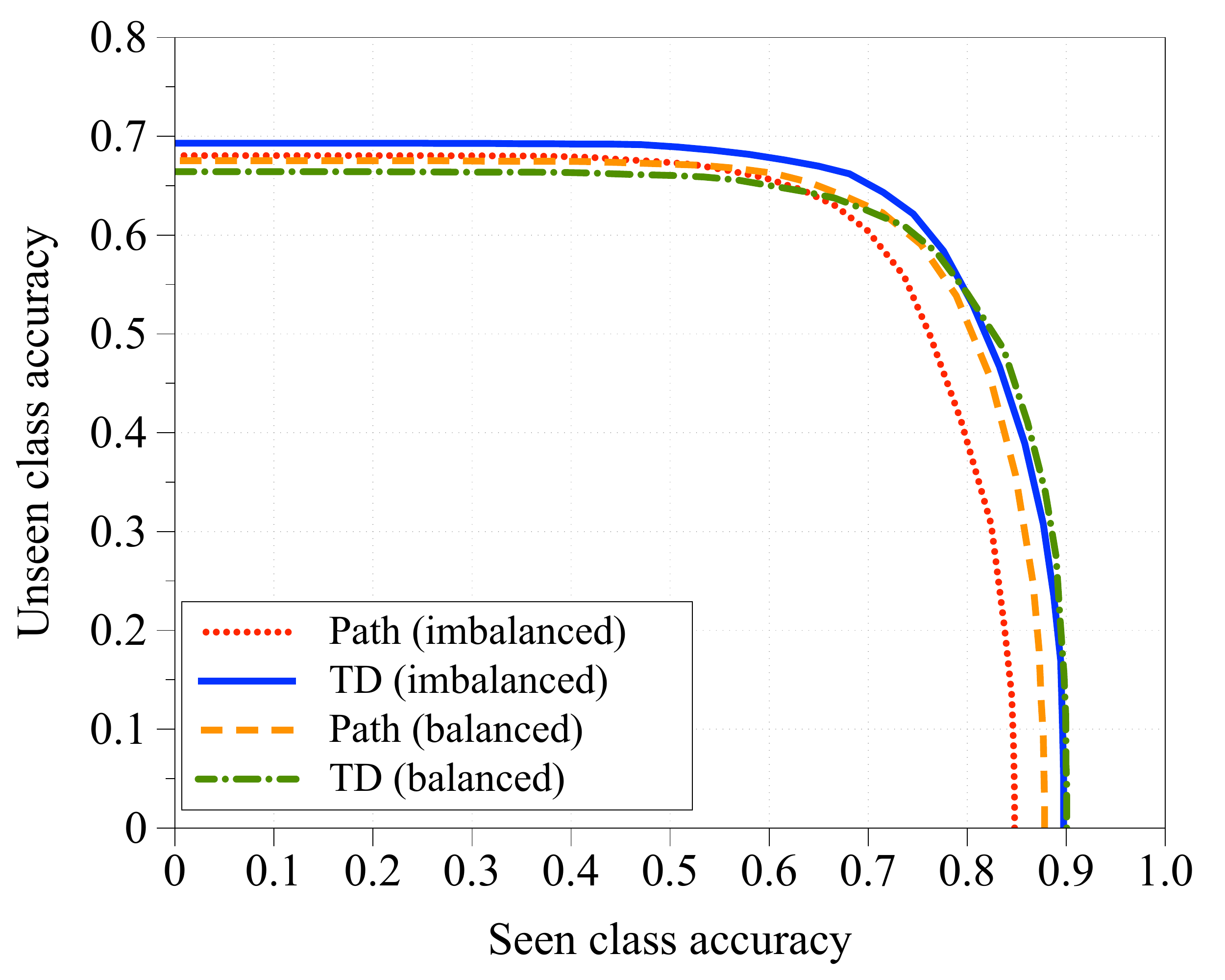} &
\includegraphics[width=.49\linewidth]{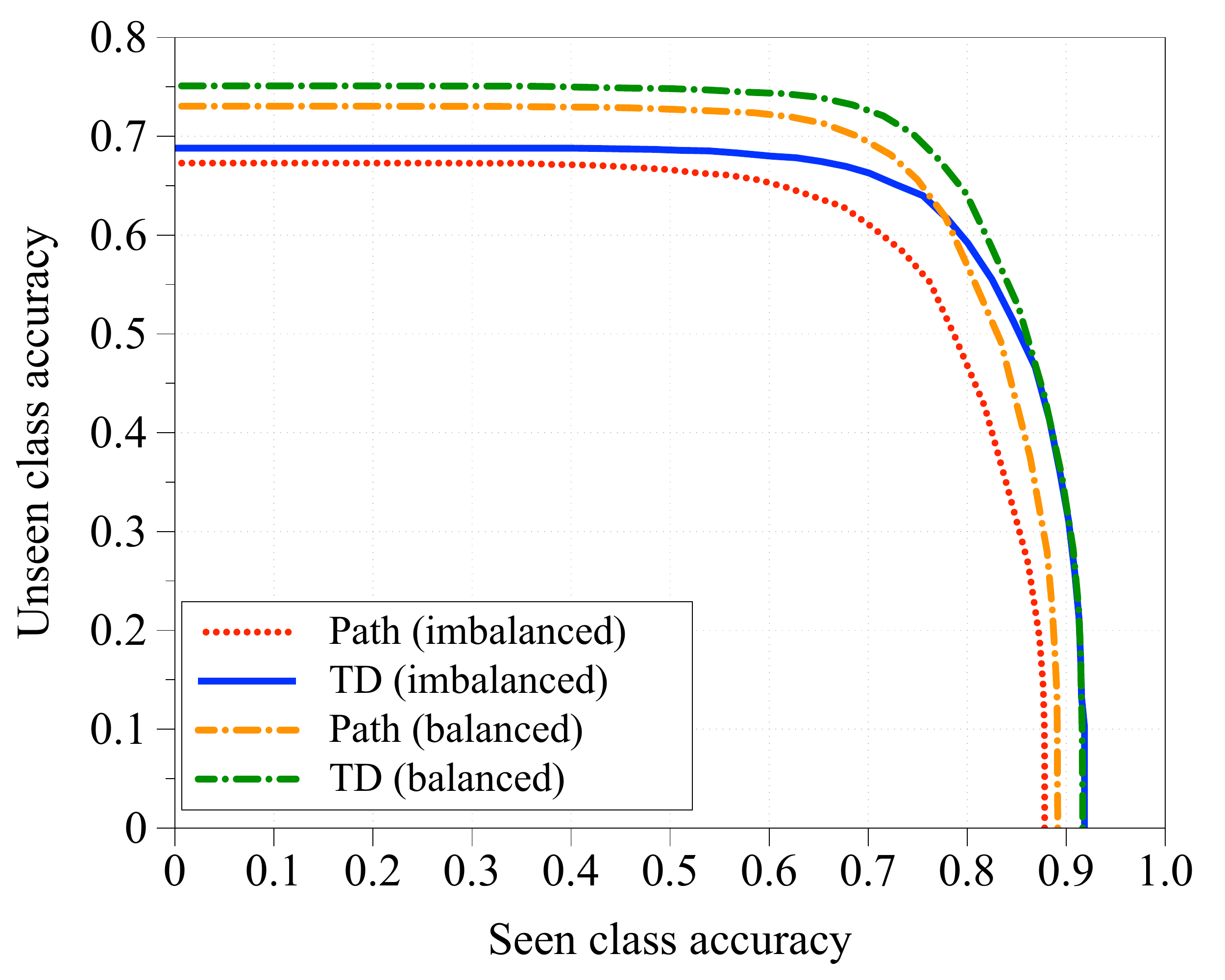} \cr
\end{tabular}
\vspace{-0.1in}
\cutcaptionup
\caption{
Seen-unseen class accuracy curves of the best combined models obtained by varying the unseen class score bias on AwA1 and AwA2, with the split of imbalanced taxonomy and that of balanced taxonomy.
``Path'' is the hierarchical embedding proposed in \cite{akata2015evaluation}, and ``TD'' is the embedding of the multiple softmax probability vector obtained from the proposed top-down method.
We remark that if the dataset has a balanced taxonomy, the overall performance can be improved.
}
\cutcaptiondown
\label{fig:a_comparison_zsl}
\end{figure}